\documentclass{article}

\usepackage{amsmath,amsfonts,bm}

\def\eqref#1{equation~\ref{#1}}

\def\1{\bm{1}}

\DeclareMathAlphabet{\mathsfit}{\encodingdefault}{\sfdefault}{m}{sl}
\SetMathAlphabet{\mathsfit}{bold}{\encodingdefault}{\sfdefault}{bx}{n}

\usepackage[final]{neurips_2022}

\usepackage[utf8]{inputenc} 
\usepackage[T1]{fontenc}    
\usepackage{hyperref}       
\usepackage{url}            
\usepackage{booktabs}       
\usepackage{amsfonts}       
\usepackage{nicefrac}       
\usepackage{microtype}      
\usepackage{xcolor}         
\usepackage{wrapfig}
\usepackage{subcaption}
\usepackage{epsfig}
\usepackage{algorithm}
\usepackage{algorithmicx}
\usepackage[noend]{algpseudocode}
\usepackage{todonotes}
\usepackage{placeins}

\title{Asynchronous Actor-Critic for Multi-Agent Reinforcement Learning}

\author{%
  Yuchen Xiao\\
  Khoury College of Computer Sciences\\
  Northeastern University\\
  Boston, MA 02115 \\
  \texttt{xiao.yuch@northeastern.edu} \\
  \And
  Weihao Tan\\
  Khoury College of Computer Sciences\\
  Northeastern University\\
  Boston, MA 02115 \\
  \texttt{w.tan@northeastern.edu} \\
  \And
  Christopher Amato\\
  Khoury College of Computer Sciences\\
  Northeastern University\\
  Boston, MA 02115 \\
  \texttt{c.amato@northeastern.edu}\\
}

\begin{document}

\maketitle

\begin{abstract}
    Synchronizing decisions across multiple agents in realistic settings is problematic since it requires agents to wait for other agents to terminate and communicate about termination reliably. 
    Ideally, 
    agents should learn and execute asynchronously instead. 
    Such asynchronous methods also allow temporally extended actions that can take different amounts of time based on the situation and action executed. 
    Unfortunately, current policy gradient methods are not applicable in asynchronous settings, as they assume that agents synchronously reason about action selection at every time step. 
    To allow asynchronous learning and decision-making, we formulate a set of asynchronous multi-agent actor-critic methods that allow agents to directly optimize asynchronous
    policies in three standard training paradigms: decentralized learning, centralized learning, and centralized training for decentralized execution. 
    Empirical results (in simulation and hardware) in a variety of realistic domains demonstrate the superiority of our approaches in large multi-agent problems and validate the effectiveness of our algorithms for learning high-quality and asynchronous solutions. 
\end{abstract}

\section{Introduction}

In recent years, multi-agent policy gradient methods using the actor-critic framework have achieved impressive success in solving a variety of cooperative and competitive domains~\citep{hideseek,LIIR,COMA,du2021learning,MAAC,M3DDPG,MADDPG,VDAC,VinyalsAlphaStar,SQDDPG,DOP,CM3,LICA}. However, as these methods assume synchronized primitive-action execution over agents, they struggle to solve large-scale real-world multi-agent problems that involve long-term reasoning and asynchronous behavior. 

Temporally-extended actions have been widely used in both learning and planning to improve scalability and reduce complexity. For example, they have come in the form of motion primitives~\citep{dalal2021raps,StulpS11}, skills~\citep{KonidarisKGB11, KonidarisKL18}, spatial action maps~\citep{wu2020spatial} or macro-actions~\citep{HeBR10, LIS123, MAGIC,NIPS2003_ea4eb493}. 
The idea of temporally-extended actions has also been incorporated into multi-agent approaches. In particular, we consider the \emph{Macro-Action Decentralized Partially Observable Markov Decision Process} (MacDec-POMDP)~\citep{AAMAS14AKK,AmatoJAIR19}.
The MacDec-POMDP is a general model for cooperative multi-agent problems with partial observability and (potentially) different action durations. As a result, agents can start and end macro-actions at different time steps so decision-making can be asynchronous. 

The MacDec-POMDP framework has shown strong scalability with planning-based methods (where the model is given)~\citep{RSS15,ICRA15MacDec,xiao_icra_2018,GDICE,IJRR17DecPOSMDP}.
In terms of multi-agent reinforcement learning (MARL), 
there have been many hierarchical approaches, 
they don't typically address asynchronicity since they assume agents' have  high-level decisions with the same  duration~\citep{schroeder:nips19,MAIntro-OptionQ,NachumAPGK19,WangK0LZITF20,wang:iclr2021,HAVEN,JiachenMAHRL}.
Only limited studies have considered asynchronicity~\citep{DOC,MendaCGBTKW19,xiao_corl_2019}, yet, none of them provides a general formulation for multi-agent policy gradients that allows agents to asynchronously learn and execute. 

In this paper, we assume a set of macro-actions has been predefined for each domain. This is well-motivated by the fact that, in real-world multi-robot systems, each robot is already equipped with certain controllers (e.g., a navigation controller, and a manipulation controller) that can be modeled as macro-actions~\citep{RSS15,IJRR17DecPOSMDP,wu2021spatial, xiao_corl_2019}. Similarly, as it is common to assume primitive actions are given in a typical RL domain, we assume the macro-actions are given in our case.  
The focus of the policy gradient methods is then on learning high-level policies over macro-actions.\footnote{Our approach could potentially also be applied to other models with temporally-extended actions \citep{IJRR17DecPOSMDP}.}

Our contributions include a set of macro-action-based multi-agent actor-critic methods that generalize their primitive-action counterparts. 
First, we formulate a \emph{macro-action-based independent actor-critic} (Mac-IAC) method. Although independent learning suffers from a theoretical curse of environmental non-stationarity, it allows fully online learning and may still work well in certain domains. Second, we introduce a \emph{macro-action-based centralized actor-critic} (Mac-CAC) method, for the case where full communication is available during execution.
We also formulate a centralized training for decentralized execution (CTDE) paradigm~\citep{KraemerB16,OliehoekSV08} variant of our method. 
CTDE has gained popularity since such methods can learn better decentralized policies by using centralized information during training. 
Current primitive-action-based multi-agent actor-critic methods typically use a centralized critic to optimize each decentralized actor. However, the asynchronous joint macro-action execution from the centralized perspective could be very different with the completion time being very different from each agent's decentralized perspective. 
To this end, we first present a \emph{Naive Independent Actor with Centralized Critic} (Naive IACC) method that naively uses a joint macro-action-value function as the critic for each actor's policy gradient estimation; and then propose a novel \emph{Independent Actor with Individual Centralized Critic} (Mac-IAICC) method that learns individual critics using centralized information to address the above challenge.    

We evaluate our proposed methods on diverse macro-action-based multi-agent problems: a benchmark Box Pushing domain~\citep{xiao_corl_2019}, a variant of the Overcooked domain~\citep{wu_wang2021too} and a larger warehouse service domain~\citep{xiao_corl_2019}. 
Experimental results show that our methods are able to learn high-quality solutions while primitive-action-based methods cannot, and show the strength of Mac-IAICC for learning decentralized policies over Naive IAICC and Mac-IAC. 
Decentralized policies learned by using Mac-IAICC are successfully deployed on real robots to solve a warehouse tool delivery task in an efficient way.
To our knowledge, this is the first general formalization of macro-action-based multi-agent actor-critic frameworks for the three state-of-the-art multi-agent training paradigms.

\section{Background}


\subsection{MacDec-POMDPs}

The macro-action decentralized partially observable Markov decision process (MacDec-POMDP)~\citep{AAMAS14AKK,AmatoJAIR19} incorporates the \emph{option} framework~\citep{Sutton:1999} into the Dec-POMDP by defining a set of macro-actions for each agent.  
Formally, a MacDec-POMDP is defined by a tuple $\langle I,S,A,M,\Omega,\zeta,T, R,O,Z,\mathbb{H},\gamma\rangle$, 
where $I$ is a set of agents;
$S$ is the environmental state space; 
$A= \times_{i\in I} A_i$ is the joint primitive-action space over each agent's primitive-action set $A_i$; 
$M=\times_{i\in I}M_i$ is the joint macro-action space over each agent's macro-action space $M_i$; 
$\Omega = \times_{i\in I} \Omega_i$ is the joint primitive-observation space over each agent's primitive-observation set $\Omega_i$;
$\zeta=\times_{i\in I}\zeta_i$ is the joint macro-observation space over each agent's macro-observation space $\zeta_i$; 
$T(s,\vec{a},s') = P(s' | s, \vec{a})$ is the environmental transition dynamics;
and $R(s,\vec{a})$ is a global reward function.
During execution, each agent independently selects a macro-action $m_i$ using a high-level policy $\Psi_i:H^M_i\times M_i\rightarrow[0,1]$ and captures a macro-observation $z_i\in\zeta_i$ according to the macro-observation probability function $Z_i(z_i,m_i,s')=P(z_i\mid m_i,s')$ when the macro-action terminates in a state $s'$. 
Each macro-action is represented as $m_i=\langle I_{m_i},\pi_{m_i}, \beta_{m_i} \rangle$,    
where the initiation set $I_{m_i}\subset H^M_i$ defines how to initiate a macro-action based on macro-observation-action history $H^M_i$ at the high-level; 
$\pi_{m_i}:H^A_i\times A_i\rightarrow [0,1]$ is the low-level policy for achieving a macro-action, and during the running, the agent receives a primitive-observation $o_i\in \Omega_i$ based on the observation function $O_i(o_i, a_i, s)=P(o_i | a_i,s)$ at every time step;
$\beta_{m_i}:H^A_i\rightarrow[0,1]$ is a stochastic termination function that determines how to terminate a macro-action based on primitive-observation-action history $H^A_i$ at the low-level.
The objective of solving MacDec-POMDPs with finite horizon is to find a joint high-level policy $\vec{\Psi}=\times_{i\in I} \Psi_i$ that maximizes the value, $V^{\vec{\Psi}}(s_{(0)})=\mathbb{E}\Big[\sum_{t=0}^{\mathbb{H}-1}\gamma^tr\big(s_{(t)},\vec{a}_{(t)}\big)\mid s_{(0)}, \vec{\pi}, \vec{\Psi}\Big]$, where $\gamma\in[0,1]$ is a discount factor, and $\mathbb{H}$ is the number of (primitive) time steps until the problem terminates (the horizon).

\subsection{Single-Agent Actor-Critic}

In single-agent reinforcement learning, the \emph{policy gradient theorem}~\citep{sutton2000policy} formulates a principled way to optimize a parameterized policy $\pi_\theta$ via gradient ascent on the policy's performance defined as $J(\theta)=\mathbb{E}_{\pi_\theta}\big[\sum_{t=0}^\infty\gamma^t r\big(s_{(t)},a_{(t)}\big)\big]$. In POMDPs, the gradient w.r.t.~parameters of a history-based policy $\pi_{\theta}(a\mid h)$ is expressed as: 
$\nabla_\theta J (\theta)=\mathbb{E}_{\pi_{\theta}}\Big[\nabla_\theta\log\pi_\theta(a\mid h)Q^{\pi_\theta}(h,a)\Big]$,
where $h$ is maintained by a recurrent neural network (RNN)~\citep{DRQN}.
The actor-critic framework~\citep{konda2000actor} learns an on-policy action-value function $Q^{\pi_{\theta}}_\phi(h,a)$ (critic) via \emph{temporal-difference} (TD) learning~\citep{Sutton88} to approximate the action-value for the policy (actor) updates. Variance reduction is commonly achieved by training a history-value function $V^{\pi_\theta}_{\mathbf{w}}(h)$ and using it as a baseline~\citep{WeaverT01} as well as bootstrapping to estimate the action-value. 
Accordingly, the policy gradient can be written as:  
\begin{equation}
        \nabla_\theta J (\theta)=\mathbb{E}_{\pi_{\theta}}\Big[\nabla_\theta\log\pi_\theta(a\mid h)\big(r + \gamma V^{\pi_\theta}_{\mathbf{w}}(h') - V^{\pi_\theta}_{\mathbf{w}}(h)\big)\Big]
        \label{AC2}
\end{equation}
where, $r$ is the immediate reward received by the agent at the corresponding time step.

\subsection{Independent Actor-Critic}
\label{IAC}

The single-agent actor-critic algorithm can be adapted to multi-agent problems in a simple way such that each agent independently learns its own actor and critic while treating other agents as part of the world~\citep{COMA}.
We consider a variance reduction version of \emph{independent actor-critic} (IAC) with the policy gradient as follows: 
\begin{equation}
        \nabla_{\theta_i} J (\theta_i)=\mathbb{E}_{\vec{\pi}_{\vec{\theta}}}\Big[\nabla_{\theta_i}\log\pi_{\theta_i}(a_i| h_i)\big(r + \gamma V^{\pi_{\theta_i}}_{\mathbf{w}_i}(h'_i) - V^{\pi_{\theta_i}}_{\mathbf{w}_i}(h_i)\big)\Big]
        \label{iac}
\end{equation}
where, $r$ is a shared reward over agents at every time step. Due to other agents' policy updating and exploring, from any agent's local perspective, the environment appears non-stationary which can lead to unstable learning of the critic without convergence guarantees~\citep{MADDPG}. This instability often prevents IAC from learning high-quality cooperative policies.     

\subsection{Independent Actor with Centralized Critic}
\label{IACC}

To address the above difficulties in independent learning approaches, centralized training for decentralized execution (CTDE) provides agents with access to global information during offline training while allowing agents to rely on only local information during online decentralized execution. Typically, the key idea of exploiting CTDE with actor-critic is to train a joint action-value function, $Q^{\vec{\pi}_{\vec{\theta}}}_\phi(\mathbf{x},\vec{a})$, as the centralized critic and use it to compute gradients w.r.t.~the parameters of each decentralized policy~\citep{COMA,MADDPG}.    
Although the centralized critic can facilitate the update of decentralized policies to optimize global collaborative performance, it also  introduces extra variance over other agents' actions~\citep{Lyu_aamas_2021,DOP}. Therefore, we consider the version of \emph{independent actor with centralized critic} (IACC) with a general variance reduction trick~\citep{COMA,VDAC}, the policy gradient of which is:  
\begin{equation}
        \nabla_{\theta_i} J (\theta_i)=\mathbb{E}_{\vec{\pi}_{\vec{\theta}}}\Big[\nabla_{\theta_i}\log\pi_{\theta_i}(a_i\mid h_i)\big(r + \gamma V^{\vec{\pi}_{\vec{\theta}}}_{\mathbf{w}}(\mathbf{x'})- V^{\vec{\pi}_{\vec{\theta}}}_{\mathbf{w}}(\mathbf{x})\big)\Big]
        \label{IACC-V}
\end{equation}
where, $\mathbf{x}$ represents the available centralized information (e.g., joint observation, joint observation-action history, or the true state). 

\subsection{Learning Macro-Action-Based Deep Q-Nets}
\label{macQ}

Previous MARL methods for Dec-POMDPs cannot work with the asynchronicity of macro-action-based agents, where agents may start and complete their macro-actions at different time steps. 
Recently, macro-action-based multi-agent DQNs have been proposed for 
MacDec-POMDPs~\citep{xiao_corl_2019}.  

For decentralized learning, a new buffer, \emph{Macro-Action Concurrent Experience Replay Trajectories} (Mac-CERTs), is designed for collecting each agent's macro-observation, macro-action, and reward information. 
In this buffer, the transition experience of each agent $i$ is represented as a tuple $\langle z_i, m_i, z_i', r_i^c\rangle$, where $r_i^c=\sum_{t=t_{m_i}}^{t_{m_i}+\tau_{m_i}-1}\gamma^{t-t_{m_i}}r_{(t)}$ is a cumulative reward of the macro-action taking $\tau_{m_i}$ time steps to be completed from its beginning time step $t_{m_i}$. 
During training, a mini-batch of concurrent sequential experiences is sampled from Mac-CERTs. Each agent independently accesses its own sampled experiences and obtains `squeezed' trajectories by removing the transitions in the middle of each macro-action execution, which produces a mini-batch of transitions when the corresponding macro-action terminates (i.e., removing time information).  
Updates for each macro-action-value function $Q_{\phi_i}(h_i,m_i)$ take place only when the agent's macro-action is complete by minimizing a TD loss over the `squeezed' data.  
In the centralized learning case, the objective is to learn a joint macro-action-value function $Q_\phi(\vec{h}, \vec{m})$. To this end, another special buffer called \emph{Macro-Action Joint Experience Replay Trajectories} (Mac-JERTs) is developed for collecting agents' joint transition experience at every time step and each is represented as a tuple $\langle\vec{z},\vec{m}, \vec{z\,}', \vec{r\,}^c \rangle$, where $\vec{r\,}^c=\sum_{t=t_{\vec{m}}}^{t_{\vec{m}}+\vec{\tau}_{\vec{m}}-1}\gamma^{t-t_{\vec{m}}}r_{(t)}$ is a shared joint cumulative reward from the beginning time step $t_{\vec{m}}$ of the joint macro-action $\vec{m}$ to its termination, defined as when \emph{any} agent finishes its own macro-action, after $\vec{\tau}_{\vec{m}}$ time steps. 
In each training iteration, the joint macro-action-value function is optimized over a mini-batch of `squeezed' (depending on each joint macro-action termination) sequential joint experiences via TD learning.  Other choices for what information to retain are also possible (e.g., the whole sequence of macro-actions or including time to complete) but this squeezing procedure was found to work well. 
In our macro-action-based actor-critic methods, we extend the above approaches to train critics on-policy, and the trajectory squeezing is changed variously for each method in order to achieve improved asynchronous macro-action-based policy updates via policy gradient.

\section{Approach}

MARL with asynchronous macro-actions is more challenging as it is difficult to determine \emph{when} to update each agent's policy and \emph{what} information to use. Although the macro-action-based DQN methods~\citep{xiao_corl_2019} (in Section~\ref{macQ}) give us the base to learn macro-action value functions, they do not directly extend to the policy gradient case, particularly in the case of centralized training for decentralized execution (CTDE). In this section, we propose principled formulations of on-policy macro-action-based multi-agent actor-critic methods for decentralized learning (Section~\ref{Mac-IAC}), centralized learning (Section~\ref{Mac-CAC}), and CTDE (Section~\ref{Mac-IACC}). 
In each case, we first introduce the version with a Q-value function as the critic and then present the variance reduction version 
\footnote{We use $h_i$ to represent an agent's local macro-observation-action history, and $\vec{h}$ to represent the joint history.}.

\subsection{Macro-Action-Based Independent Actor-Critic (Mac-IAC)}
\label{Mac-IAC}

Similar to the idea of IAC with primitive-actions (Section~\ref{IAC}), a straightforward extension is to have each agent independently optimize its own macro-action-based policy (actor) using a local macro-action-value function (critic). 
Hence, we start with deriving a \emph{macro-action-based policy gradient theorem} in Appendix~\ref{MacPGT} by incorporating the general Bellman equation for the state values of a macro-action-based policy~\citep{Sutton:1999} into the \emph{policy gradient theorem} in MDPs~\citep{sutton2000policy}, and then extend it to MacDec-POMDPs so that each agent can have the following policy gradient w.r.t.~the parameters of its macro-action-based policy $\Psi_{\theta_i}(m_i|h_i)$ as: $\nabla_{\theta_i} J (\theta_i)=\mathbb{E}_{\vec{\Psi}_{\vec{\theta}}}\biggr[\nabla_{\theta_i}\log\Psi_{\theta_i}(m_i\mid h_i)Q^{\Psi_{\theta_i}}_{\phi_i}(h_i,m_i)\biggr]$.
During training, each agent accesses its own trajectories and squeezes them in the same way as the decentralized case mentioned in Section~\ref{macQ} to train the critic $Q_{\phi_i}^{\Psi_{\theta_i}}(h_i,m_i)$ via on-policy TD learning and perform gradient ascent to update the policy when the agent's macro-action terminates. In our case, we train a local history value function $V^{\Psi_{\theta_i}}_{\mathbf{w}_i}(h_i)$ as each agent's critic and use it as a baseline to achieve variance reduction. The corresponding policy gradient is as follows: 
\begin{equation}
    \nabla_{\theta_i} J (\theta_i)=\mathbb{E}_{\vec{\Psi}_{\vec{\theta}}}\biggr[\nabla_{\theta_i}\log\Psi_{\theta_i}(m_i\mid h_i)\big(r^c_i + \gamma^{\tau_{m_i}}V^{\Psi_{\theta_i}}_{\mathbf{w}_i}(h_i')-V^{\Psi_{\theta_i}}_{\mathbf{w}_i}(h_i)\big)\biggr]
        \label{Mac-IAC-V}
\end{equation}
where, the cumulative reward $r^c_i$ is w.r.t. the execution of agent $i$'s macro-action $m_i$. 

\subsection{Macro-Action-Based Centralized Actor-Critic (Mac-CAC)}
\label{Mac-CAC}

In the fully centralized learning case, we treat all agents as a single joint agent to learn a centralized actor $\Psi_\theta(\vec{m}\mid\vec{h})$ with a centralized critic $Q^{\Psi_\theta}_\phi(\vec{h},\vec{m})$, and the policy gradient can be expressed as: $\nabla_{\theta} J (\theta)=\mathbb{E}_{\Psi_\theta}\biggr[\nabla_{\theta}\log\Psi_{\theta}(\vec{m}\mid \vec{h})Q^{\Psi_{\theta}}_{\phi}(\vec{h},\vec{m})\biggr]$.
Similarly, to achieve a lower variance optimization for the actor, we learn a centralized history value function $V^{\Psi_\theta}_{\mathbf{w}}(\vec{h})$ by minimizing a TD-error loss over joint trajectories that are squeezed w.r.t. each joint macro-action termination (when \emph{any} agent terminates its macro-action, defined in the centralized case in Section~\ref{macQ}). Accordingly, the policy's updates are performed when each joint macro-action is completed by ascending the following gradient:
\begin{equation}
    \nabla_{\theta} J (\theta)=\mathbb{E}_{\Psi_\theta}\biggr[\nabla_{\theta}\log\Psi_{\theta}(\vec{m}\mid \vec{h})\big(\vec{r\,}^c + \gamma^{\vec{\tau}_{\vec{m}}}V^{\Psi_\theta}_{\mathbf{w}}(\vec{h}')-V^{\Psi_\theta}_{\mathbf{w}}(\vec{h})\big)\biggr]
        \label{Mac-IAC-V}
\end{equation}
where the cumulative reward $\vec{r\,}^c$ is w.r.t.~the execution of the joint macro-action $\vec{m}$.

\subsection{Macro-Action-Based Independent Actor with Centralized Critic (Mac-IACC)}
\label{Mac-IACC}

As mentioned earlier, fully centralized learning requires perfect online communication that is often hard to guarantee, and fully decentralized learning suffers from environmental non-stationarity due to agents' changing policies. In order to learn better decentralized macro-action-based policies, in this section, we propose two macro-action-based actor-critic algorithms using the CTDE paradigm. The difference between a joint macro-action termination from the centralized perspective and a macro-action termination from each agent's local perspective gives rise to a new challenge: \emph{what kind of centralized critic should be learned and how should it be used to optimize decentralized policies where some have completed and some have not}, which we investigate below.

\textbf{Naive Mac-IACC.}
\label{N-Mac-IACC}
A naive way of incorporating macro-actions into a CTDE-based actor-critic framework is to directly adapt the idea of the primitive-action-based IACC (Section~\ref{IACC}) to have a shared joint macro-action-value function $Q^{\vec{\Psi}_{\vec{\theta}}}_\phi(\mathbf{x},\vec{m})$ in each agent's decentralized macro-action-based policy gradient as: $\nabla_{\theta_i} J (\theta_i)=\mathbb{E}_{\vec{\Psi}_{\vec{\theta}}}\biggr[\nabla_{\theta_i}\log\Psi_{\theta_i}(m_i\mid h_i)Q^{\vec{\Psi}_{\vec{\theta}}}_\phi(\mathbf{x},\vec{m})\biggr]$.
To reduce variance, with a value function $V^{\vec{\Psi}_{\vec{\theta}}}_\mathbf{w}(\mathbf{x})$ as the centralized critic, the policy gradient w.r.t.~the parameters of each agent's high-level policy can be rewritten as: 
\begin{equation}
    \nabla_{\theta_i}J(\theta_i) = \mathbb{E}_{\vec{\Psi}_{\vec{\theta}}}\biggr[\nabla_{\theta_i}\log\Psi_{\theta_i}(m_i \mid h_i)\big(\vec{r}^{\,c} + \gamma^{\vec{\tau}_{\vec{m}}}V_{\mathbf{w}}^{\vec{\Psi}_{\vec{\theta}}}(\mathbf{x}') - V_{\mathbf{w}}^{\vec{\Psi}_{\vec{\theta}}}(\mathbf{x})\big)\biggr]
        \label{N-Mac-IACC-V}
\end{equation}
Here, the critic 
is trained in the fully centralized manner described in Section~\ref{Mac-CAC} while allowing it to access additional global information (e.g., joint macro-observation-action history, ground truth state or both) represented by the symbol $\mathbf{x}$. 
However, updates of each agent's policy $\Psi_{\theta_i}(m_i\mid h_i)$ only occur at the agent's own macro-action termination time steps rather than depending on joint macro-action terminations in the centralized critic training.

\textbf{Independent Actor with Individual Centralized Critic (Mac-IAICC).}
\label{Mac-IAICC}
Note that naive Mac-IACC is technically incorrect. 
The cumulative reward $\vec{r\,}^c $ in Eq.~\ref{N-Mac-IACC-V} is based on the corresponding joint macro-action's termination that is defined as when \emph{any} agent finishes its own macro-action, which produces two potential issues: a) $\vec{r}^{\,c} +\gamma^{\vec{\tau}_{\vec{m}}} V_{\mathbf{w}}^{\vec{\Psi}_{\vec{\theta}}}(\mathbf{x}')$ may not estimate the value of the macro-action $m_i$ well as the reward does not depend on $m_i$'s termination; b) from agent $i$'s perspective, its policy gradient estimation may involve higher variance associated with the asynchronous macro-action terminations of other agents.  

To tackle the aforementioned issues, we propose to learn a separate centralized critic $V_{\mathbf{w}_i}^{\vec{\Psi}_{\vec{\theta}}}(\mathbf{x}')$ for each agent via TD-learning. In this case, the TD-error for updating $V_{\mathbf{w}_i}^{\vec{\Psi}_{\vec{\theta}}}(\mathbf{x}')$ is computed by using the reward $r^c_i$ that is accumulated purely based on the execution of the agent $i$'s macro-action $m_i$. With this TD-error estimation, each agent's decentralized macro-action-based policy gradient becomes: 
\begin{equation}
    \nabla_{\theta_i}J(\theta_i) = \mathbb{E}_{\vec{\Psi}_{\vec{\theta}}}\biggr[\nabla_{\theta_i}\log\Psi_{\theta_i}(m_i \mid h_i)\big(r^{c}_i + \gamma^{\tau_{m_i}}V_{\mathbf{w}_i}^{\vec{\Psi}_{\vec{\theta}}}(\mathbf{x}') - V_{\mathbf{w}_i}^{\vec{\Psi}_{\vec{\theta}}}(\mathbf{x})\big)\biggr]
            \label{Mac-IAICC}
\end{equation}

\vspace{-3mm}
Now, from agent $i$'s perspective, $r^{c}_i + \gamma^{\tau_{m_i}}V_{\mathbf{w}_i}^{\vec{\Psi}_{\vec{\theta}}}(\mathbf{x}')$ is able to offer a more accurate value prediction for the macro-action $m_i$, since both the reward, $r^{c}_i$ and the value function $V_{\mathbf{w}_i}^{\vec{\Psi}_{\vec{\theta}}}(\mathbf{x}') $ depend on agent $i$'s macro-action termination. Also, unlike the case in Naive Mac-IACC, other agents' terminations cannot lead to extra noisy estimated rewards w.r.t. $m_i$ anymore so that the variance on policy gradient estimation gets reduced. 
Then, updates for both the critic and the actor occur when the corresponding agent's macro-action ends and take the advantage of information sharing.
The pseudocode and detailed trajectory squeezing process for each proposed method are presented in Appendix~\ref{PCode}.

\section{Simulation Experiments}

\subsection{Domain Setup}
 
\begin{figure}[h!]
    \centering
    \captionsetup[subfigure]{labelformat=empty}
    \centering
    \subcaptionbox{(a) Box Pushing\vspace{2mm}\label{domain_BP}}
        [0.22\linewidth]{\includegraphics[height=1.6cm, width=1.6cm]{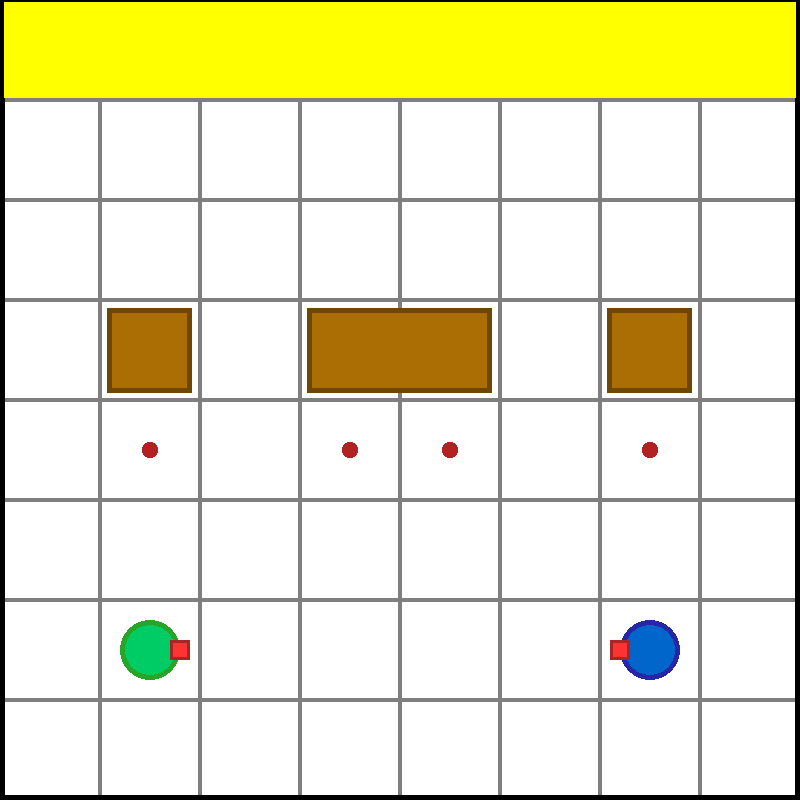}}
    ~
    \centering
    \subcaptionbox{(b) Overcooked-A\vspace{2mm}\label{domain_OA}}
        [0.22\linewidth]{\includegraphics[height=1.6cm, width=1.6cm]{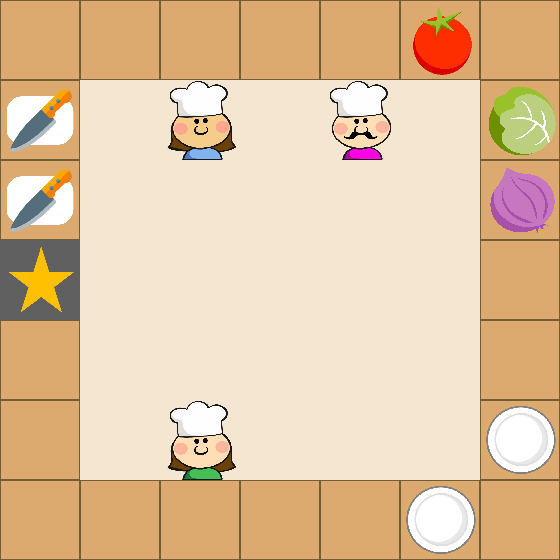}}
    ~
    \centering
    \subcaptionbox{(c) Overcooked-B\vspace{2mm}\label{domain_OB}}
        [0.22\linewidth]{\includegraphics[height=1.6cm, width=1.6cm]{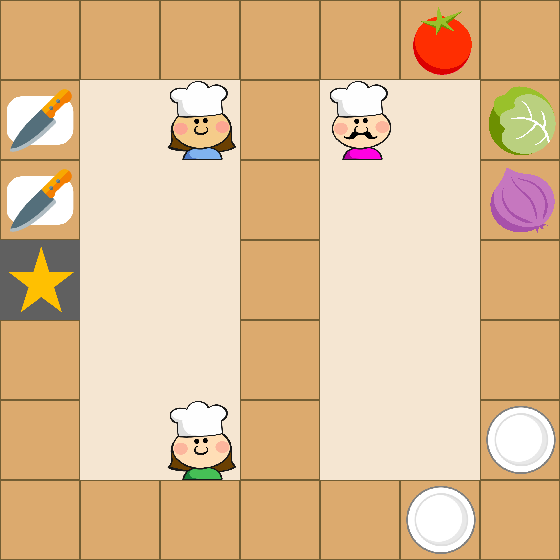}}
    ~
    \centering
    \subcaptionbox{(d) Overcooked Salad Recipe\vspace{1mm}\label{domain_OB_recipe}}
        [0.26\linewidth]{\includegraphics[height=1.6cm, width=2.1cm]{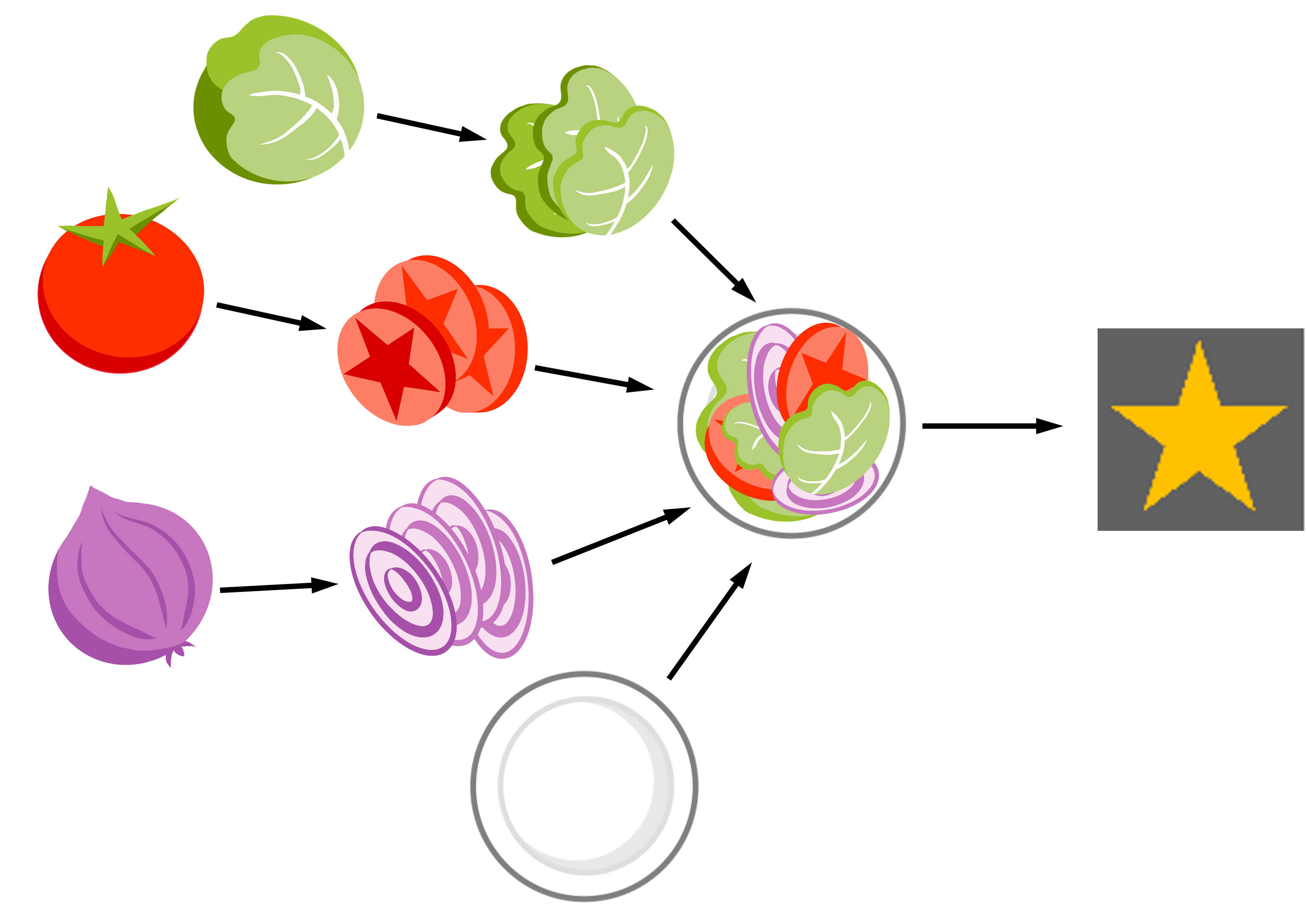}}
    ~
    \centering
    \subcaptionbox{(e) Warehouse-A\vspace{1mm}\label{domain_wtdA}}
        [0.22\linewidth]{\includegraphics[height=1.6cm]{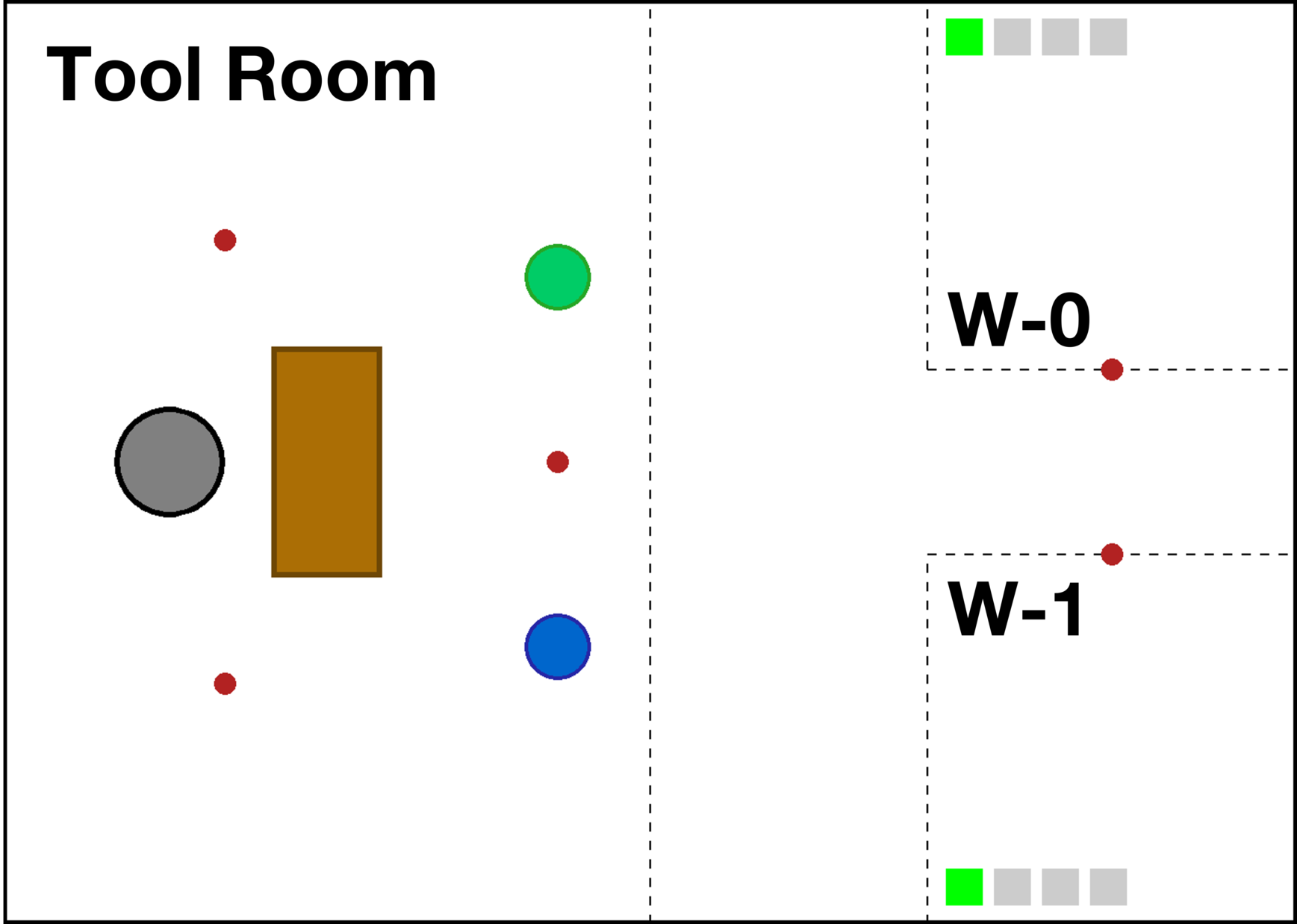}}
     ~
    \centering
    \subcaptionbox{(f) Warehouse-B\vspace{1mm}\label{domain_wtdC}}
        [0.23\linewidth]{\includegraphics[height=1.6cm]{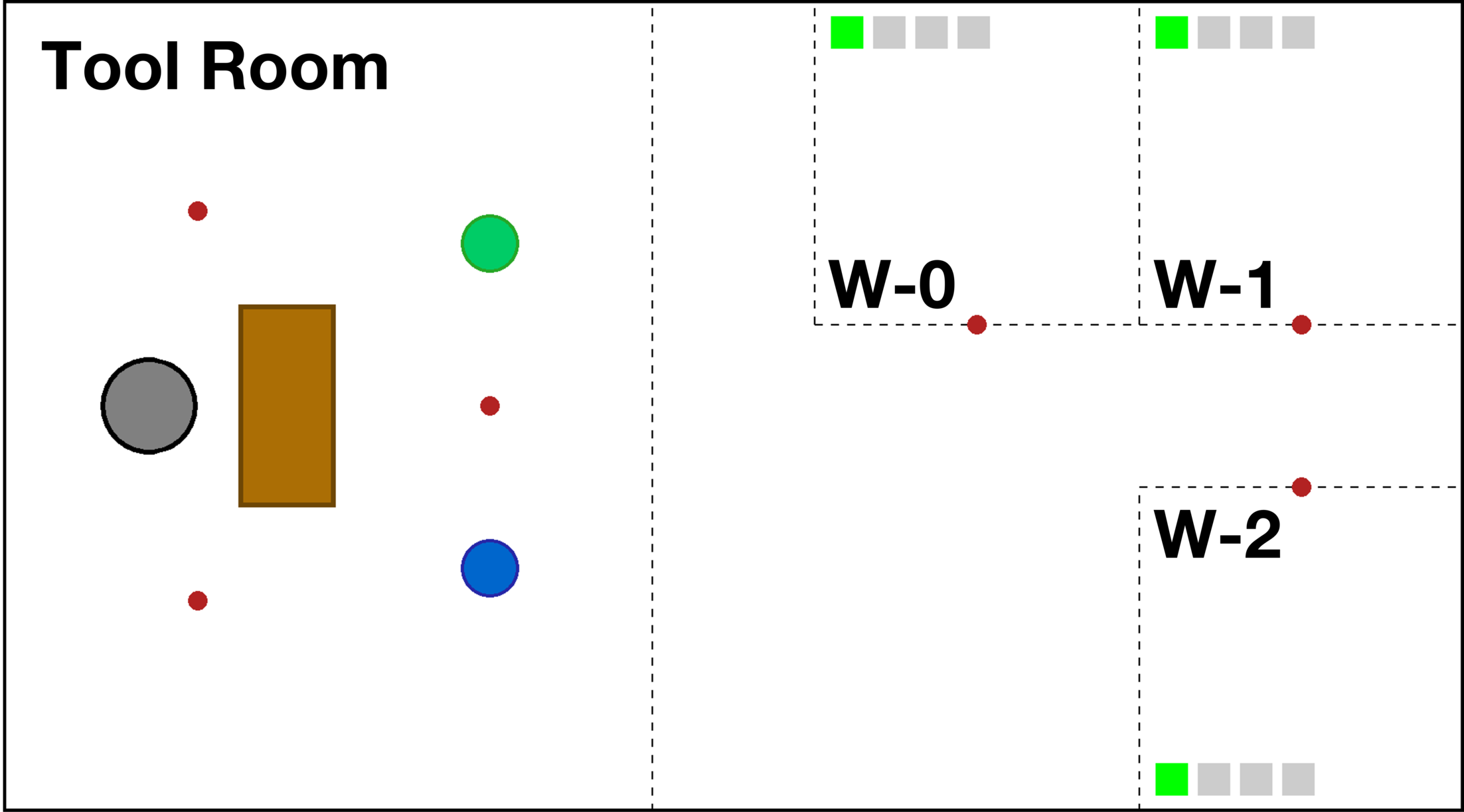}}
     ~
    \centering
    \subcaptionbox{(g) Warehouse-C\vspace{1mm}\label{domain_wtdD}}
        [0.23\linewidth]{\includegraphics[height=1.6cm]{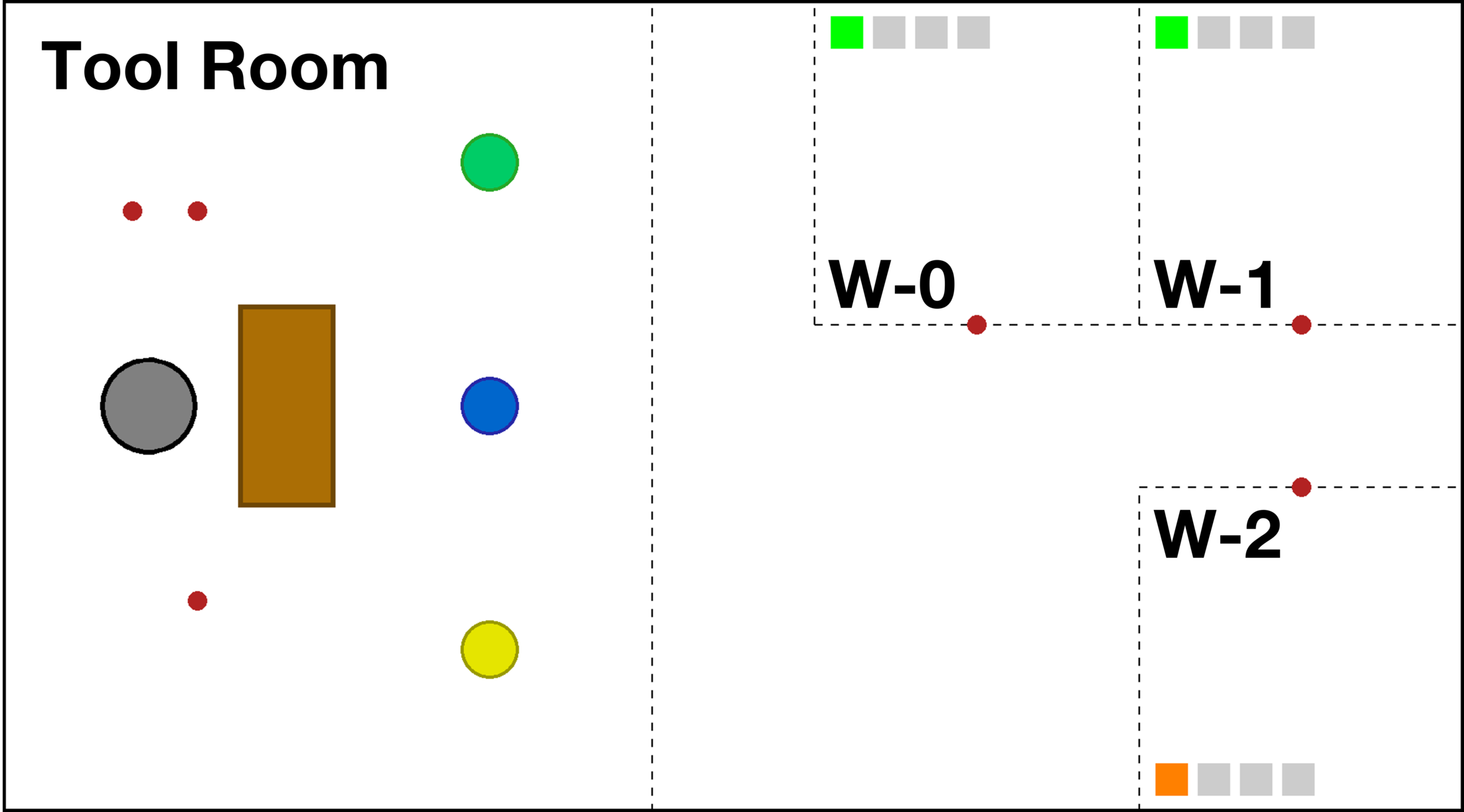}}
    ~
    \centering
    \subcaptionbox{(h) Warehouse-D\label{domain_wtdE}}
        [0.23\linewidth]{\includegraphics[height=1.6cm]{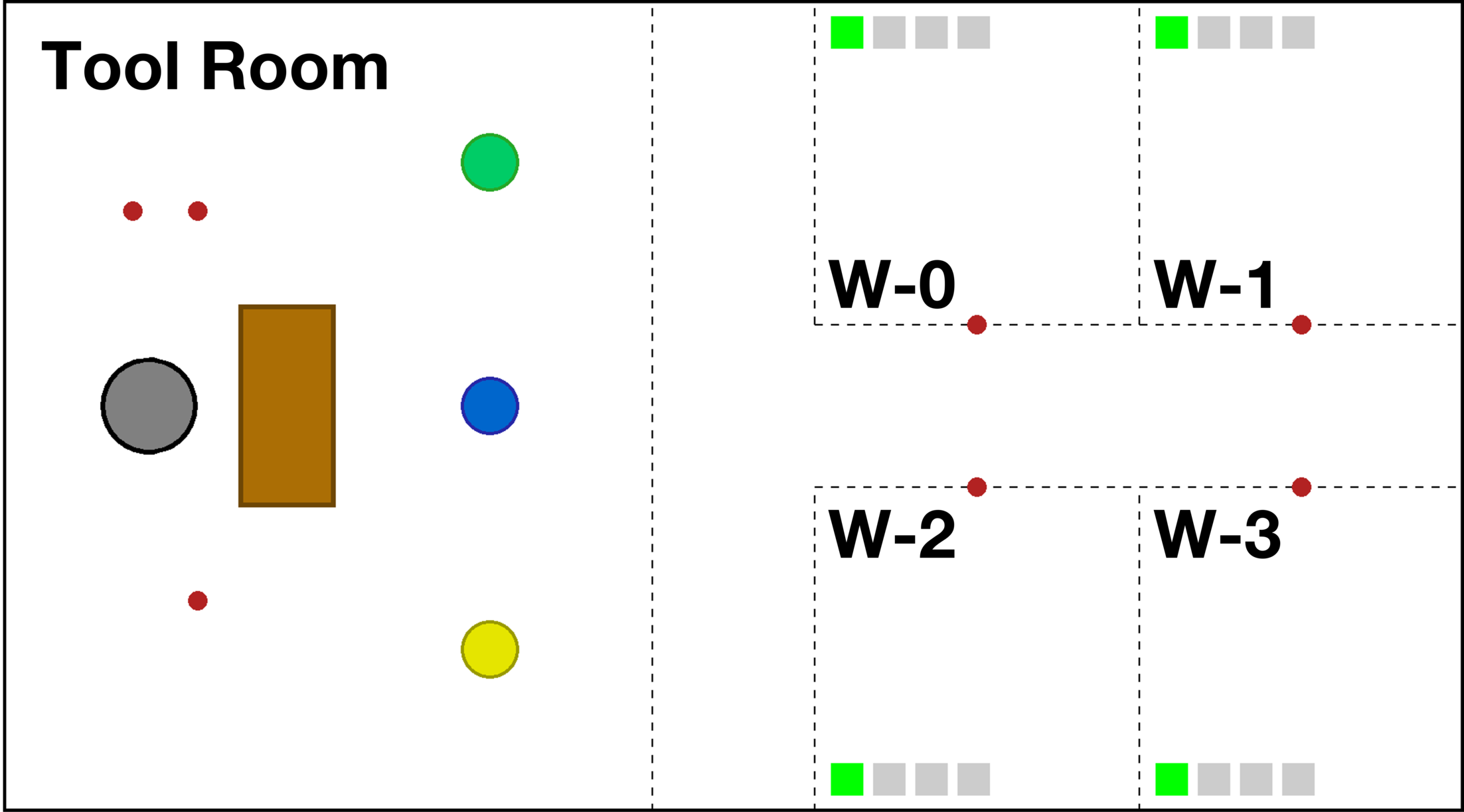}}
    \caption{Experimental environments.}
    \label{domains}
\end{figure}

We investigate the performance of our algorithms over a variety of multi-agent problems with macro-actions (Fig.~\ref{domains}): Box Pushing~\citep{xiao_corl_2019}, Overcooked~\citep{wu_wang2021too},
and a larger Warehouse Tool Delivery~\citep{xiao_corl_2019} domain. 
Macro-actions are defined by using prior domain knowledge as they are straightforward in these tasks. Typically, we also include primitive-actions into macro-action set (as one-step macro-actions), which gives agents the chance to learn more complex policies that use both when it is necessary.
We describe the domains' key properties here and have more details in Appendix~\ref{A-Domain}.  

\textbf{Box Pushing} (Fig.~\ref{domain_BP}). 
The optimal solution for the two agents is to cooperatively push the big box to the yellow goal area for a terminal reward, but partial observability makes this difficult. 
Specifically, robots have four primitive-actions: \emph{move forward}, \emph{turn-left}, \emph{turn-right} and \emph{stay}. 
In the macro-action case, each robot has three one-step macro-actions: \emph{\textbf{Turn-left}}, \emph{\textbf{Turn-right}}, and \emph{\textbf{Stay}}, as well as three multi-step macro-actions: \emph{\textbf{Move-to-small-box(i)}} and
\emph{\textbf{Move-to-big-box(i)}} navigate the robot to the red spot below the corresponding box and terminate with the robot facing the box; 
\emph{\textbf{Push}} causes the robot to keep moving forward until arriving at the world's boundary (potentially pushing the small box or trying to push the big one). 
The big box only moves if both agents push it together. 
Each robot can only observe the status (\emph{empty}, \emph{teammate}, \emph{boundary}, \emph{small or big box}) of the cell in front of it. 
A penalty is issued when any robot hits the boundary or pushes the big box alone.    

\textbf{Overcooked} (Fig.~\ref{domain_OA} - \ref{domain_OB}).
Three agents must learn to cooperatively prepare a lettuce-tomato-onion salad and deliver it to the `star' cell.
The challenge is that the salad's recipe (Fig.~\ref{domain_OB_recipe}) is unknown to agents.
With primitive-actions (\emph{move up}, \emph{down}, \emph{left}, \emph{right}, and \emph{stay}), agents can move around and achieve picking, placing, chopping and delivering by standing next to the corresponding cell and moving against it (e.g., in Fig.~\ref{domain_OA}, the pink agent can \emph{move right} and then \emph{move up} to pick up the tomato).
We describe the major function of macro-actions below and full details (e.g., termination conditions) are included in Appendix~\ref{A-Overcooked}.
Each agent's macro-action set consists of: 
a) five one-step macro-actions that are the same as the primitive ones; 
b) \emph{\textbf{Chop}}, 
cuts a raw vegetable into pieces when the agent stands next to a cutting board and an unchopped vegetable is on the board, otherwise it does nothing;
c) long-term navigation macro-actions: \emph{\textbf{Get-Lettuce}}, \emph{\textbf{Get-Tomato}},
\emph{\textbf{Get-Onion}},
\emph{\textbf{Get-Plate-1/2}}, \emph{\textbf{Go-Cut-Board-1/2}} and \emph{\textbf{Deliver}}, which navigate the agent to the location of the corresponding object with various possible terminal effects (e.g., holding a vegetable in hand, placing a chopped vegetable on a plate, arriving at the cell next to a cutting board, delivering an item to the star cell, or immediately terminating when any property condition does not hold, e.g., no path is found or the vegetable/plate is not found); 
d) \emph{\textbf{Go-Counter}} (only available in Overcook-B, Fig.~\ref{domain_OB}), navigates an agent to the center cell in the middle of the map when the cell is not occupied, otherwise, it moves to an adjacent cell. 
If the agent is holding an object or one is at the cell, the object will be placed or picked up. 
Each agent only observes the \emph{positions} and \emph{status} of the entities within a $5\times5$ square  centered on the robot.

\textbf{Warehouse Tool Delivery} (Fig.~\ref{domain_wtdA} - \ref{domain_wtdE}). 
In each workshop (e.g., W-0), a human is working on an assembly task (involving 4 sub-tasks that each takes a number of time steps to complete) and requires three different tools for future sub-tasks to continue. 
A robot arm (grey) must find tools for each human on the table (brown) and pass them to mobile robots (green, blue and yellow) who are responsible for delivering tools to humans. 
Note that, the correct tools needed by each human are unknown to robots, which has to be learned during training in order to perform efficient delivery. A delayed delivery leads to a penalty.
We consider variants with two or three mobile robots and two to four humans to examine the scalability of our methods
(Fig.~\ref{domain_wtdC} - \ref{domain_wtdE}). We also consider one faster human (orange) to check if robots can prioritize him (Fig.~\ref{domain_wtdD}). 
Mobile robots have the following macro-actions:
\emph{\textbf{Go-W(i)}}, moves to the waypoint (red) at workshop $i$; 
\emph{\textbf{Go-TR}}, goes to the waypoint at the right side of the tool room (covered by the blue robot in Fig.~\ref{domain_wtdD} and \ref{domain_wtdE});
and \emph{\textbf{Get-Tool}}, navigates to a pre-allocated waypoint (that is different for each robot to avoid collisions) next to the robot arm and waits there until either receiving a tool or 10 time steps have passed. 
The robot arm's applicable macro-actions are: \emph{\textbf{Search-Tool(i)}}, finds tool $i$ and places it in a staging area (containing at most two tools) on the table, and
otherwise, it freezes the robot for the amount of time the action would take when the area is fully occupied; 
\emph{\textbf{Pass-to-M(i)}}, passes the first staged tool to mobile robot $i$; 
and \emph{\textbf{Wait-M}}, waits for 1 time step. 
The robot arm only observes the \emph{type} of each tool in the staging area and \emph{which mobile robot} is waiting at the adjacent waypoints. 
Each mobile robot always knows its \emph{position} and the \emph{type} of tool that it is carrying, and can observe the \emph{number} of tools in the staging area or the \emph{sub-task} a human is working on only when at the tool room or the workshop respectively. 


\subsection{Results and Discussions}
\label{RandD}

We evaluate performance 
of one training trial with a mean discounted return measured by periodically (every 100 episodes) evaluating the learned policies over 10 testing episodes. 
We plot the average performance of each method over 20 independent trials with one standard error and smooth the curves over 10 neighbors. We also show the optimal expected return in Box Pushing domain as a dash-dot line.
More training details are in Appendix~\ref{A-Train}.

\begin{figure}[t!]
    \centering
    \captionsetup[subfigure]{labelformat=empty}
    \centering
    \subcaptionbox{}
        [0.9\linewidth]{\includegraphics[scale=0.13]{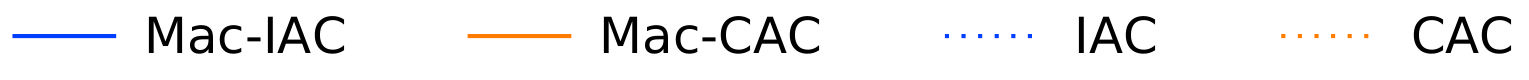}}
    ~
    \centering
    \subcaptionbox{}
        [0.23\linewidth]{\includegraphics[height=2.5cm]{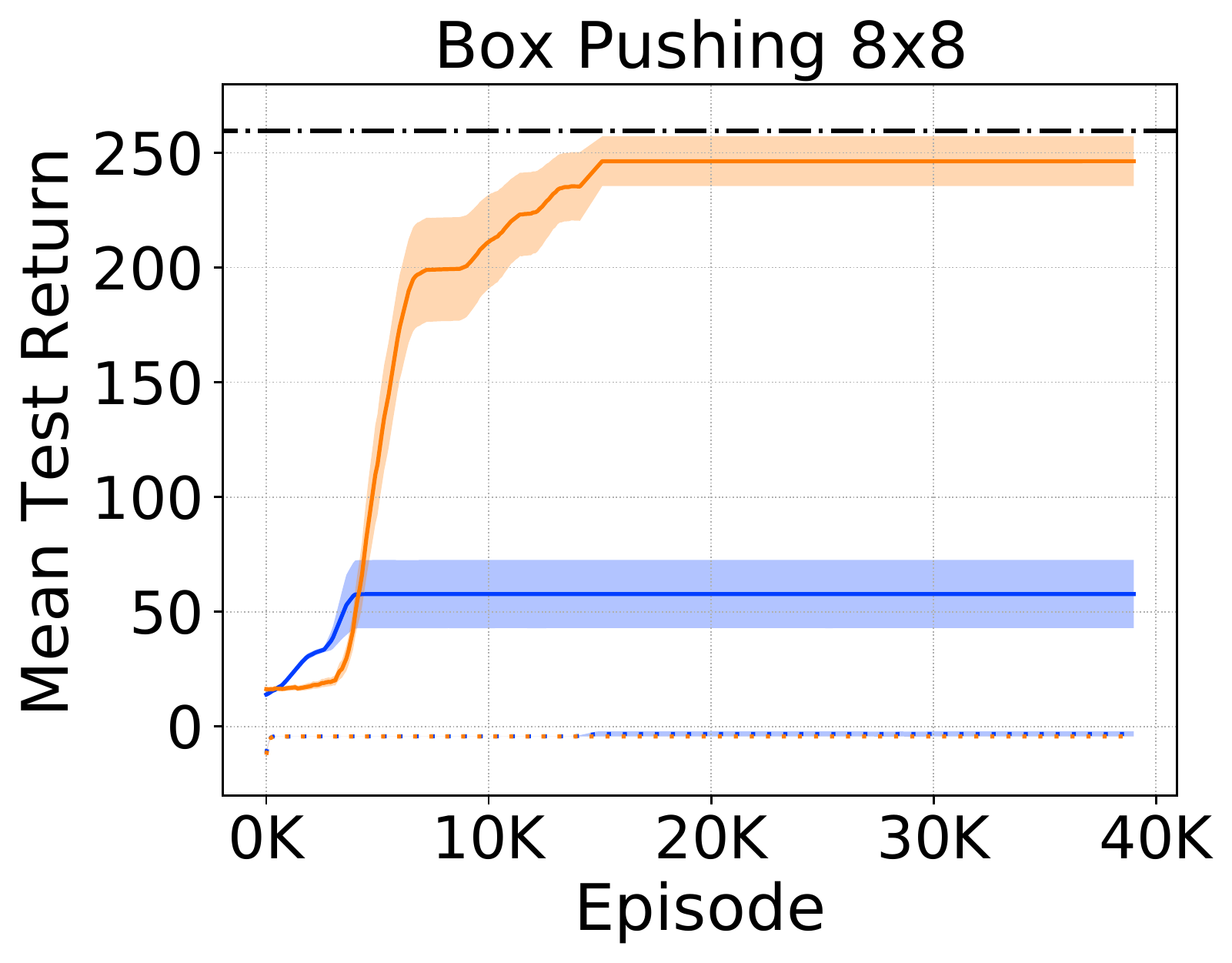}}
    ~
    \centering
    \subcaptionbox{}
        [0.23\linewidth]{\includegraphics[height=2.5cm]{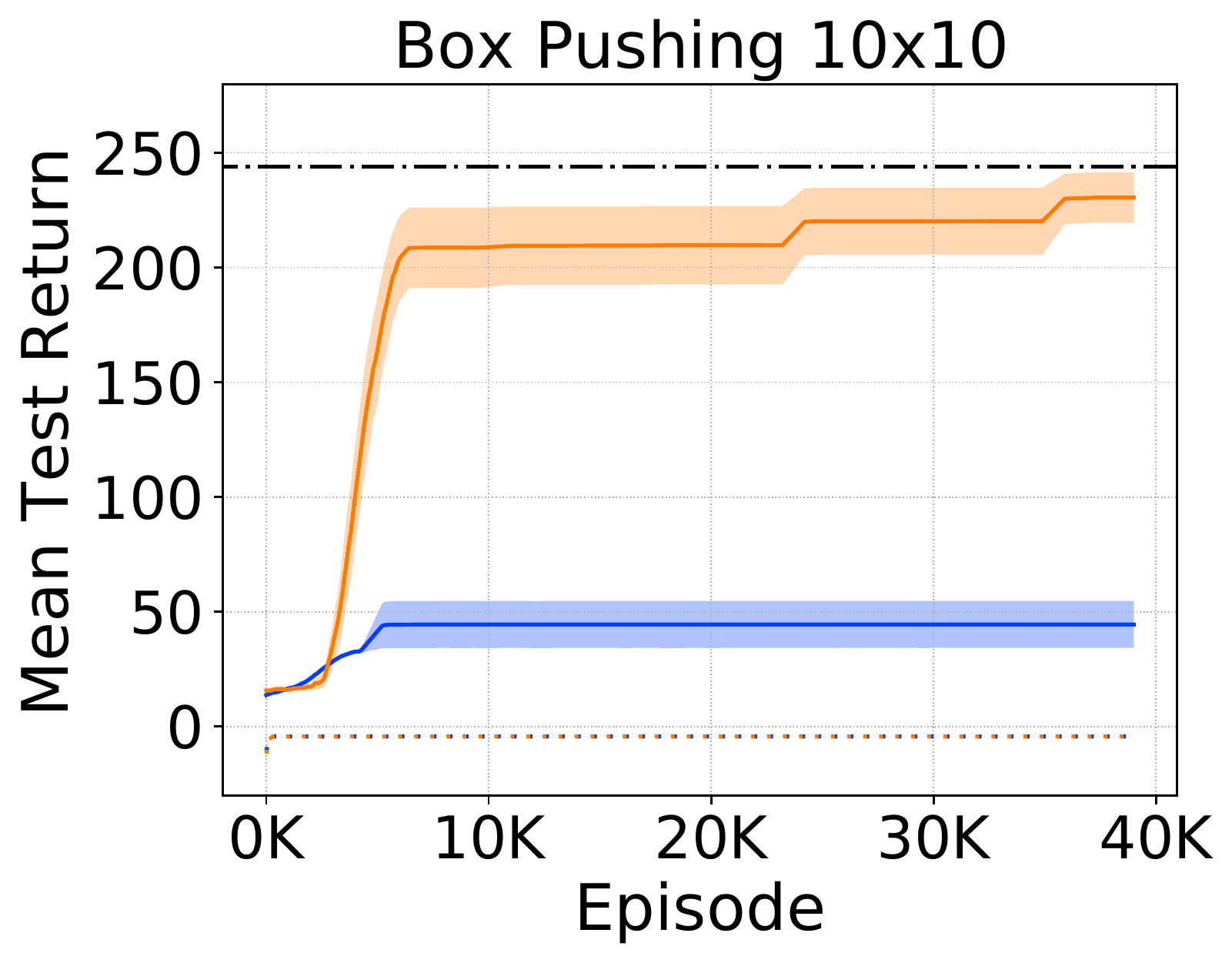}}
    ~
    \centering
    \subcaptionbox{}
        [0.23\linewidth]{\includegraphics[height=2.5cm]{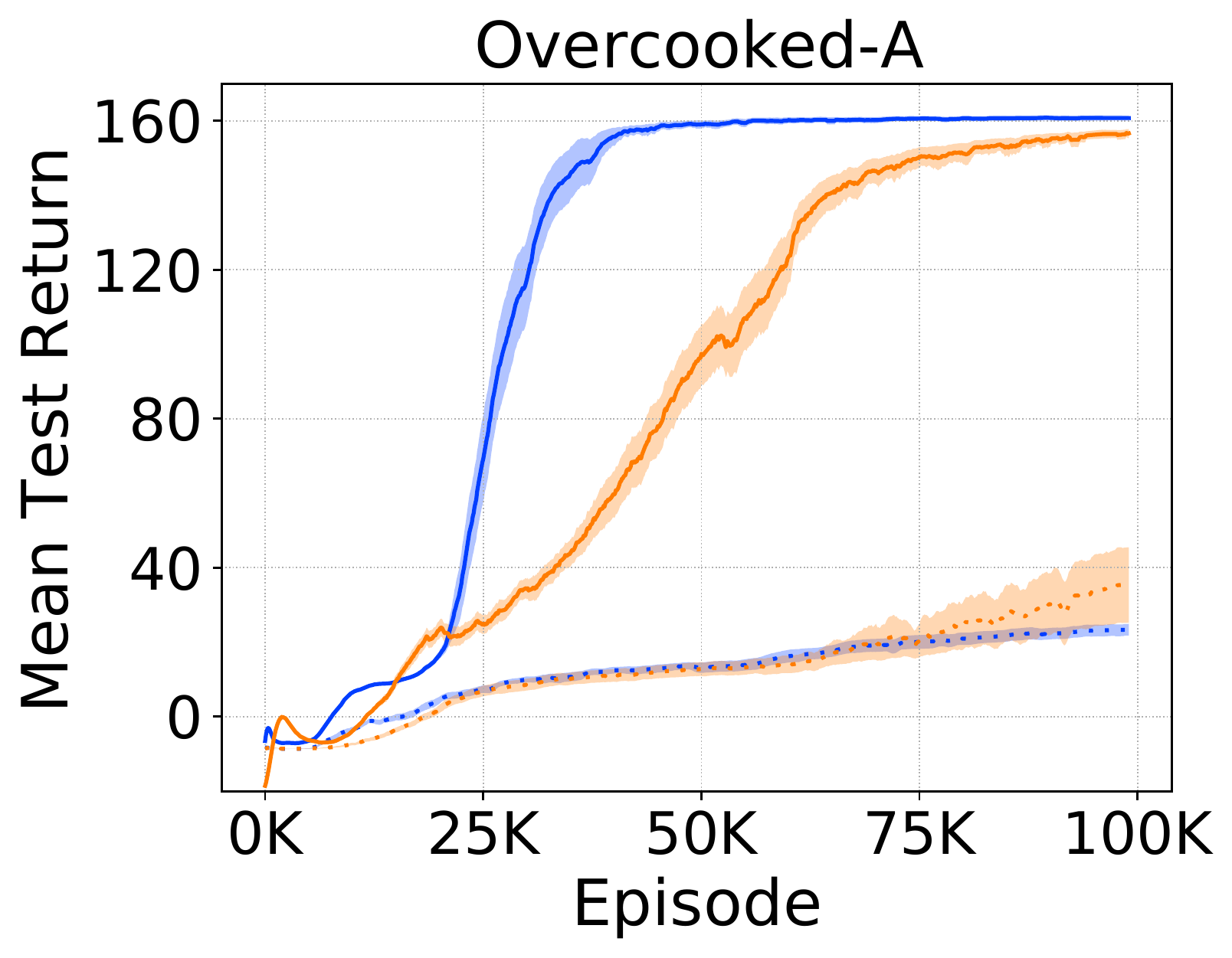}}
    ~
    \centering
    \subcaptionbox{}
        [0.23\linewidth]{\includegraphics[height=2.5cm]{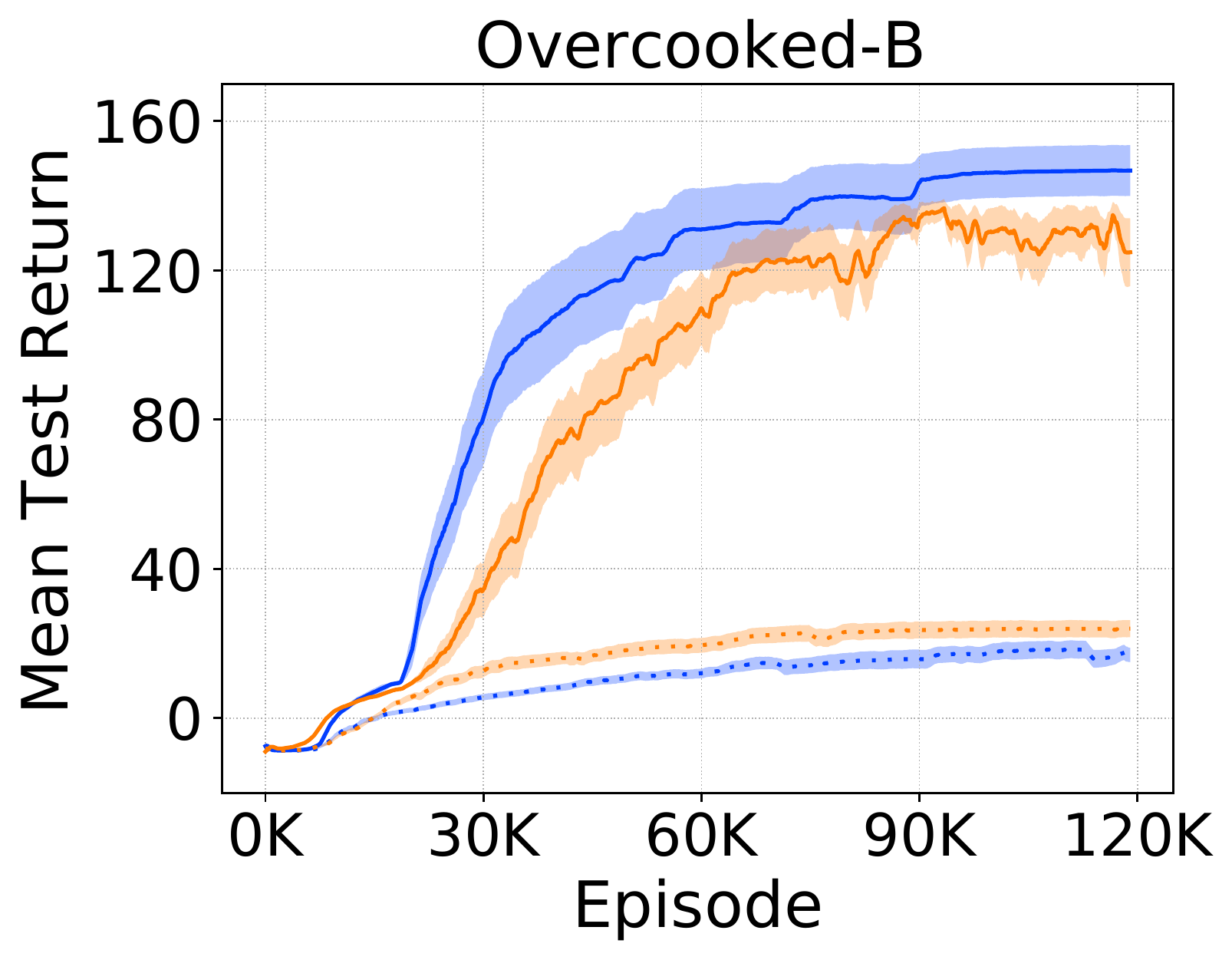}}
    \caption{Decentralized learning and centralized learning with macro-actions vs primitive-actions.}
    \label{mac_adv}
\end{figure}

\textbf{Advantages of learning with macro-actions}. We first present a comparison of our macro-action-based actor-critic methods against the primitive-action-based methods in fully decentralized and fully centralized cases. 
We consider various grid world sizes of the  Box Pushing domain (top row in Fig.~\ref{mac_adv} and two Overcooked scenarios (bottom row in Fig.~\ref{mac_adv}). The results show  significant performance improvements of using macro-actions over primitive-actions. More concretely, in the Box Pushing domain, reasoning about primitive movements at every time step 
makes the problem intractable so the robots cannot learn any good behaviors in primitive-action-based approaches other than to keep moving around. 
Conversely, 
Mac-CAC reaches near-optimal performance, enabling the robots to push the big box together.
Unlike the centralized critic which can access joint information, even in the macro-action case, it is hard for each robot's decentralized critic to correctly measure the responsibility for a penalty caused by a teammate pushing the big box alone. Mac-IAC thus converges to a local-optima of pushing two small boxes in order to avoid getting the penalty. 

In the Overcooked domain, an efficient solution requires the robots to asynchronously work on independent subtasks (e.g., in scenario A, one robot gets a plate while another two robots pick up and chop vegetables; and in scenario B, the right robot transports items while the left two robots prepare the salad). This large amount of independence explains why Mac-IAC can solve the task well. 
This also indicates that using local information is enough for robots to achieve high-quality behaviors. As a result,  Mac-CAC learns slower because it must figure out the redundant part of joint information in much larger joint macro-level history and action spaces than the spaces in the decentralized case. 
The primitive-action-based methods begin to learn, but perform poorly in such long-horizon tasks.

\begin{figure*}[t!]
    \centering
    \captionsetup[subfigure]{labelformat=empty}
    \centering
    \subcaptionbox{}
        [0.9\linewidth]{\includegraphics[scale=0.15]{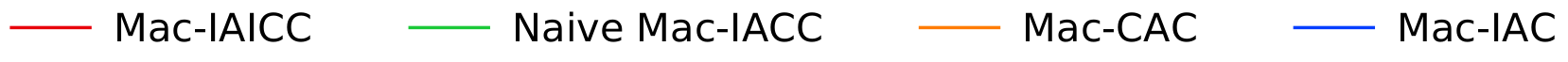}}
    ~
    \centering
    \subcaptionbox{}
        [0.23\linewidth]{\includegraphics[height=2.5cm]{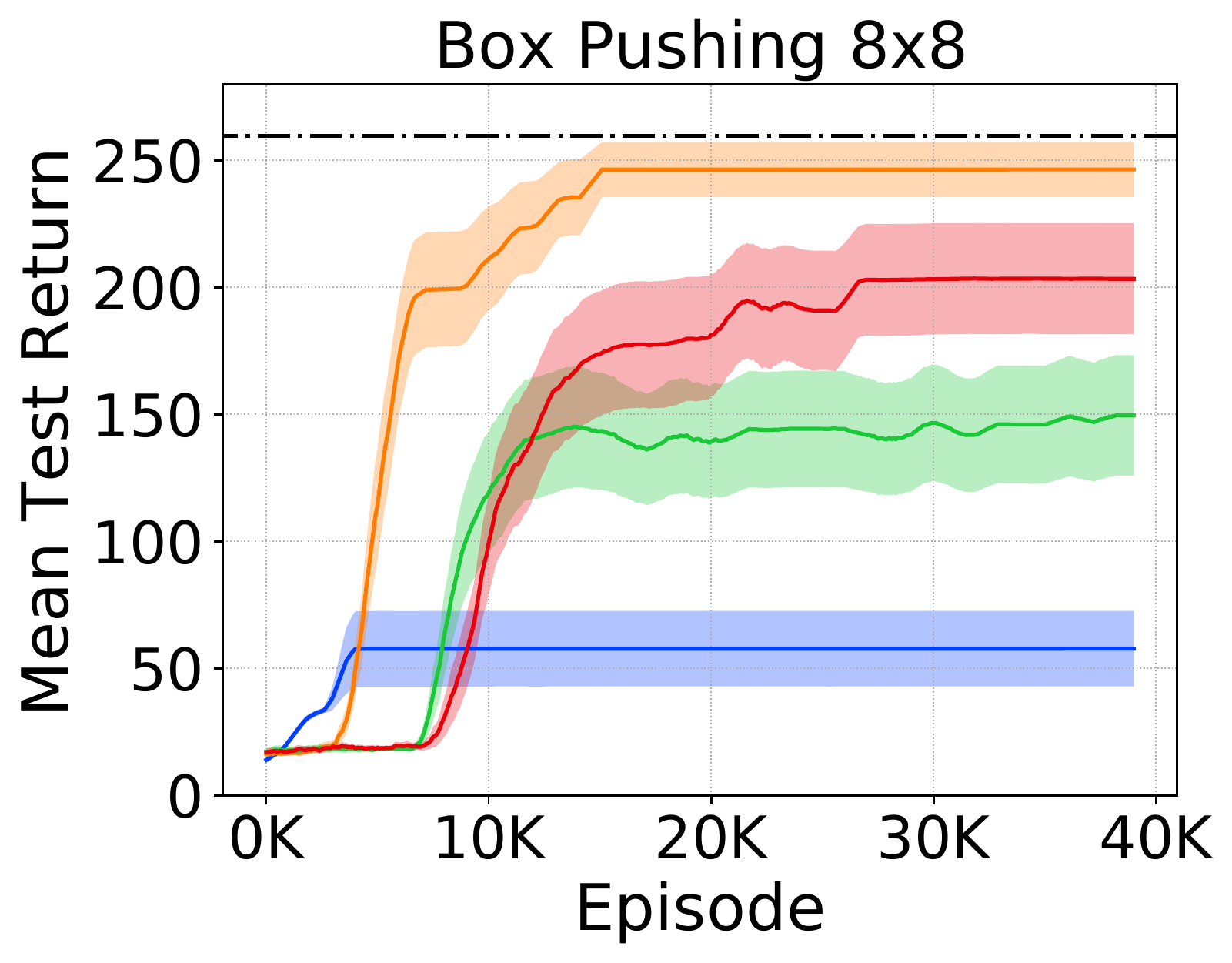}}
    ~
    \centering
    \subcaptionbox{}
        [0.23\linewidth]{\includegraphics[height=2.5cm]{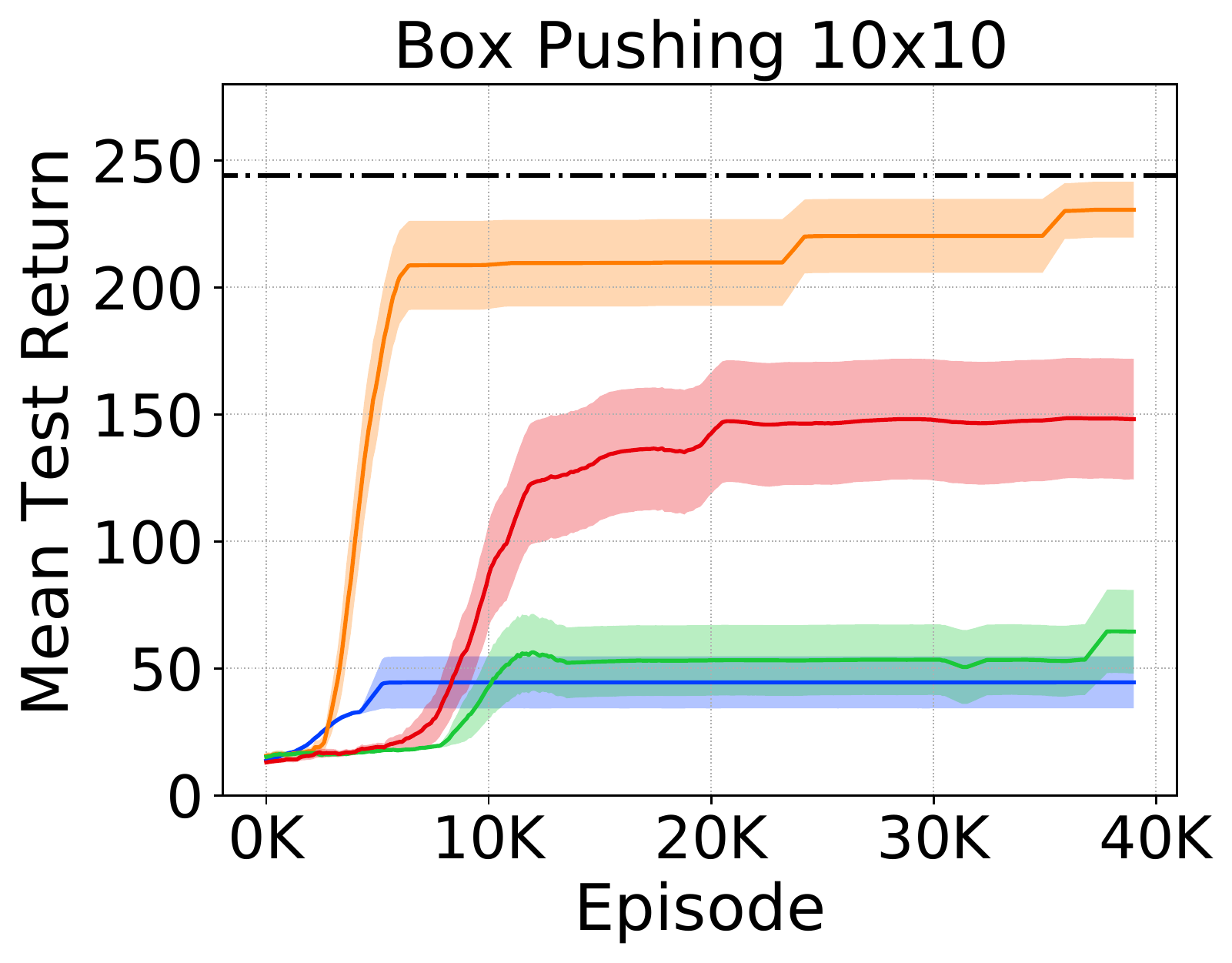}}
    ~
    \centering
    \subcaptionbox{}
        [0.23\linewidth]{\includegraphics[height=2.5cm]{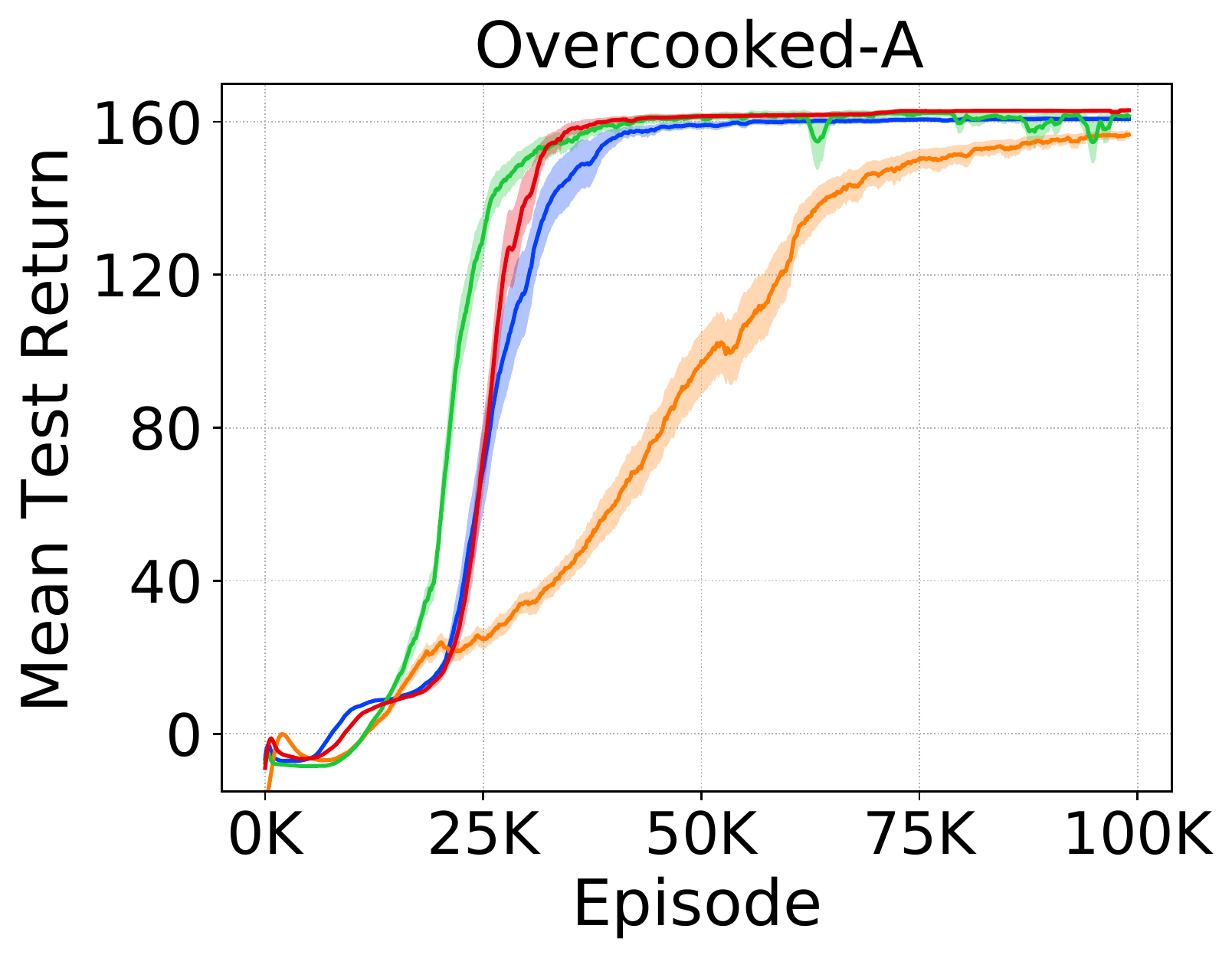}}
    ~
    \centering
    \subcaptionbox{}
        [0.23\linewidth]{\includegraphics[height=2.5cm]{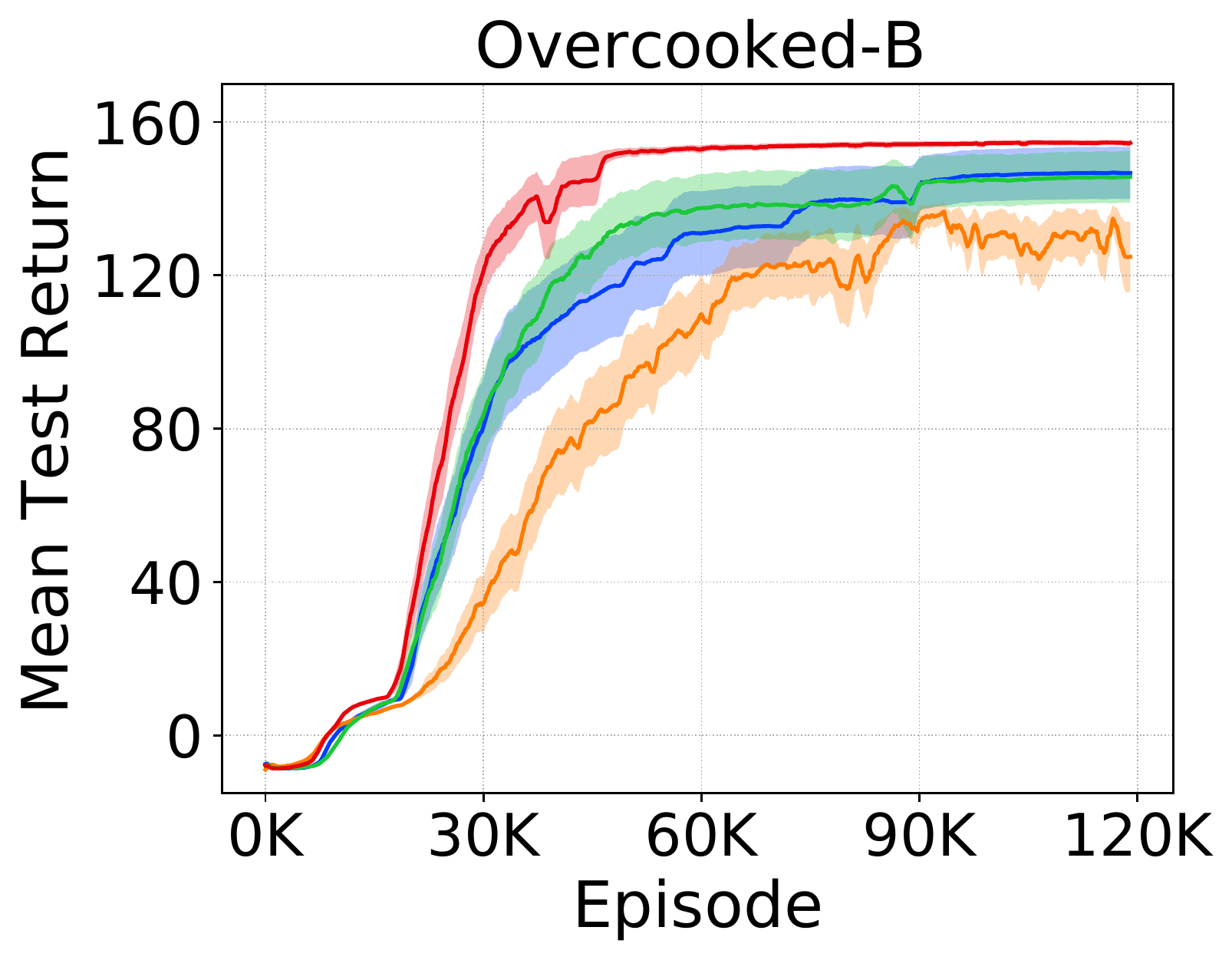}}
    ~
    \centering
    \subcaptionbox{}
        [0.23\linewidth]{\includegraphics[height=2.5cm]{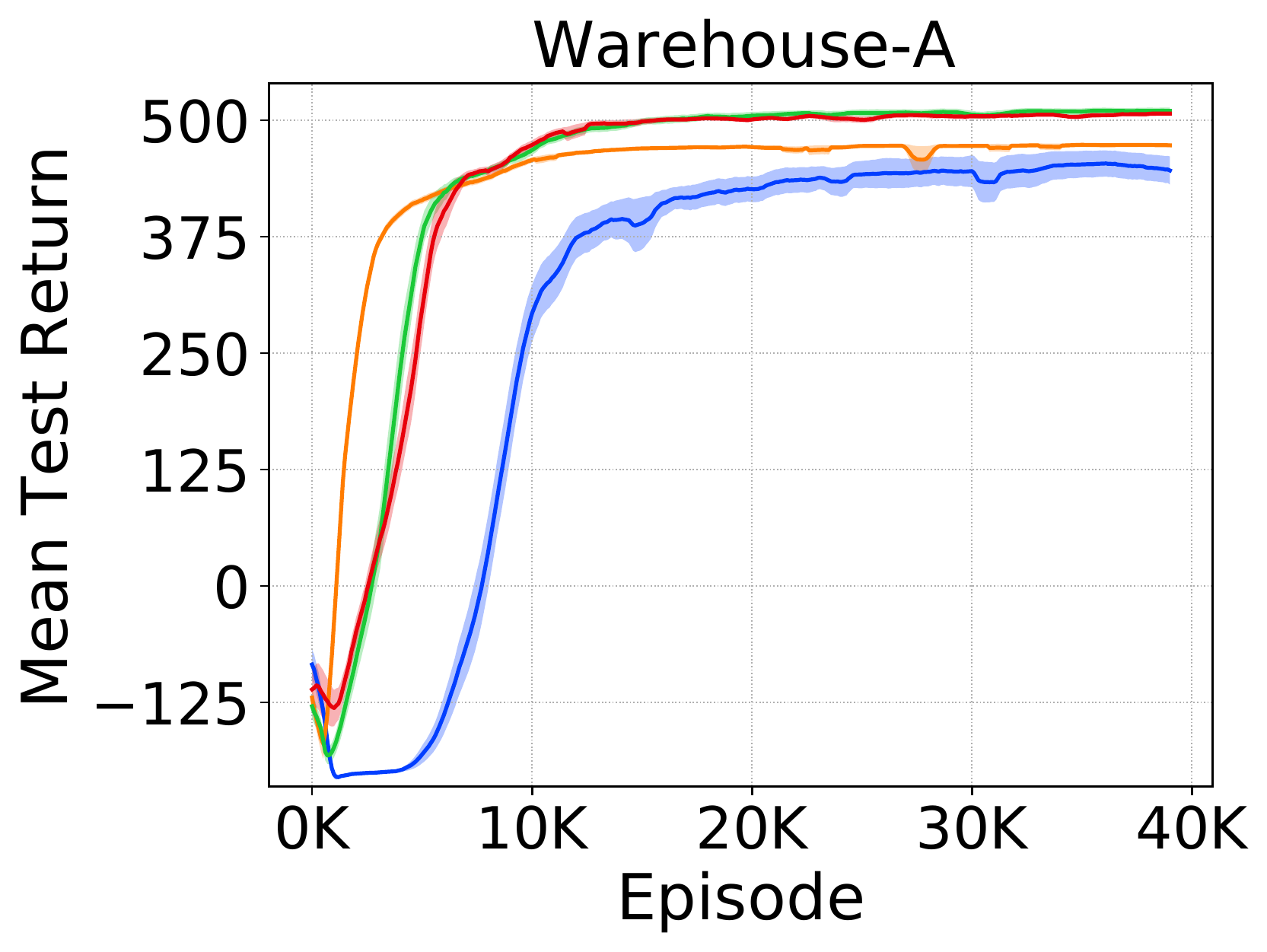}}
    ~    
    \centering
    \subcaptionbox{}
        [0.23\linewidth]{\includegraphics[height=2.5cm]{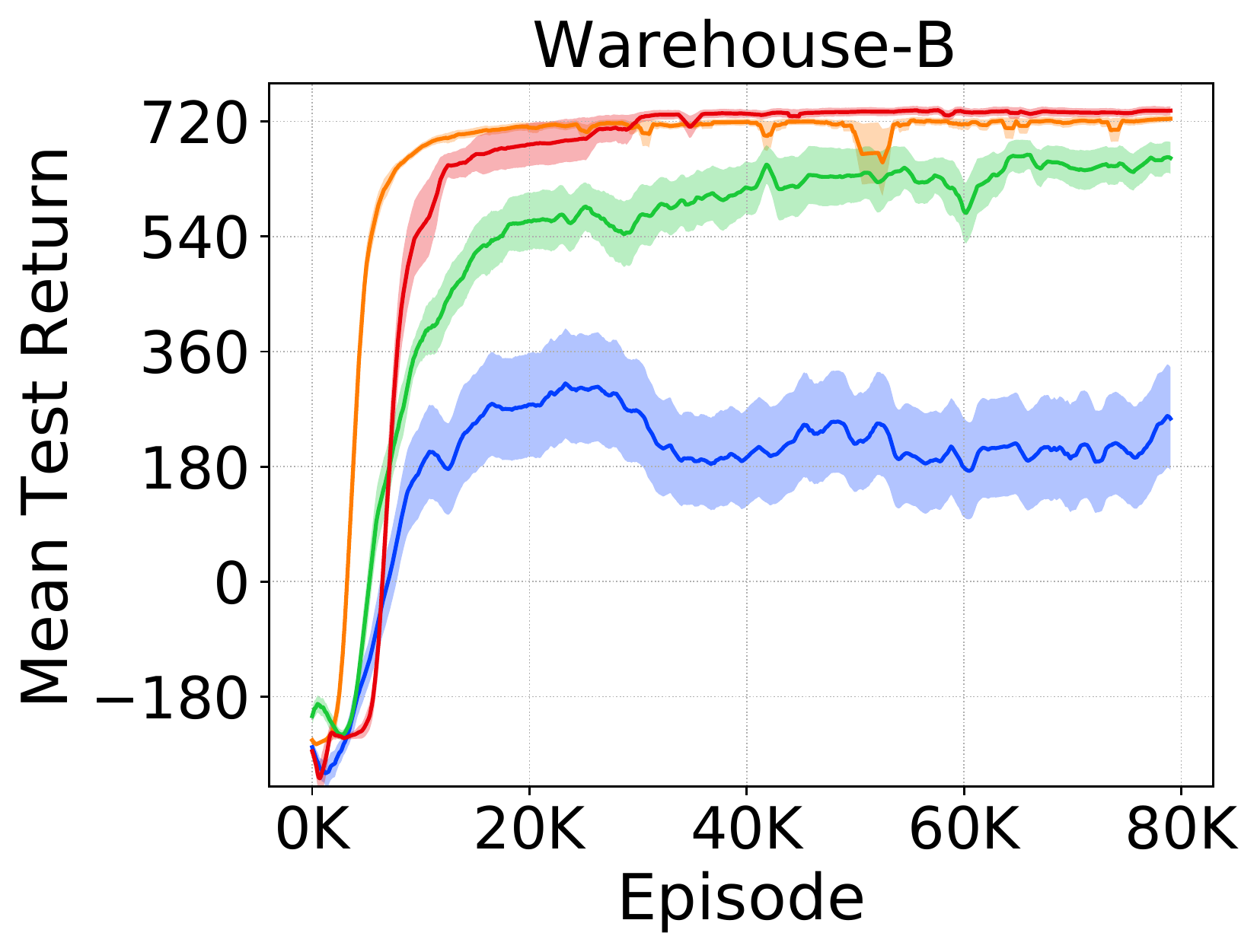}}
    ~
    \centering
    \subcaptionbox{}
        [0.23\linewidth]{\includegraphics[height=2.5cm]{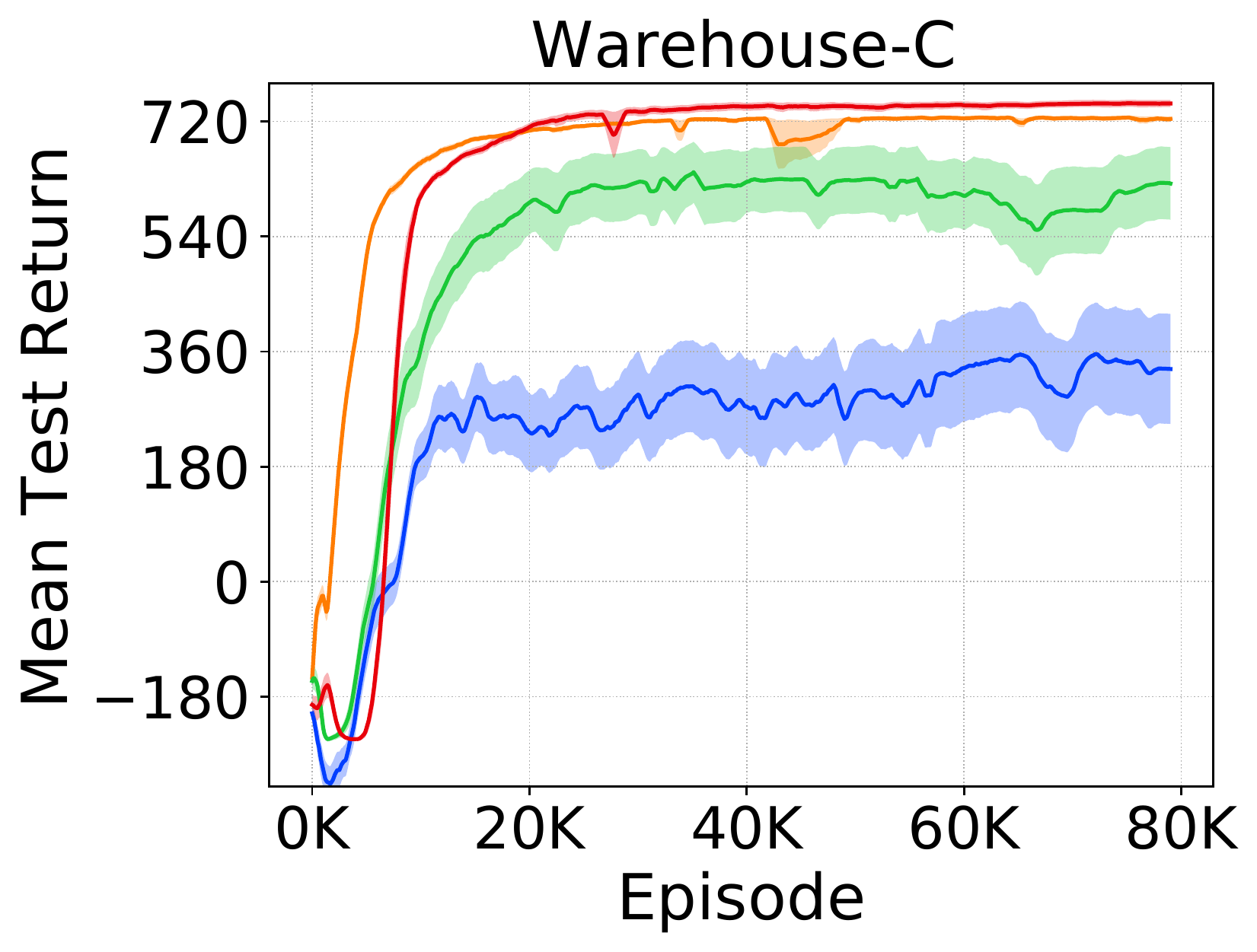}}
    ~
    \centering
    \subcaptionbox{}
        [0.23\linewidth]{\includegraphics[height=2.5cm]{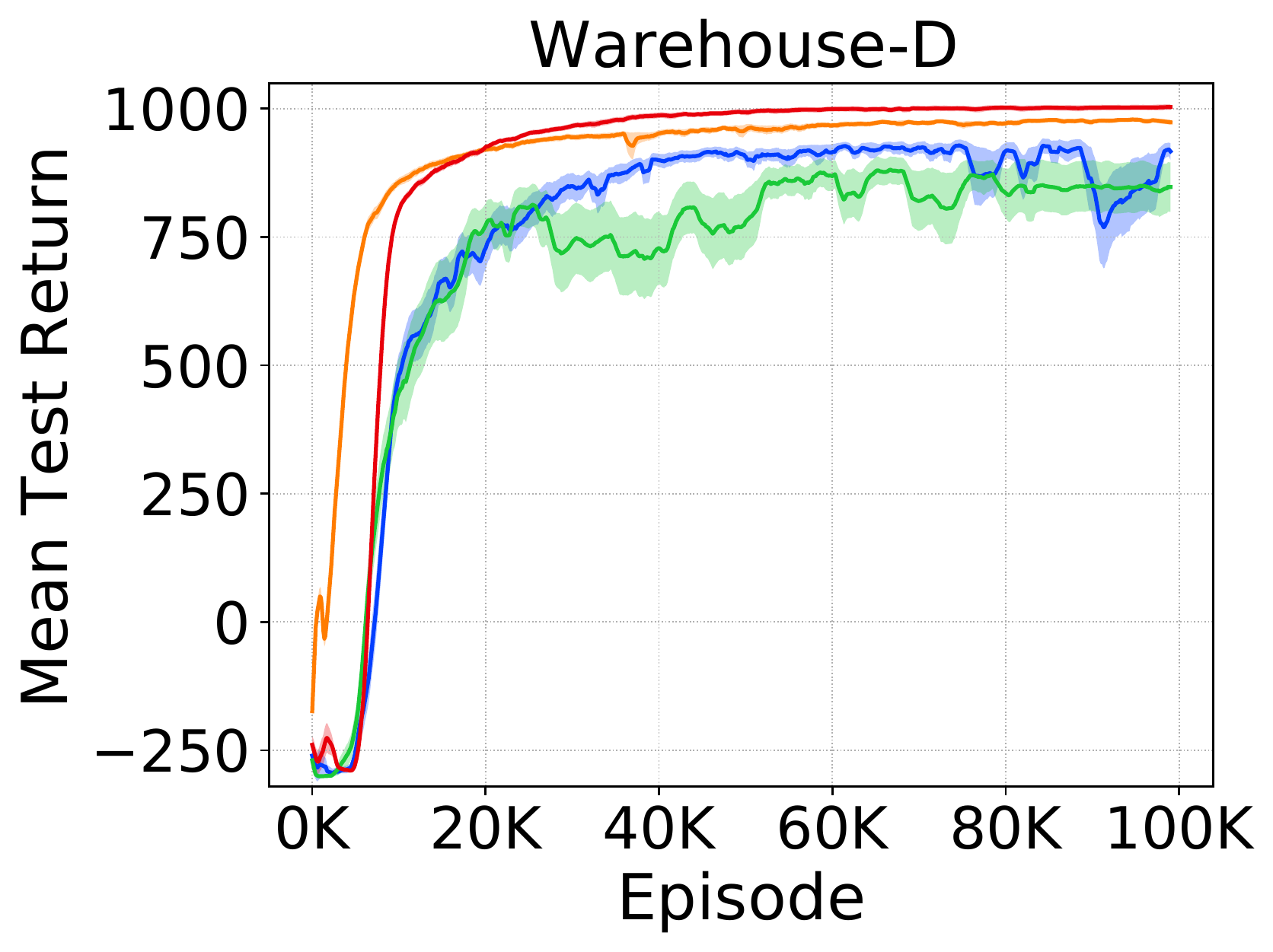}}
    \caption{Comparison of macro-action-based asynchronous actor-critic methods.}
    \label{mac_ctde}
\end{figure*}

\textbf{Advantages of having individual centralized critics.} Fig.~\ref{mac_ctde} shows the evaluation of our methods in all three domains. 
As each agent's observation is extremely limited in Box Pushing, we allow centralized critics in both Mac-IAICC and Naive Mac-IACC to access the state (agents' poses and boxes' positions), but use the joint macro-observation-action history in the other two domains.

In the Box Pushing task (the left two in the top row in Fig.~\ref{mac_ctde}), Naive Mac-IACC (green) can learn policies almost as good as the ones for Mac-IAICC (red) for the smaller domain, but as the grid world size grows, Naive Mac-IACC performs poorly while Mac-IAICC keeps its performance near the centralized approach. 
From each agent's perspective, the bigger the world size is, the more time steps a macro-action could take, and the less accurate the critic of Naive Mac-IACC becomes since it is trained depending on any agent's macro-action termination. Conversely, Mac-IAICC gives each agent a separate centralized critic trained with the reward associated with its own macro-action execution. 

In Overcooked-A (the third one at the top row in Fig.~\ref{mac_ctde}), as Mac-IAICC's performance is determined by the training of three agents' critics, it learns slower than Naive Mac-IACC in the early stage but converges to a slightly higher value and has better learning stability than Naive Mac-IACC in the end. 
The result of scenario B (the last one at the top row in Fig.~\ref{mac_ctde}) shows that Mac-IAICC outperforms other methods in terms of achieving  better sample efficiency, a higher final return and a lower variance. 
The middle wall in scenario B limits each agent's moving space and leads to a higher frequency of  macro-action terminations. The shared centralized critic in Naive Mac-IACC thus provides more noisy value estimations for each agent's actions. Because of this, Naive Mac-IACC performs worse with more variance. Mac-IAICC, however, does not get hurt by such environmental dynamics change. 
Both Mac-CAC and Mac-IAC are not competitive with Mac-IAICC in this domain.  

In the Warehouse scenarios (the bottom row in Fig.~\ref{mac_ctde}), Mac-IAC (blue) performs the worst due to its natural limitations and the domain's partial observability. 
In particular, it is difficult for the gray robot (arm) to learn an efficient way to find the correct tools purely based on local information and very delayed rewards that depend on the mobile robots' behaviors. In contrast, in the fully centralized Mac-CAC (orange), both the actor and the critic have global information so  it can learn faster in the early training stage. However, Mac-CAC eventually gets stuck at a local-optimum in all five scenarios due to the exponential dimensionality of joint history and action spaces over robots.   
By leveraging the CTDE paradigm, both Mac-IAICC and Naive Mac-IACC perform the best in warehouse A. Yet, the weakness of Naive Mac-IACC is clearly exposed when the problem is scaled up in Warehouse B, C and D. In these larger cases, the robots' asynchronous macro-action executions (e.g., traveling between rooms) become more complex and cause more mismatching between the termination from each agent's local perspective and the termination from the centralized perspective, and therefore, Naive Mac-IACC's performance significantly deteriorates, even getting worse than Mac-IAC in Warehouse-D. 
In contrast, Mac-IAICC can maintain its outstanding performance, converging to a higher value with much lower variance, compared to other methods. 
This outcome confirms not only Mac-IAICC's scalability but also the effectiveness of having an individual critic for each agent to handle variable degrees of asynchronicity in agents' high-level decision-making. 


\begin{figure}[h!]
    \centering
    \captionsetup[subfigure]{labelformat=empty}
    \centering
    \subcaptionbox{}
        [0.9\linewidth]{\includegraphics[scale=0.14]{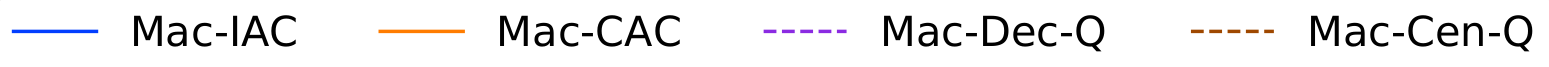}}
    ~
    \centering
    \subcaptionbox{}
        [0.23\linewidth]{\includegraphics[height=2.5cm]{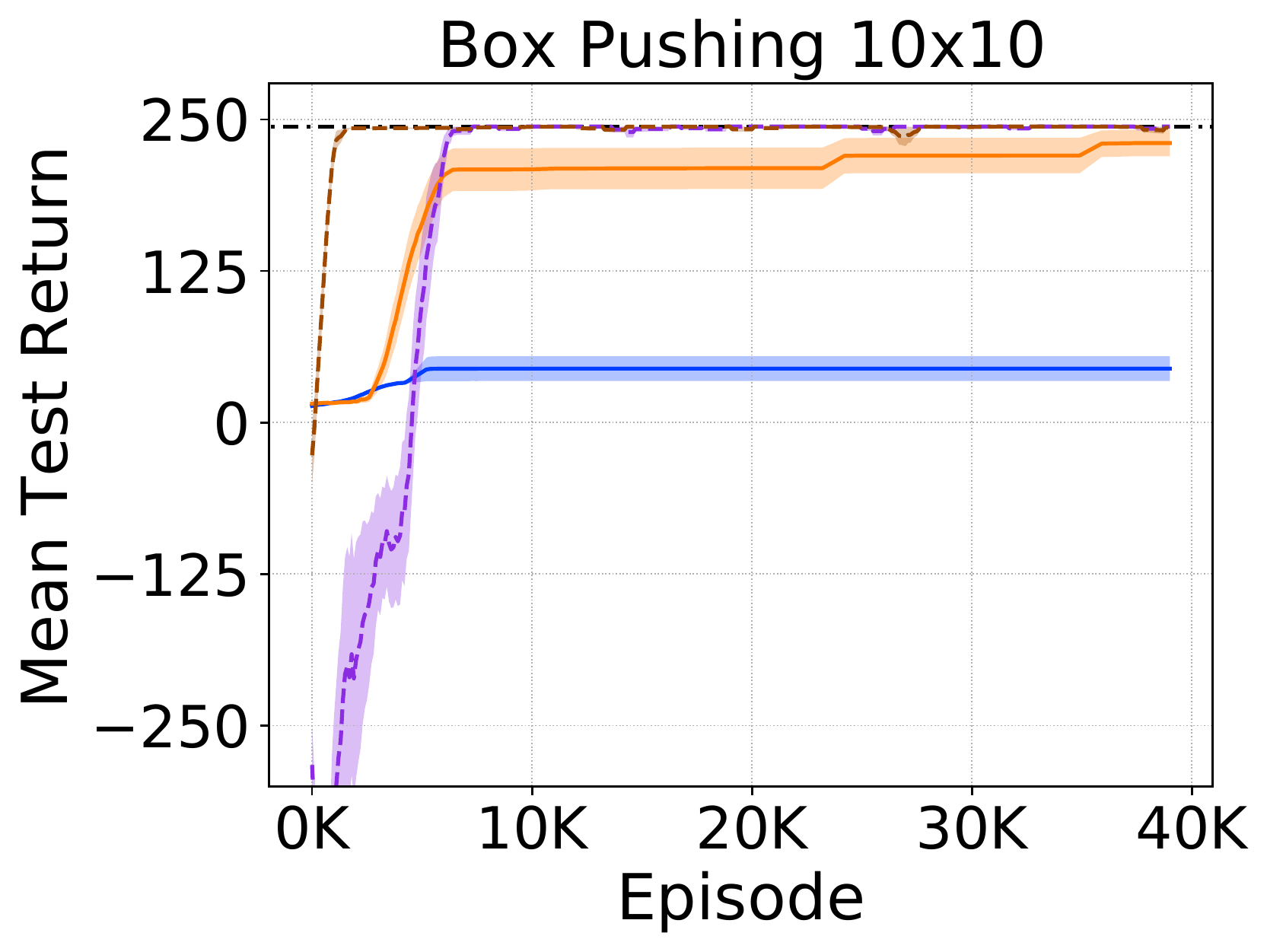}}
    ~
    \centering
    \subcaptionbox{}
        [0.23\linewidth]{\includegraphics[height=2.5cm]{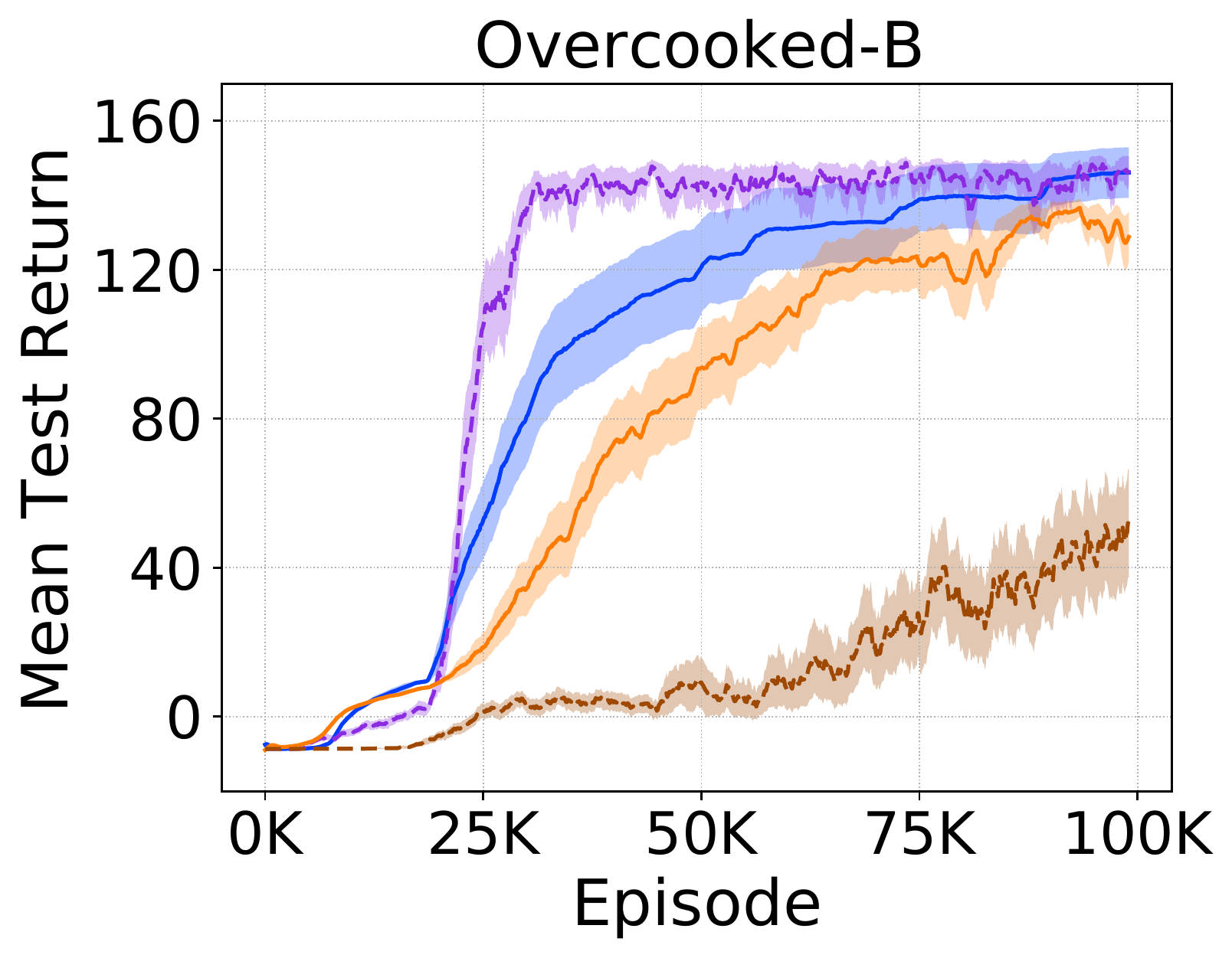}}
    ~
    \centering
    \subcaptionbox{}
        [0.23\linewidth]{\includegraphics[height=2.5cm]{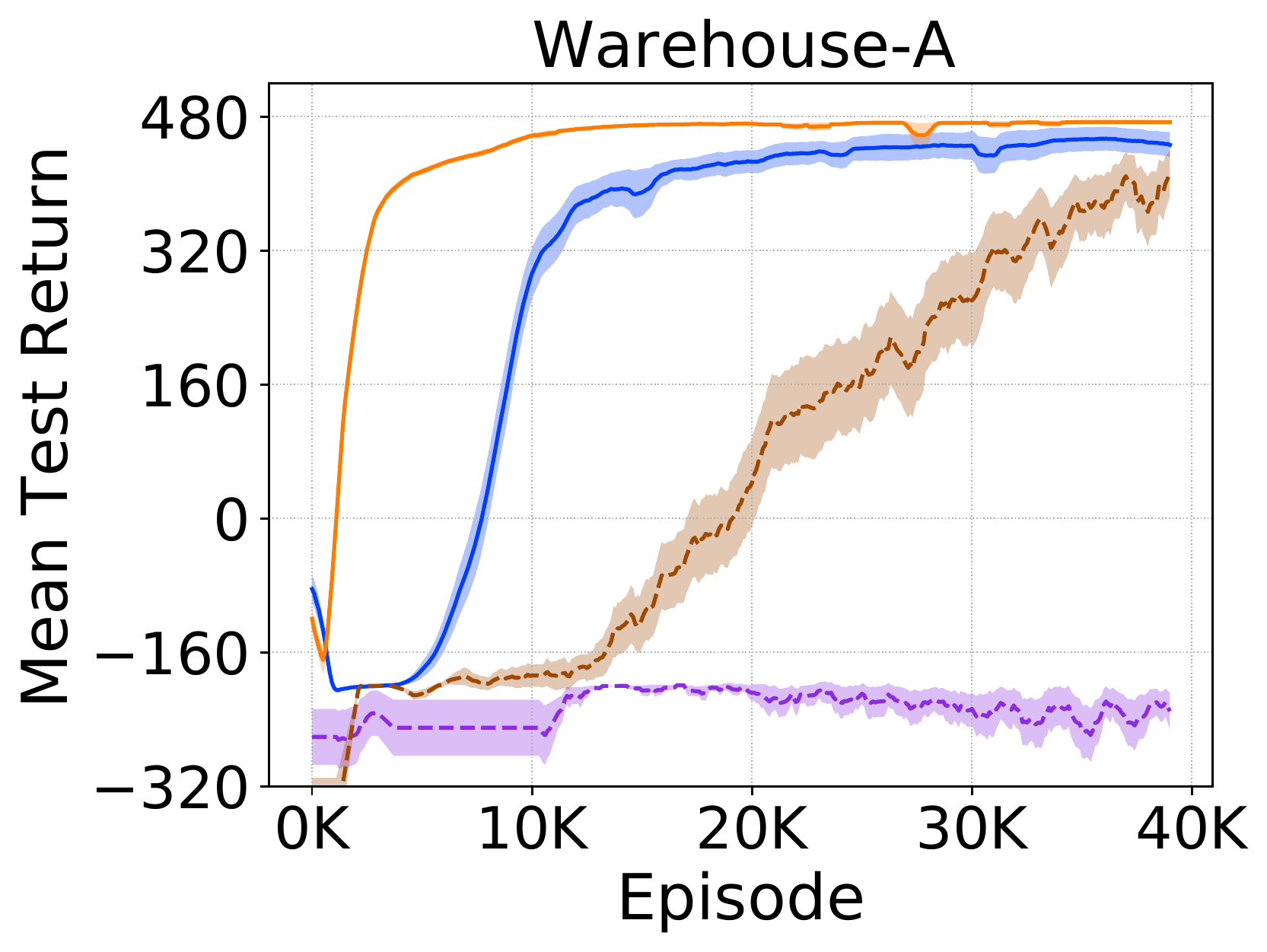}}
    ~
    \centering
    \subcaptionbox{}
        [0.23\linewidth]{\includegraphics[height=2.5cm]{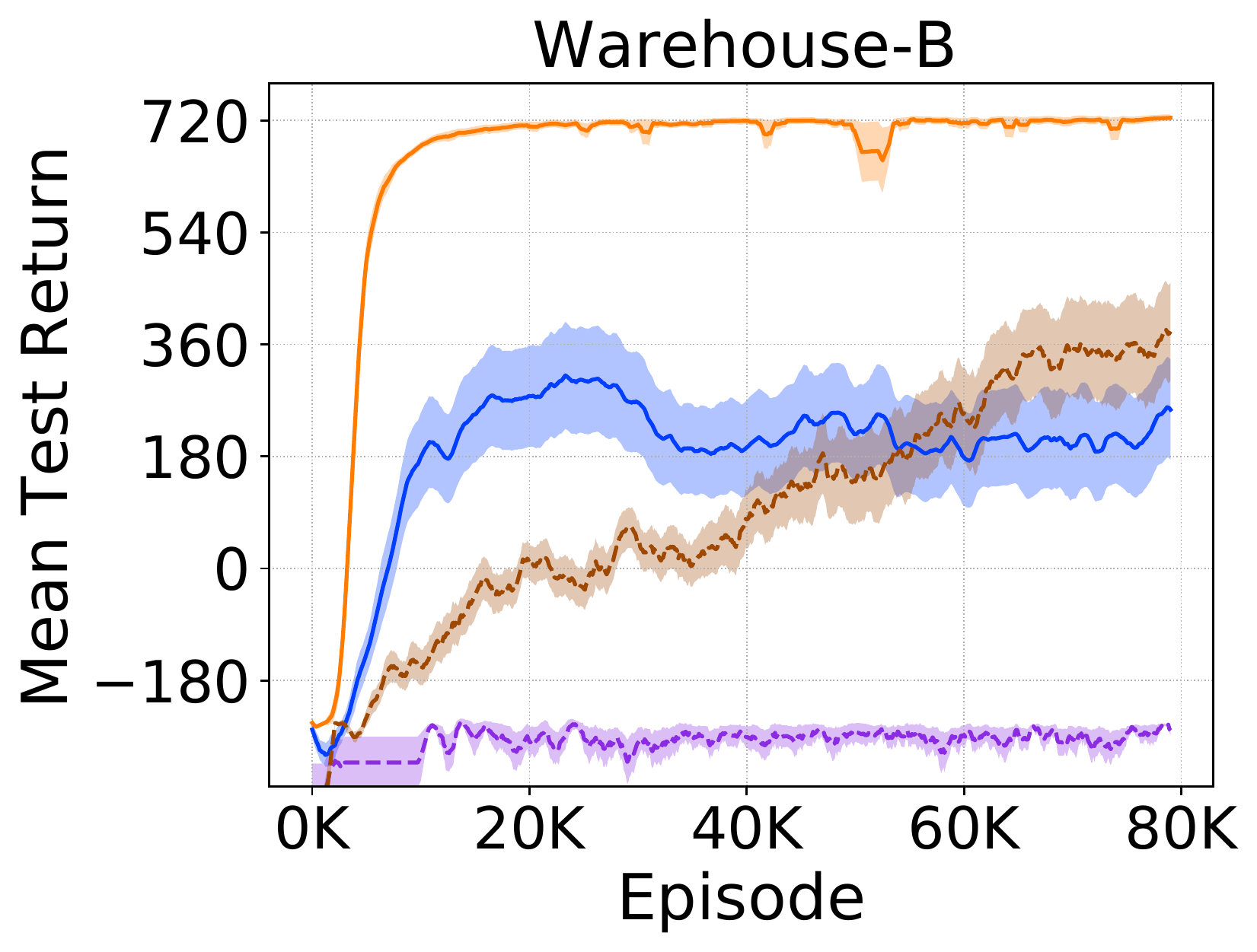}}
    \caption{Comparisons of macro-action-based actor-critic methods and value-based methods.}
    \label{pg_vs_ac}
\end{figure}

\textbf{Comparative analysis between actor-critic and value-based approaches}.
We also compare our actor-critic methods (Mac-IAC and Mac-CAC) with the current state-of-the-art asynchronous decentralized and centralized MARL methods, the value-based approaches (Mac-Dec-Q and Mac-Cen-Q)~\citep{xiao_corl_2019}, shown in Fig.~\ref{pg_vs_ac}. 
The Box Pushing task requires agents to simultaneously reach the big box and push it together. 
This consensus is rarely achieved when agents independently sample actions using stochastic policies in Mac-IAC and is hard to learn from pure on-policy data. 
By having a replay-buffer, value-based approaches show much stronger sample efficiency than on-policy actor-critic approaches in this domain with a small action space (left figure).   
Such an advantage is sustained by the decentralized value-based method (Mac-Dec-Q) but gets lost in the centralized one (Mac-Cen-Q) in the Overcooked domains due to a huge joint macro-action space ($15^3$).
On the contrary, our actor-critic methods can scale to large domains and learn high-quality solutions. This is particularly noticeable on Warehouse-A, where the policy gradient methods quickly learn a high-quality policy while the centralized Mac-Cen-Q is slow to learn and the decentralized Mac-Dec-Q is unable to learn. 
In addition, the stochastic policies in actor-critic methods potentially have better exploration property so that, in Warehouse domains, Mac-IAC can bypass an obvious local-optima that Mac-Dec-Q falls into, where the robot arm greedily chooses \emph{\textbf{Wait-M}} to avoid more penalties. 
\section{Hardware Experiments}

\begin{figure*}[h!]
    \centering
    \captionsetup[subfigure]{labelformat=empty}
    \centering
    \subcaptionbox{(a)\label{real_a}}
        [0.23\linewidth]{\includegraphics[height=1.8cm]{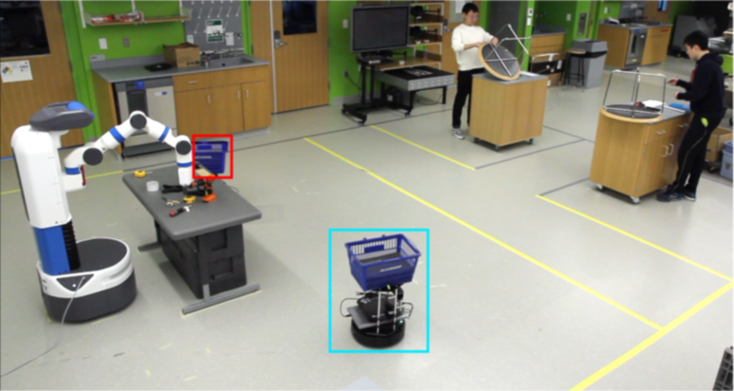}}
    ~
    \centering
    \subcaptionbox{(b)\label{real_b}}
        [0.23\linewidth]{\includegraphics[height=1.8cm]{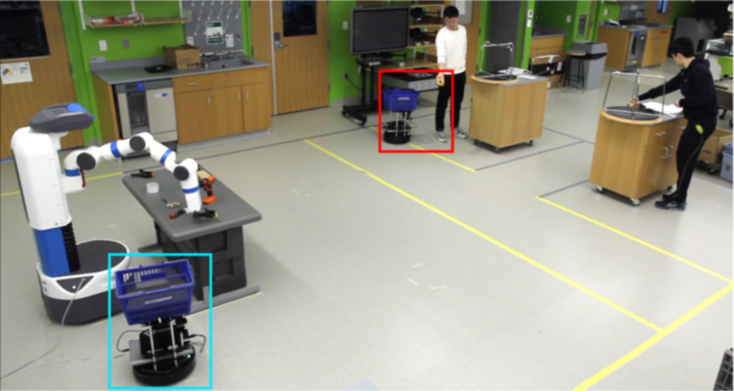}}
    ~
    \centering
    \subcaptionbox{(c)\label{real_c}}
        [0.23\linewidth]{\includegraphics[height=1.8cm]{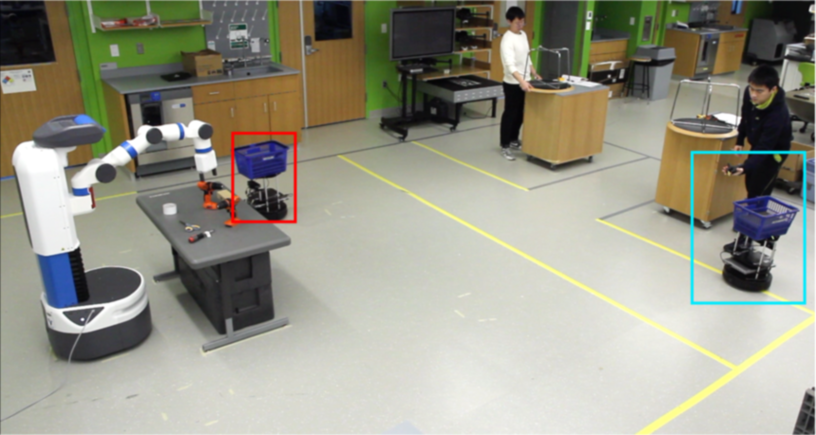}}
    ~
    \centering
    \subcaptionbox{(d)\label{real_d}}
        [0.23\linewidth]{\includegraphics[height=1.8cm]{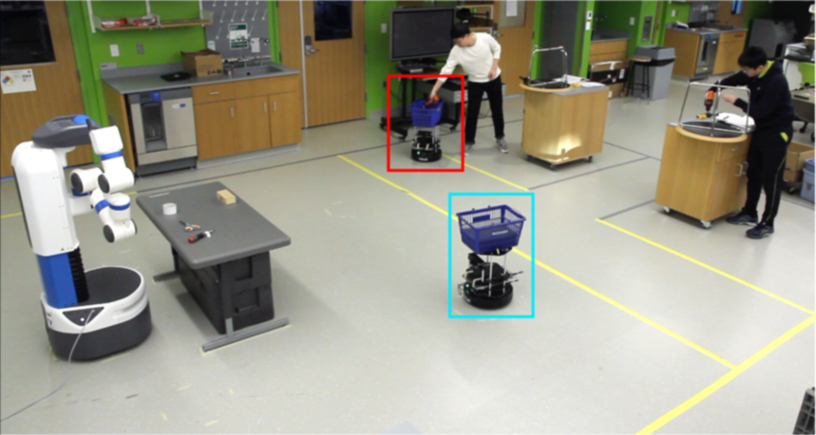}}
    \caption{Collaborative behaviors generated by running the decentralized policies learned by Mac-IAICC where Turtlebot-0 (T-0) is bounded in red and Turtlebot-1 (T-1) is bounded in blue. (a) After staging a tape measure at the left, Fetch looks for the 2nd one while Turtlebots approach the table; (b) T-0 deliveries a tap measure to W-0 and T-1 waits for a clamp from Fetch; (c)  T-1 deliveries a clamp to W-1, while T-0 carries the other clamp and goes to W-0, and Fetch searches for an electric drill; (d) T-0 deliveries an electric drill (the last tool) to W-0 and the entire delivery task is completed.}
    \label{real_exp}
\end{figure*}
We also extend scenario A of the Warehouse Tool Delivery task to a hardware domain (details of experimental setup are referred to Appendix~\ref{HE}). 
Fig.~\ref{real_exp} shows the sequential collaborative behaviors of the robots in one hardware trial. 
Fetch was able to find tools in parallel such that two tape measures (Fig.~\ref{real_a}), two clamps (Fig.~\ref{real_b}) and two electric drills, were found instead of finding all three types of tool for one human and then moving on to the other which would result in one of the humans waiting. 
Fetch's efficiency is also reflected in the behaviors such that it passed a tool to the Turtelbot who arrived first (Fig.~\ref{real_b}) and continued to find the next tool when there was no Turtlebot waiting beside it (Fig.~\ref{real_c}). 
Meanwhile, Turtlebots were clever such that they successfully avoid delayed delivery by sending tools one by one to the nearby workshop (e.g., T-0 focused on W-0 shown in Fig.~\ref{real_b} and \ref{real_d}, and T-1 focused on W-1 shown in Fig.~\ref{real_c}), rather than waiting for all tools before delivering, traveling a longer distance to serve the human at the diagonal, or prioritizing one of the humans altogether.       

\section{Conclusion} 
\label{sec:conclusion}

This paper introduces a general formulation for asynchronous multi-agent macro-action-based policy gradients under partial observability along with proposing a decentralized actor-critic method (Mac-IAC), a centralized actor-critic method (Mac-CAC), and two CTDE-based actor-critic methods (Naive Mac-IACC and Mac-IAICC). 
 These are the first approaches to be able to incorporate controllers that may require different amounts of time to complete (macro-actions) in a general asynchronous multi-agent actor-critic framework. 
Empirically, our methods are able to learn high-quality macro-action-based policies allowing agents to perform asynchronous collaborations in large and long-horizon problems. Importantly, our most advanced method, Mac-IAICC, allows agents to have individual centralized critics tailored to the agent's own macro-action execution. 
Additionally, the practicality of our approach is validated in a real-world multi-robot setup based on a warehouse domain.
This work provides a foundation for future macro-action-based MARL algorithm development, including other policy gradient-based methods as well as methods which also learn the macro-actions.         

\section*{Acknowledgments}

We thank Chengguang Xu and Tian Xia for their participation in hardware experiments. This research is supported in part
by the U.S. Office of Naval Research under award number
N00014-19-1-2131, Army Research Office award W911NF20-1-0265 and NSF CAREER Award 2044993.

\bibliographystyle{unsrtnat}
\bibliography{references}

\clearpage
\appendix


\section{Related Work}

MARL has been used for solving multi-robot problems~\cite{AlonMARL,KhanTR019,MitchellFPP20,CoRL20-Park,AB05,StricklandCV19,TangICCV2019} and hierarchy has also been introduced into multi-robot scenarios~\cite{Chen14_Masters,2018Multi,Oliehoek06HAAMAS,Omidshafiei17_macobs_ICRA,AAMAS17-Zhang}, but hierarchical MARL is still novel for multi-robot systems~\cite{NachumAPGK19, WangK0LZITF20, wu2021spatial, xiao_corl_2019}.

One line of hierarchical MARL is still focusing on learning primitive-action-based policy for each agent, while leveraging a hierarchical structure to achieve knowledge transfer~\cite{tianpei:neurips21}, credit assignment~\cite{AhilanFedual} and low-level policy factorization over agents~\cite{OptionResq}.   
In these works, as the decision-making over agents is still limited at a single low-level, none of them has been evaluated in large-scale realistic domains. 
Instead, by having macro-actions, our methods equip agents with the potential capability of exploiting abstracted skills, sub-task allocation and problem decomposition via hierarchical decision-making, which is critical for scaling up to real-world multi-robot tasks. 

Another line of the research allow agents to learn both a high-level policy and a low-level policy, but the  methods either force agents to perform a high-level choice at every time step~\cite{schroeder:nips19,MAIntro-OptionQ} or require all agents' high-level decisions
have the same time duration~\cite{NachumAPGK19,WangK0LZITF20,wang:iclr2021,HAVEN,JiachenMAHRL}, where agents are actually synchronized at both levels. In contrast, our frameworks are more general and applicable to real-world multi-robot systems because they allow agents to asynchronously execute at a high-level without synchronization or waiting for all agents to terminate. 

Recently, some asynchronous hierarchical approaches have been developed. \cite{xiao_corl_2019} and \cite{wu2021spatial} extend DQN~\cite{DQN} to learn macro-action-value functions and spatial-action-value maps for agents respectively. Our work, however, focuses on policy gradient algorithms that have different theoretical properties than value-based approaches (e.g., our methods are more scalable in the action space). Both classes of methods can co-exit and fit well with different sets of tasks. 
\cite{MendaCGBTKW19} frame multi-agent asynchronous decision-making problems as event-driven processes with one assumption on the acceptable of losing the ability to capture low-level interaction between agents within an event duration and the other on homogeneous agents, but our frameworks rely on the time-driven simulator used for general multi-agent and single-agent RL problems and do not have the above assumptions. 
\cite{DOC} adapt a single-agent~\emph{option-critic} framework~\cite{OptionAC} to multi-agent domains to learn all components (e.g., low-level policy, high-level abstraction, high-level policy) from scratch, but learning at both levels is difficult and the proposed method does not perform well even in small TeamGrid~\cite{TeamGrid} scenarios. More important to note is that 
none of the existing works provides a principled way for directly optimizing parameterized macro-action-based policies via asynchronous policy gradients to solve general multi-agent problems with macro-actions, and our work in this paper seeks to fill this gap.
\newpage
\section{Macro-Action-Based Policy Gradient Theorem}
\label{MacPGT}

As POMDPs can always be transformed to history-based MDPs, we can directly adapt the general Bellman equation for the state values of a hierarchical policy~\citep{Sutton:1999} to a macro-action-based POMDP by replacing the state $s$ with a history $h$ as follows (for keeping the notaion simple, we use $\tau$ to represent the number of timesteps taken by the corresponding macro-action $m$, and we use $h$ to represent macro-observation-action history):
\begin{equation}
    V^\Psi(h) = \sum_{m}\Psi(m|h)Q^{\Psi}(h,m) 
\end{equation}
\begin{equation}
    Q^\Psi(h,m) = r^c(h,m) + \sum_{h'}P(h'| h, m)V^{\Psi}(h') 
\end{equation}
where, 
\begin{equation}
    r^c(h,m)=\mathbb{E}_{\tau\sim\beta_m,s_{t_m}| h}\Big[\sum_{t=t_m}^{t_m+\tau-1}\gamma^tr_t\Big]
\end{equation}
\begin{align}
    P(h'|h,m) = P(z'|h,m) &= \sum_{\tau=1}^\infty\gamma^\tau P(z', \tau | h,m) \\ 
    & = \sum_{\tau=1}^\infty\gamma^\tau P(\tau | h,m)P(z'| h, m, \tau) \\ 
    & = \sum_{\tau=1}^\infty\gamma^\tau P(\tau | h,m)P(z'| h, m, \tau) \\ 
    & = \mathbb{E}_{\tau\sim\beta_m}\Big[\gamma^\tau\mathbb{E}_{s|h}\big[\mathbb{E}_{s' | s,m,\tau}[P(z'|m,s')]\big]\Big] 
\end{align}
Next, we follow the proof of the policy gradient theorem~\citep{sutton2000policy}:
\begin{align}
    \nabla_\theta V^{\Psi_\theta}(h)&=\nabla_\theta\Biggr[\sum_{m}\Psi_{\theta}(m| h)Q^{\Psi_\theta}(h,m)\Biggr] \\
    &=\sum_m\Big[\nabla_{\theta}\Psi_\theta(m| h)Q^{\Psi_\theta}(h,m) + \Psi_{\theta}(m| h)\nabla_{\theta}Q^{\Psi_\theta}(h,m)\Big]\\
    &=\sum_m\Big[\nabla_{\theta}\Psi_\theta(m| h)Q^{\Psi_\theta}(h,m) + \Psi_{\theta}(m| h)\nabla_{\theta}\big(r^c(h,m) +\sum_{h'}P(h'| h, m)V^{\Psi_\theta}(h')\big)\Big]\\
    &=\sum_m\Big[\nabla_{\theta}\Psi_\theta(m| h)Q^{\Psi_\theta}(h,m) + \Psi_{\theta}(m| h)\sum_{h'}P(h'| h, m)\nabla_{\theta}V^{\Psi_\theta}(h')\big)\Big]\\
    &=\sum_{\hat{h}\in H}\sum_{k=0}^\infty P(h\rightarrow \hat{h}, k, \Psi_\theta)\sum_{m}\nabla_\theta\Psi_\theta(m|\hat{h})Q^{\Psi_\theta}(\hat{h},m)\,\,\,\,(\text{after repeated unrolling})
\end{align}
Then, we can have:
\begin{align}
    \nabla_\theta J(\theta) &= \nabla_\theta V^{\Psi_\theta}(h_0)\\
    &=\sum_{h\in H}\sum_{k=0}^\infty P(h_0\rightarrow h, k, \Psi_\theta)\sum_{m}\nabla_\theta\Psi_\theta(m|h)Q^{\Psi_\theta}(h,m)\\
    &=\sum_{h}\rho^{\Psi_\theta}(h)\sum_{m}\nabla_\theta\Psi_\theta(m|h)Q^{\Psi_\theta}(h,m)\\
    &=\sum_{h}\rho^{\Psi_\theta}(h)\sum_{m}\Psi_\theta(m|h)\nabla_\theta\log\Psi_\theta(m|h)Q^{\Psi_\theta}(h,m)\\
    &=\mathbb{E}_{h\sim\rho^{\Psi_\theta}, m\sim\Psi_\theta}\Big[\nabla_\theta\log\Psi_\theta(m|h)Q^{\Psi_\theta}(h,m)\Big]
\end{align}

\newpage
\section{Asynchronous Acotr-Critic Algorithms}
\label{PCode}

In this section, we present the pesudo code of each proposed macro-action-based actor-critic algorithm with an example to show how the sequential experiences are squeezed for training the critic and the actor. We describe all methods in the on-policy learning manner while off-policy learning can be achieved by applying importance sampling weights and not resetting the buffer.\\ 

\noindent
\textbf{Macro-Action-Based Independent Actor-Critic (Mac-IAC):}

\begin{figure}[h!]
    \centering
    \includegraphics[height=4.9cm]{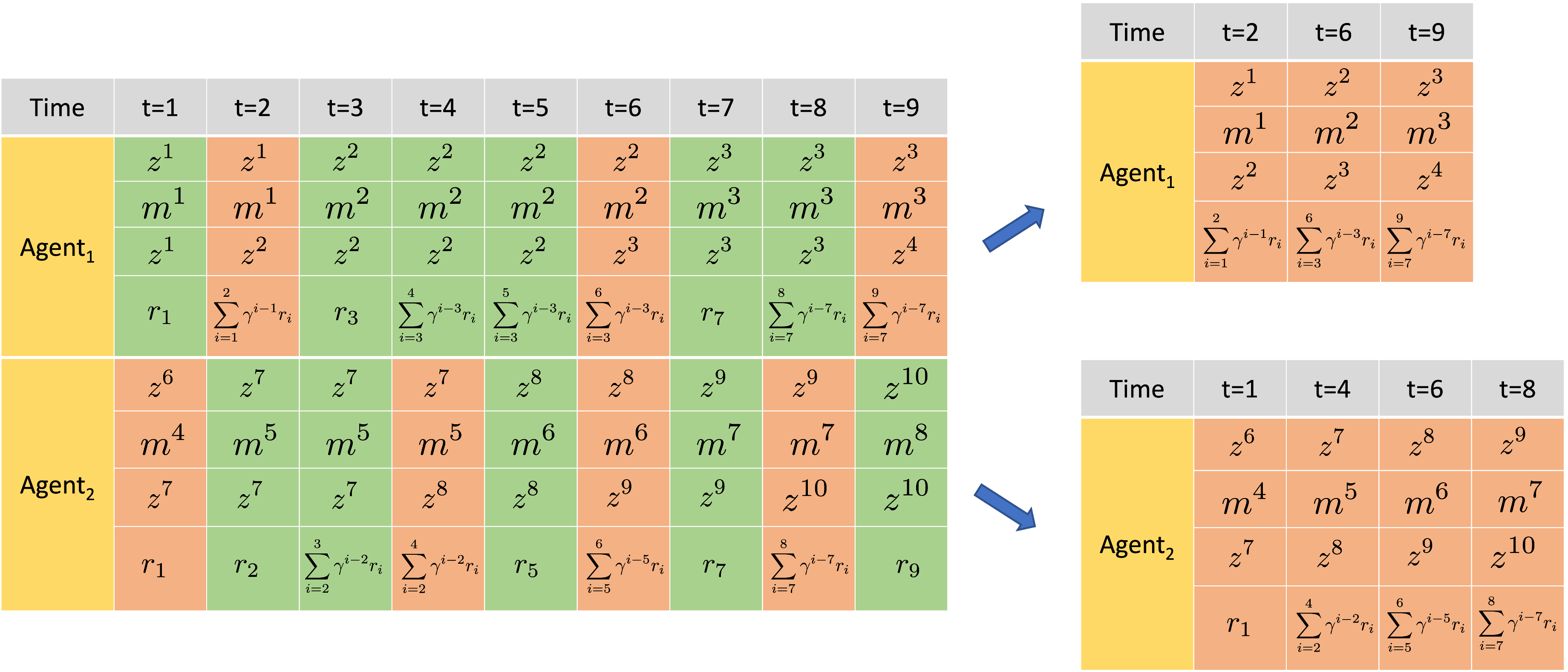}
    \caption{An example of the trajectory squeezing process in Mac-IAC. We collect each agent's high-level transition tuple at every primitive-step. Each agent is allowed to obtain a new macro-observation if and only if the current macro-action terminates, otherwise, the next macro-observation is set as same as the previous one. Each agent separately squeezes its sequential experiences by picking out the transitions when its macro-action terminates (red cells). Each agent independently train the critic and the policy using the squeezed trajectory.}
    \label{Mac-IAC-Traj}
\end{figure}
\vspace{-3mm}
\begin{algorithm}[h!]
    \footnotesize
    \caption{Mac-IAC}
    \label{alg1}
        \begin{algorithmic}[1]
            \State Initialize a decentralized policy network for each agent $i$: $\Psi_{\theta_i}$
            \State Initialize decentralized critic networks for each agent $i$: $V_{\mathbf{w}_i}^{\Psi_{\theta_i}}$, $V_{\mathbf{w}_i^-}^{\Psi_{\theta_i}}$ 
            \State Initialize a buffer $\mathcal{D}$
            \For{\emph{episode} = $1$ to $M$}
                \State $t=0$
                \State Reset env
                \While{not reaching a terminal state \textbf{and} $t < \mathbb{H}$}
                    \State $t \leftarrow t + 1$
                    \For{each agent $i$}
                        \If{the macro-action $m_i$ is terminated}
                            \State $m_{i} \sim \Psi_{\theta_i}(\cdot \mid h_i; \epsilon)$
                        \Else
                            \State Continue running current macro-action $m_i$
                        \EndIf
                    \EndFor
                    \For{each agent $i$}
                        \State Get cumulative reward $r^c_i$, next macro-observation $z'_i$ 
                        \State Collect $\langle z_i,m_i, z'_i, r^c_i \rangle$ into the buffer $\mathcal{D}$
                    \EndFor
                \EndWhile
                \If{\emph{episode} mod $I_{\text{train}} = 0$}
                    \For{each agent $i$}
                        \State Squeeze agent $i$'s trajectories in the buffer $\mathcal{D}$
                        \State Perform a gradient decent step on $L(\mathbf{w}_i)=\big(y-V^{\Psi_{\theta_i}}_{\mathbf{w}_i}(h_i)\big)^2_\mathcal{D}$, where  $y = r^c_i + \gamma^{\tau_{m_i}} V^{\Psi_{\theta_i}}_{\mathbf{w}_i^-}(h_i')$ 
                        \State Perform a gradient ascent on:
                        \State $\nabla_{\theta_i} J(\theta_i) = \mathbb{E}_{\vec{\Psi}_{\vec{\theta}}}\Big[\nabla_{\theta_i}\log\Psi_{\theta_i}(m_i|h_i)\big(r^c_i + \gamma^{\tau_{m_i}} V^{\Psi_{\theta_i}}_{\mathbf{w}_i^-}(h_i')-V^{\Psi_{\theta_i}}_{\mathbf{w}_i}(h_i)\big)\Big]$
                    \EndFor
                    \State Reset buffer $\mathcal{D}$
                \EndIf
                \If{\emph{episode} mod $I_{\text{TargetUpdate}} = 0$}
                    \For{each agent $i$}
                        \State Update the critic target network $\mathbf{w}_i^-\leftarrow\mathbf{w}_i$
                    \EndFor
                \EndIf
            \EndFor
        \end{algorithmic}
\end{algorithm}

\newpage
\noindent
\textbf{Macro-Action-Based Centralized Actor-Critic (Mac-CAC):}

\begin{figure}[h!]
    \centering
    \includegraphics[height=3.5cm]{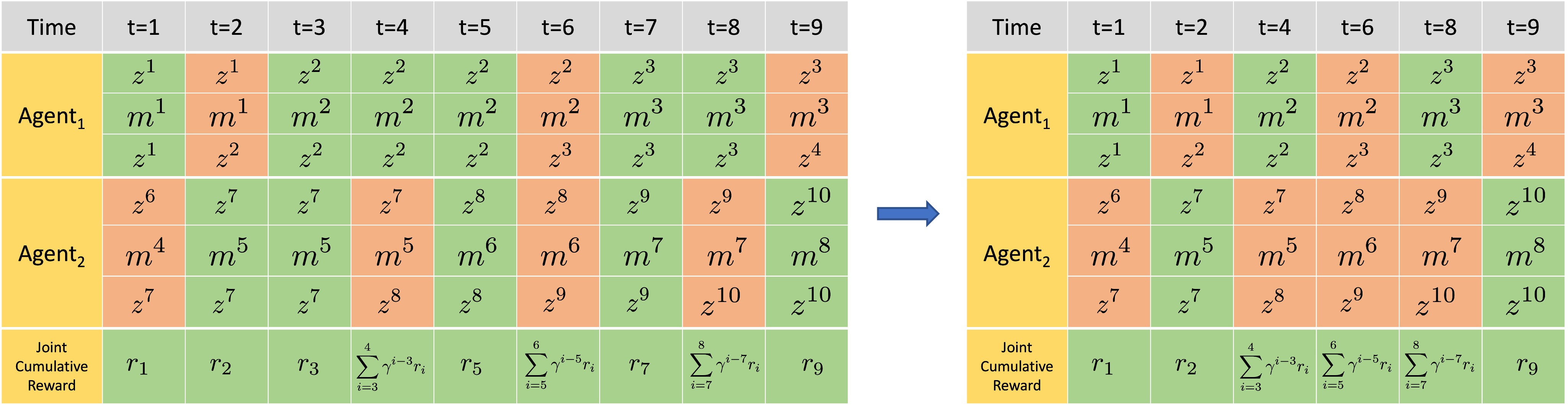}
    \caption{An example of the trajectory squeezing process in Mac-CAC. Joint sequential experiences are squeezed by picking out joint transition tuples when the joint macro-action terminates, in that, $any$ agent's macro-action termination (marked in red) ends the joint macro-action at the timestep. For example, at $t=1$, agents execute a joint macro-action $\vec{m}=\langle m^1, m^4\rangle$ for one timestep; at $t=2$, the joint macro-action becomes $\langle m^1,m^5\rangle$ as $\text{Agent}_2$ finished $m^4$ at last step and chooses a new macro-action $m^5$; $\text{Agent}_1$ finished its macro-action $m_1$ at $t=2$ and selects a new macro-action $m^2$ at $t=3$ so that the joint macro-action switches to $\langle m^2, m^5 \rangle$ which keeps running until the 4th timestep. Therefore, the first two joint macro-actions have two single-step reward respectively, and reward of joint macro-action $\langle m^2, m^5 \rangle$ is an accumulative reward over two consecutive timesteps.} 
    \label{Mac-CAC-Traj}
\end{figure}

\begin{algorithm}[h!]
    \footnotesize
    \caption{Mac-CAC}
    \label{alg2}
        \begin{algorithmic}[1]
            \State Initialize a centralized policy network: $\Psi_{\theta}$
            \State Initialize centralized critic networks: $V_{\mathbf{w}}^{\Psi_{\theta}}$, $V_{\mathbf{w}^-}^{\Psi_{\theta}}$ 
            \State Initialize a centralized buffer $\mathcal{D}\leftarrow\text{Mac-JERTs}$,
            \For{\emph{episode} = $1$ to $M$}
                \State $t=0$
                \State Reset env
                \While{not reaching a terminal state \textbf{and} $t < \mathbb{H}$}
                    \State $t \leftarrow t + 1$
                    \If{the joint macro-action $\vec{m}$ is terminated}
                        \State $\vec{m} \sim \Psi_{\theta}(\cdot \mid \vec{h}, \vec{m}^{\text{undone}}; \epsilon)$
                    \Else
                        \State Continue running current joint macro-action $\vec{m}$
                    \EndIf
                    \State Get a joint cumulative reward $\vec{r\,}^c$, next joint macro-observation $\vec{z\,}'$
                    \State Collect $\langle \vec{z},\vec{m}, \vec{z\,}', \vec{r\,}^c \rangle$ into the buffer $\mathcal{D}$
                \EndWhile
                \If{\emph{episode} mod $I_{\text{train}} = 0$}
                    \State Squeeze joint macro-level trajectories in the buffer $\mathcal{D}$ according to joint macro-action terminations
                    \State Perform a gradient decent step on $L(\mathbf{w})=\big(y-V^{\Psi_{\theta}}_{\mathbf{w}}(\vec{h})\big)^2_\mathcal{D}$, where  $y = \vec{r\,}^c + \gamma^{\vec{\tau}_{\vec{m}}} V^{\Psi_{\theta}}_{\mathbf{w}^-}(\vec{h}')$ 
                    \State Perform a gradient ascent on $\nabla_{\theta} J(\theta) = \mathbb{E}_{\Psi_\theta}\Big[\nabla_{\theta}\log\Psi_{\theta}(\vec{m}\mid \vec{h})\big(\vec{r\,}^c + \gamma^{\vec{\tau}_{\vec{m}}} V^{\Psi_{\theta}}_{\mathbf{w}^-}(\vec{h}')-V^{\Psi_{\theta}}_{\mathbf{w}}(\vec{h})\big)\Big]$
                    \State Reset buffer $\mathcal{D}$
                \EndIf
                \If{\emph{episode} mod $I_{\text{TargetUpdate}} = 0$}
                    \State Update the critic target network $\mathbf{w}^-\leftarrow\mathbf{w}$
                \EndIf
            \EndFor
        \end{algorithmic}
\end{algorithm}
where, $\vec{m}^\text{undone}$ is the sub-joint-macro-action over the agents who have not terminated their macro-actions and will continue running.

\clearpage
\noindent
\textbf{Naive Mac-IACC:}

In the pseudo code of Naive Mac-IACC presented below, we assume the accessible centralized information $\mathbf{x}$ is joint macro-observation-action history in the centralized critic. 

\begin{figure}[h!]
    \centering
    \includegraphics[height=5.2cm]{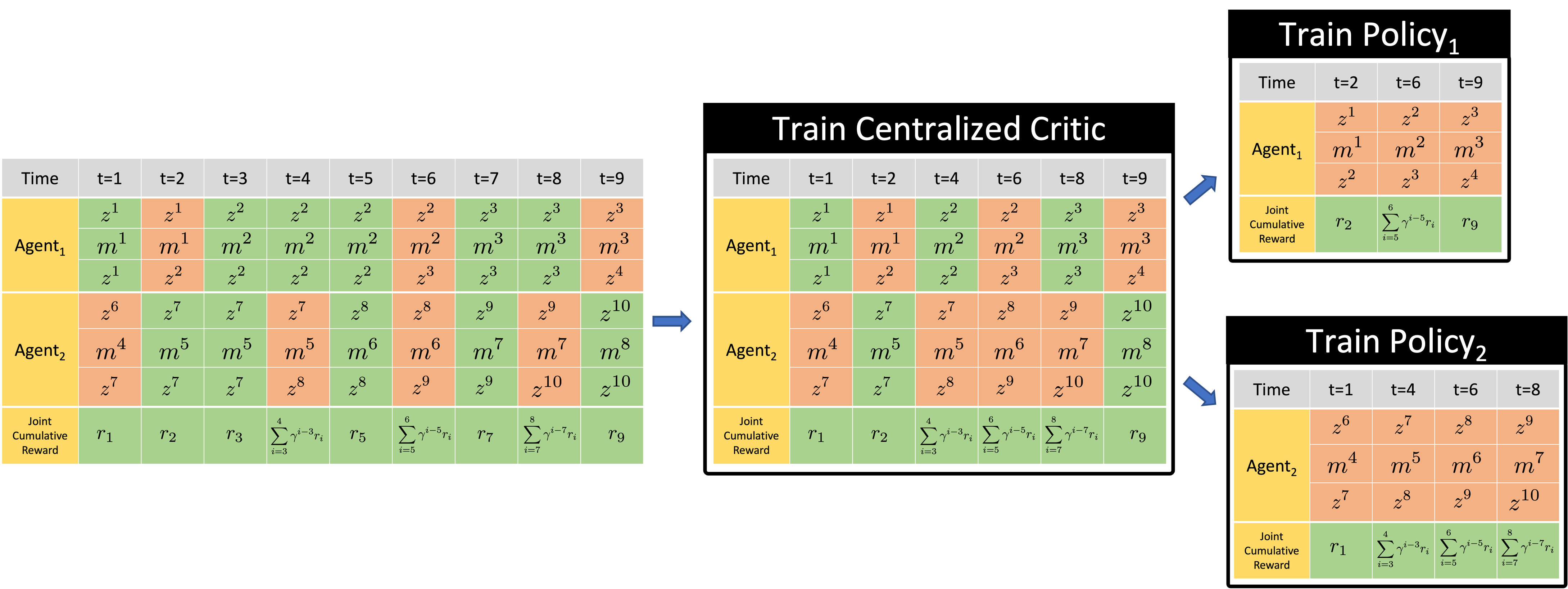}
    \caption{An example of the trajectory squeezing process in Navie Mac-IACC.The joint trajectory is first squeezed depending on joint macro-action termination for training the centralized critic (line 18-19 in Algorithm~\ref{alg3}). Then, the trajectory is further squeezed for each agent depending on each agent's own macro-action termination for training the decentralized policy (line 20-23 in Algorithm~\ref{alg3}).}
    \label{Mac-IAC-Traj}
\end{figure}

\begin{algorithm}[h!]
    \footnotesize
    \caption{Naive Mac-IACC}
    \label{alg3}
        \begin{algorithmic}[1]
            \State Initialize a decentralized policy network for each agent $i$: $\Psi_{\theta_i}$
            \State Initialize centralized critic networks: $V_{\mathbf{w}}^{\vec{\Psi}_{\vec{\theta}}}$, $V_{\mathbf{w}^-}^{\vec{\Psi}_{\vec{\theta}}}$ 
            \State Initialize a decentralized buffer $\mathcal{D}\leftarrow\text{Mac-JERTs}$,
            \For{\emph{episode} = $1$ to $M$}
                \State $t=0$
                \State Reset env
                \While{not reaching a terminal state \textbf{and} $t < \mathbb{H}$}
                    \State $t \leftarrow t + 1$
                    \For{each agent $i$}
                        \If{the macro-action $m_i$ is terminated}
                            \State $m_{i} \sim \Psi_{\theta_i}(\cdot \mid h_i; \epsilon)$
                        \Else
                            \State Continue running current macro-action $m_i$
                        \EndIf
                    \EndFor
                    \State Get a reward $\vec{r\,}^c$ accumulated based on current joint macro-action termination
                    \State Get next joint macro-observations $\vec{z\,}'$
                    \State Collect $\langle \vec{z},\vec{m}, \vec{z\,}', \vec{r\,}^c \rangle$ into the buffer $\mathcal{D}$
                \EndWhile
                \If{\emph{episode} mod $I_{\text{train}} = 0$}
                    \State Squeeze joint macro-level trajectories in the buffer $\mathcal{D}$ according to joint macro-action terminations
                    \State Perform a gradient decent step on $L(\mathbf{w})=\big(y-V^{\vec{\Psi}_{\vec{\theta}}}_{\mathbf{w}}(\vec{h})\big)^2_\mathcal{D}$, where  $y = \vec{r\,}^c + \gamma^{\vec{\tau}_{\vec{m}}} V^{\vec{\Psi}_{\vec{\theta}}}_{\mathbf{w}^-}(\vec{h}')$ 
                    \For{each agent $i$}
                        \State Squeeze agent $i$'s trajectories in the buffer $\mathcal{D}$ according to its own macro-action terminations
                        \State Perform a gradient ascent on: 
                        \State $\nabla_{\theta_i} J(\theta_i) = \mathbb{E}_{\vec{\Psi}_{\vec{\theta}}}\Big[\nabla_{\theta_i}\log\Psi_{\theta_i}(m_i|h_i)\big(\vec{r\,}^c + \gamma^{\vec{\tau}_{\vec{m}}} V^{\vec{\Psi}_{\vec{\theta}}}_{\mathbf{w}^-}(\vec{h}')-V^{\vec{\Psi}_{\vec{\theta}}}_{\mathbf{w}}(\vec{h})\big)\Big]$
                    \EndFor
                    \State Reset buffer $\mathcal{D}$
                \EndIf
                \If{\emph{episode} mod $I_{\text{TargetUpdate}} = 0$}
                    \State Update the critic target network $\mathbf{w}^-\leftarrow\mathbf{w}$
                \EndIf
            \EndFor
        \end{algorithmic}
\end{algorithm}

\clearpage
\noindent
\textbf{Macro-Action-Based Independent Actor with Individual Centralized Critic (Mac-IAICC):}

In the pseudo code of Mac-IAICC presented below, we assume the accessible centralized information $\mathbf{x}$ is joint macro-observation-action history in the centralized critic.

\begin{figure}[h!]
    \centering
    \includegraphics[height=7.4cm]{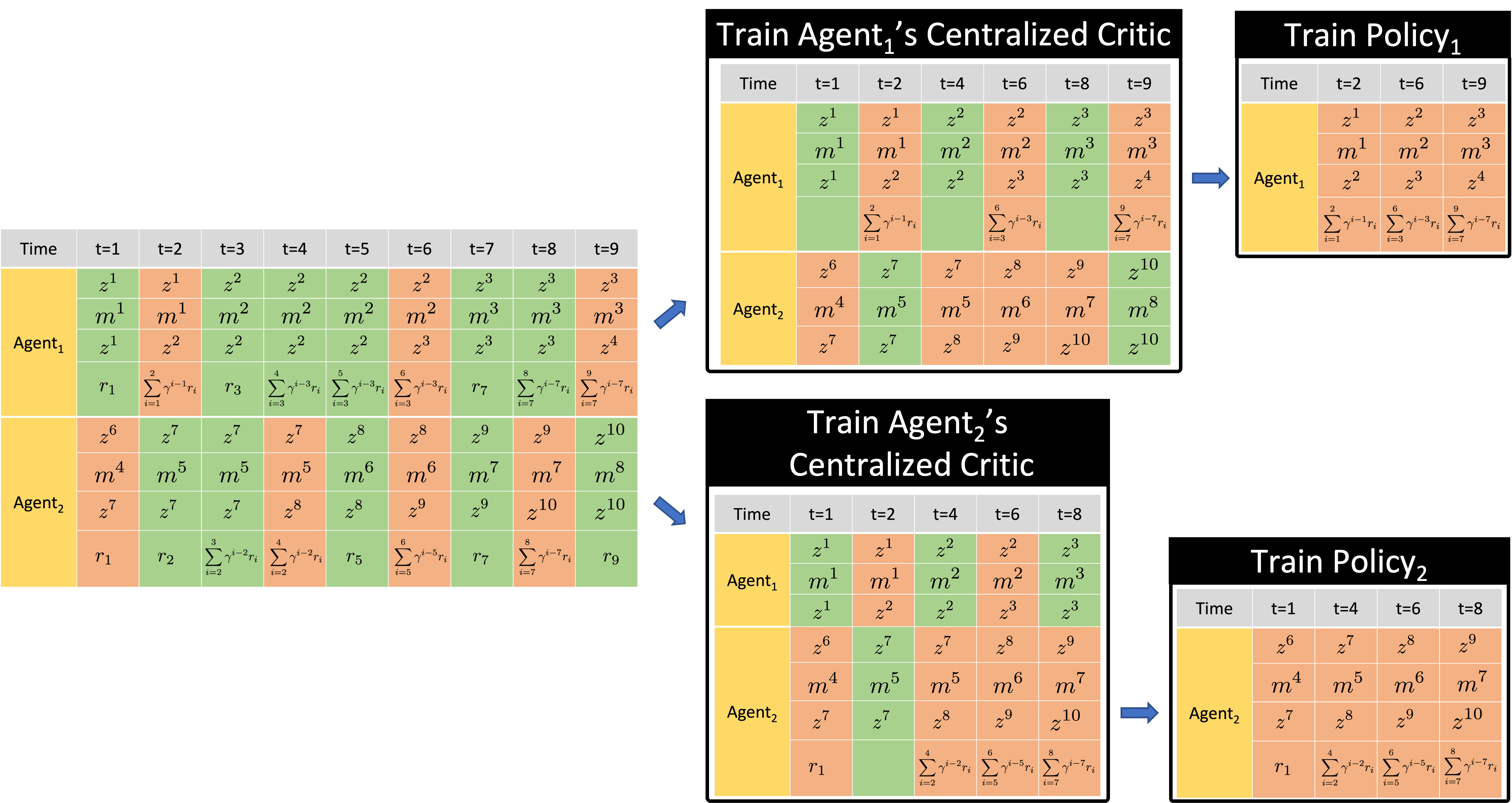}
    \caption{An example of the trajectory squeezing process in Mac-IAICC: each agent learns an individual centralized critic for the decentralized policy optimization. In order to achieve a better use of centralized information, the recurrent layer in each critic's neural network should receive all the valid joint macro-observation-action information (when $any$ agent terminates its macro-action (line 20-22) and obtain a new joint macro-observation). However, the critic's TD updates and the policy's updates still rely on each agent's individual macro-action termination and the accumulative reward at the corresponding timestep (line 23-26). Hence, the trajectory squeezing process for training each critic still depends on joint-macro-action termination but only retaining the accumulative rewards w.r.t. the corresponding agent's macro-action termination for computing the TD loss (the middle part in the above picture). Then, each agent's trajectory is further squeezed depending on its macro-action termination to update the decentralized policy.}
    \label{Mac-IAC-Traj}
\end{figure}

\begin{algorithm}[h!]
    \footnotesize
    \caption{Mac-IAICC}
    \label{alg4}
        \begin{algorithmic}[1]
            \State Initialize a decentralized policy network for each agent $i$: $\Psi_{\theta_i}$
            \State Initialize centralized critic networks for each agent $i$: $V_{\mathbf{w}_i}^{\vec{\Psi}_{\vec{\theta}}}$, $V_{\mathbf{w}^-_i}^{\vec{\Psi}_{\vec{\theta}}}$ 
            \State Initialize a decentralized buffer $\mathcal{D}$
            \For{\emph{episode} = $1$ to $M$}
                \State $t=0$
                \State Reset env
                \While{not reaching a terminal state \textbf{and} $t < \mathbb{H}$}
                    \State $t \leftarrow t + 1$
                    \For{each agent $i$}
                        \If{the macro-action $m_i$ is terminated}
                            \State $m_{i} \sim \Psi_{\theta_i}(\cdot \mid h_i; \epsilon)$
                        \Else
                            \State Continue running current macro-action $m_i$
                        \EndIf
                    \EndFor
                    \For{each agent $i$}
                        \State Get a reward $r^c_i$ accumulated based on agent $i$'s macro-action termination
                    \EndFor
                    \State Get next joint macro-observations $\vec{z\,}'$
                    \State Collect $\langle \vec{z},\vec{m}, \vec{z\,}', \{r^c_1,\dots, r^c_n \} \rangle$ into the buffer $\mathcal{D}$
                \EndWhile
                \If{\emph{episode} mod $I_{\text{train}} = 0$}
                    \For{each agent $i$}
                        \State Squeeze trajectories in the buffer $\mathcal{D}$ according to joint macro-action terminations
                        \State Compute the TD-error of each timestep in the squeezed experiences:
                        \State $L(\mathbf{w}_i)=\big(y-V^{\vec{\Psi}_{\vec{\theta}}}_{\mathbf{w}_i}(\vec{h})\big)^2_\mathcal{D}$, where  $y = r^c_i + \gamma^{\tau_{m_i}} V^{\vec{\Psi}_{\vec{\theta}}}_{\mathbf{w}^-_i}(\vec{h}')$ 
                        \State Perform a gradient descent only over the TD-errors when agent $i$'s macro-action is terminated
                        \State Squeeze agent $i$'s trajectories in the buffer $\mathcal{D}$ according to its own macro-action terminations
                        \State Perform a gradient ascent on: 
                        \State $\nabla_{\theta_i} J(\theta_i) = \mathbb{E}_{\vec{\Psi}_{\vec{\theta}}}\Big[\nabla_{\theta_i}\log\Psi_{\theta_i}(m_i|h_i)\big(r^c_i + \gamma^{\tau_{m_i}} V^{\vec{\Psi}_{\vec{\theta}}}_{\mathbf{w}^-_i}(\vec{h}')-V^{\vec{\Psi}_{\vec{\theta}}}_{\mathbf{w}_i}(\vec{h})\big)\Big]$
                    \EndFor
                    \State Reset buffer $\mathcal{D}$
                \EndIf
                \If{\emph{episode} mod $I_{\text{TargetUpdate}} = 0$}
                    \For{each agent $i$}
                        \State Update the critic target network $\mathbf{w}_i^-\leftarrow\mathbf{w}_i$
                    \EndFor
                \EndIf
            \EndFor
        \end{algorithmic}
\end{algorithm}

\clearpage
\section{Domain Descriptions and Results}
\label{A-Domain}
\subsection{Box Pushing}
\label{A-BP}\hfill

\begin{figure*}[h!]
    \centering
    \captionsetup[subfigure]{labelformat=empty}
    \centering
    \subcaptionbox{(a) 8x8}
        [0.3\linewidth]{\includegraphics[scale=0.12]{results/BP/bpma.png}}
    ~
    \centering
    \subcaptionbox{(b) 10x10}
        [0.3\linewidth]{\includegraphics[scale=0.14]{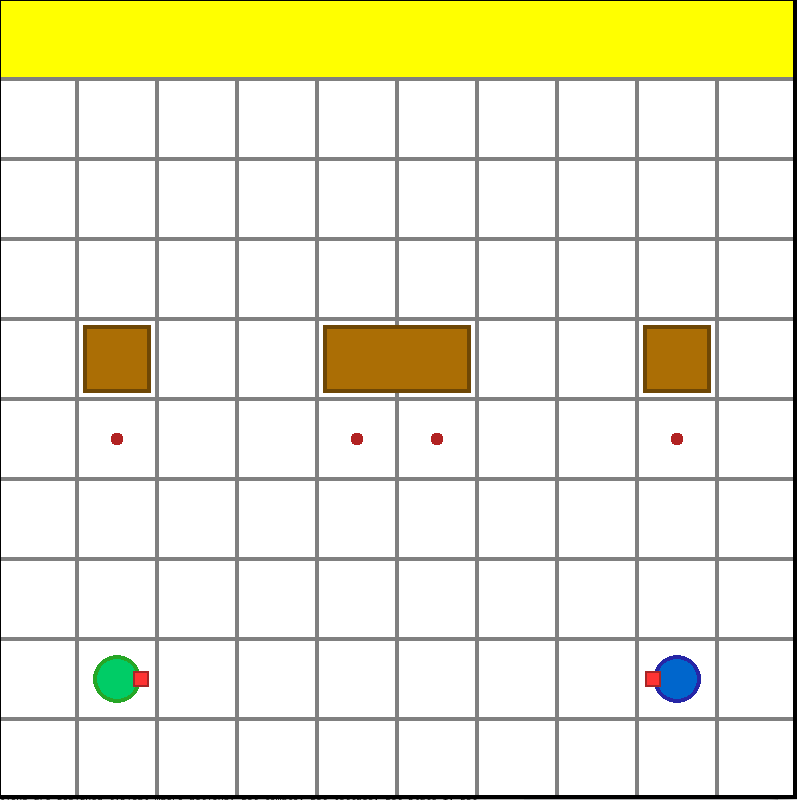}}
    ~
    \centering
    \subcaptionbox{(c) 12x12}
        [0.3\linewidth]{\includegraphics[scale=0.16]{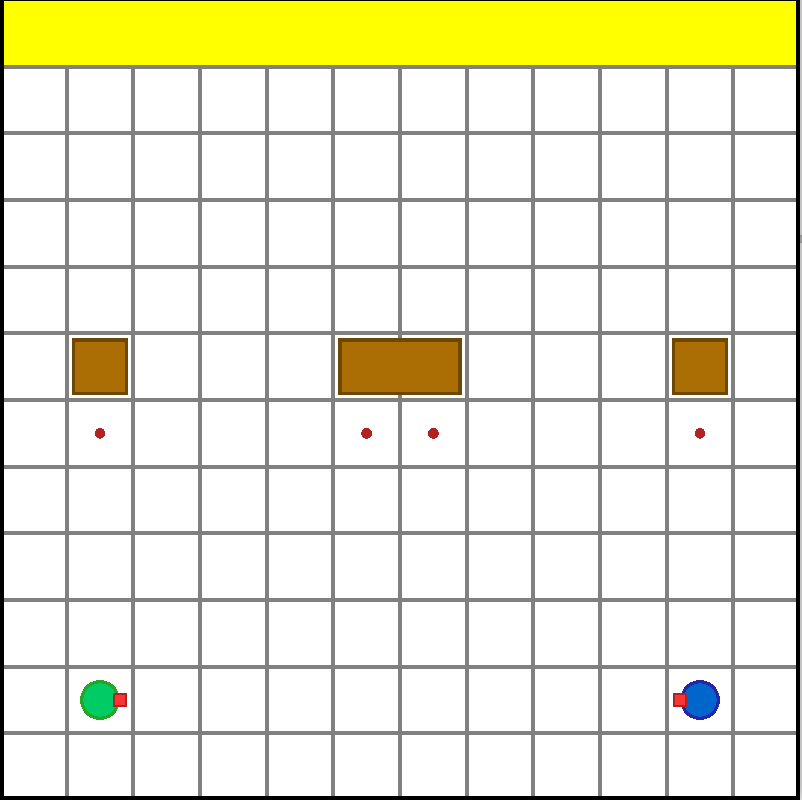}}
    \caption{Experimental environments.}
\end{figure*}

\textbf{Goal}.
The objective of the two robots is to learn collaboratively push the middle big box to the goal area at the top rather than pushing a small box on each own.\\

\textbf{State}. 
The global state information consists of the position and orientation of each robot and each box’s position in a grid world.\\

\textbf{Primitive-Action Space}. 
\emph{move forward}, \emph{turn-left}, \emph{turn-right} and \emph{stay}.\\

\textbf{Macro-Action Space}. 

$\bullet$ One-step macro-actions: \emph{\textbf{Turn-left}}, \emph{\textbf{Turn-right}}, and \emph{\textbf{Stay}}. 

$\bullet$ Multi-step macro-actions: \emph{\textbf{Move-to-small-box(i)}} that navigates the robot to the red spot below the corresponding small box and terminate with robot facing the box; \emph{\textbf{Move-to-big-box(i)}} that navigates the robot to a red spot below the big box and terminate with robot facing the big box; \emph{\textbf{Push}} that operates the robot to keep moving forward and terminate while arriving the world's boundary, touching the big box along or pushing a small box to the goal.\\ 

\textbf{Observation Space}. 
In both the primitive-observation and macro-observation, each robot is only allowed to capture one of five states of the cell in front of it: \emph{empty}, \emph{teammate}, \emph{boundary}, \emph{small box}, \emph{big box}.\\ 

\textbf{Dynamics}.
The transition in this task is deterministic. Boxes can only be moved towards the north when the robot faces the box and moves forward. The small box can be moved by a single robot while the big box require two robots to move it together.\\ 

\textbf{Rewards}.
The team receives $+300$ for pushing big box to the goal area and $+20$ for pushing a small box to the goal area. A penalty $-10$ is issued when any robot hits the boundary or pushes the big box on its own.\\  

\textbf{Episode Termination}.
Each episode terminates when any box is pushed to the goal area, or when 100 timesteps has elapsed.

\clearpage
\textbf{Results.}

\begin{figure*}[h!]
    \centering
    \captionsetup[subfigure]{labelformat=empty}
    \centering
    \subcaptionbox{}
        [0.9\linewidth]{\includegraphics[scale=0.13]{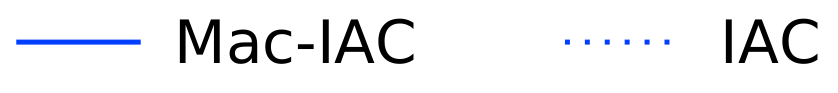}}
    ~
    \centering
    \subcaptionbox{}
        [0.31\linewidth]{\includegraphics[height=3cm]{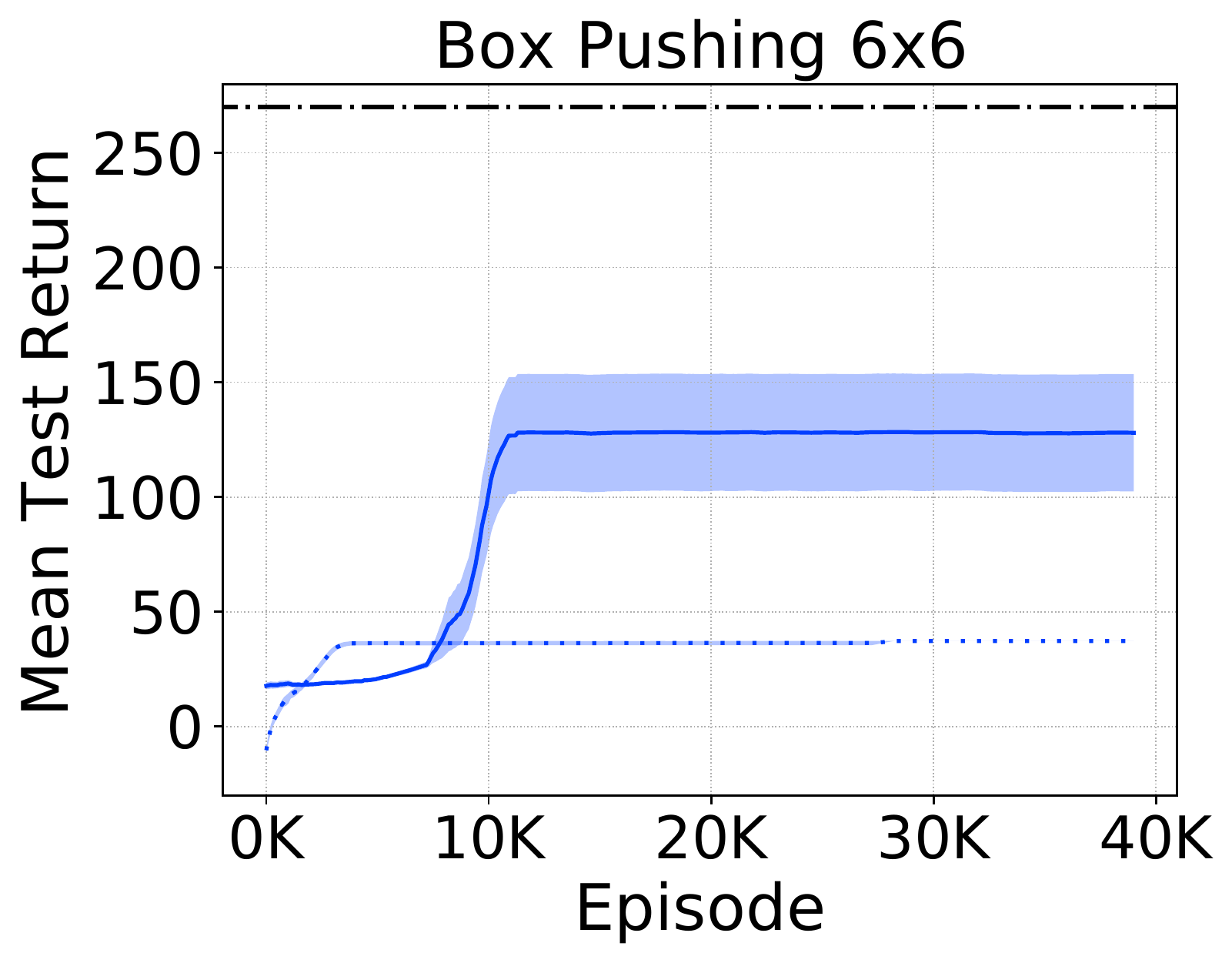}}
    ~
    \centering
    \subcaptionbox{}
        [0.31\linewidth]{\includegraphics[height=3cm]{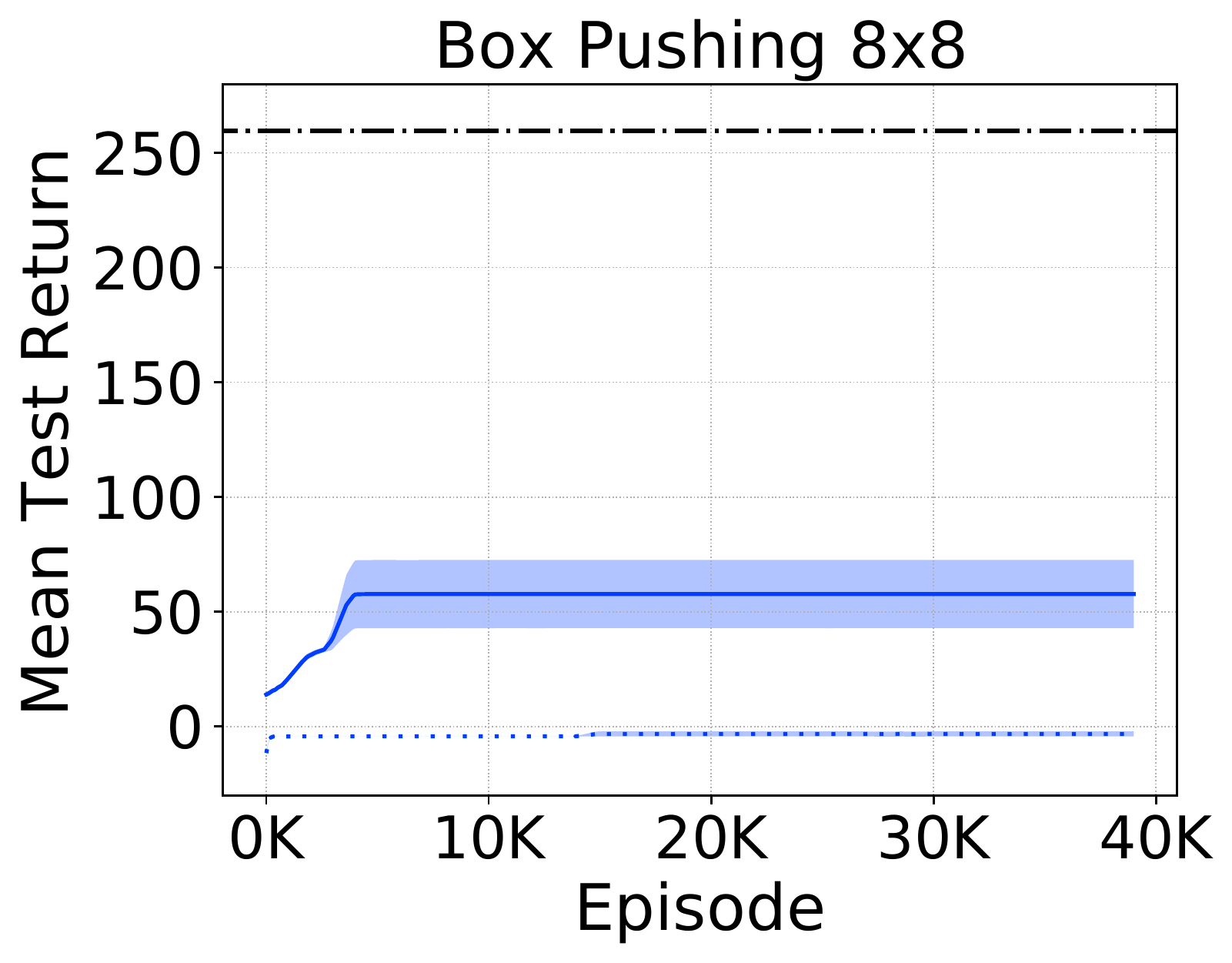}}
    ~
    \centering
    \subcaptionbox{}
        [0.31\linewidth]{\includegraphics[height=3cm]{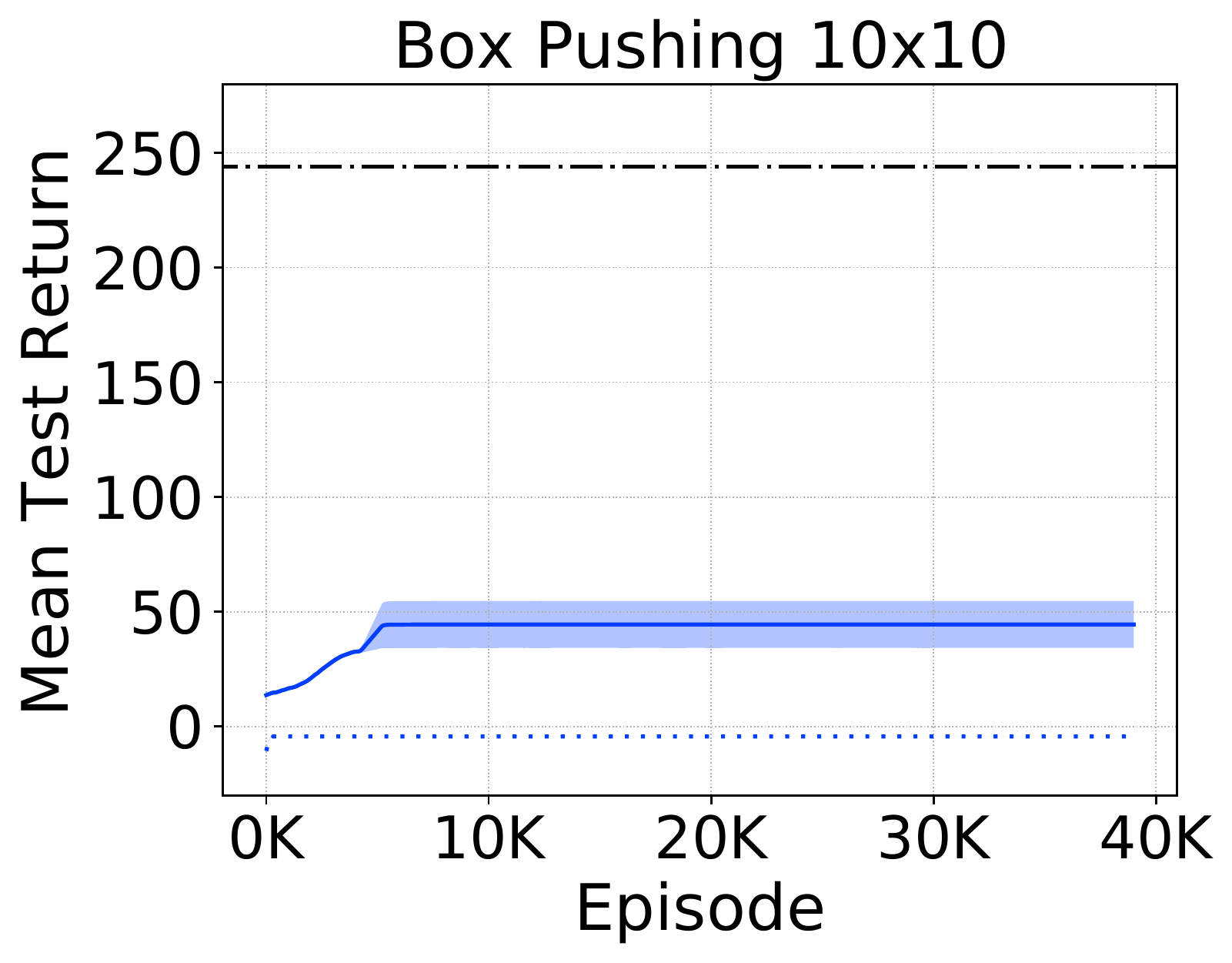}}
    ~
    \centering
    \subcaptionbox{}
        [0.31\linewidth]{\includegraphics[height=3cm]{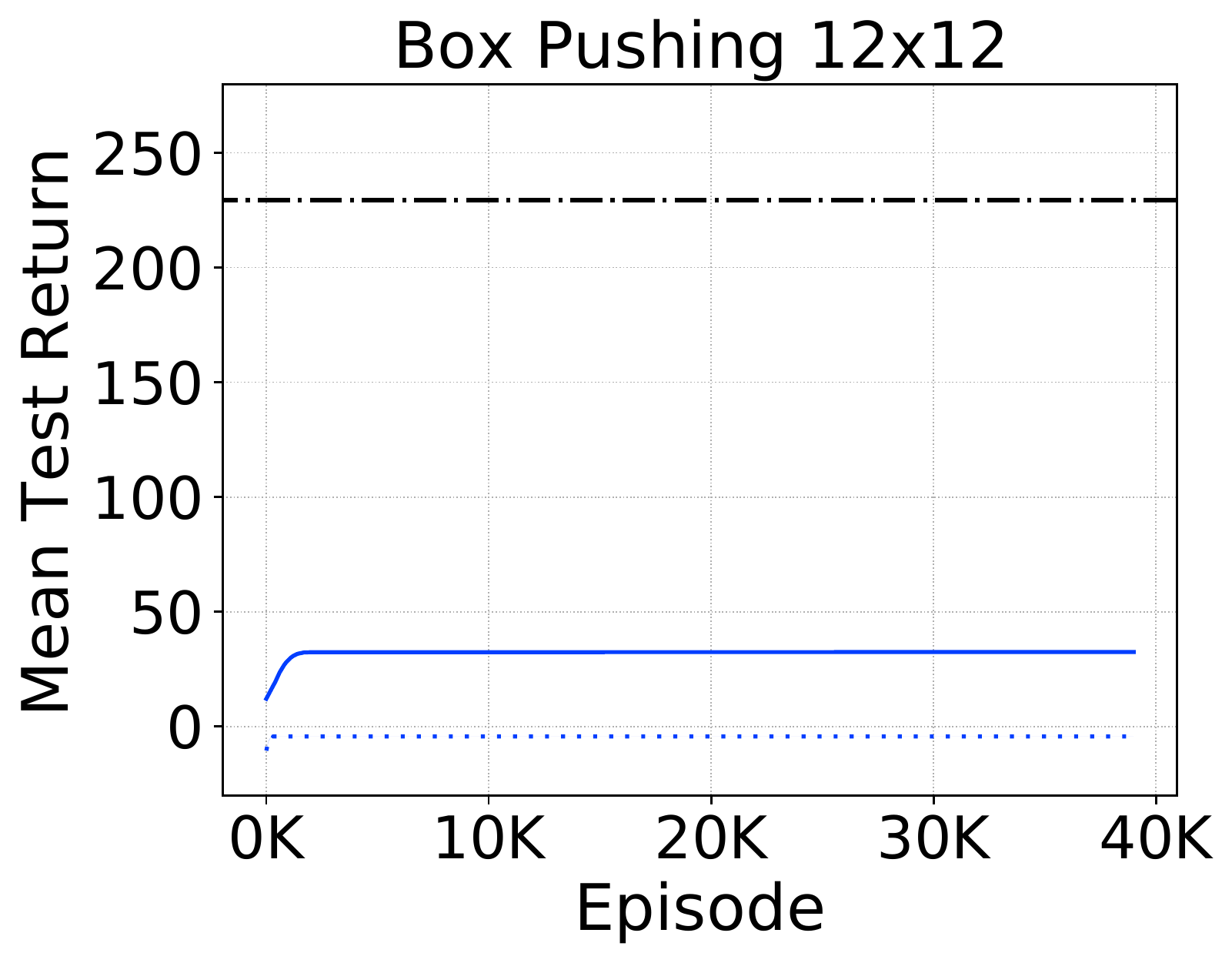}}
    ~
    \centering
    \subcaptionbox{}
        [0.31\linewidth]{\includegraphics[height=3cm]{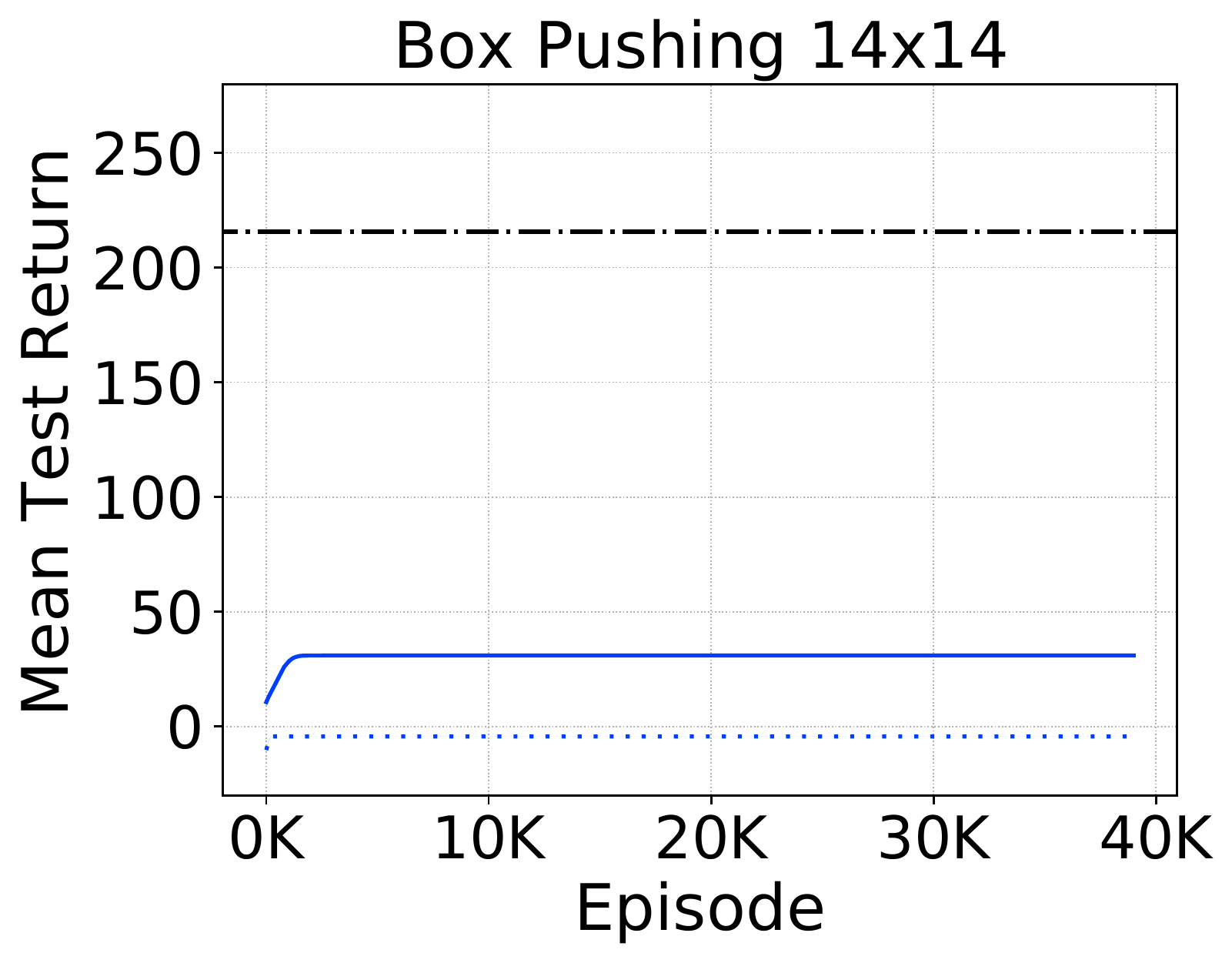}}
    \caption{Decentralized learning with macro-actions vs primitive-actions.}
    \label{bp_dec_comp}
\end{figure*}

\begin{figure*}[h!]
    \centering
    \captionsetup[subfigure]{labelformat=empty}
    \centering
    \subcaptionbox{}
        [0.9\linewidth]{\includegraphics[scale=0.13]{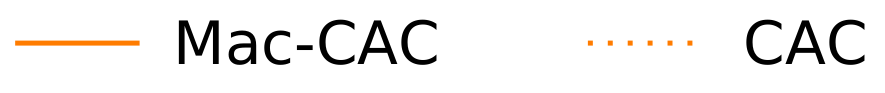}}
    ~
    \centering
    \subcaptionbox{}
        [0.31\linewidth]{\includegraphics[height=3cm]{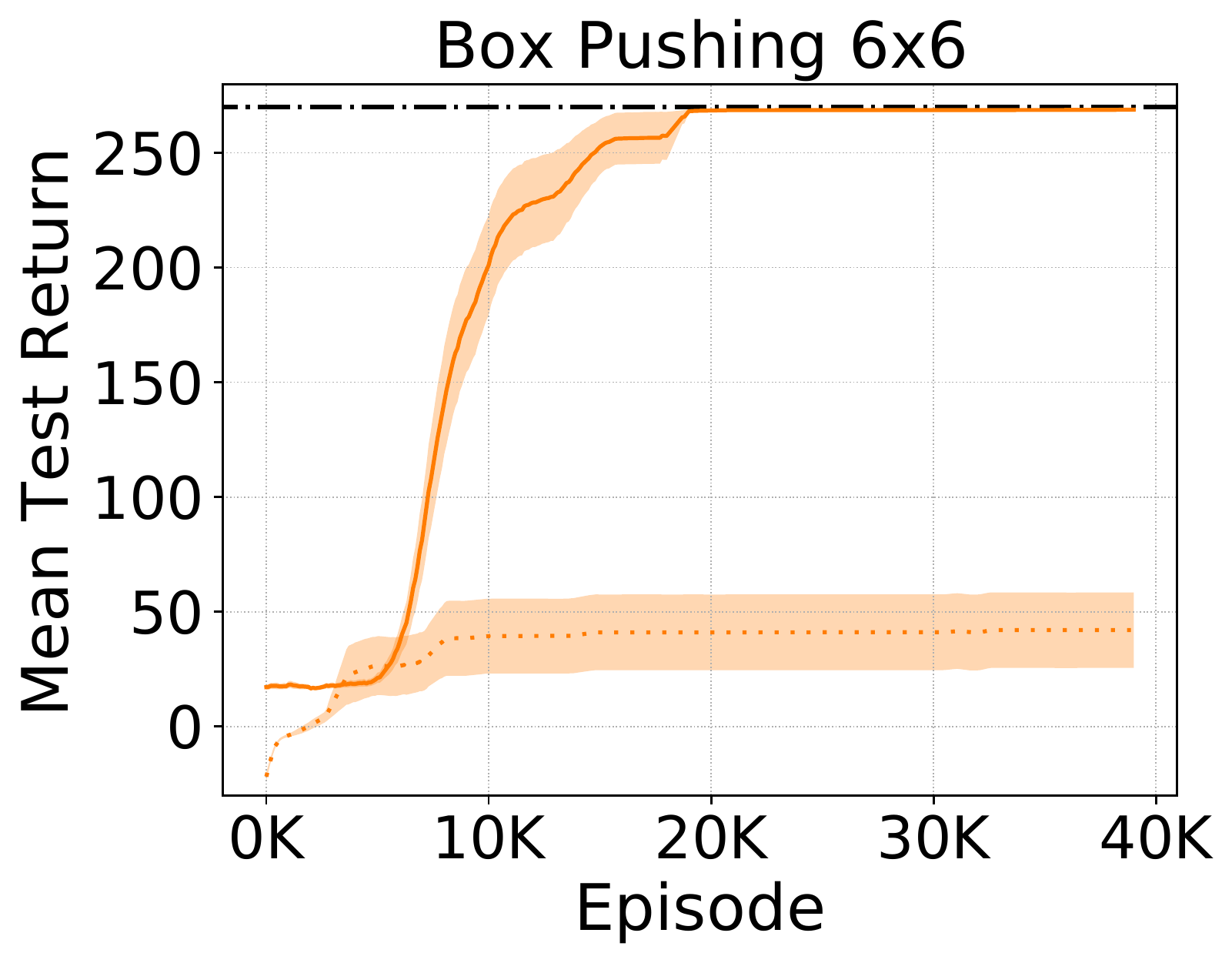}}
    ~
    \centering
    \subcaptionbox{}
        [0.31\linewidth]{\includegraphics[height=3cm]{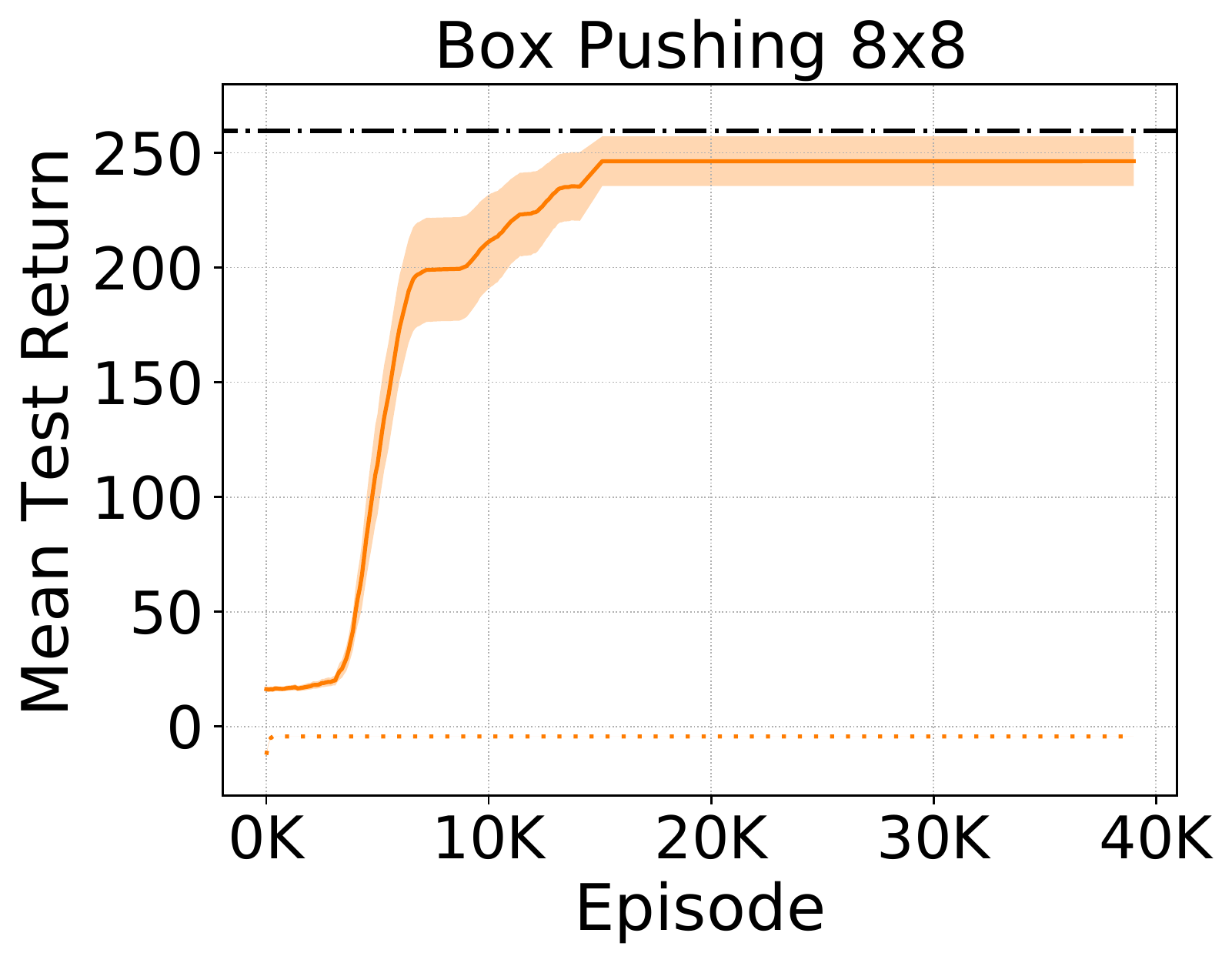}}
    ~
    \centering
    \subcaptionbox{}
        [0.31\linewidth]{\includegraphics[height=3cm]{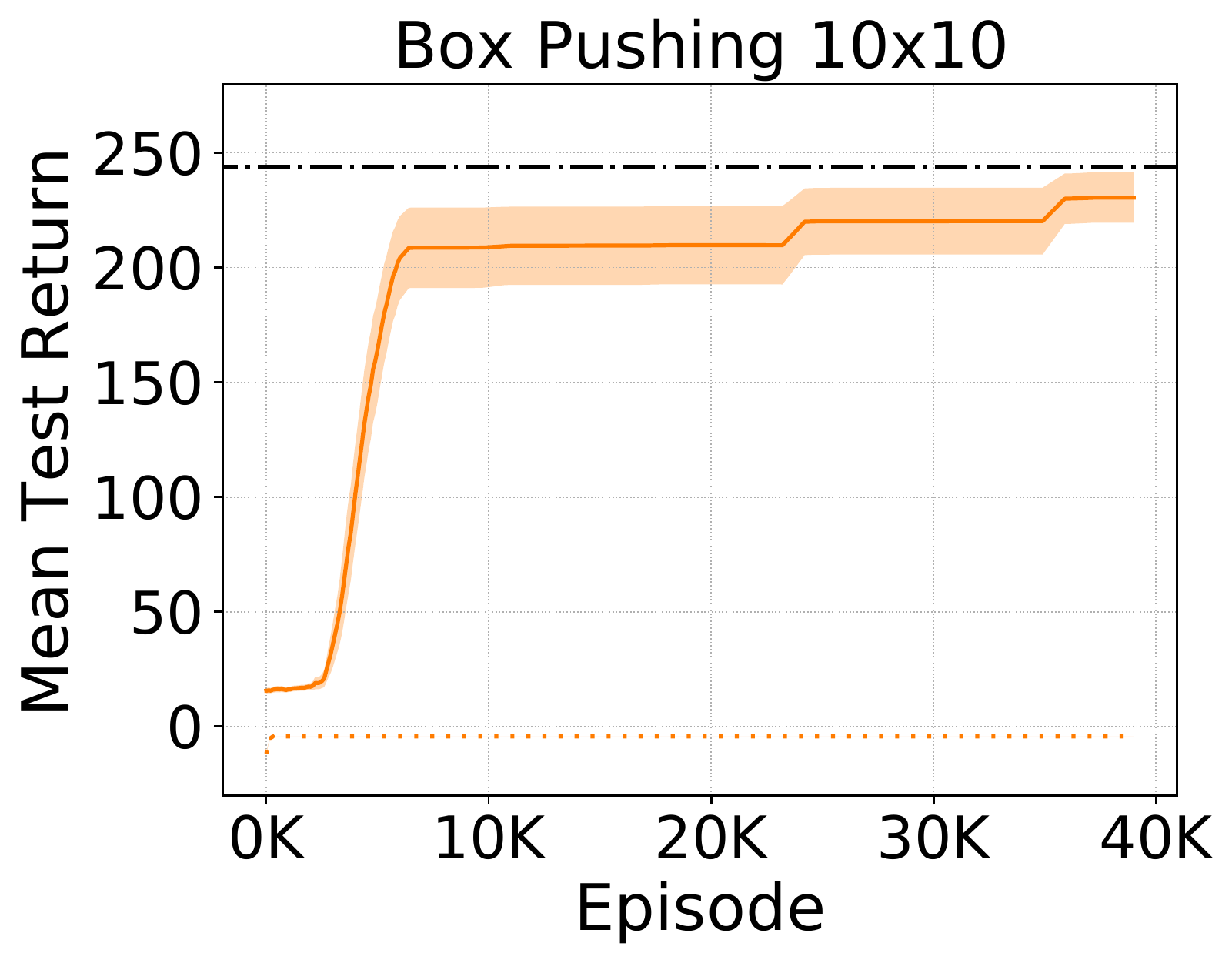}}
    ~
    \centering
    \subcaptionbox{}
        [0.31\linewidth]{\includegraphics[height=3cm]{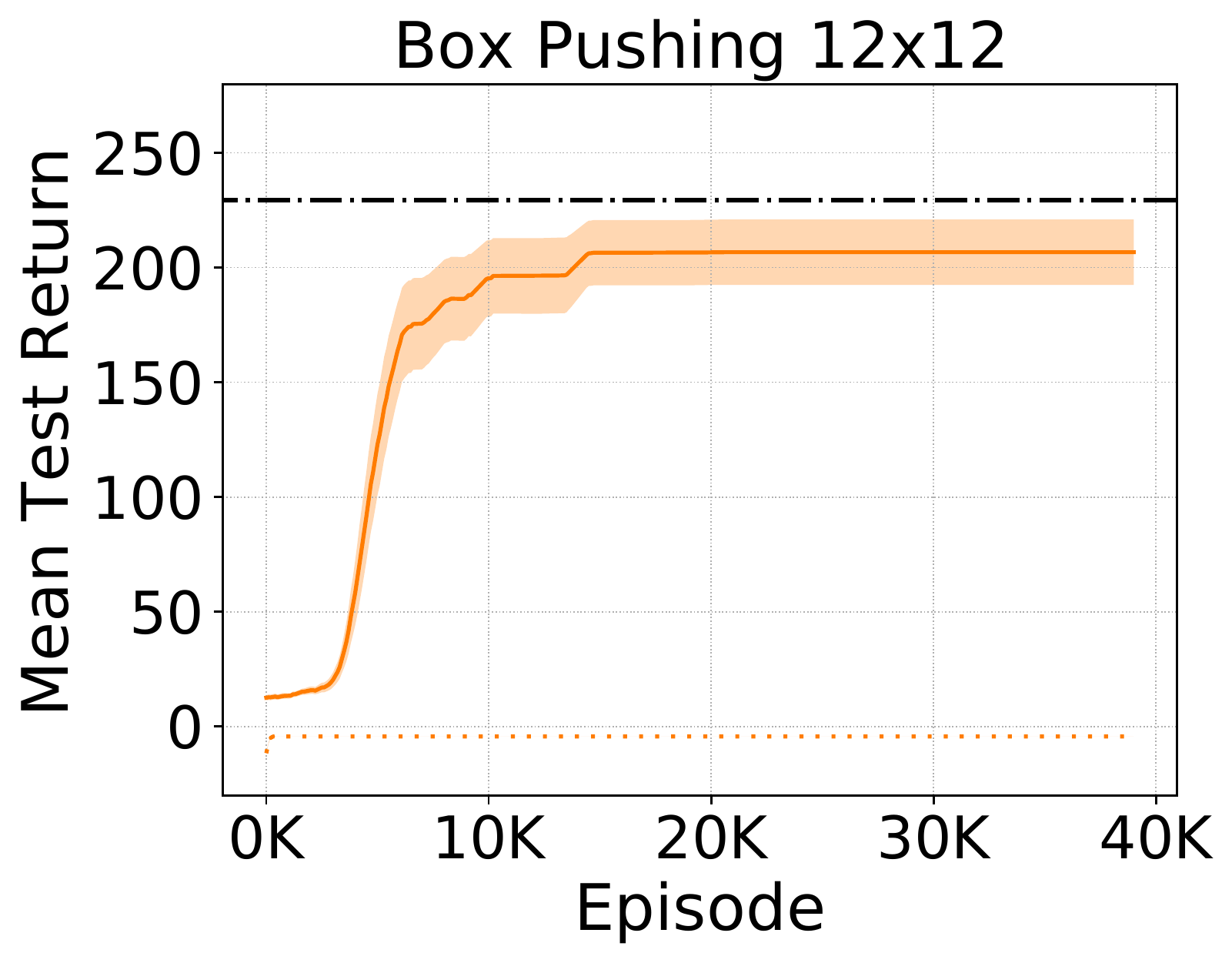}}
    ~
    \centering
    \subcaptionbox{}
        [0.31\linewidth]{\includegraphics[height=3cm]{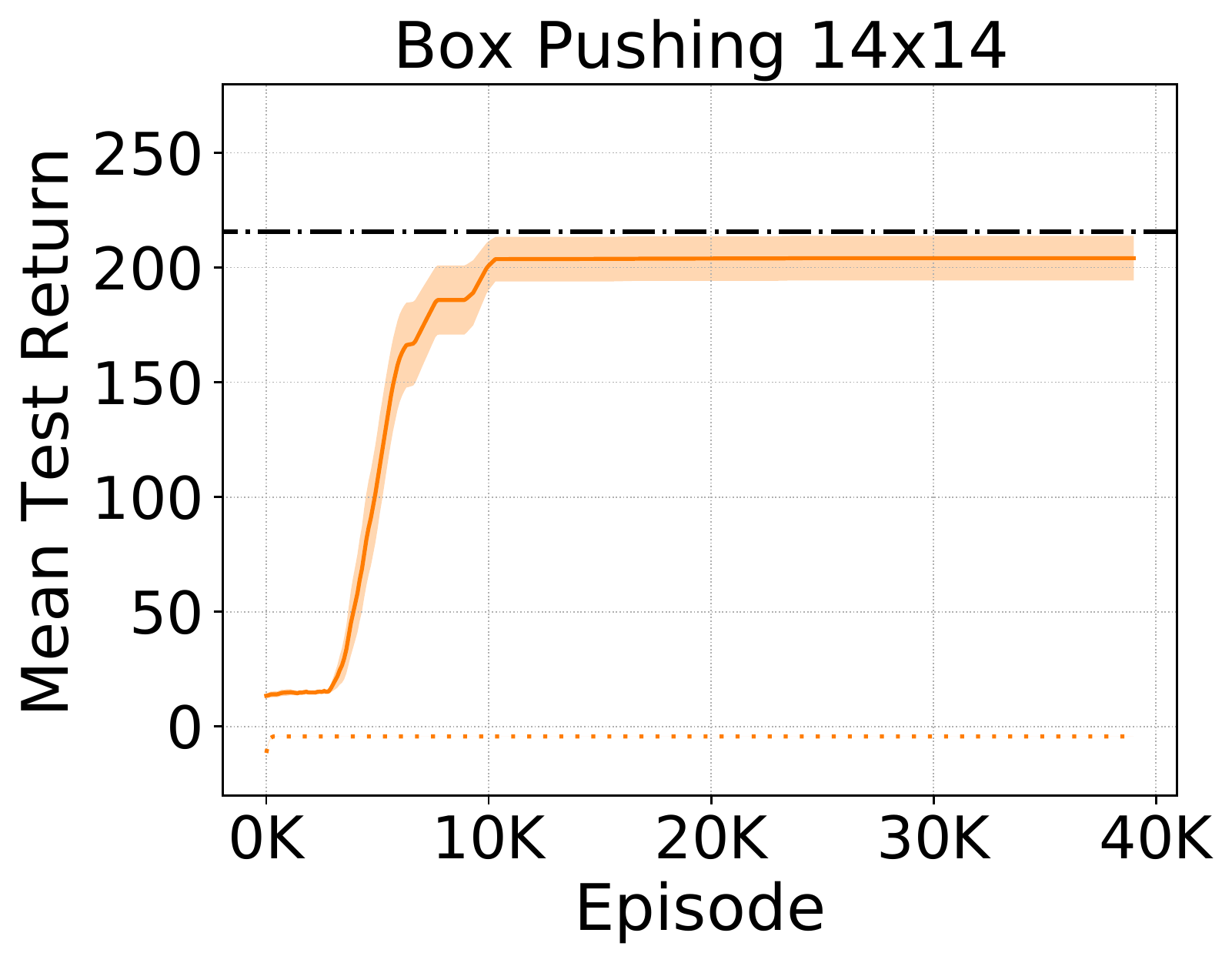}}
    \caption{Centralized learning with macro-actions vs primitive-actions.}
    \label{bp_cen_comp}
\end{figure*}

\begin{figure*}[t!]
    \centering
    \captionsetup[subfigure]{labelformat=empty}
    \centering
    \subcaptionbox{}
        [0.9\linewidth]{\includegraphics[scale=0.17]{results/legend/ctde_comp.png}}
    ~
    \centering
    \subcaptionbox{}
        [0.31\linewidth]{\includegraphics[height=3cm]{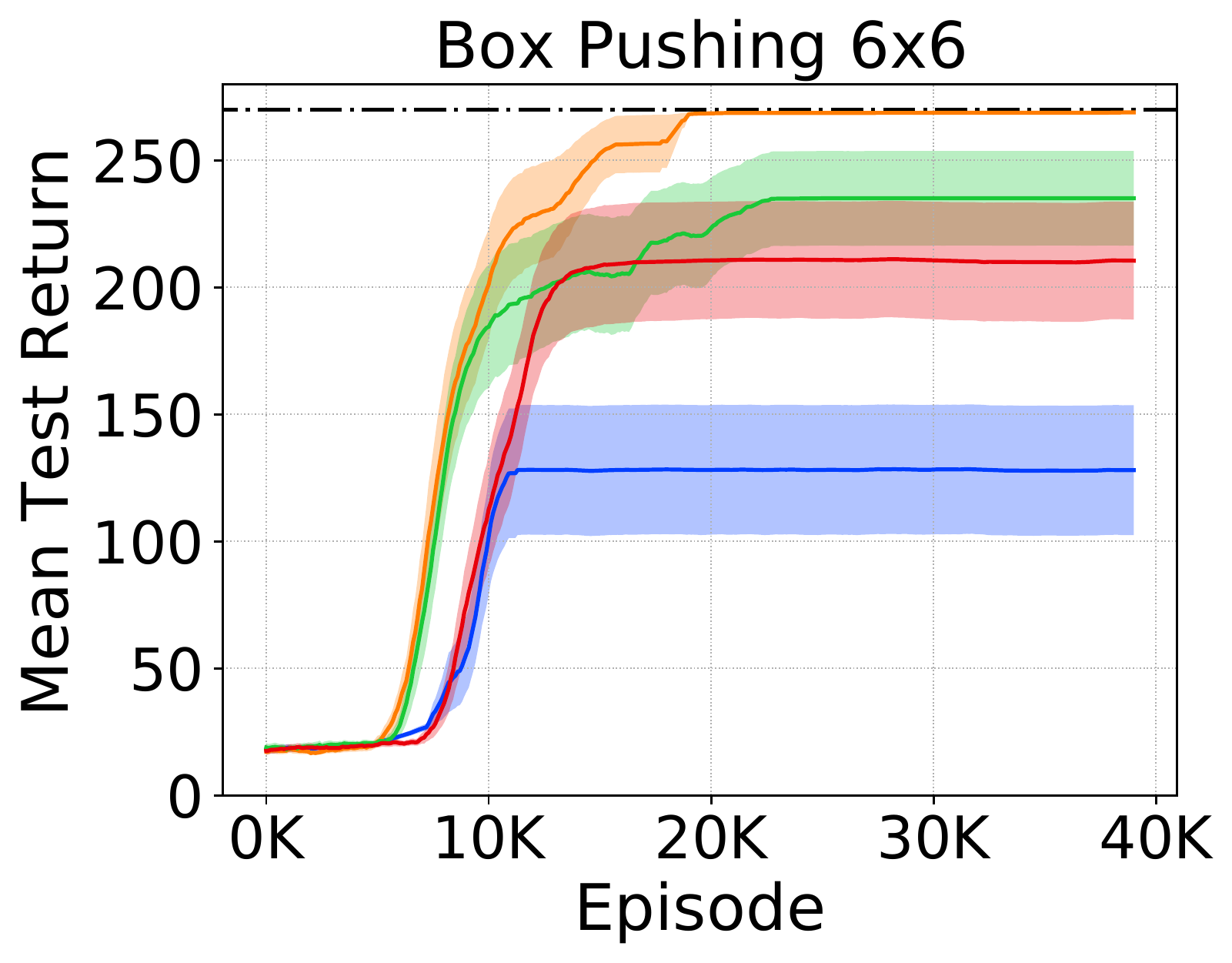}}
    ~
    \centering
    \subcaptionbox{}
        [0.31\linewidth]{\includegraphics[height=3cm]{results/BP/bp8_ctde_comp.pdf}}
    ~
    \centering
    \subcaptionbox{}
        [0.31\linewidth]{\includegraphics[height=3cm]{results/BP/bp10_ctde_comp.pdf}}
    ~
    \centering
    \subcaptionbox{}
        [0.31\linewidth]{\includegraphics[height=3cm]{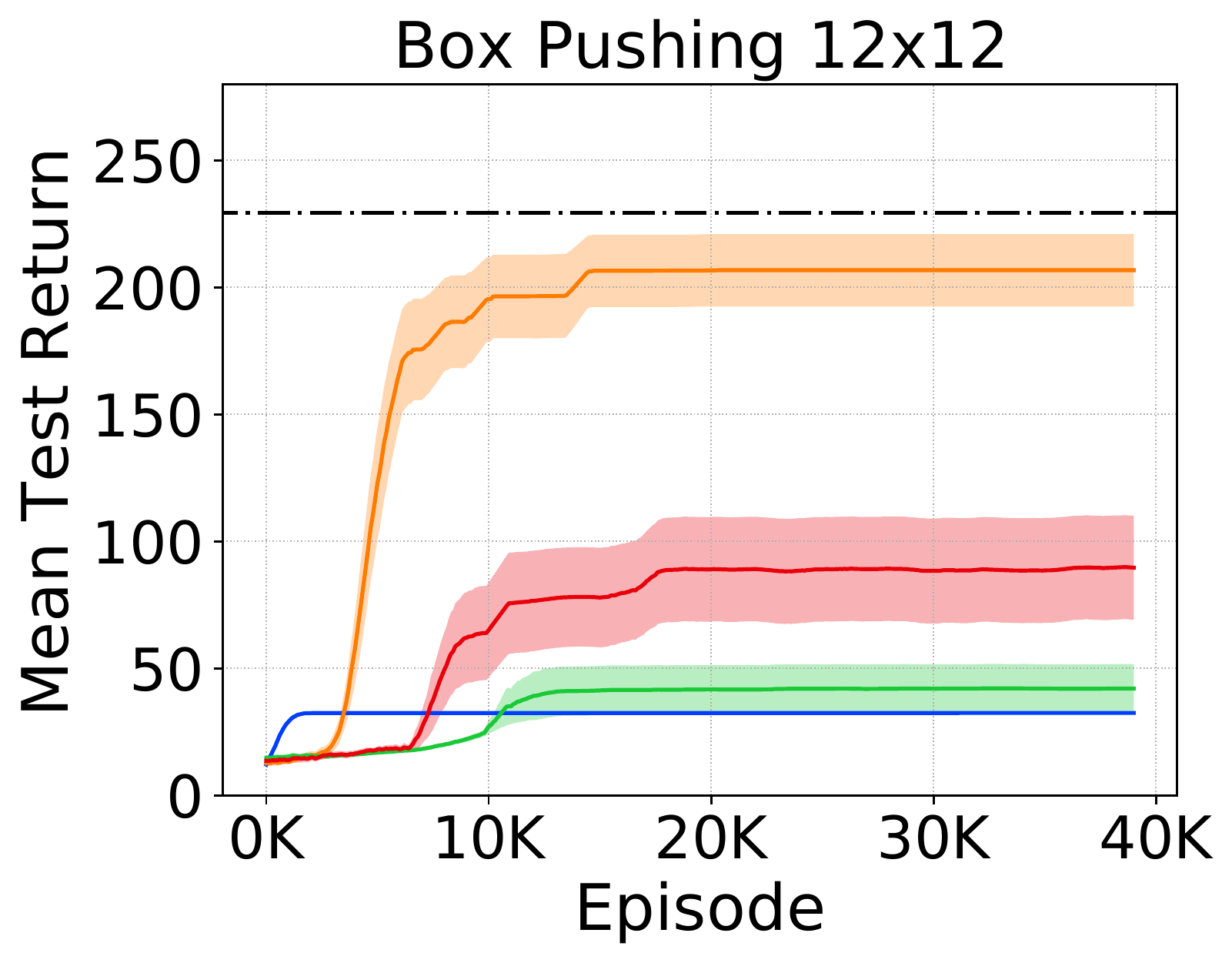}}
    ~
    \centering
    \subcaptionbox{}
        [0.31\linewidth]{\includegraphics[height=3cm]{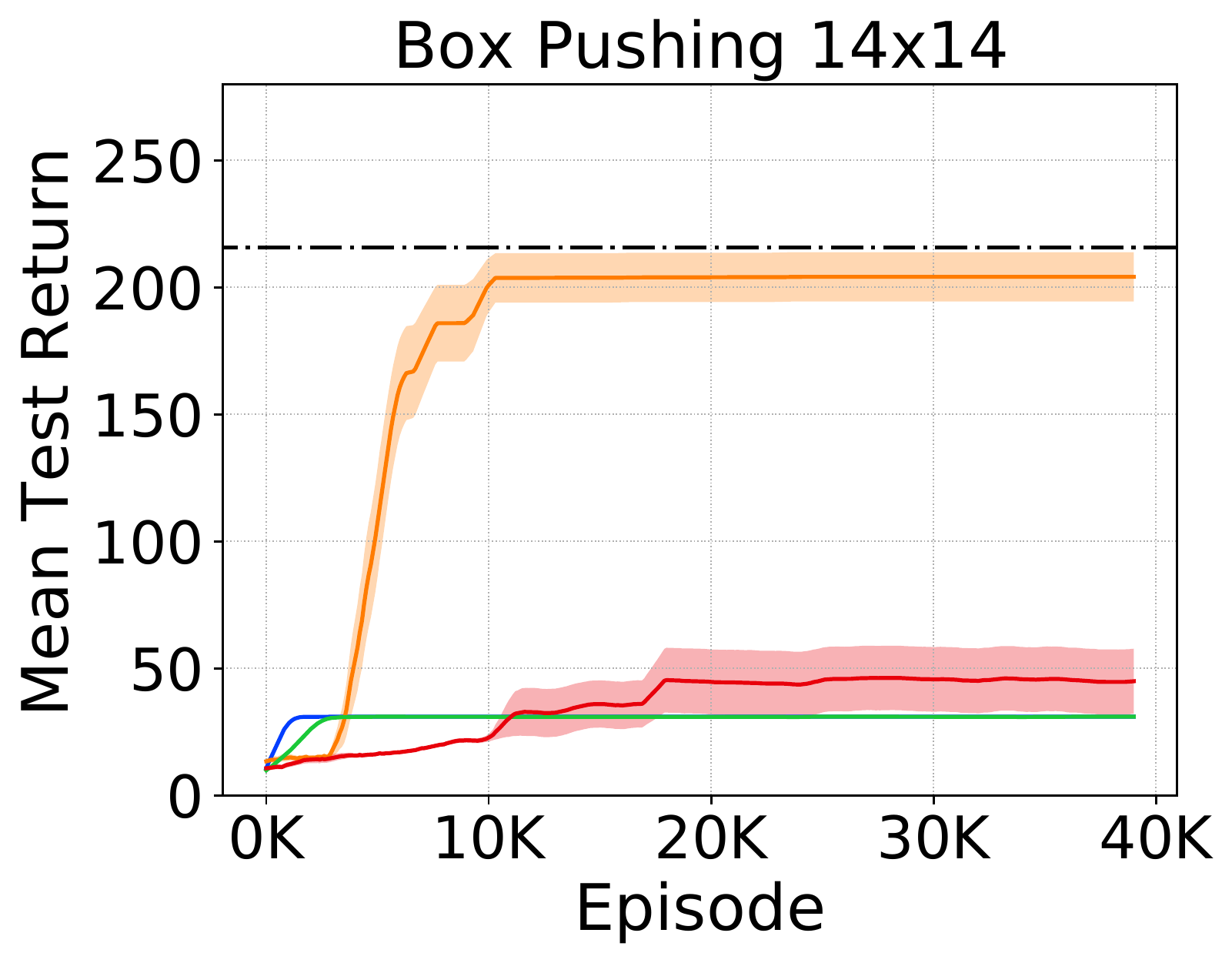}}
    \caption{Comparison of macro-action-based multi-agent actor-critic methods.}
    \label{bp_ctde_comp}
\end{figure*}

\begin{figure*}[t!]
    \centering
    \captionsetup[subfigure]{labelformat=empty}
    \centering
    \subcaptionbox{}
        [0.9\linewidth]{\includegraphics[scale=0.17]{results/legend/qvspg_colors.png}}
    ~
    \centering
    \subcaptionbox{}
        [0.31\linewidth]{\includegraphics[height=3cm]{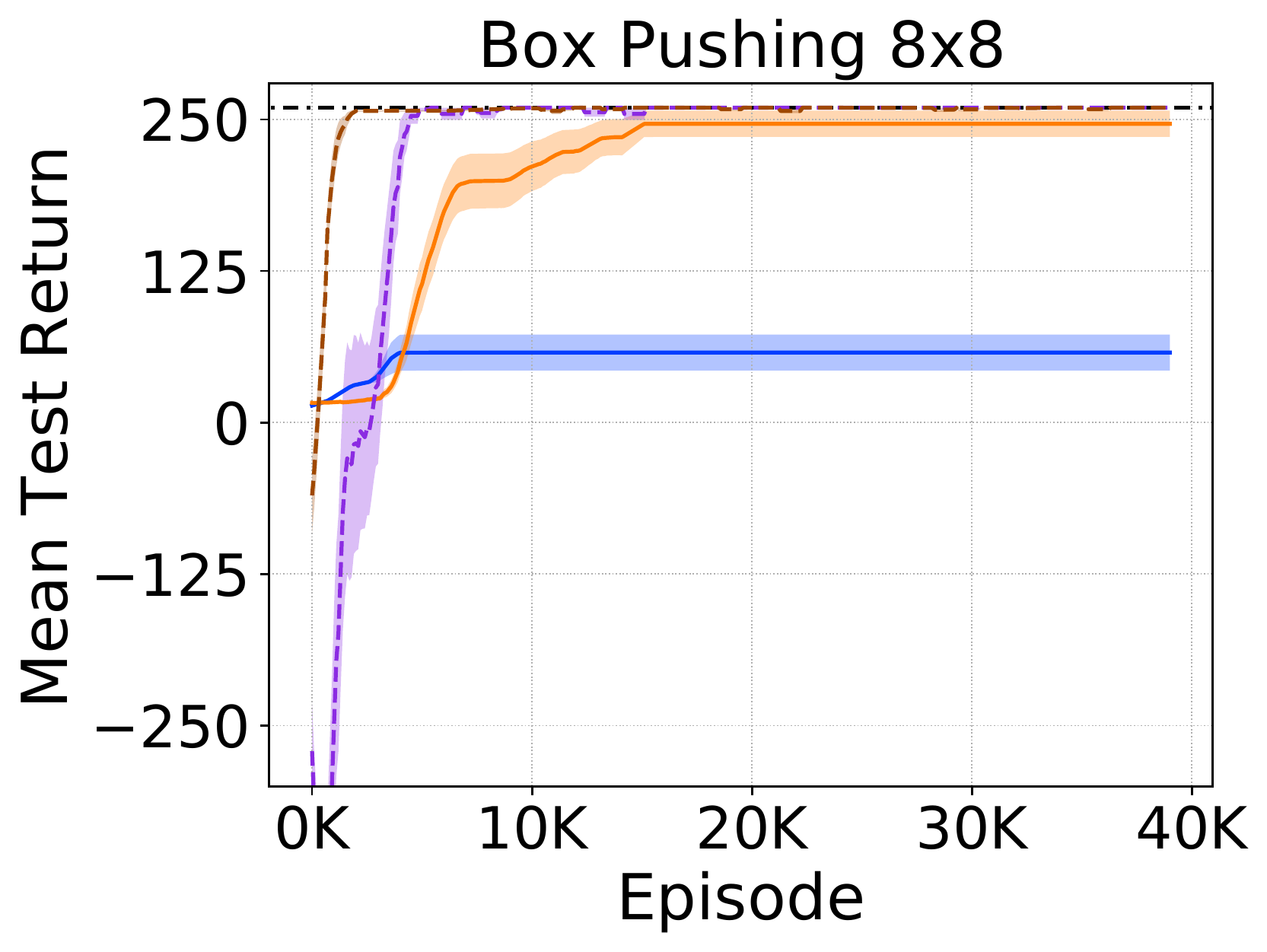}}
    ~
    \centering
    \subcaptionbox{}
        [0.31\linewidth]{\includegraphics[height=3cm]{results/BP/bp10_qvspg_colors.pdf}}
    ~
    \centering
    \subcaptionbox{}
        [0.31\linewidth]{\includegraphics[height=3cm]{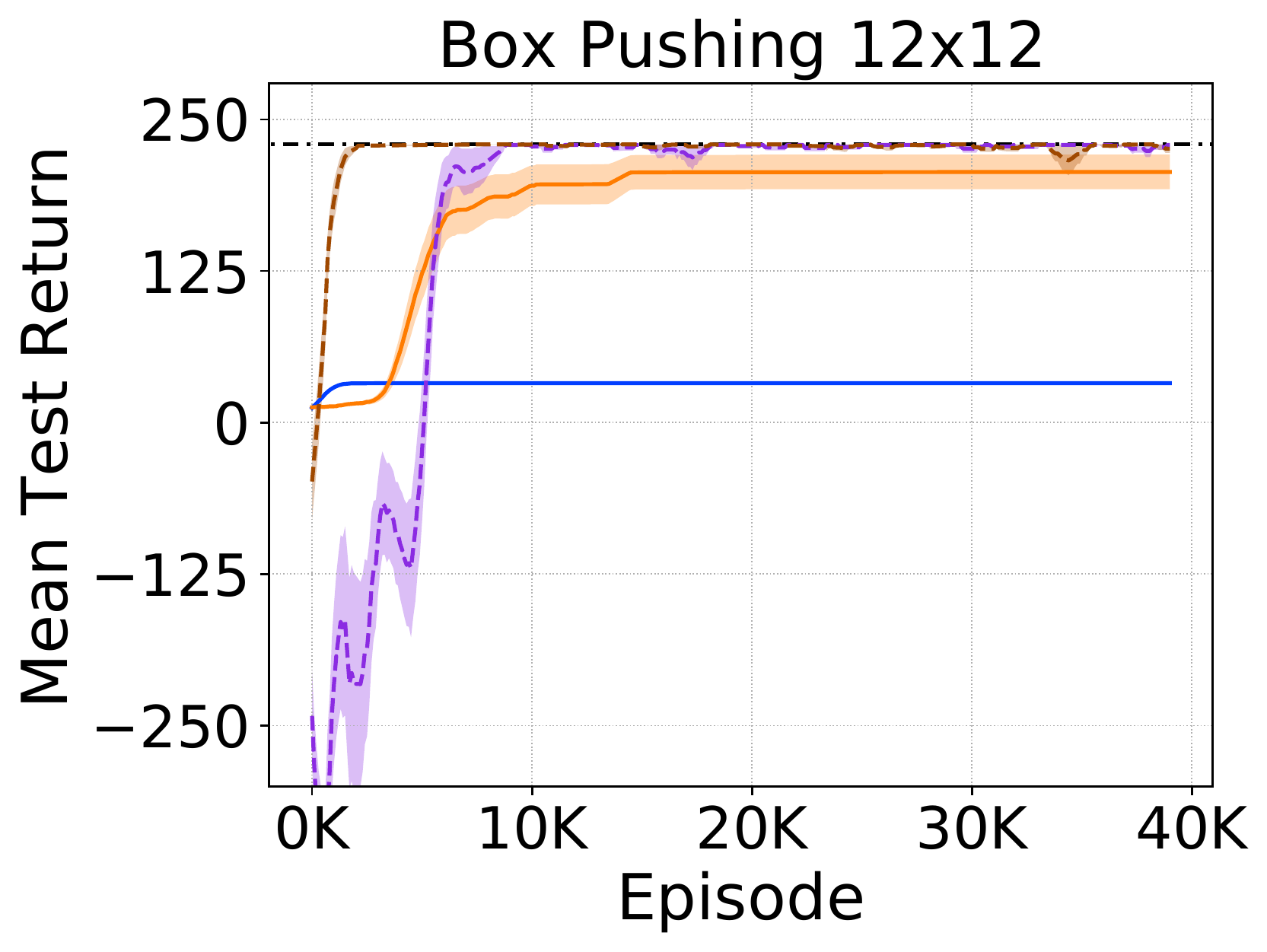}}
    ~
    \centering
    \subcaptionbox{}
        [0.31\linewidth]{\includegraphics[height=3cm]{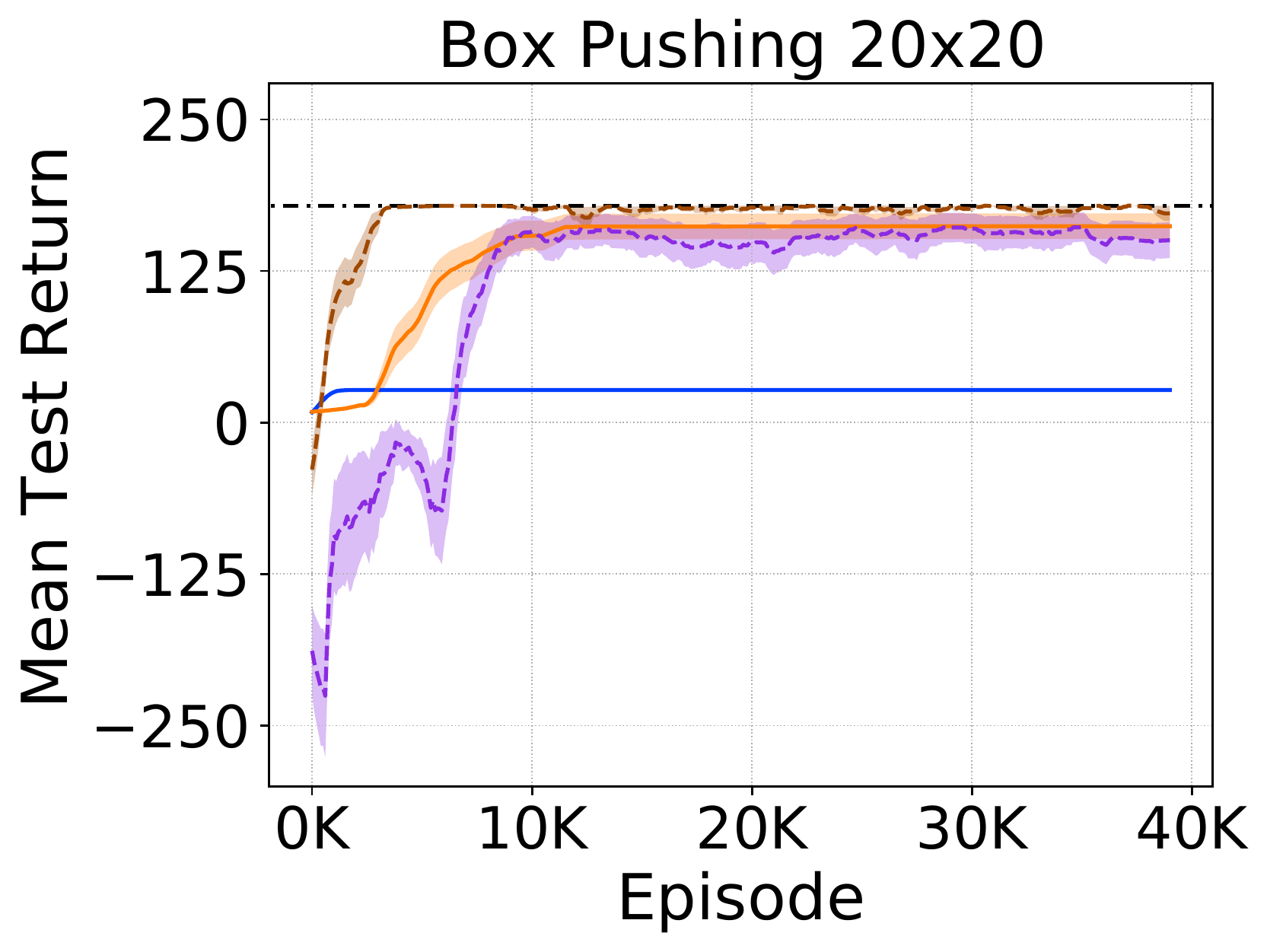}}
    ~
    \centering
    \subcaptionbox{}
        [0.31\linewidth]{\includegraphics[height=3cm]{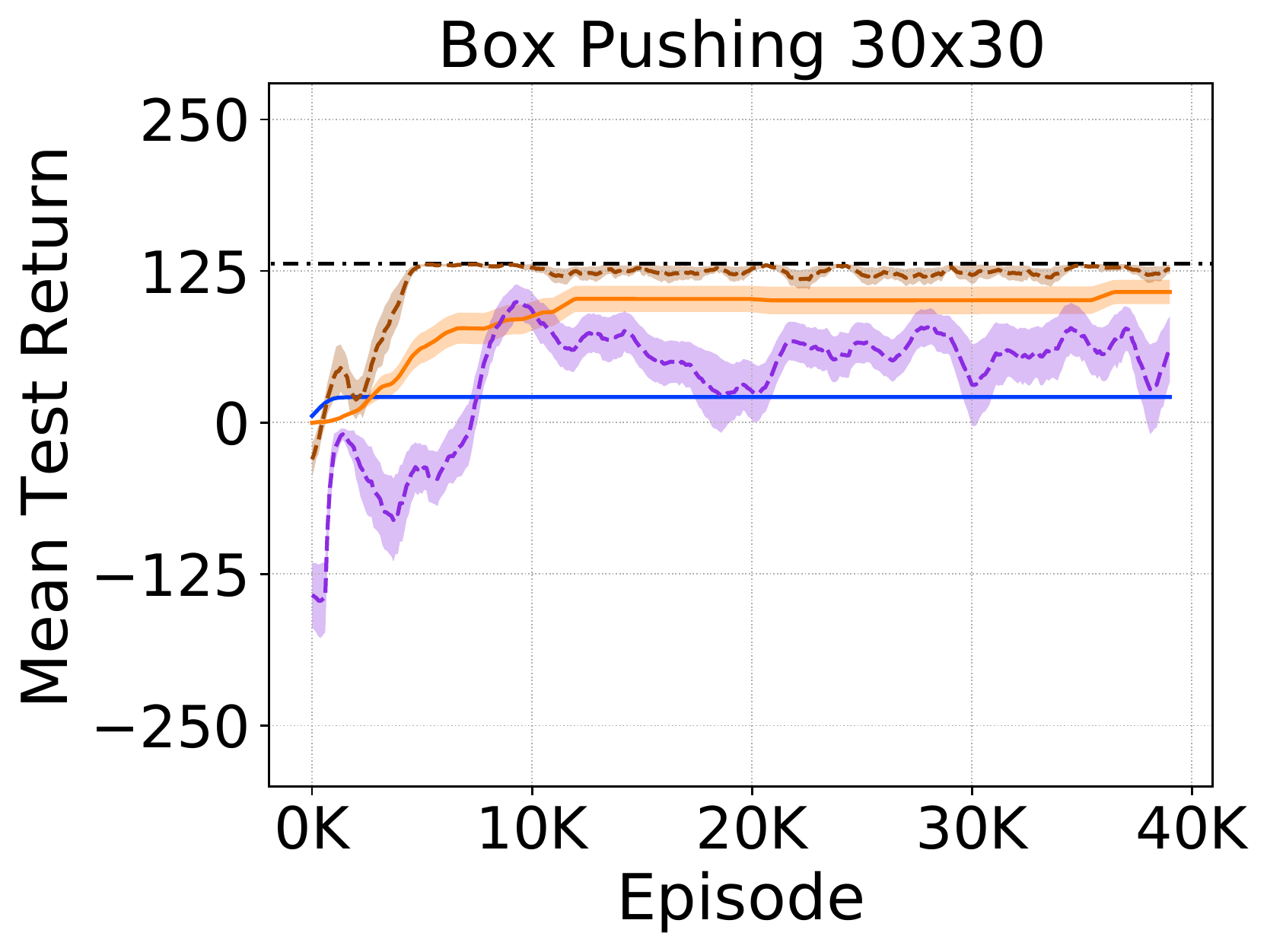}}
    \caption{Comparison of macro-action-based actor-critic methods and value-based methods}
    \label{bp_qvspg_comp}
\end{figure*}

\clearpage
\subsection{Overcooked}
\label{A-Overcooked}\hfill

\begin{figure*}[h!]
    \centering
    \captionsetup[subfigure]{labelformat=empty}
    \centering
    \subcaptionbox{(a) Overcooked-A}
        [0.20\linewidth]{\includegraphics[scale=0.14]{results/Overcooked/3_agent_D.png}}
    ~
    \centering
    \subcaptionbox{(b) Overcooked-B}
        [0.20\linewidth]{\includegraphics[scale=0.14]{results/Overcooked/3_agent_F.png}}
    ~
    \centering
    \subcaptionbox{(c) Overcooked-C}
        [0.20\linewidth]{\includegraphics[scale=0.14]{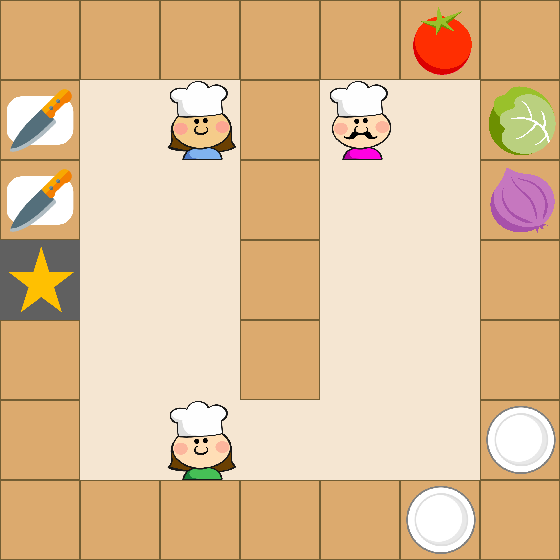}}
        ~
    \centering
    \subcaptionbox{(d) Lettuce-Onion-Tomato salad recipe}
        [0.3\linewidth]{\includegraphics[scale=0.13]{results/Overcooked/recipe.png}}
    \caption{Experimental environments.}
    \label{app:overcooked}
\end{figure*}

\textbf{Goal}. Three agents need to learn cooperating with each other to prepare a Tomato-Lettuce-Onion salad and deliver it to the `star' counter cell as soon as possible. The challenge is that the recipe of making a tomato-lettuce-onion salad is unknown to agents. Agents have to learn the correct procedure in terms of picking up raw vegetables, chopping, and merging in a plate before delivering.\\

\textbf{State Space}. The environment is a 7×7 grid world involving three agents, one tomato, one lettuce, one onion, two plates, two cutting boards and one delivery cell. The global state information consists of the positions of each agent and above items, and the status of each vegetable: chopped, unchopped, or the progress under chopping.\\

\textbf{Primitive-Action Space}. 
Each agent has five primitive-actions: \emph{up}, \emph{down}, \emph{left}, \emph{right} and \emph{stay}. Agents can move around and achieve picking, placing, chopping and delivering by standing next to the corresponding cell and moving against it (e.g., in Fig.~\ref{app:overcooked}a, the pink agent can \emph{move right} and then \emph{move up} to pick up the tomato).\\

\textbf{Macro-Action Space}. Here, we first describe the main function of each macro-action and then list the corresponding termination conditions.
\begin{itemize}
        \item{Five one-step macro-actions that are the same as the primitive ones;} 
            \vspace{-1mm}
        \item{\textbf{\emph{Chop}}, cuts a raw vegetable into pieces (taking three time steps) when the agent stands next to a cutting board and an unchopped vegetable is on the board, otherwise it does nothing; and it terminates when:
            \begin{itemize}
                    \vspace{-1mm}
                \item{The vegetable on the cutting board has been chopped into pieces;}
                    \vspace{-1mm}
                \item{The agent is not next to a cutting board;}
                    \vspace{-1mm}
                \item{There is no unchopped vegetable on the cutting board;}
                    \vspace{-1mm}
                \item{The agent holds something in hand.}
            \end{itemize}}
            \vspace{-1mm}
        \item{\emph{\textbf{Get-Lettuce}}, \emph{\textbf{Get-Tomato}}, and \emph{\textbf{Get-Onion}}, navigate the agent to the latest observed position of the vegetable, and pick the vegetable up if it is there; otherwise, the agent moves to check the initial position of the vegetable. The corresponding termination conditions are listed below:
            \begin{itemize}
                    \vspace{-1mm}
                \item{The agent successfully picks up a chopped or unchopped vegetable;}
                    \vspace{-1mm}
                \item{The agent observes the target vegetable is held by another agent or itself;}
                    \vspace{-1mm}
                \item{The agent is holding something else in hand;}
                    \vspace{-1mm}
                \item{The agent's path to the vegetable is blocked by another agent;}
                    \vspace{-1mm}
                \item{The agent does not find the vegetable either at the latest observed location or the initial location;}
                    \vspace{-1mm}
                \item{The agent attempts to enter the same cell with another agent, but has a lower priority than another agent.}
            \end{itemize}}
            \vspace{-1mm}
        \item{\emph{\textbf{Get-Plate-1/2}}, navigates the agent to the latest observed position of the plate, and picks the vegetable up if it is there; otherwise, the agent moves to check the initial position of the vegetable. The corresponding termination conditions are listed below:
            \begin{itemize}
                    \vspace{-1mm}
                \item{The agent successfully picks up a plate;}
                    \vspace{-1mm}
                \item{The agent observes the target plate is held by another agent or itself;}
                    \vspace{-1mm}
                \item{The agent is holding something else in hand;}
                    \vspace{-1mm}
                \item{The agent's path to the plate is blocked by another agent;}
                    \vspace{-1mm}
                \item{The agent does not find the plate either at the latest observed location or at the initial location;}
                    \vspace{-1mm}
                \item{The agent attempts to enter the same cell with another agent but has a lower priority than another agent.}
            \end{itemize}}
            \vspace{-1mm}
        \item{\emph{\textbf{Go-Cut-Board-1/2}}, navigates the agent to the corresponding cutting board with the following termination conditions:
            \begin{itemize}
                    \vspace{-1mm}
                \item{The agent stops in front of the corresponding cutting board, and places an in-hand item on it if the cutting board is not occupied;}
                    \vspace{-1mm}
                \item{If any other agent is using the target cutting board, the agent stops next to the teammate;}
                    \vspace{-1mm}
                \item{The agent attempts to enter the same cell with another agent but has a lower priority than another agent.}
            \end{itemize}}
            \vspace{-1mm}
        \item{\emph{\textbf{Go-Counter}} (only available in Overcook-B, Fig.~\ref{domain_OB}), navigates the agent to the center cell in the middle of the map when the cell is not occupied, otherwise it moves to an adjacent cell. If the agent is holding an object the object will be placed. If an object is in the cell, the object will be picked up.} 
            \vspace{-1mm}
        \item{\emph{\textbf{Deliver}}, navigates the agent to the `star' cell for delivering with several possible termination conditions:
            \begin{itemize}
                    \vspace{-1mm}
                \item{The agent places the in-hand item on the cell if it is holding any item;}
                    \vspace{-1mm}
                \item{If any other agent is standing in front of the `star' cell, the agent stops next to the teammate;}
                    \vspace{-1mm}
                \item{The agent attempts to enter the same cell with another agent, but has a lower priority than another agent.}
            \end{itemize}}
\end{itemize}

\textbf{Observation Space}: The macro-observation space for each agent is the same as the primitive observation space. 
Agents are only allowed to observe the \emph{positions} and \emph{status} of the entities within a $5\times5$ view centered on the agent.
The initial position of all the items are known to agents.\\

\textbf{Dynamics}: The transition in this task is deterministic. If an agent delivers any wrong item, the item will be reset to its initial position. From the low-level perspective, to chop a vegetable into pieces on a cutting board, the agent needs to stand next to the cutting board and executes \emph{left} three times. Only the chopped vegetable can be put on a plate.\\

\textbf{Reward}: $+10$ for chopping a vegetable, $+200$ terminal reward for delivering a tomato-lettuce-onion salad, $-5$ for delivering any wrong entity, and $-0.1$ for every timestep.\\

\textbf{Episode Termination}: Each episode terminates either when agents successfully deliver a tomato-lettuce-onion salad or reaching the maximal time steps, 200. 

\clearpage

\textbf{Results}
\begin{figure*}[h!]
    \centering
    \captionsetup[subfigure]{labelformat=empty}
    \subcaptionbox{}
        [0.9\linewidth]{\includegraphics[height=0.36cm]{results/legend/dec_comp.png}\vspace{-2mm}}
    ~
    \centering
    \subcaptionbox{}
        [0.32\linewidth]{\includegraphics[height=3cm]{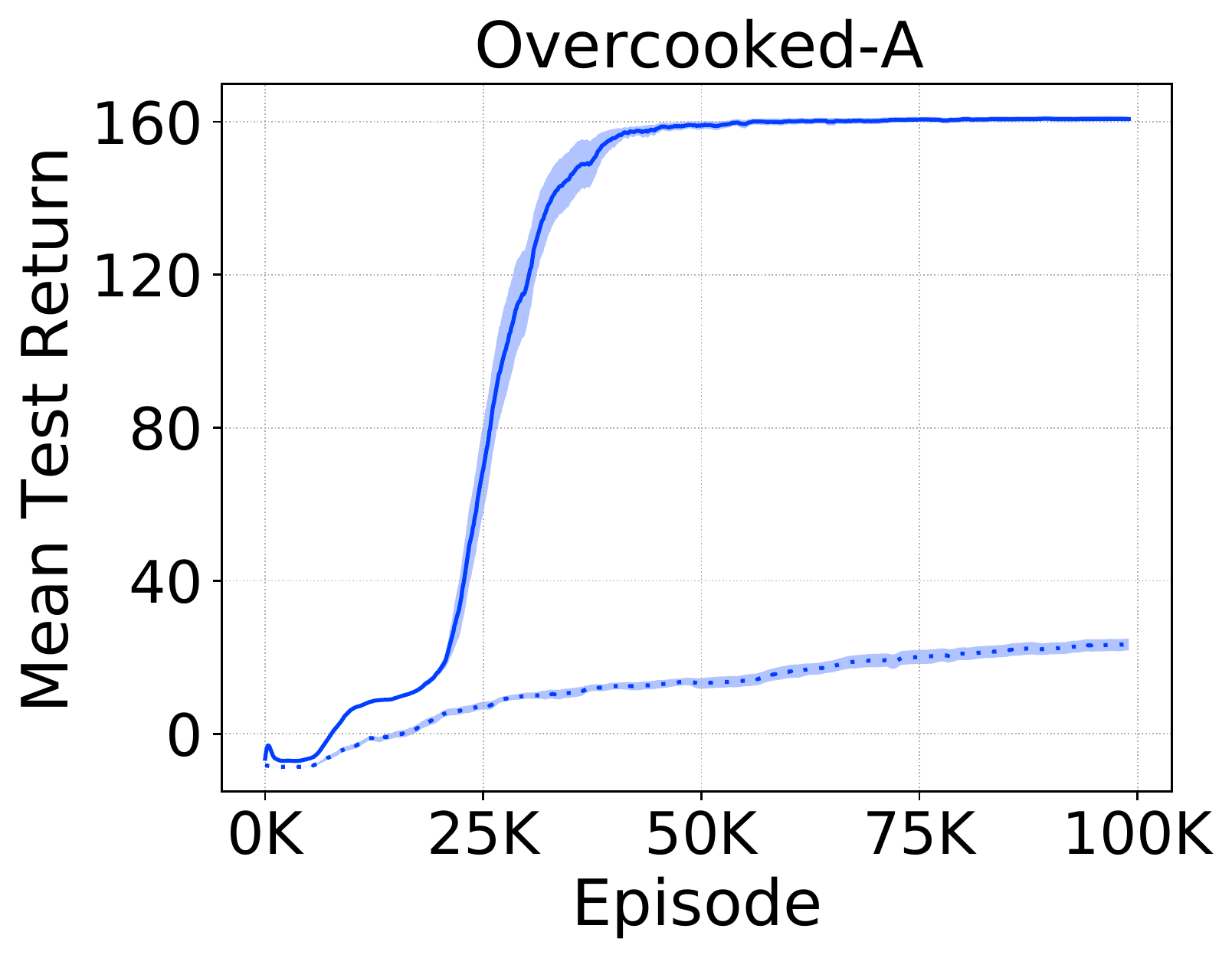}\vspace{-2mm}}
    ~
    \centering
    \subcaptionbox{}
        [0.32\linewidth]{\includegraphics[height=3cm]{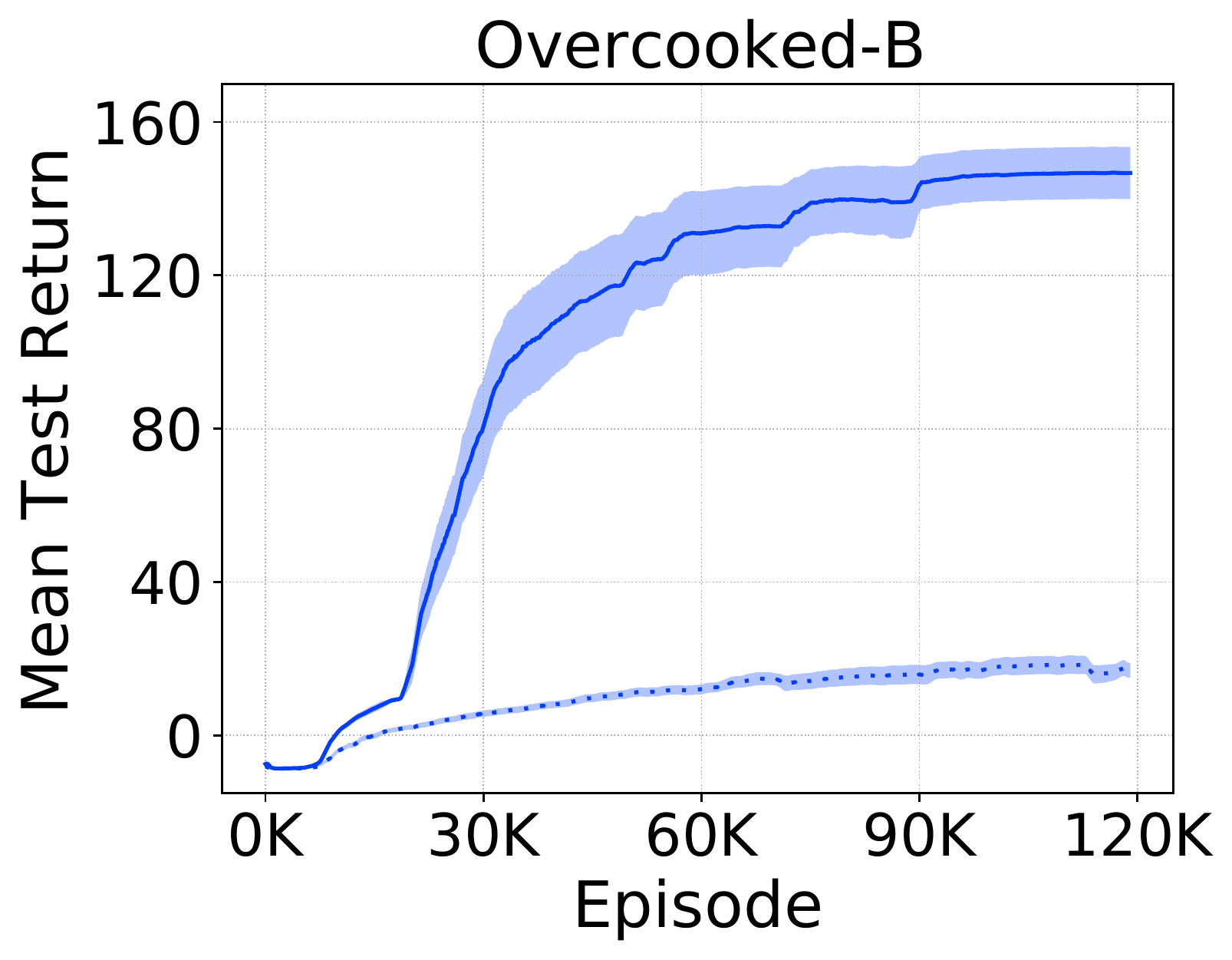}\vspace{-2mm}}
    ~
    \centering
    \subcaptionbox{}
        [0.32\linewidth]{\includegraphics[height=3cm]{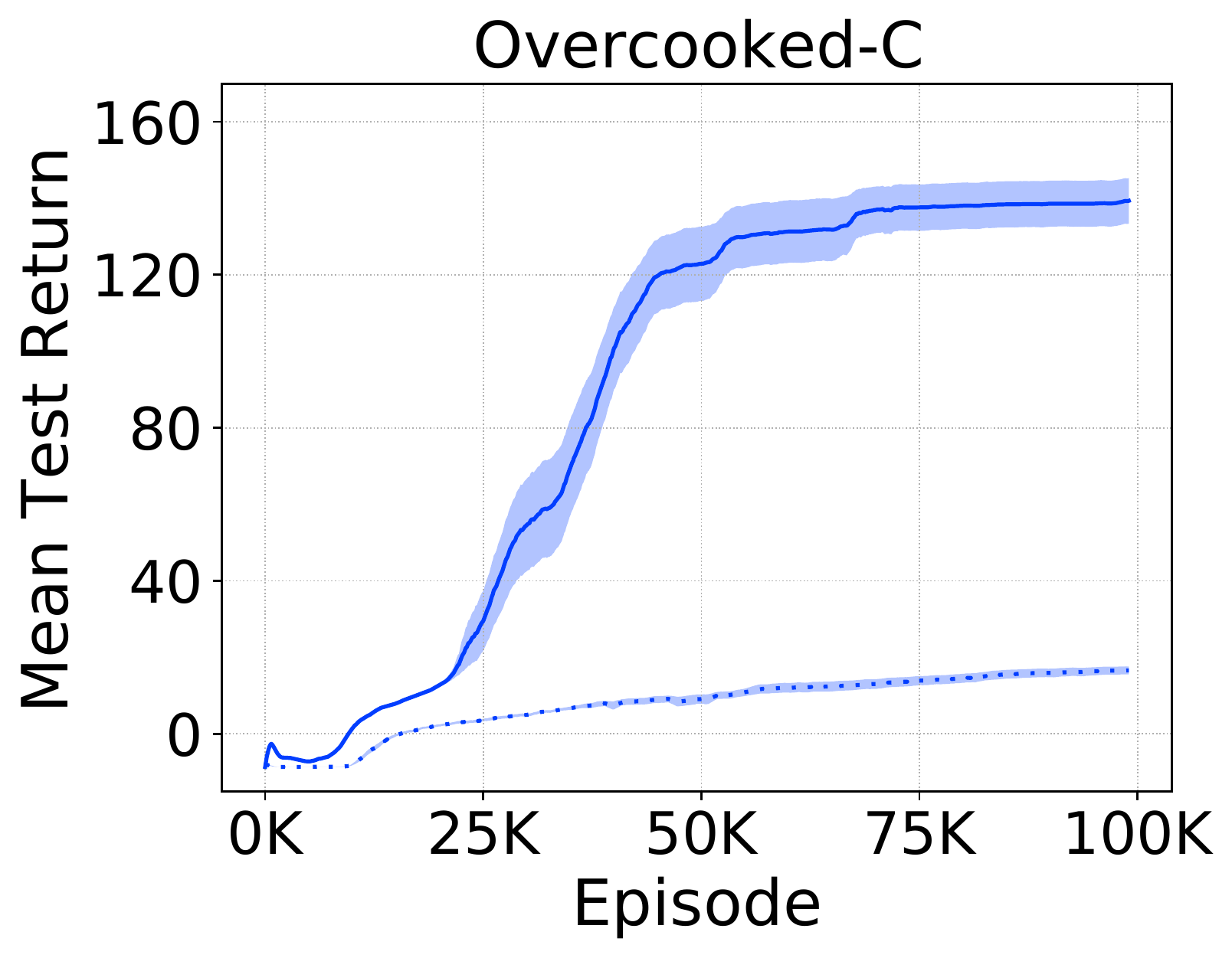}\vspace{-2mm}}
    \caption{Decentralized learning with macro-actions vs primitive-actions.}
    \label{fig:Overcooked_dec_comp}
\end{figure*}

\begin{figure*}[h!]
    \centering
    \captionsetup[subfigure]{labelformat=empty}
    \subcaptionbox{}
        [0.9\linewidth]{\includegraphics[height=0.36cm]{results/legend/cen_comp.png}\vspace{-2mm}}
    ~
    \centering
    \subcaptionbox{}
        [0.31\linewidth]{\includegraphics[height=3cm]{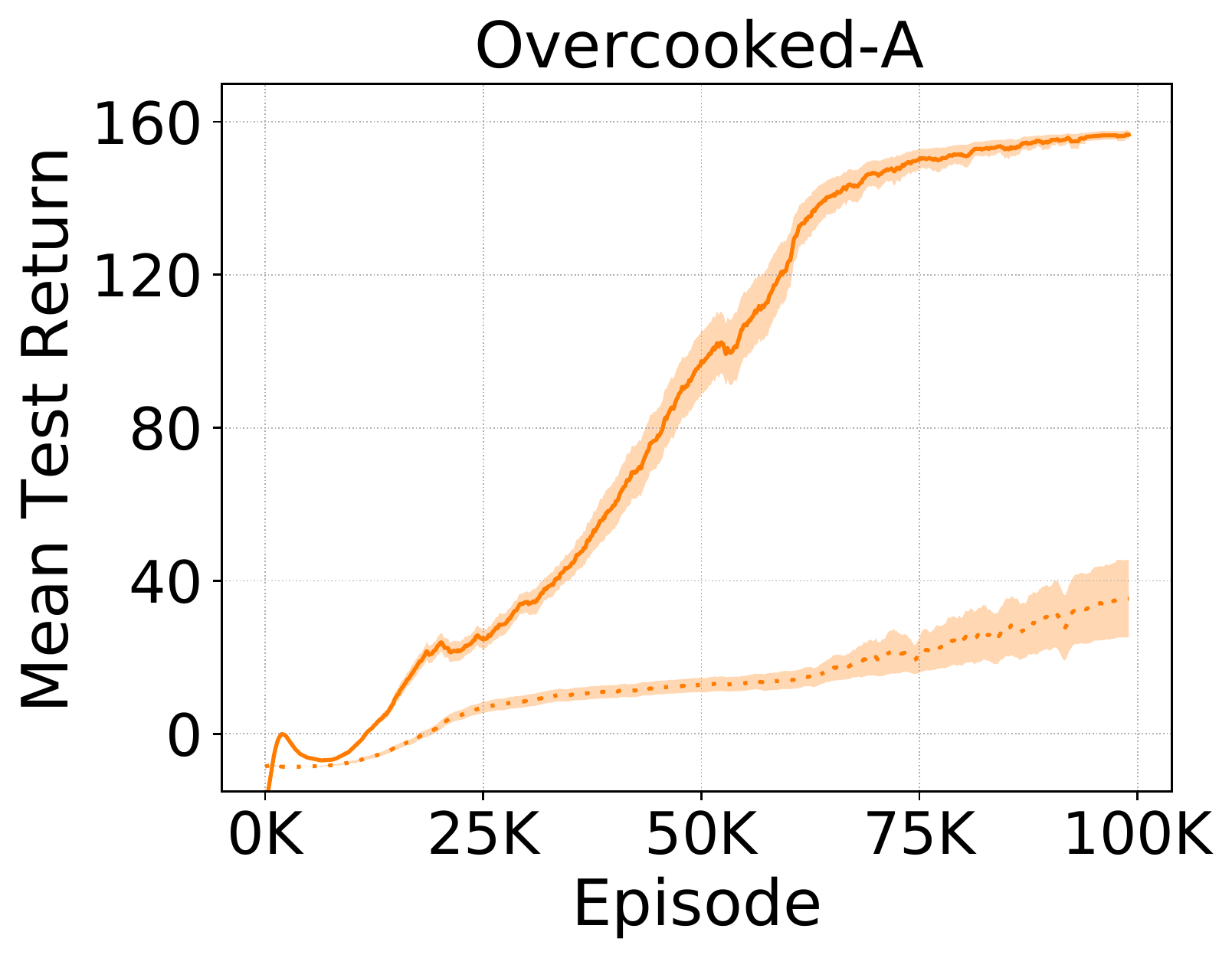}\vspace{-2mm}}
    ~
    \centering
    \subcaptionbox{}
        [0.31\linewidth]{\includegraphics[height=3cm]{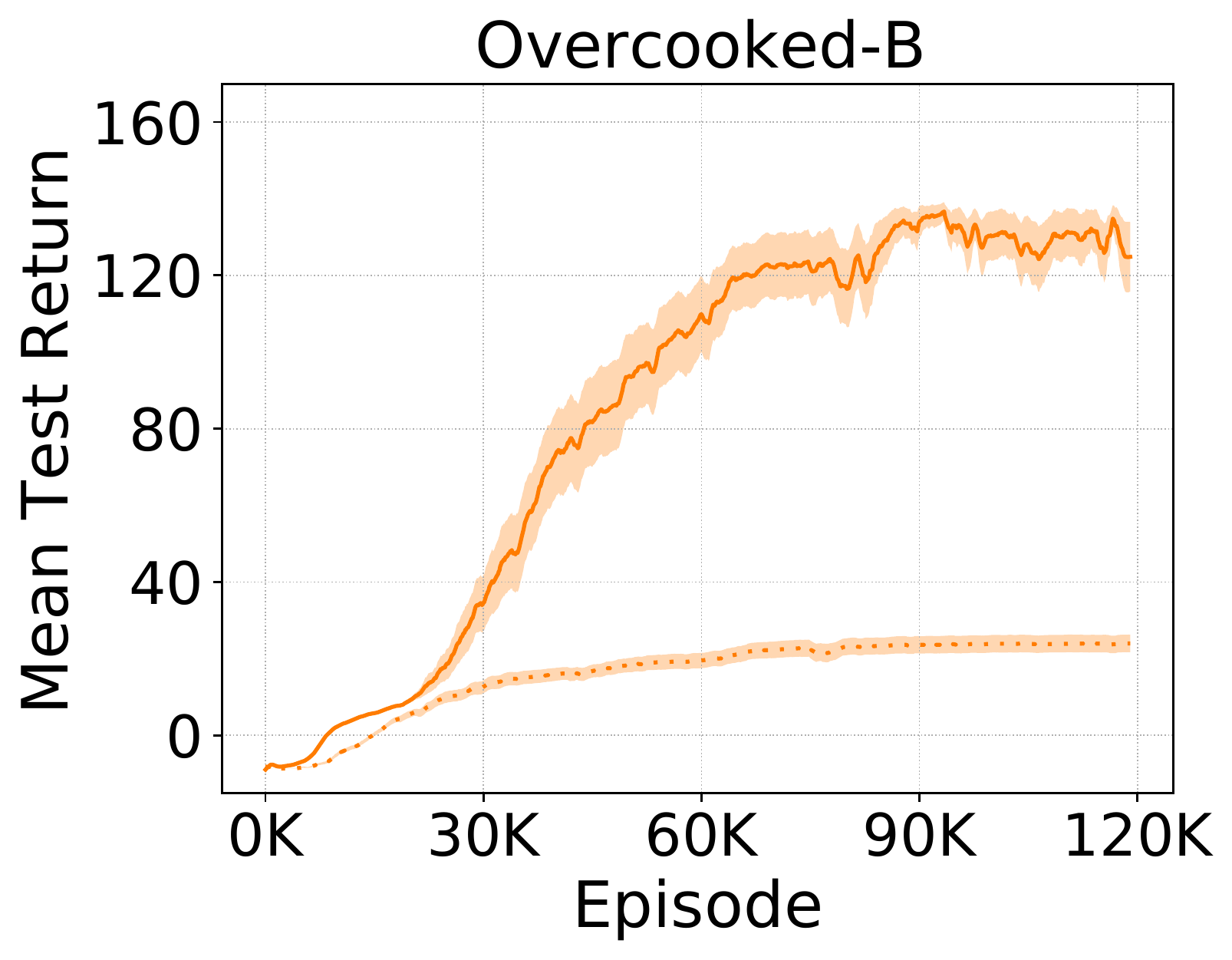}\vspace{-2mm}}
    ~
    \centering
    \subcaptionbox{}
        [0.31\linewidth]{\includegraphics[height=3cm]{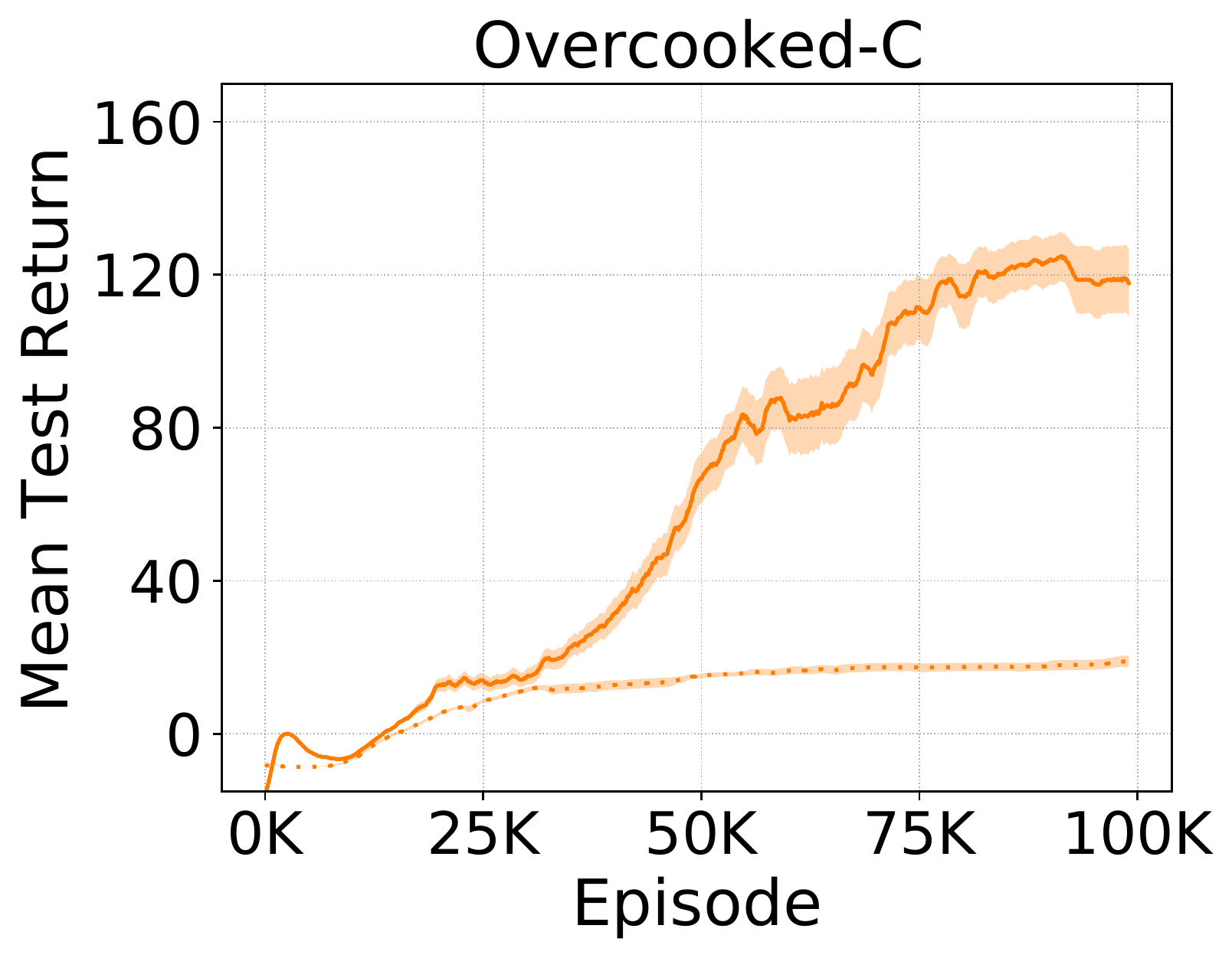}\vspace{-2mm}}
    \caption{Centralized learning with macro-actions vs primitive-actions.}
    \label{fig:Overcooked_cen_comp}
\end{figure*}

\begin{figure*}[h!]
    \centering
    \captionsetup[subfigure]{labelformat=empty}
    \subcaptionbox{}
        [0.9\linewidth]{\includegraphics[height=0.36cm]{results/legend/ctde_comp.png}\vspace{-2mm}}
    ~
    \centering
    \subcaptionbox{}
        [0.31\linewidth]{\includegraphics[height=3cm]{results/Overcooked/mapA_ctde_v1_large.pdf}\vspace{-2mm}}
    ~
    \centering
    \subcaptionbox{}
        [0.31\linewidth]{\includegraphics[height=3cm]{results/Overcooked/mapB_ctde_v1_large.pdf}\vspace{-2mm}}
    ~
    \centering
    \subcaptionbox{}
        [0.31\linewidth]{\includegraphics[height=3cm]{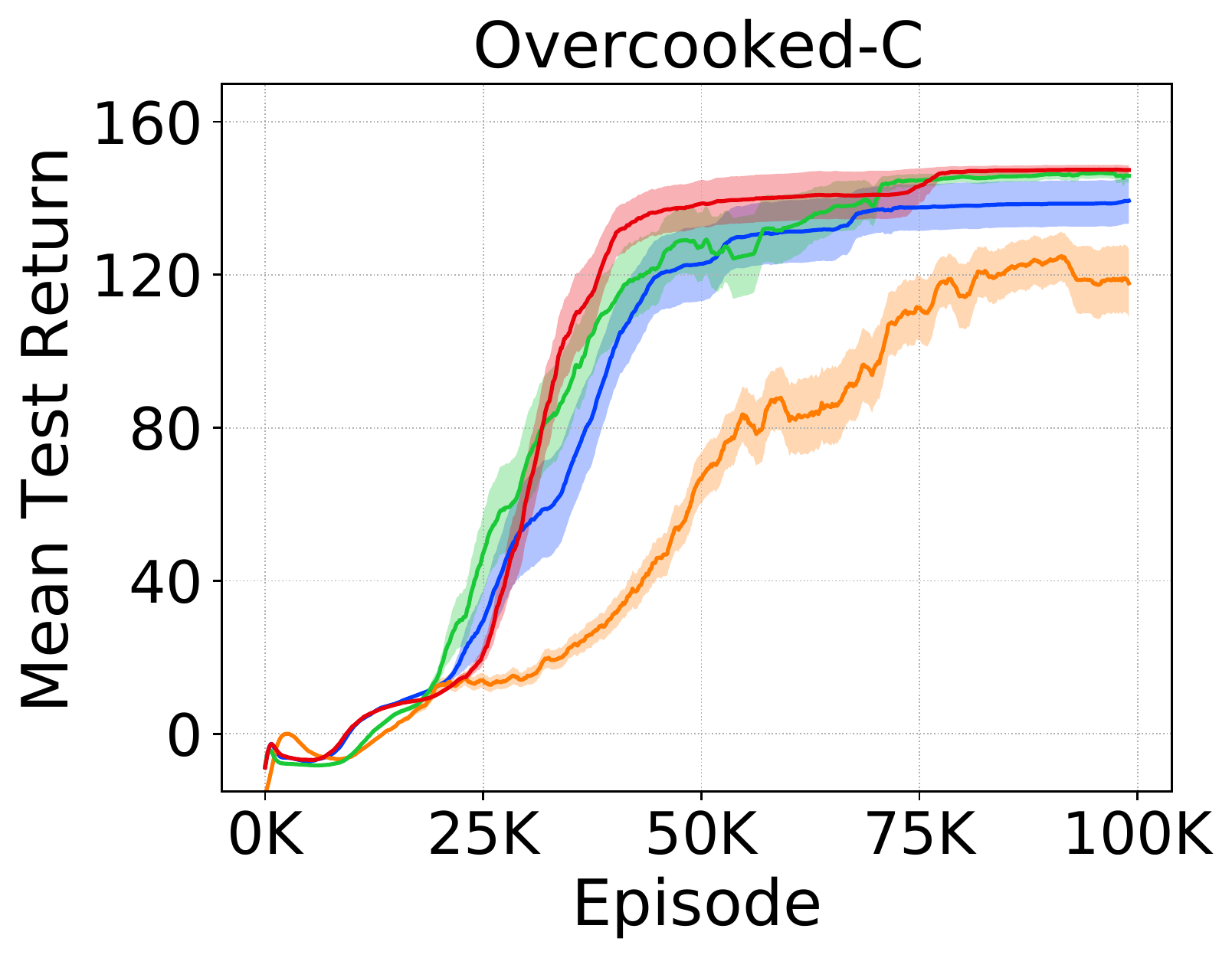}\vspace{-2mm}}
     \caption{Comparison of macro-action-based multi-agent actor-critic methods.}
    \label{fig:Overcooked_cen_comp}
\end{figure*}

\begin{figure*}[h!]
    \centering
    \captionsetup[subfigure]{labelformat=empty}
    \subcaptionbox{}
        [0.9\linewidth]{\includegraphics[height=0.36cm]{results/legend/qvspg_colors.png}\vspace{-2mm}}
    ~
    \centering
    \subcaptionbox{}
        [0.31\linewidth]{\includegraphics[height=3cm]{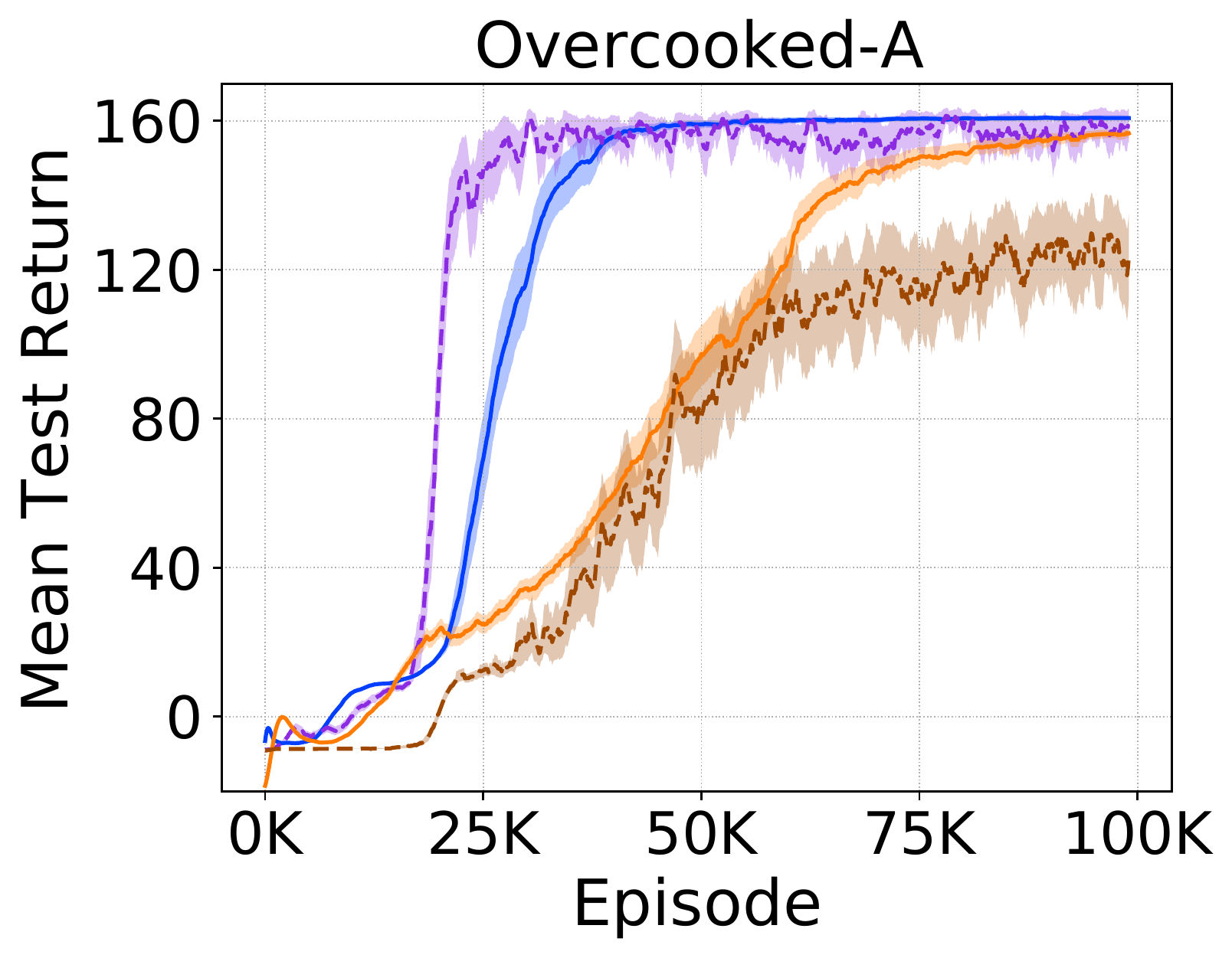}\vspace{-2mm}}
    ~
    \centering
    \subcaptionbox{}
        [0.31\linewidth]{\includegraphics[height=3cm]{results/Overcooked/mapB_q_ac_3agent42_colors.pdf}\vspace{-2mm}}
     \caption{Comparisons of macro-action-based actor-critic methods and value-based methods.}
    \label{fig:Overcooked_q_ac}
\end{figure*}
\clearpage

\clearpage
\subsection{Warehouse Tool Delivery}
\label{A-WTD}\hfill

\begin{figure*}[h!]
    \centering
    \captionsetup[subfigure]{labelformat=empty}
    \centering
    \subcaptionbox{(a) Warehouse-A\vspace{2mm}}
        [0.4\linewidth]{\includegraphics[height=2.6cm]{results/WTD/wtd_a_small.png}}
    ~
    \centering
    \subcaptionbox{(b) Warehouse-B\vspace{2mm}}
        [0.4\linewidth]{\includegraphics[height=2.6cm]{results/WTD/wtd_c_small.png}}
    ~
    \centering
    \subcaptionbox{(c) Warehouse-C}
        [0.4\linewidth]{\includegraphics[height=2.6cm]{results/WTD/wtd_e_small.png}}
    ~
    \centering
    \subcaptionbox{(d) Warehouse-D}
        [0.4\linewidth]{\includegraphics[height=2.6cm]{results/WTD/wtd_d_small.png}}
    ~
    \centering
    \subcaptionbox{(e) Warehouse-E\vspace{2mm}}
        [0.28\linewidth]{\includegraphics[height=2.6cm]{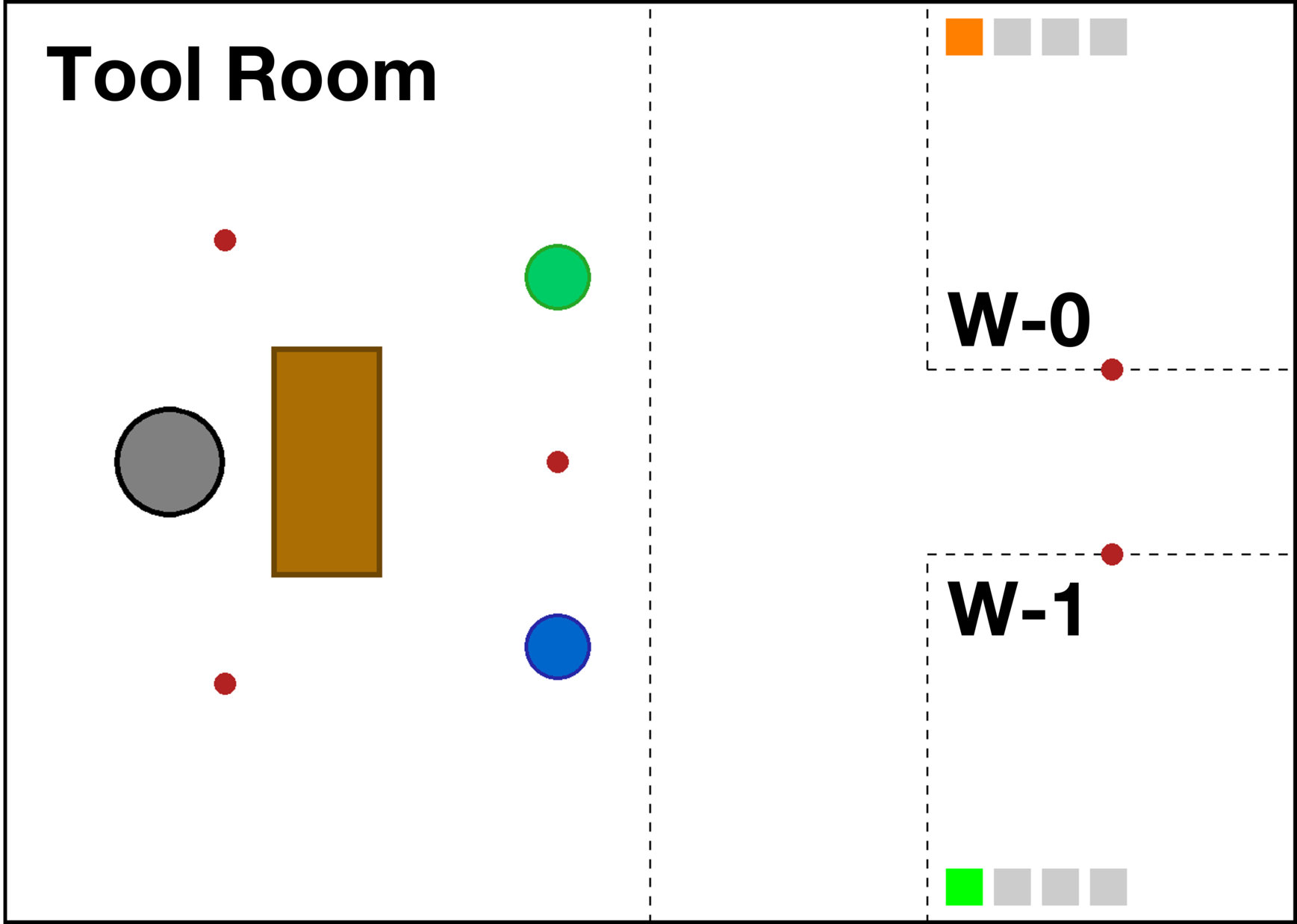}}
    \caption{Experimental environments.}
    \label{wtd_domains}
\end{figure*}

In this Warehouse Tool Delivery domain, we consider five different scenarios shown in Fig.~\ref{wtd_domains}. To further examine the scalability of our methods and the effectiveness of Mac-IAICC on handling more noisy asynchronous terminations over robots, we consider many variants in terms of both the number of robots and the number of humans as well as having faster human(orange) in the environment.\\ 

\textbf{Goal.}
Under all scenarios, in each workshop, a human is working on an assembly task involving 4 subtasks to be finished (each subtask takes amount of primitive time steps). At the beginning, each human has already got the tool for the first subtask and immediately starts. In order to continue, the human needs a particular tool for each following subtask. In the scenarios, humans either work in the same speed (Fig.~\ref{wtd_domains}a, \ref{wtd_domains}b, \ref{wtd_domains}d) or have one of them working faster (the orange one in Fig.\ref{wtd_domains}c and \ref{wtd_domains}e). 
A team of robots includes a robot arm (gray) with the duty of finding tools for each human on the table (brown) and passing them to mobile robots (green, blue and yellow) who are responsible for delivering tools to the humans. 
The objective of the robots is to assist the humans to finish their assembly tasks as soon as possible by finding and delivering the correct tools in the proper order. 
To make this problem more challenging, the correct tools needed by each human are unknown to robots, which has to be learned during training in order to perform timely delivery without letting humans wait.\\ 

\textbf{State}. The environment is either a $5\times7$ (Fig.~\ref{wtd_domains}a and ~\ref{wtd_domains}e) or a $5\times9$ (Fig.~\ref{wtd_domains}b - \ref{wtd_domains}d) continuous space. A global state consists of the 2D position of each mobile robots, the execution status of the arm robot's current macro-action (e.g how munch steps are left for completing the macro-action, but in real-world, this should be the angle and speed of each arm's joint), the subtask each human is working with a percentage indicating the progress of the subtask, and the position of each tools (either on the brown table or carried by a mobile robot). The initial state of every episode is deterministic as shown in Fig.~\ref{wtd_domains}, where humans always start from the first step.\\ 

\textbf{Macro-Action Space}. 

The available macro-actions for each mobile robot include:

$\bullet$ \emph{\textbf{Go-W(i)}}, navigates to the red waypoint at the corresponding workshop; 

$\bullet$ \emph{\textbf{Go-TR}}, navigates to the red waypoint (covered by the blue robot in Fig.~\ref{wtd_domains}c and~\ref{wtd_domains}d) at the right side of the tool room; 

$\bullet$ \emph{\textbf{Get-Tool}}, navigates to a pre-allocated waypoint besides the arm robot and waits over there until either 10 timesteps have passed  or receiving a tool from the gray robot. 

The available macro-actions for the arm robot include:

$\bullet$ \emph{\textbf{Search-Tool(i)}}, takes 6 timesteps to find tool $i$ and place it in a staging area (containing at most two tools) when the area is not fully occupied, otherwise freezes the robot for the same amount of time; 

$\bullet$ \emph{\textbf{Pass-to-M(i)}}, takes 4 timesteps to pass the first found tool to a mobile robot from the staging area; 

$\bullet$ \emph{\textbf{Wait-M}}, takes 1 timestep to wait for mobile robots coming.\\

\textbf{Macro-Observation Space}. 

The arm robot's macro-observation include the information about \emph{the type} of each tool in the staging area and \emph{which mobile robot} is waiting beside. 

Each mobile robot always observes its own \emph{position} and \emph{the type} of each tool carried by itself, while observes \emph{the number} of tools in the staging area or \emph{the subtask} a human working on only when locating at the tool room or the workshop respectively.\\  

\textbf{Dynamics}. Transitions are deterministic. Each mobile robot moves in a fixed velocity 0.8 and is only allowed to receive tools from the arm robot rather than from humans. 
Note that each human is only allowed to possess the tool for the next subtask from a mobile robot when the robot locates at the corresponding workshop and carries the correct tool. 
Humans are not allowed to pass tool back to mobile robots. There are enough tools for humans on the table in tool room, such that the number of each type of tool exactly matches with the number of humans in the environment.
Human cannot start the next subtask without obtaining the correct tool. Humans' dynamics on their tasks are shown in Table~\ref{table:humanDyn}.


\begin{table}[h!]
\caption {The number of time steps taken by each human on each subtask in scenarios.}
\label{table:humanDyn}
\centering
\begin{tabular}{lccccc}
\toprule
Scenarios       & Warehouse-A     & Warehouse-B     & Warehouse-C  & Warehouse-D & Warehouse-E \\
\cmidrule(r){1-6}
Human-0         & $[27,20,20,20]$   & $[40,40,40,40]$   & $[38,38,38,38]$  & $[40,40,40,40]$ & $[18,15,15,15]$\\ 
Human-1         & $[27,20,20,20]$   & $[40,40,40,40]$   & $[38,38,38,38]$  & $[40,40,40,40]$ & $[48,18,15,15]$\\
Human-2         & N/A             & $[40,40,40,40]$   & $[27,27,27,27]$  & $[40,40,40,40]$  &  N/A\\
Human-3         & N/A             & N/A             & N/A            & $[40,40,40,40]$  &  N/A\\
\bottomrule
\end{tabular}
\end{table}

\textbf{Rewards}.
The team receives a $+100$ reward when a correct tool is delivered to a human in time while getting an extra $-20$ penalty for a delayed delivery such that the human has paused over there. A $-10$ reward occurs when the gray robot does \emph{\textbf{Pass-to-M(i)}} but the mobile robot $i$ is not next to it, and a $-1$ reward is issued every time step.\\ 

\textbf{Episode Termination}. Each episode terminates when all humans obtained all the correct tools for all subtasks, otherwise, the episode will run until the maximal time steps (200 for Warehouse-A and E, 250 for Warehouse-B and C, 300 for Warehous-D).

\clearpage
\textbf{Results}

\begin{figure*}[h!]
    \centering
    \captionsetup[subfigure]{labelformat=empty}
    \centering
    \subcaptionbox{}
        [0.9\linewidth]{\includegraphics[height=0.36cm]{results/legend/ctde_comp.png}}
    ~
    \centering
    \subcaptionbox{}
        [0.3\linewidth]{\includegraphics[height=3cm]{results/WTD/wtdA_ctde_comp.pdf}}
    ~
    \centering
    \subcaptionbox{}
        [0.3\linewidth]{\includegraphics[height=3cm]{results/WTD/wtdCB_ctde_comp.pdf}}
    ~
    \centering
    \subcaptionbox{}
        [0.3\linewidth]{\includegraphics[height=3cm]{results/WTD/wtdDC_ctde_comp.pdf}}
    ~
    \centering
    \subcaptionbox{}
        [0.3\linewidth]{\includegraphics[height=3cm]{results/WTD/wtdED_ctde_comp.pdf}}
    ~
    \centering
    \subcaptionbox{}
        [0.3\linewidth]{\includegraphics[height=3cm]{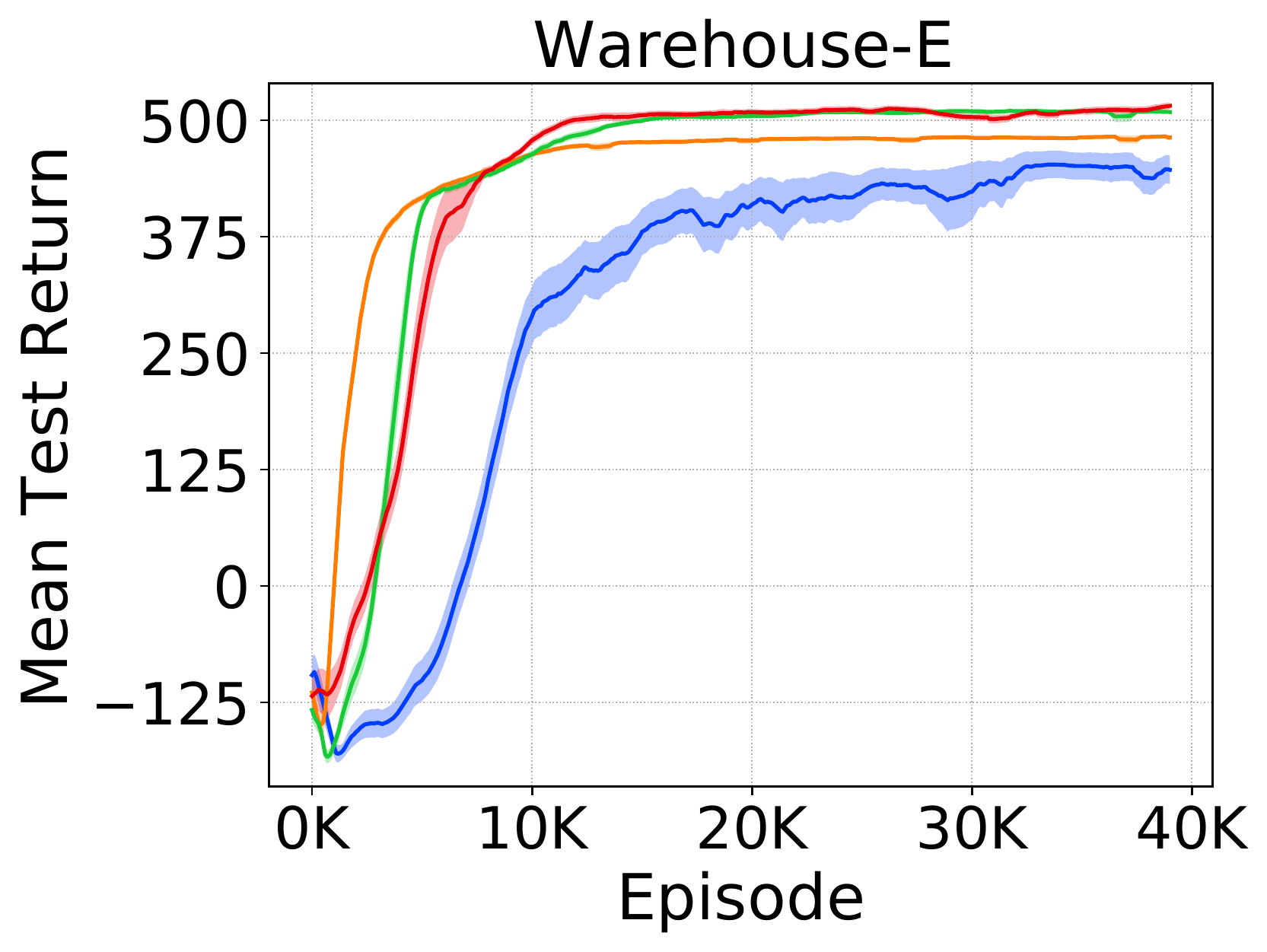}}
    \caption{Comparison of macro-action-based multi-agent actor-critic methods.}
    \label{wtd_ctde_comp}
\end{figure*}

\textbf{Ablation Study}



\begin{minipage}{\textwidth}
  \begin{minipage}[b]{0.49\textwidth}
    \centering
    \includegraphics[height=3cm]{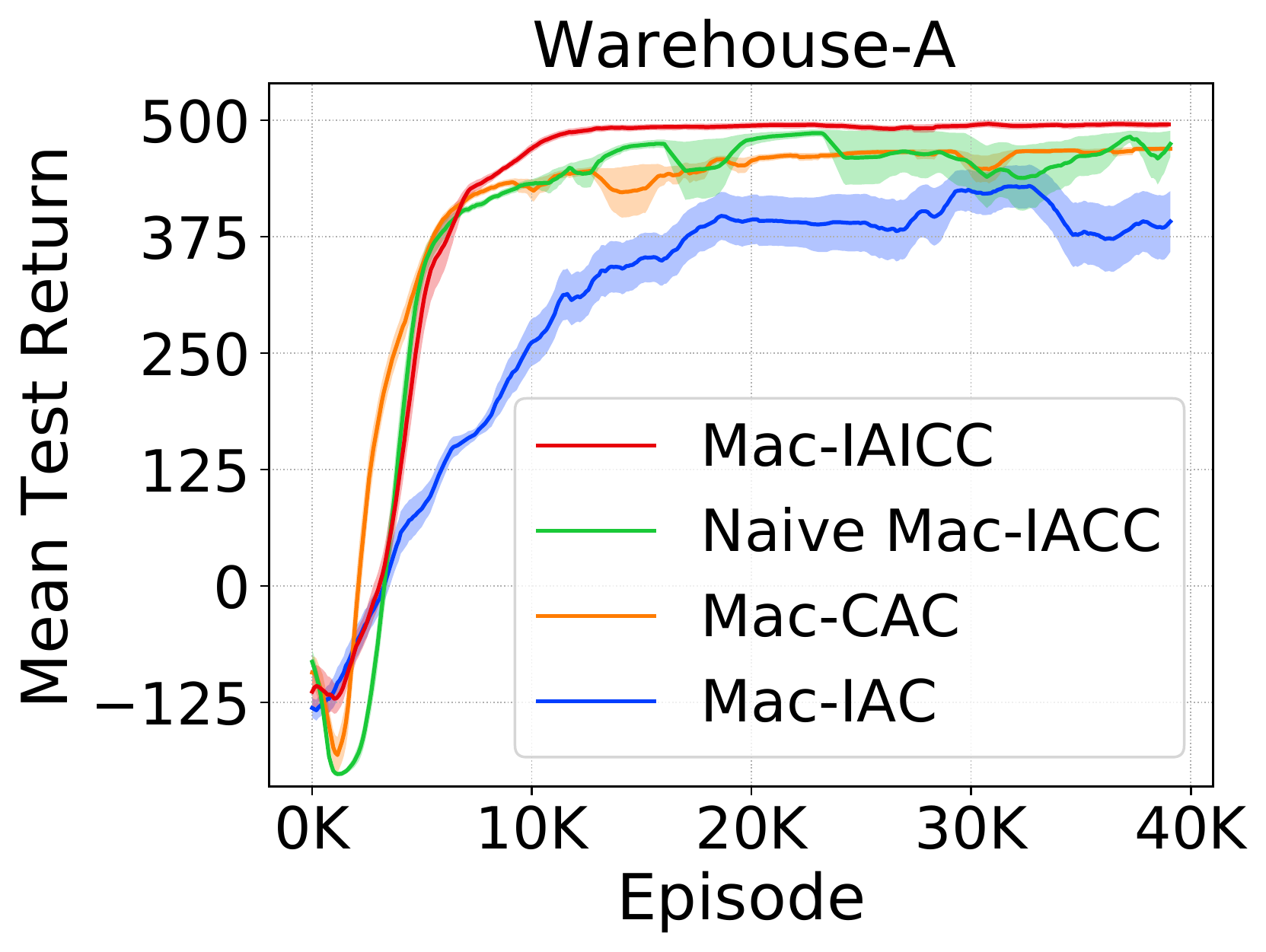}
    \captionof{figure}{Results of an ablation study.}
    \label{ablation}
  \end{minipage}
  \hfill
  \begin{minipage}[b]{0.49\textwidth}
    \centering
        \begin{tabular}{lcccccc}
        \toprule
        Scenarios & Ablation Experiment \\
        \cmidrule(r){1-1}
        Human-0         & $[18,18,18,18]$ \\ 
        Human-1         & $[18,18,18,18]$\\
        Human-2         & N/A \\
        Human-3         & N/A\\
        \bottomrule
        \end{tabular}
      \captionof{table}{The number of time steps taken by each human in the ablation study. }
    \end{minipage}
\end{minipage}

We also conducted an ablation experiment in Warehouse-A, where two humans still operate at the same speed on their tasks but faster than the original setting. Such a change makes agents' learning more difficult, because the probability of having a delayed delivery for each tool grows,
especially when agents are exploring. Agents likely receives more penalty during training.    
Fig.~\ref{ablation} shows the learning quality of Naive Mac-IACC degrades markedly and becomes much less stable with higher variance than its performance in the original domain configuration (shown in Fig.~\ref{wtd_ctde_comp}). In contrast, Mac-IAICC remains its high-quality performance, which reveals its robustness to noisy penalty signals and further proves the advantage of separately training a centralized critic depending on each agent's own macro-action terminations. Both Mac-CAC and Mac-IAC still cannot rival Mac-IAICC. 

\begin{figure*}[h!]
    \centering
    \captionsetup[subfigure]{labelformat=empty}
    \centering
    \subcaptionbox{}
        [0.9\linewidth]{\includegraphics[height=0.36cm]{results/legend/qvspg_colors.png}}
    ~
    \centering
    \subcaptionbox{}
        [0.3\linewidth]{\includegraphics[height=3cm]{results/WTD/wtd_A_qvspg_colors.pdf}}
    ~
    \centering
    \subcaptionbox{}
        [0.3\linewidth]{\includegraphics[height=3cm]{results/WTD/wtd_cb_qvspg.pdf}}
    ~
    \centering
    \subcaptionbox{}
        [0.3\linewidth]{\includegraphics[height=3cm]{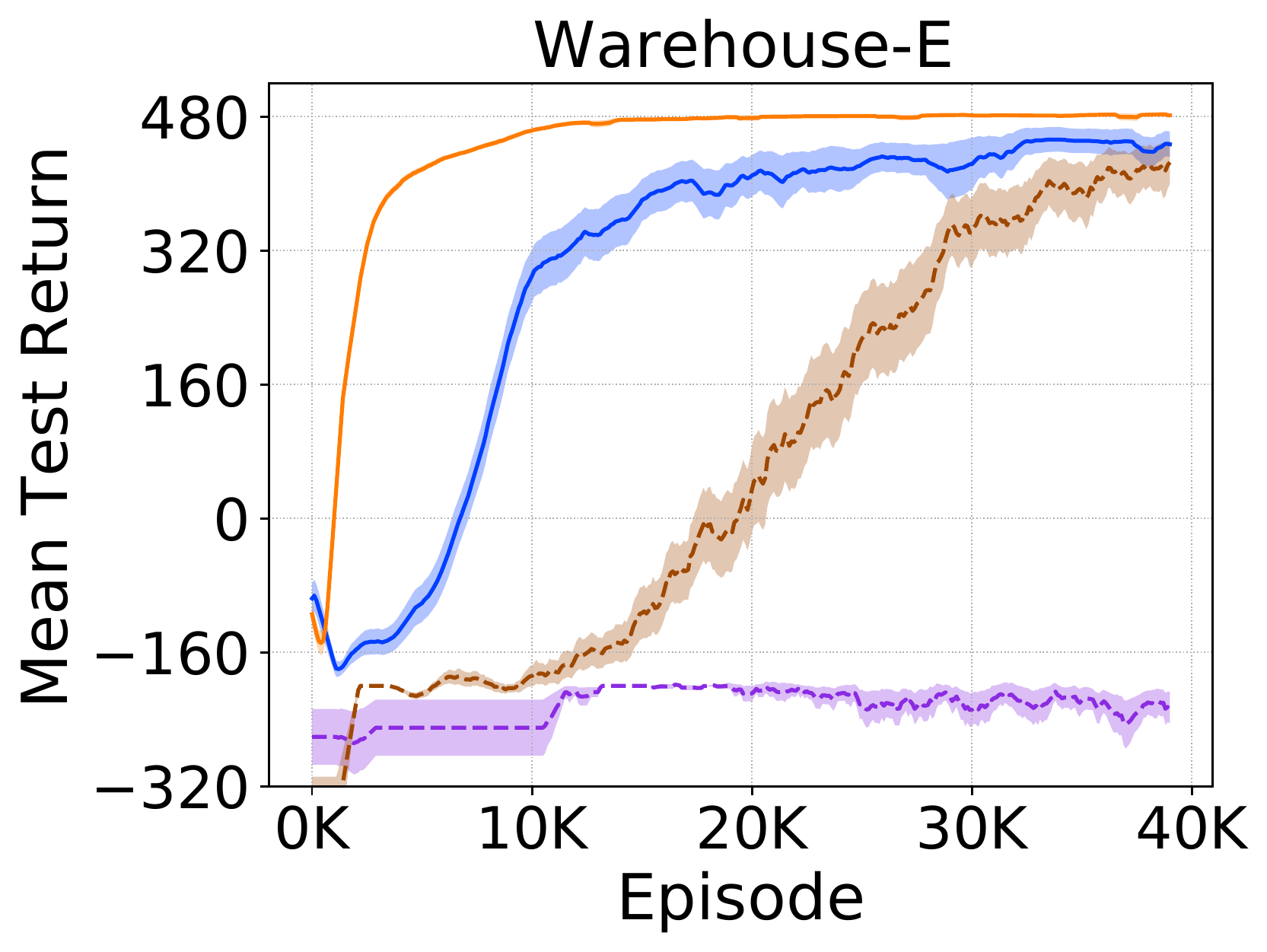}}
    \caption{Comparison of macro-action-based actor-critic methods and value-based methods.}
    \label{wtd_qvspg}
\end{figure*}
\clearpage
\section{Training Details}
Our results are generated by running on a cluster of computer nodes under "CentOS Linux" operating system. We use the CPUs including "Dual Intel Xeon E5-2650", "Dual Intel Xeon E5-2680 v2", "Dual Intel Xeon E5-2690 v3".
\label{A-Train}
\subsection{Network Architecture}
\label{A-A}\hfill

For all domains, all methods apply the same neural network architecture for both actor \& critic network and Q-network. Each of them consists of two fully connected (FC) layers with Leaky-Relu activation function, one GRU layer~\citep{GRU} and one more FC layer followed by an output layer. The number of neurons in each layer for Decentralized(Dec) or Centralized(Cen) actor, critic or Q-network are shown in Table \ref{table:architecture}. Empirical experiments show that centralized actor and critic usually need more neurons to deal with larger joint macro-observation and macro-action spaces. 

\begin{table}[h!]
\caption {Number of neurons on each layer in networks for all methods in domains}
\label{table:architecture}
\centering
\begin{tabular}{ccccccc}
\toprule
Domain          & \multicolumn{2}{c}{Box Pushing} & \multicolumn{2}{c}{Overcooked} & \multicolumn{2}{c}{Warehouse} \\
\midrule
Actor \& Critic \& Q-network & Dec   & Cen   & Dec   & Cen  & Dec  & Cen  \\
\cmidrule{1-7}
MLP-1           & 32              & 32            & 32              & 128          & 32             & 32           \\
MLP-2           & 32              & 32            & 32              & 128           & 32             & 32           \\
GRU             & 32              & 64            & 32              & 64           & 32             & 64           \\
MLP-3           & 32              & 32            & 32              & 64           & 32             & 32           \\ 
\bottomrule
\end{tabular}
\end{table}






\subsection{Hyper-Parameters for macro-action-based actor-critic methods}
\label{A-Hyper}\hfill

In following subsections, we first list the hyper-parameter candidates used for tuning each method via grid search in the corresponding domain, and then show the hyper-parameter table with the parameters used by each method achieving the best performance. We choose the best performance of each method depending on its final converged value as the first priority and the sample efficiency as the second.\\

\indent$\bullet$ Box Pushing:

\begin{table}[h!]
    \caption {Hyper-parameter candidates for grid search tuning.}
    \centering
    \begin{tabular}{lc}
    \toprule
        Learning rate pair (actor,critic) & (1e-3,3e-3), (1e-3,1e-3) (5e-4,3e-3), (5e-4,1e-3) \\ 
        & (5e-4,5e-4), (3e-4,3e-3)\\
        Episodes per train & 8, 16, 32 \\
        Target-net update freq (episode) & 32, 64, 128 \\
        N-step TD & 0, 3, 5\\
    \bottomrule
    \end{tabular} 
\end{table}

\begin{table}[h!]
    \caption {Hyper-parameter candidates for grid search tuning.}
    \centering
    \begin{tabular}{lc}
    \toprule
        Learning rate pair (actor,critic) & (1e-3,3e-3), (1e-3,1e-3) (5e-4,3e-3), (5e-4,1e-3) \\ 
        & (5e-4,5e-4), (3e-4,3e-3)\\
        Episodes per train & 48 \\
        Target-net update freq (episode) & 48, 96, 144 \\
        N-step TD & 0, 3, 5\\
    \bottomrule
    \end{tabular} 
\end{table}

\begin{table}[h!]
    \caption {Hyper-parameters used for methods in Box Pushing $6\times6$.}
    \centering
    \begin{tabular}{lcccccc}
    \toprule
        Parameter & IAC & CAC & Mac-IAC & Mac-CAC & Mac-NIACC & Mac-IAICC \\
    \cmidrule(r){2-7}
        Training Episodes & 40K & 40K & 40K & 40K & 40K & 40K\\
        Actor Learning rate &0.001 &0.0005 &0.0005 &0.0003 &0.0005 &0.0003\\
        Critic Learning rate &0.003 &0.0005 &0.001 &0.003 &0.001 &0.003\\
        Episodes per train &8 &8 &48 &48 &48 &48\\
        Target-net update freq &32 &64 &48 &144 &144 &96\\
        $\,\,\,\,\,\,\,\,\,\,\,\,$ (episode) & \\
        N-step TD &5 &5 &5 &5 &0 &0\\
        $\epsilon_{\text{start}}$ & 1 & 1 & 1 & 1 & 1 & 1\\
        $\epsilon_{\text{end}}$ & 0.01 & 0.01 & 0.01 & 0.01 & 0.01 & 0.01\\
        $\epsilon_{\text{decay}}$ (episode) & 4K & 4K & 4K & 4K & 4K & 4K \\
    \bottomrule
    \end{tabular} 
\end{table}

\begin{table}[h!]
    \caption {Hyper-parameters used for methods in Box Pushing $8\times8$.}
    \centering
    \begin{tabular}{lcccccc}
    \toprule
        Parameter & IAC & CAC & Mac-IAC & Mac-CAC & Mac-NIACC & Mac-IAICC \\
    \cmidrule(r){2-7}
        Training Episodes & 40K & 40K & 40K & 40K & 40K & 40K\\
        Actor Learning rate &0.001 &0.001 &0.001 &0.0005 &0.0005 &0.0003\\
        Critic Learning rate &0.003 &0.003 &0.003 &0.003 &0.001 &0.003\\
        Episodes per train &8 &8 &16 &48 &48 &48\\
        Target-net update freq &32 &32 &32 &48 &144 &144\\
        $\,\,\,\,\,\,\,\,\,\,\,\,$ (episode) & \\
        N-step TD &3 &0 &5 &3 &0 &0\\
        $\epsilon_{\text{start}}$ & 1 & 1 & 1 & 1 & 1 & 1\\
        $\epsilon_{\text{end}}$ & 0.01 & 0.01 & 0.01 & 0.01 & 0.01 & 0.01\\
        $\epsilon_{\text{decay}}$ (episode) & 4K & 4K & 4K & 4K & 4K & 4K \\
    \bottomrule
    \end{tabular} 
\end{table}

\begin{table}[h!]
    \caption {Hyper-parameters used for methods in Box Pushing $10\times10$.}
    \centering
    \begin{tabular}{lcccccc}
    \toprule
        Parameter & IAC & CAC & Mac-IAC & Mac-CAC & Mac-NIACC & Mac-IAICC \\
    \cmidrule(r){2-7}
        Training Episodes & 40K & 40K & 40K & 40K & 40K & 40K\\
        Actor Learning rate &0.001 &0.001 &0.001 &0.001 &0.0005 &0.0003\\
        Critic Learning rate &0.003 &0.003 &0.001 &0.003 &0.001 &0.003\\ 
        Episodes per train &8 &8 &32 &48 &48 &32\\
        Target-net update freq &64 &32 &32 &96 &144 &64\\
        $\,\,\,\,\,\,\,\,\,\,\,\,$ (episode) & \\
        N-step TD &0 &0 &5 &3 &0 &0\\
        $\epsilon_{\text{start}}$ & 1 & 1 & 1 & 1 & 1 & 1\\
        $\epsilon_{\text{end}}$ & 0.01 & 0.01 & 0.01 & 0.01 & 0.01 & 0.01\\
        $\epsilon_{\text{decay}}$ (episode) & 6K & 6K & 6K & 6K & 6K & 6K \\
    \bottomrule
    \end{tabular} 
\end{table}

\begin{table}[h!]
    \caption {Hyper-parameters used for methods in Box Pushing $12\times12$.}
    \centering
    \begin{tabular}{lcccccc}
    \toprule
        Parameter & IAC & CAC & Mac-IAC & Mac-CAC & Mac-NIACC & Mac-IAICC \\
    \cmidrule(r){2-7}
        Training Episodes & 40K & 40K & 40K & 40K & 40K & 40K\\
        Actor Learning rate &0.001 &0.001 &0.001 &0.0005 &0.0005 &0.0003\\
        Critic Learning rate &0.003 &0.003 &0.003 &0.0005 &0.001 &0.003\\
        Episodes per train &8 &8 &8 &32 &48 &32\\
        Target-net update freq &128 &128 &64 &64 &96 &128\\
        $\,\,\,\,\,\,\,\,\,\,\,\,$ (episode) & \\
        N-step TD &0 &0 &5 &3 &0 &0\\
        $\epsilon_{\text{start}}$ & 1 & 1 & 1 & 1 & 1 & 1\\
        $\epsilon_{\text{end}}$ & 0.01 & 0.01 & 0.01 & 0.01 & 0.01 & 0.01\\
        $\epsilon_{\text{decay}}$ (episode) & 6K & 6K & 6K & 6K & 6K & 6K \\
    \bottomrule
    \end{tabular} 
\end{table}

\clearpage

\begin{table}[h!]
    \caption {Hyper-parameters used for methods in Box Pushing $14\times14$.}
    \centering
    \begin{tabular}{lcccccc}
    \toprule
        Parameter & IAC & CAC & Mac-IAC & Mac-CAC & Mac-NIACC & Mac-IAICC \\
    \cmidrule(r){2-7}
        Training Episodes & 40K & 40K & 40K & 40K & 40K & 40K\\
        Actor Learning rate &0.001 &0.001 &0.001 &0.001 &0.001 &0.0003\\
        Critic Learning rate &0.003 &0.003 &0.003 &0.001 &0.003 &0.003\\
        Episodes per train &8 &8 &8 &48 &16 &32\\
        Target-net update freq &128 &64 &32 &96 &32 &64\\
        $\,\,\,\,\,\,\,\,\,\,\,\,$ (episode) & \\
        N-step TD &0 &0 &3 &3 &5 &0\\
        $\epsilon_{\text{start}}$ & 1 & 1 & 1 & 1 & 1 & 1\\
        $\epsilon_{\text{end}}$ & 0.01 & 0.01 & 0.01 & 0.01 & 0.01 & 0.01\\
        $\epsilon_{\text{decay}}$ (episode) & 8K & 8K & 8K & 8K & 8K & 8K \\
    \bottomrule
    \end{tabular} 
\end{table}

\indent$\bullet$ Overcooked:

\begin{table}[h!]
    \caption {Hyper-parameter candidates for grid search tuning.}
    \centering
    \begin{tabular}{lc}
    \toprule
        Learning rate pair (actor,critic) & (1e-4, 3e-3) (3e-4,3e-3)\\
        Episodes per train & 4 \\
        Target-net update freq (episode) & 8, 16, 32 \\
        N-step TD & 3, 5\\
    \bottomrule
    \end{tabular} 
\end{table}

\begin{table}[h!]
    \caption {Hyper-parameter candidates for grid search tuning.}
    \centering
    \begin{tabular}{lc}
    \toprule
        Learning rate pair (actor,critic) & (1e-4, 3e-3) (3e-4,3e-3)\\
        Episodes per train & 8, 16 \\
        Target-net update freq (episode) & 16, 32, 64 \\
        N-step TD & 3, 5\\
    \bottomrule
    \end{tabular} 
\end{table}

\begin{table}[h!]
    \caption {Hyper-parameters used for methods in Overcooked-A.}
    \centering
    \begin{tabular}{lcccccc}
    \toprule
        Parameter & IAC & CAC & Mac-IAC & Mac-CAC & Mac-NIACC & Mac-IAICC \\
    \cmidrule(r){2-7}
        Training Episodes & 100K & 100K & 100K & 100K & 100K & 100K\\
        Actor Learning rate &0.0003 &0.0003 &0.0003 &0.0001 &0.0003 &0.0003\\
        Critic Learning rate &0.003 &0.003 &0.003 &0.003 &0.003 &0.003\\     
        Episodes per train &4 &8 &4 &8 &4 &8\\
        Target-net update freq &8 &16 &8 &32 &16 &32\\
        $\,\,\,\,\,\,\,\,\,\,\,\,$ (episode) & \\
        N-step TD &5 &5 &5 &5 &5 &5\\
        $\epsilon_{\text{start}}$ & 1 & 1 & 1 & 1 & 1 & 1\\
        $\epsilon_{\text{end}}$ & 0.05 & 0.05 & 0.05 & 0.05 & 0.05 & 0.05 \\
        $\epsilon_{\text{decay}}$ (episode) & 20K & 20K & 20K & 20K & 20K & 20K  \\
    \bottomrule
    \end{tabular} 
\end{table}

\begin{table}[h!]
    \caption {Hyper-parameters used for methods in Overcooked-B.}
    \centering
    \begin{tabular}{lcccccc}
    \toprule
        Parameter & IAC & CAC & Mac-IAC & Mac-CAC & Mac-NIACC & Mac-IAICC \\
    \cmidrule(r){2-7}
        Training Episodes & 120K & 120K & 120K & 120K & 120K & 120K\\
        Actor Learning rate &0.0003 &0.0003 &0.0003 &0.0001 &0.0003 &0.0003\\
        Critic Learning rate &0.003 &0.003 &0.003 &0.003 &0.003 &0.003\\     
        Episodes per train &4 &4 &4 &4 &8 &4\\
        Target-net update freq &8 &16 &8 &16 &16 &32\\
        $\,\,\,\,\,\,\,\,\,\,\,\,$ (episode) & \\
        N-step TD &5 &5 &5 &3 &5 &5\\
        $\epsilon_{\text{start}}$ & 1 & 1 & 1 & 1 & 1 & 1\\
        $\epsilon_{\text{end}}$ & 0.05 & 0.05 & 0.05 & 0.05 & 0.05 & 0.05 \\
        $\epsilon_{\text{decay}}$ (episode) & 20K & 20K & 20K & 20K & 20K & 20K  \\
    \bottomrule
    \end{tabular} 
\end{table}

\begin{table}[h!]
    \caption {Hyper-parameters used for methods in Overcooked-C.}
    \centering
    \begin{tabular}{lcccccc}
    \toprule
        Parameter & IAC & CAC & Mac-IAC & Mac-CAC & Mac-NIACC & Mac-IAICC \\
    \cmidrule(r){2-7}
        Training Episodes & 100K & 100K & 100K & 100K & 100K & 100K\\
        Actor Learning rate &0.0003 &0.0003 &0.0003 &0.0001 &0.0003 &0.0003\\
        Critic Learning rate &0.003 &0.003 &0.003 &0.003 &0.003 &0.003\\
        Episodes per train &8 &8 &8 &8 &8 &8\\
        Target-net update freq &32 &32 &32 &32 &16 &32\\
        $\,\,\,\,\,\,\,\,\,\,\,\,$ (episode) & \\
        N-step TD &5 &5 &5 &3 &5 &5\\
        $\epsilon_{\text{start}}$ & 1 & 1 & 1 & 1 & 1 & 1\\
        $\epsilon_{\text{end}}$ & 0.05 & 0.05 & 0.05 & 0.05 & 0.05 & 0.05 \\
        $\epsilon_{\text{decay}}$ (episode) & 20K & 20K & 20K & 20K & 20K & 20K  \\
    \bottomrule
    \end{tabular} 
\end{table}

\clearpage
\indent$\bullet$ Warehouse Tool Delivery:

\begin{table}[h!]
    \caption {Hyper-parameter candidates for grid search tuning.}
    \centering
    \begin{tabular}{lc}
    \toprule
        Learning rate pair (actor,critic) & (1e-3,1e-3), (5e-4,1e-3) (5e-4,5e-4) (3e-4,3e-3)\\
        Episodes per train & 4, 8 \\
        Target-net update freq (episode) & 8, 16, 32, 64 \\
        N-step TD & 0, 3, 5\\
    \bottomrule
    \end{tabular} 
\end{table}

\begin{table}[h!]
    \caption {Hyper-parameter candidates for grid search tuning.}
    \centering
    \begin{tabular}{lc}
    \toprule
        Learning rate pair (actor,critic) & (1e-3,1e-3), (5e-4,1e-3) (5e-4,5e-4) (3e-4,3e-3)\\
        Episodes per train & 16 \\
        Target-net update freq (episode) & 16, 32, 64 \\
        N-step TD & 0, 3, 5\\
    \bottomrule
    \end{tabular} 
\end{table}

\begin{table}[h!]
    \caption {Hyper-parameters used for methods in Warehouse-A.}
    \centering
    \begin{tabular}{lcccc}
    \toprule
        Parameter & Mac-IAC & Mac-CAC & Mac-NIACC & Mac-IAICC \\
    \cmidrule(r){2-5}
        Training Episodes & 40K & 40K & 40K & 40K\\
        Actor Learning rate &0.0003 &0.0003 &0.0003 &0.0005\\
        Critic Learning rate &0.003 &0.003 &0.003 &0.0005\\  
        Episodes per train &4 &4 &4 &4\\
        Target-net update freq &32 &32 &32 &32\\
        $\,\,\,\,\,\,\,\,\,\,\,\,$ (episode) & \\
        N-step TD &5 &5 &3 &5\\
        $\epsilon_{\text{start}}$ & 1 & 1 & 1 & 1\\
        $\epsilon_{\text{end}}$ & 0.01 & 0.01 & 0.01 & 0.01\\
        $\epsilon_{\text{decay}}$ (episode) & 10K & 10K & 10K & 10K \\
    \bottomrule
    \end{tabular} 
\end{table}

\begin{table}[h!]
    \caption {Hyper-parameters used for methods in Warehouse-A for ablation.}
    \centering
    \begin{tabular}{lcccc}
    \toprule
        Parameter & Mac-IAC & Mac-CAC & Mac-NIACC & Mac-IAICC \\
    \cmidrule(r){2-5}
        Training Episodes & 40K & 40K & 40K & 40K\\
        Actor Learning rate &0.0005 &0.0005 &0.0003 &0.0005\\
        Critic Learning rate &0.0005 &0.001 &0.003 &0.0005\\ 
        Episodes per train &16 &8 &8 &4\\
        Target-net update freq &16 &64 &64 &64\\
        $\,\,\,\,\,\,\,\,\,\,\,\,$ (episode) & \\
        N-step TD &5 &5 &5 &5\\
        $\epsilon_{\text{start}}$ & 1 & 1 & 1 & 1\\
        $\epsilon_{\text{end}}$ & 0.05 & 0.05 & 0.05 & 0.05\\
        $\epsilon_{\text{decay}}$ (episode) & 10K & 10K & 10K & 10K \\
    \bottomrule
    \end{tabular} 
\end{table}

\begin{table}[h!]
    \caption {Hyper-parameters used for methods in Warehouse-B.}
    \centering
    \begin{tabular}{lcccc}
    \toprule
        Parameter & Mac-IAC & Mac-CAC & Mac-NIACC & Mac-IAICC \\
    \cmidrule(r){2-5}
        Training Episodes & 40K & 40K & 40K & 40K\\
        Actor Learning rate &0.0005 &0.0005 &0.0003 &0.0003\\
        Critic Learning rate &0.0005 &0.001 &0.003 &0.003\\
        Episodes per train &8 &4 &16 &4\\
        Target-net update freq &64 &64 &64 &32\\
        $\,\,\,\,\,\,\,\,\,\,\,\,$ (episode) & \\
        N-step TD &5 &5 &5 &5\\
        $\epsilon_{\text{start}}$ & 1 & 1 & 1 & 1\\
        $\epsilon_{\text{end}}$ & 0.01 & 0.01 & 0.01 & 0.01\\
        $\epsilon_{\text{decay}}$ (episode) & 10K & 10K & 10K & 10K \\
    \bottomrule
    \end{tabular} 
\end{table}

\begin{table}[h!]
    \caption {Hyper-parameters used for methods in Warehouse-C.}
    \centering
    \begin{tabular}{lcccc}
    \toprule
        Parameter & Mac-IAC & Mac-CAC & Mac-NIACC & Mac-IAICC \\
    \cmidrule(r){2-5}
        Training Episodes  & 80K & 80K & 80K & 80K\\
        Actor Learning rate &0.0005 &0.0003 &0.0003 &0.0003\\
        Critic Learning rate &0.001 &0.003 &0.003 &0.003\\
        Episodes per train &8 &8 &8 &8\\
        Target-net update freq &64 &64 &64 &64\\
        $\,\,\,\,\,\,\,\,\,\,\,\,$ (episode) & \\
        N-step TD &5 &5 &5 &5\\
        $\epsilon_{\text{start}}$ & 1 & 1 & 1 & 1\\
        $\epsilon_{\text{end}}$ & 0.01 & 0.01 & 0.01 & 0.01\\
        $\epsilon_{\text{decay}}$ (episode) & 10K & 10K & 10K & 10K \\
    \bottomrule
    \end{tabular} 
\end{table}

\begin{table}[h!]
    \caption {Hyper-parameters used for methods in Warehouse-D.}
    \centering
    \begin{tabular}{lcccc}
    \toprule
        Parameter & Mac-IAC & Mac-CAC & Mac-NIACC & Mac-IAICC \\
    \cmidrule(r){2-5}
        Training Episodes  & 80K & 80K & 80K & 80K\\
        Actor Learning rate &0.0003 &0.0003 &0.0005 &0.0003\\
        Critic Learning rate &0.003 &0.003 &0.005 &0.003\\
        Episodes per train &4 &8 &4 &8\\
        Target-net update freq &16 &64 &32 &64\\
        $\,\,\,\,\,\,\,\,\,\,\,\,$ (episode) & \\
        N-step TD &5 &5 &5 &5\\
        $\epsilon_{\text{start}}$ & 1 & 1 & 1 & 1\\
        $\epsilon_{\text{end}}$ & 0.01 & 0.01 & 0.01 & 0.01\\
        $\epsilon_{\text{decay}}$ (episode) & 10K & 10K & 10K & 10K \\
    \bottomrule
    \end{tabular} 
\end{table}

\begin{table}[h!]
    \caption {Hyper-parameters used for methods in Warehouse-E.}
    \centering
    \begin{tabular}{lcccc}
    \toprule
        Parameter & Mac-IAC & Mac-CAC & Mac-NIACC & Mac-IAICC \\
    \cmidrule(r){2-5}
        Training Episodes  & 100K & 100K & 100K & 100K\\
        Actor Learning rate &0.0005 &0.0003 &0.0003 &0.0005\\
        Critic Learning rate &0.001 &0.003 &0.003 &0.0005\\
        Episodes per train &4 &4 &4 &4\\
        Target-net update freq &32 &16 &32 &32\\
        $\,\,\,\,\,\,\,\,\,\,\,\,$ (episode) & \\
        N-step TD &5 &5 &5 &5\\
        $\epsilon_{\text{start}}$ & 1 & 1 & 1 & 1\\
        $\epsilon_{\text{end}}$ & 0.05 & 0.05 & 0.05 & 0.05\\
        $\epsilon_{\text{decay}}$ (episode) & 10K & 10K & 10K & 10K \\
    \bottomrule
    \end{tabular} 
\end{table}

\clearpage
\subsection{Hyper-Parameters for macro-action-based value-based methods}

\indent$\bullet$ Box Pushing:

\begin{table}[h!]
    \caption {Hyper-parameter candidates for grid search tuning.}
    \centering
    \begin{tabular}{lc}
    \toprule
        Learning rate & 5e-4, 1e-3 \\ 
        batch size & 32, 64, 128 \\
    \bottomrule
    \end{tabular} 
\end{table}

\begin{table}[h!]
    \caption {Hyper-parameters used in Box Pushing $8\times8$.}
    \centering
    \begin{tabular}{lcc}
    \toprule
        Parameter & Mac-Dec-Q & Mac-Cen-Q \\
    \cmidrule(r){2-3}
        Training Episodes  & 40K & 40K\\
        Learning rate & 0.001 & 0.001\\
        Batch size & 64 & 64\\
        Replay-buffer size (step) & 100K & 100K\\
        Train freq (step) & 10 & 10\\
        Trace length & 10 & 10\\
        Target-net update freq (step) & 5K & 5K\\
        $\epsilon_{\text{start}}$ & 1 & 1\\
        $\epsilon_{\text{end}}$ & 0.05 & 0.05\\
        $\epsilon_{\text{decay}}$ (episode) & 4K & 4K  \\
    \bottomrule
    \end{tabular} 
\end{table}

\begin{table}[h!]
    \caption {Hyper-parameters used in Box Pushing $10\times10$.}
    \centering
    \begin{tabular}{lcc}
    \toprule
        Parameter & Mac-Dec-Q & Mac-Cen-Q \\
    \cmidrule(r){2-3}
        Training Episodes  & 40K & 40K\\
        Learning rate & 0.001 & 0.001\\
        Batch size & 32 & 128\\
        Replay-buffer size (step) & 100K & 100K\\
        Train freq (step) & 14 & 14\\
        Trace length & 14 & 14\\
        Target-net update freq (step) & 5K & 5K\\
        $\epsilon_{\text{start}}$ & 1 & 1\\
        $\epsilon_{\text{end}}$ & 0.05 & 0.05\\
        $\epsilon_{\text{decay}}$ (episode) & 6K & 6K  \\
    \bottomrule
    \end{tabular} 
\end{table}

\clearpage

\begin{table}[h!]
    \caption {Hyper-parameters used in Box Pushing $12\times12$.}
    \centering
    \begin{tabular}{lcc}
    \toprule
        Parameter & Mac-Dec-Q & Mac-Cen-Q \\
    \cmidrule(r){2-3}
        Training Episodes  & 40K & 40K\\
        Learning rate & 0.001 & 0.001\\
        Batch size & 32 & 128\\
        Replay-buffer size (step) & 100K & 100K\\
        Train freq (step) & 20 & 20\\
        Trace length & 20 & 20\\
        Target-net update freq (step) & 5K & 5K\\
        $\epsilon_{\text{start}}$ & 1 & 1\\
        $\epsilon_{\text{end}}$ & 0.05 & 0.05\\
        $\epsilon_{\text{decay}}$ (episode) & 6K & 6K  \\
    \bottomrule
    \end{tabular} 
\end{table}

\begin{table}[h!]
    \caption {Hyper-parameters used in Box Pushing $20\times20$.}
    \centering
    \begin{tabular}{lcc}
    \toprule
        Parameter & Mac-Dec-Q & Mac-Cen-Q \\
    \cmidrule(r){2-3}
        Training Episodes  & 40K & 40K\\
        Learning rate & 0.001 & 0.001\\
        Batch size & 32 & 64\\
        Replay-buffer size (step) & 100K & 100K\\
        Train freq (step) & 35 & 35\\
        Trace length  & 35 & 35\\
        Target-net update freq (step) & 5K & 5K\\
        $\epsilon_{\text{start}}$ & 1 & 1\\
        $\epsilon_{\text{end}}$ & 0.05 & 0.05\\
        $\epsilon_{\text{decay}}$ (episode) & 8K & 8K  \\
    \bottomrule
    \end{tabular} 
\end{table}

\begin{table}[h!]
    \caption {Hyper-parameters used in Box Pushing $30\times30$.}
    \centering
    \begin{tabular}{lcc}
    \toprule
        Parameter & Mac-Dec-Q & Mac-Cen-Q \\
    \cmidrule(r){2-3}
        Training Episodes  & 40K & 40K\\
        Learning rate & 0.0005 & 0.001\\
        Batch size & 32 & 32\\
        Replay-buffer size (step) & 100K & 100K\\
        Train freq (step) & 45 & 45\\
        Trace length  & 45 & 45\\
        Target-net update freq (step) & 5K & 5K\\
        $\epsilon_{\text{start}}$ & 1 & 1\\
        $\epsilon_{\text{end}}$ & 0.05 & 0.05\\
        $\epsilon_{\text{decay}}$ (episode) & 8K & 8K  \\
    \bottomrule
    \end{tabular} 
\end{table}

\indent$\bullet$ Overcooked:

\begin{table}[h!]
    \caption {Hyper-parameter candidates for grid search tuning.}
    \centering
    \begin{tabular}{lc}
    \toprule
        Learning rate & 3e-5, 5e-5, 1e-4, 3e-4, 5e-4 \\ 
        batch size & 32, 64 \\
        Train freq (step) & 64, 128\\
        Replay-buffer size (episode) & 500, 1K, 2K, 3K \\
    \bottomrule
    \end{tabular} 
\end{table}

\clearpage

\begin{table}[h!]
    \caption {Hyper-parameters used in Overcooked-A.}
    \centering
    \begin{tabular}{lcc}
    \toprule
        Parameter & Mac-Dec-Q & Mac-Cen-Q \\
    \cmidrule(r){2-3}
        Training Episodes  & 100K & 100K\\
        Learning rate & 0.0005 & 0.00003\\
        Batch size & 64 & 64\\
        Replay-buffer size (episode) & 1K & 1K\\
        Train freq (step) & 64 & 64\\
        Target-net update freq (step) & 5K & 5K\\
        $\epsilon_{\text{start}}$ & 1 & 1\\
        $\epsilon_{\text{end}}$ & 0.05 & 0.05\\
        $\epsilon_{\text{decay}}$ (episode) & 20K & 20K  \\
    \bottomrule
    \end{tabular} 
\end{table}

\begin{table}[h!]
    \caption {Hyper-parameters used in Overcooked-B.}
    \centering
    \begin{tabular}{lcc}
    \toprule
        Parameter & Mac-Dec-Q & Mac-Cen-Q \\
    \cmidrule(r){2-3}
        Training Episodes  & 100K & 100K\\
        Learning rate & 0.0005 & 0.0001\\
        Batch size & 32 & 32\\
        Replay-buffer size (episode) & 3K & 500\\
        Train freq (step) & 64 & 64\\
        Target-net update freq (step) & 5K & 5K\\
        $\epsilon_{\text{start}}$ & 1 & 1\\
        $\epsilon_{\text{end}}$ & 0.05 & 0.05\\
        $\epsilon_{\text{decay}}$ (episode) & 20K & 20K  \\
    \bottomrule
    \end{tabular} 
\end{table}

\indent$\bullet$ Warehouse Tool Delivery:

\begin{table}[h!]
    \caption {Hyper-parameter candidates for grid search tuning.}
    \centering
    \begin{tabular}{lc}
    \toprule
        Learning rate & 5e-5, 1e-4 \\ 
        batch size & 32, 64 \\
        Train freq (step) & 64, 128\\
        Replay-buffer size (episode) & 1K, 2K\\
    \bottomrule
    \end{tabular} 
\end{table}

\begin{table}[h!]
    \caption {Hyper-parameters used in Warehouse-A.}
    \centering
    \begin{tabular}{lcc}
    \toprule
        Parameter & Mac-Dec-Q & Mac-Cen-Q \\
    \cmidrule(r){2-3}
        Training Episodes  & 40K & 40K\\
        Learning rate & 0.0001 & 0.0001\\
        Batch size & 64 & 64\\
        Replay-buffer size (episode) & 2K & 2K\\
        Train freq (step) & 128 & 128\\
        Target-net update freq (step) & 5K & 5K\\
        $\epsilon_{\text{start}}$ & 1 & 1\\
        $\epsilon_{\text{end}}$ & 0.05 & 0.05\\
        $\epsilon_{\text{decay}}$ (episode) & 10K & 10K  \\
    \bottomrule
    \end{tabular} 
\end{table}

\begin{table}[h!]
    \caption {Hyper-parameters used in Warehouse-B.}
    \centering
    \begin{tabular}{lcc}
    \toprule
        Parameter & Mac-Dec-Q & Mac-Cen-Q \\
    \cmidrule(r){2-3}
        Training Episodes  & 40K & 40K\\
        Learning rate & 0.00005 & 0.00005\\
        Batch size & 64 & 64\\
        Replay-buffer size (episode) & 2K & 2K\\
        Train freq (step) & 128 & 128\\
        Target-net update freq (step) & 5K & 5K\\
        $\epsilon_{\text{start}}$ & 1 & 1\\
        $\epsilon_{\text{end}}$ & 0.05 & 0.05\\
        $\epsilon_{\text{decay}}$ (episode) & 10K & 10K  \\
    \bottomrule
    \end{tabular} 
\end{table}

\begin{table}[h!]
    \caption {Hyper-parameters used in Warehouse-E.}
    \centering
    \begin{tabular}{lcc}
    \toprule
        Parameter & Mac-Dec-Q & Mac-Cen-Q \\
    \cmidrule(r){2-3}
        Training Episodes  & 100K & 100K\\
        Learning rate & 0.0001 & 0.0001\\
        Batch size & 64 & 64\\
        Replay-buffer size (episode) & 2K & 2K\\
        Train freq (step) & 128 & 128\\
        Target-net update freq (step) & 5K & 5K\\
        $\epsilon_{\text{start}}$ & 1 & 1\\
        $\epsilon_{\text{end}}$ & 0.05 & 0.05\\
        $\epsilon_{\text{decay}}$ (episode) & 10K & 10K  \\
    \bottomrule
    \end{tabular} 
\end{table}


\clearpage
\section{Hardware Experiments}
\label{HE}

\begin{figure}[h!]
    \centering
    \includegraphics[scale=0.85]{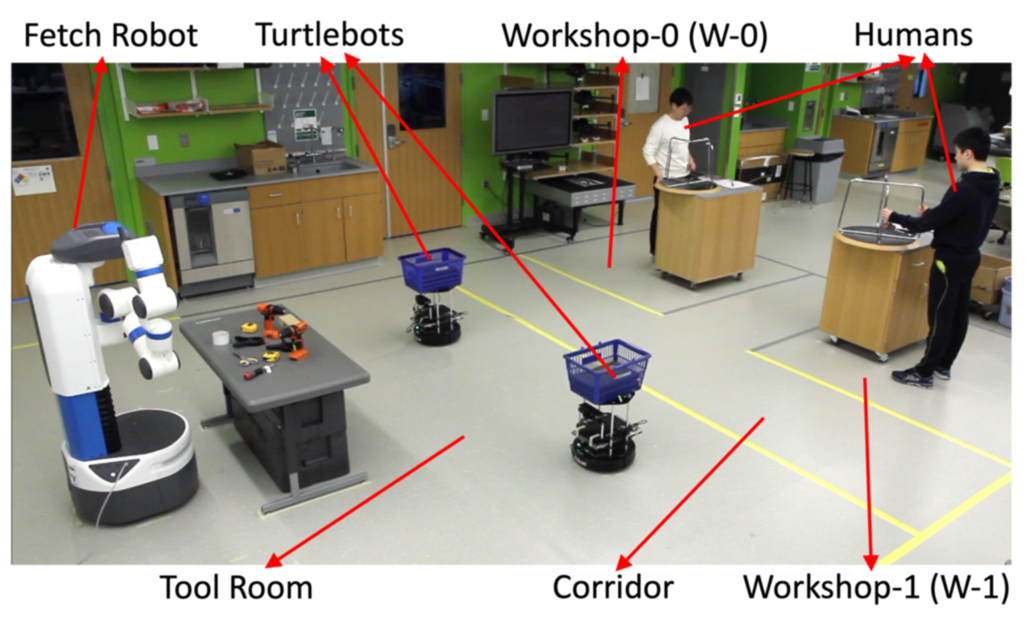}
    \caption{Overview of Warehouse-A hardware domain.}
    \vspace{-1mm}
    \label{hardware_domain}
\end{figure}

While simulation results validate that the proposed Mac-IAICC approach achieves the best performance for learning decentralized policies in various macro-action-based domains, we also extend scenario A of the Warehouse Tool Delivery task to a hardware domain. Fig.~\ref{hardware_domain} provides an overview of the real-world experimental setup. An open area is divided into regions, a tool room, a corridor, and two workshops, to resemble the configuration shown in Fig.~\ref{domain_wtdA}. This mission involves one Fetch Robot~\cite{FetchRobot} and two Turtlebots~\cite{Turtlebot} to cooperatively find and deliver three YCB tools~\cite{YCB}, in the order: a tape measure, a clamp and an electric drill, required by each human in order to assemble an IKEA table. 

The Turtlebot's navigation macro-actions were executed by using the ROS navigation stack~\cite{ROSNavigation}. For Fetch's manipulation macro-actions, we combined PCL bindings for Python~\cite{PCLPython}, MoveIt~\cite{MoveIt} and the OpenRave simulator~\cite{Diankov:R} with an OMPL~\cite{OMPL} plugin to achieve picking and placing of tools. The information about the number of tools in staging areas and each human's working status was tracked and broadcast by ROS services but were only observable in the tool room and the corresponding workshop area respectively (to simulate possible visual information). 

For the visualization of the real-robot experiment, please check the video in our supplementary.  

\clearpage
\section{Behavior Visualization in Simulation}
\label{A-BV}





In this section, we display the decentralized behaviors learned by using Mac-IAICC under all considered domains.

\subsection{Box Pushing}
\begin{figure*}[h!]
    \centering
    \captionsetup[subfigure]{labelformat=empty}
    \centering
    \subcaptionbox{(a) Green robot executes \emph{\textbf{Move-to-big-box(1)}} to move to the left waypoint below the big box while the blue robot runs \emph{\textbf{Move-to-big-box(2)}} to move to the right waypoint below the big box.}
        [0.30\linewidth]{\includegraphics[scale=0.13]{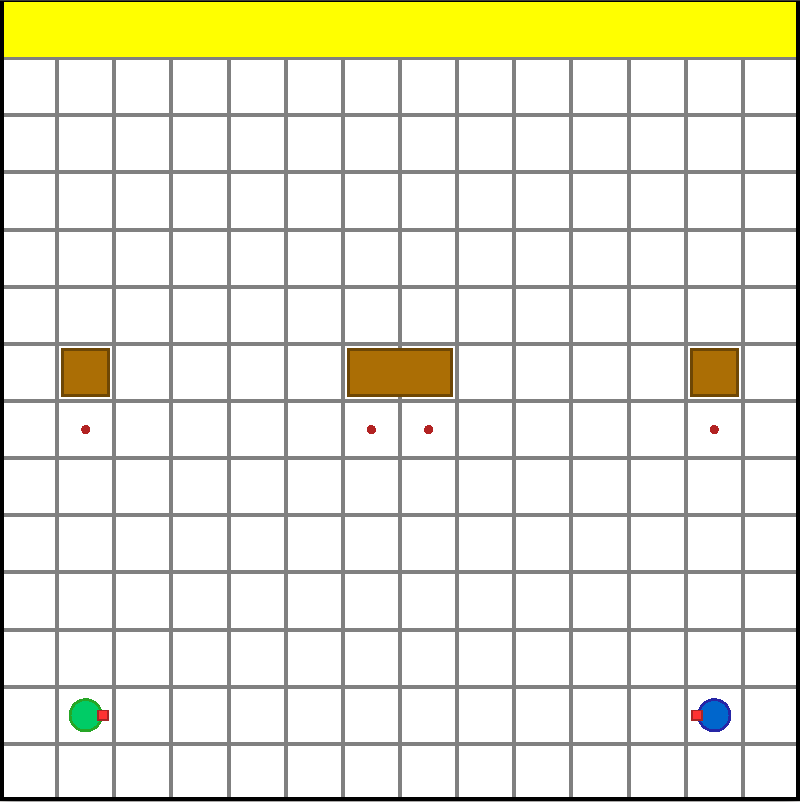}}
    \quad
    \centering
    \subcaptionbox{(b) After completing the previous macro-actions, robots choose \emph{\textbf{Push}} to move the big box towards the goal together.}
        [0.30\linewidth]{\includegraphics[scale=0.13]{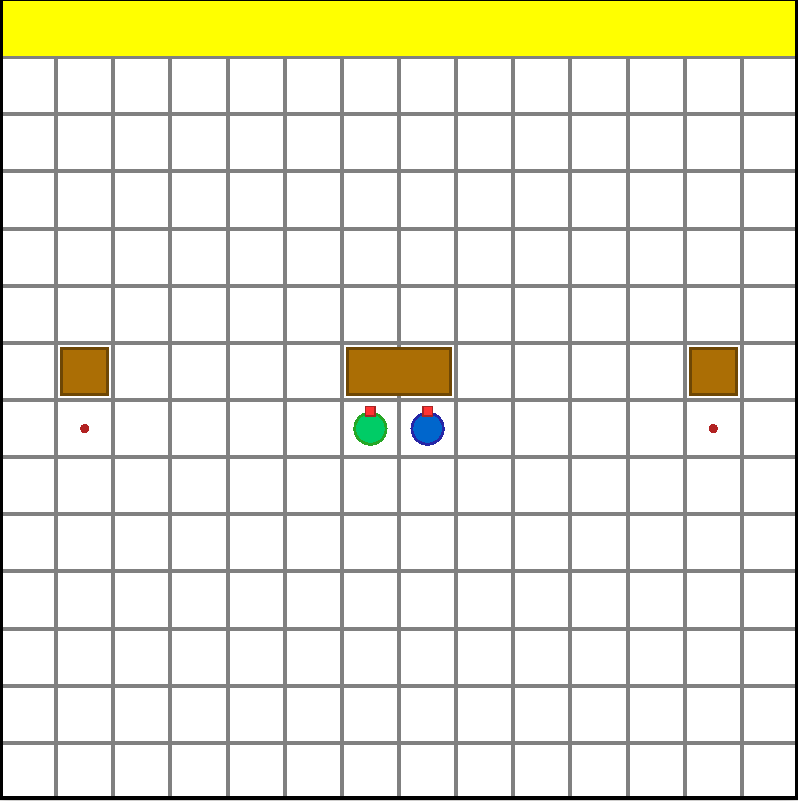}}
    \quad
    \centering
    \subcaptionbox{(c) Robots finish the task by pushing the big box to the goal area.}
        [0.30\linewidth]{\includegraphics[scale=0.13]{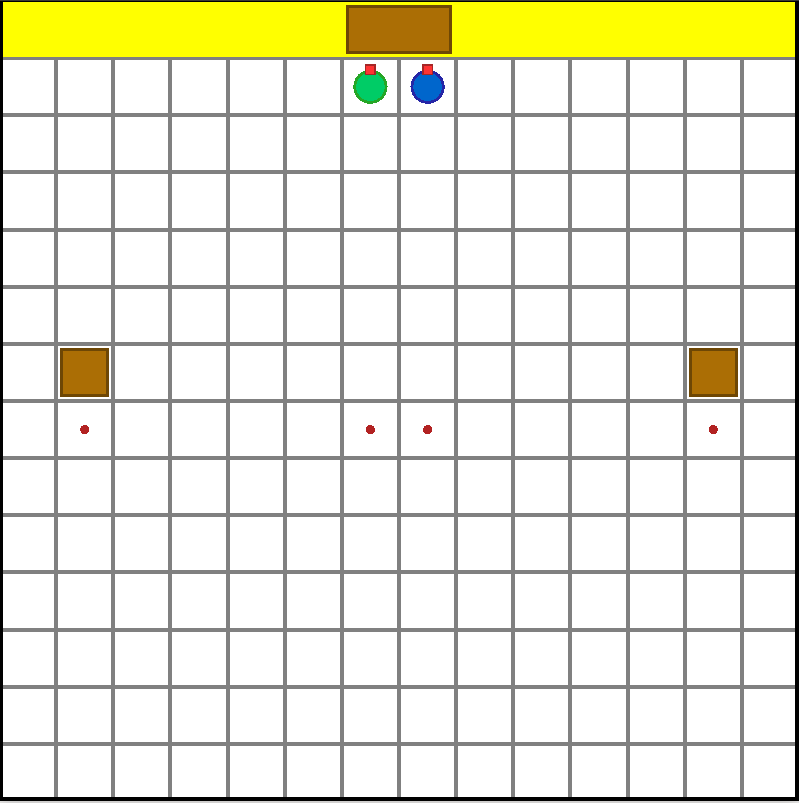}}
    \caption{Visualization of the optimal macro-action-based behaviors learned using Mac-IAICC in the Box Pushing domain under a $14\times14$ grid world.}
    \label{bp_behavior}
\end{figure*}

We show the behaviors learned under the grid world size $14\times14$ in Fig.~\ref{bp_behavior}. Although the averaged performance of the training is not near-optimal (Fig.~\ref{bp_ctde_comp}), several runs can learn the optimal behavior. 

\clearpage
\subsection{Overcooked}
\label{B-Overcooked}

\textbf{\emph{Map A:}} In this map, our method learns an efficient collaboration such as three agents separately get three different vegetables, and then go to the cutting board and chop them respectively. Especially, the pink agent leans to take away the chopped lettuce in order to make room for the incoming green agent to chop the onion (Fig.~\ref{fig:Overcooked_visual_mapA}h - \ref{fig:Overcooked_visual_mapA}i). Details are shown below. 
  
\begin{figure*}[h!]
    \centering
    \captionsetup[subfigure]{labelformat=empty}
    \centering
    \subcaptionbox{(a) The blue agent executes \textbf{\emph{Get-Lettuce}}. The pink agent executes \textbf{\emph{Get-Tomato}}. The green agent executes \textbf{\emph{Get-Onion}}.}
        [0.23\linewidth]{\includegraphics[scale = 0.15]{results/Overcooked/3_agent_D.png}}
    ~    
    \subcaptionbox{(b) After getting the lettuce, the pink agent executes \textbf{\emph{Go-Cut-Board-2}}.}
        [0.23\linewidth]{\includegraphics[scale = 0.20]{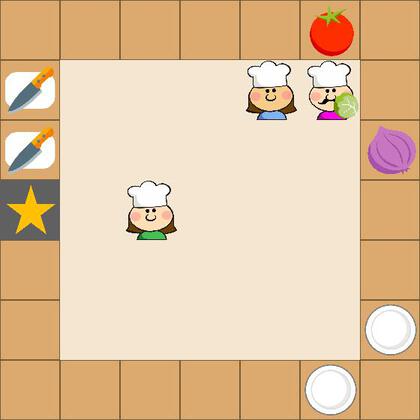}}
    ~    
    \subcaptionbox{(c) After getting the tomato, the blue agent executes \textbf{\emph{Go-Cut-Board-1}}.}
        [0.23\linewidth]{\includegraphics[scale = 0.20]{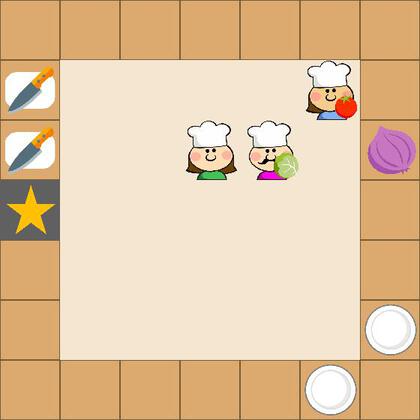}}
    ~    
    \subcaptionbox{(d) After getting the onion, the green agent executes \textbf{\emph{Go-Cut-Board-2}}.}
        [0.23\linewidth]{\includegraphics[scale = 0.20]{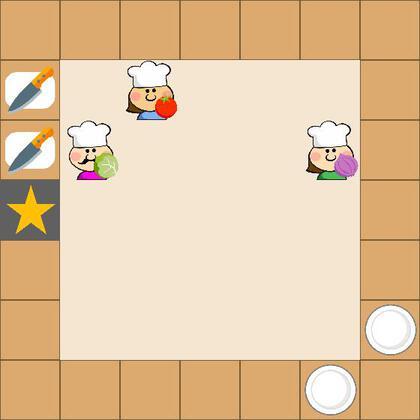}}
    ~    
    \subcaptionbox{(e) After placing the lettuce on the cutting board, the pink agent executes \textbf{\emph{Chop}}.  }
        [0.23\linewidth]{\includegraphics[scale = 0.20]{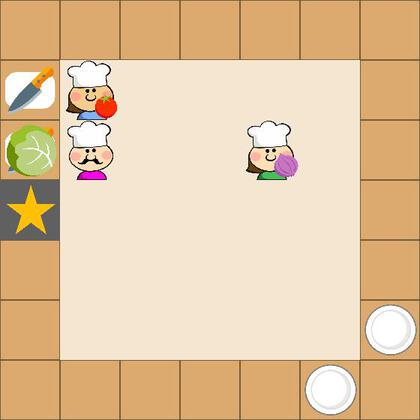}}
    ~    
    \subcaptionbox{(f) After placing the tomato on the cutting board, the blue agent executes \textbf{\emph{Chop}}.  }
        [0.23\linewidth]{\includegraphics[scale = 0.20]{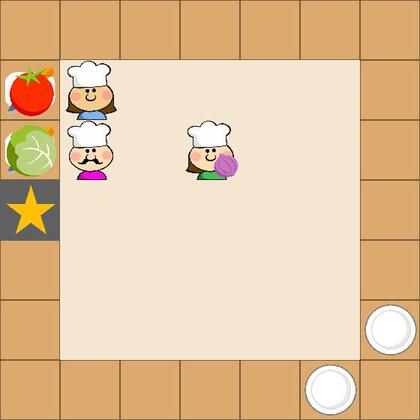}}
    ~    
    \subcaptionbox{(g) After finishing chopping the lettuce, the pink agent executes \textbf{\emph{Get-Lettuce}} to pick it up.}
        [0.23\linewidth]{\includegraphics[scale = 0.20]{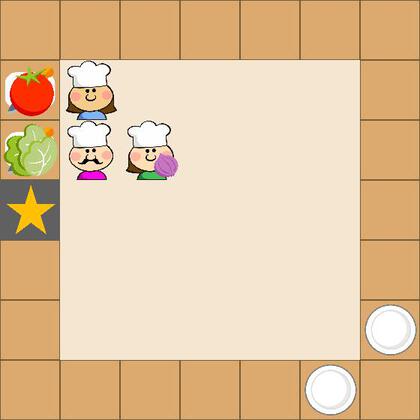}}
    ~    
    \subcaptionbox{(h) With the lettuce in hand, the pink agent executes \textbf{\emph{Get-Plate-1}}.}
        [0.23\linewidth]{\includegraphics[scale = 0.20]{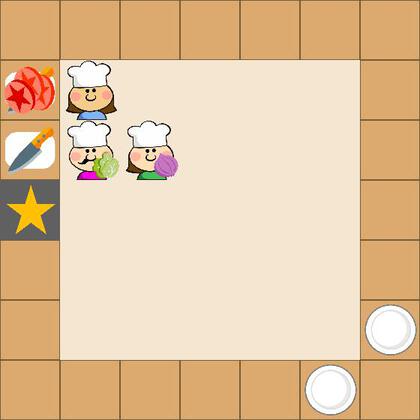}}
    ~    
    \subcaptionbox{(i) After placing the onion on the cutting board, the green agent executes \textbf{\emph{Chop}}.}
        [0.23\linewidth]{\includegraphics[scale = 0.20]{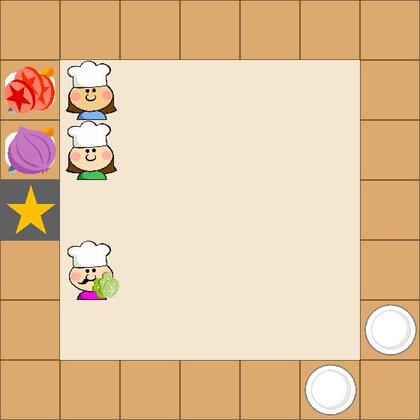}}
    ~    
    \subcaptionbox{(j) The green agent executes \textbf{\emph{Get-Plate-2}}, and the blue agent keep running \textbf{\emph{Move-Down}} to make room for the pink agent to merge the chopped vegetables later on.}
        [0.23\linewidth]{\includegraphics[scale = 0.20]{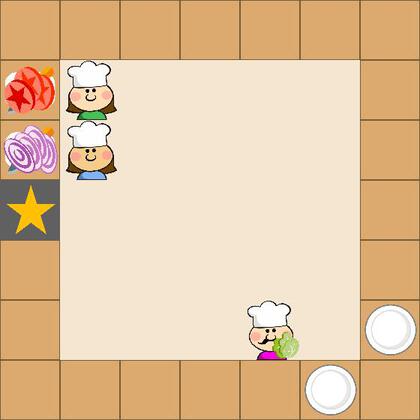}}
    ~    
    \subcaptionbox{(k) The pink agent reaches the plate and it is going to put the lettuce on the plate.}
        [0.23\linewidth]{\includegraphics[scale = 0.20]{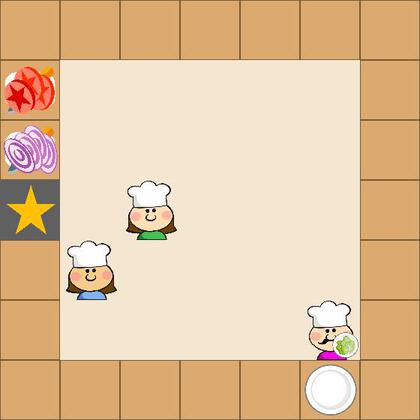}}
    ~    
    \subcaptionbox{(l) After putting the lettuce on the plate, the pink agent merges the onion in the plate by executing executes \textbf{\emph{Get-Onion}}.}
        [0.23\linewidth]{\includegraphics[scale = 0.20]{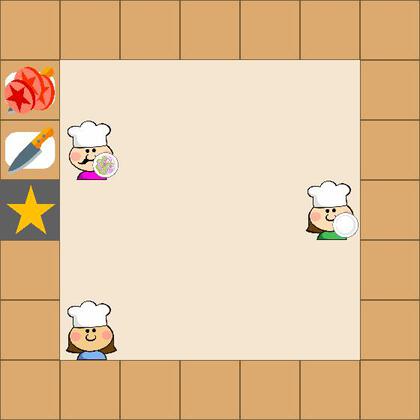}}
\end{figure*}

\clearpage

\begin{figure*}[h!]
    \centering
    \captionsetup[subfigure]{labelformat=empty}
    \centering
    ~
    \subcaptionbox{(m) The pink agent gets the chopped tomato into the plate by executing \emph{Get-Tomato}.}
        [0.23\linewidth]{\includegraphics[scale = 0.20]{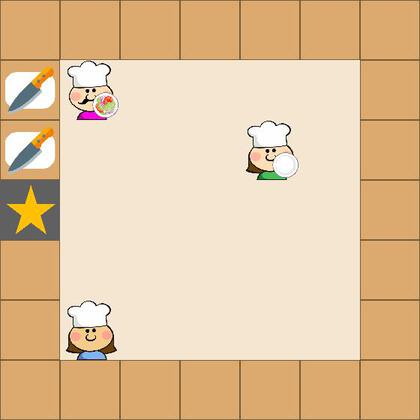}}
    ~    
    \subcaptionbox{(n) The pink agent successfully delivers the tomato-lettuce-onion salad by running \emph{Deliver}.}
        [0.23\linewidth]{\includegraphics[scale = 0.20]{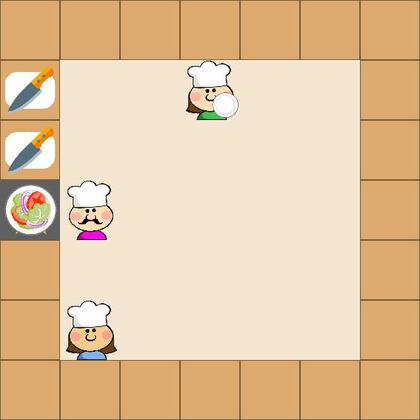}}
    ~
    \caption{Visualization of running decentralized policies learned by Mac-IAICC in Overcooked-A.}
    \label{fig:Overcooked_visual_mapA}
\end{figure*}

\textbf{\emph{Map B:}} In this map, the decentralized policies trained by our method learns the collaboration such that the pink agent focuses on transporting items from right to left, while the other two agents cooperatively prepare the salad.

\begin{figure*}[h!]
    \centering
    \captionsetup[subfigure]{labelformat=empty}
    \centering
    \subcaptionbox{(a) The blue agent executes \textbf{\emph{Go-Cut-board-1}}. The green agent executes \textbf{\emph{Go-Cut-board-2}}. The pink agent executes \textbf{\emph{Get-Lettuce}}.}
        [0.23\linewidth]{\includegraphics[scale = 0.15]{results/Overcooked/3_agent_F.png}}
    ~    
    \subcaptionbox{(b) After getting the lettuce, the pink agent executes \textbf{\emph{Go-Counter}}.}
        [0.23\linewidth]{\includegraphics[scale = 0.20]{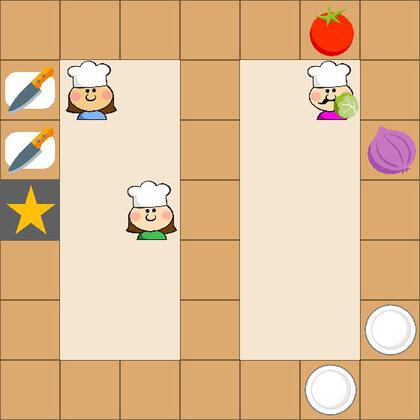}}
    ~    
    \subcaptionbox{(c) After putting the lettuce on the counter, the pink agent executes \textbf{\emph{Get-Onion}}.}
        [0.23\linewidth]{\includegraphics[scale = 0.20]{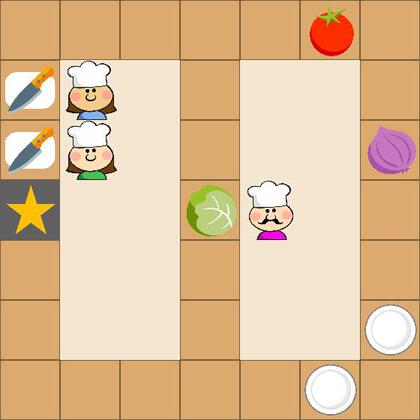}}
    ~    
    \subcaptionbox{(d) The green agent executes \textbf{\emph{Get-Lettuce}}.}
        [0.23\linewidth]{\includegraphics[scale = 0.20]{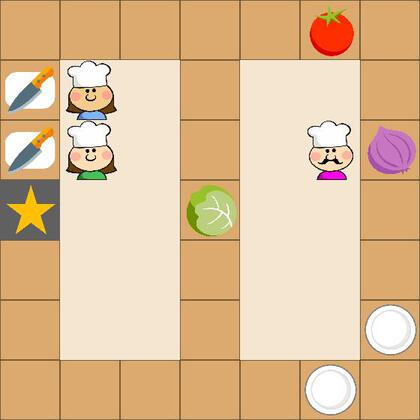}}
    ~    
    \subcaptionbox{(e) After getting the onion, the pink agent executes \textbf{\emph{Go-Counter}}.}
        [0.23\linewidth]{\includegraphics[scale = 0.20]{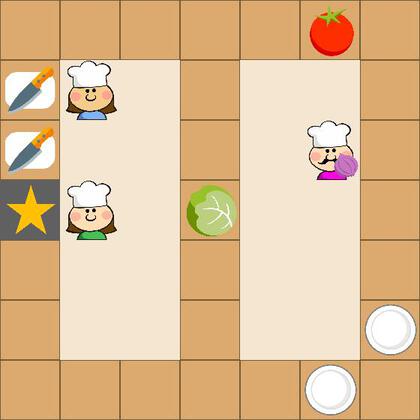}}
    ~    
    \subcaptionbox{(f) After getting the lettuce, the green agent executes \textbf{\emph{Go-Cut-Board-2}}. Meanwhile the pink agent puts the onion on the counter, and then executes \textbf{\emph{Get-Tomato}}. The blue agent executes \textbf{\emph{Get-Onion}}. }
        [0.23\linewidth]{\includegraphics[scale = 0.20]{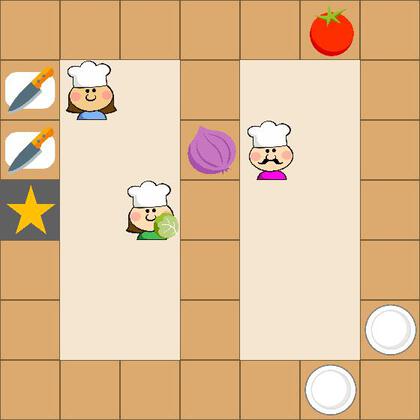}}
    ~    
    \subcaptionbox{(g) After putting the lettuce on the cutting board, the green executes \textbf{\emph{Chop}}. Blue agent executes \textbf{\emph{Go-Cut-Board-1}} with onion in hand.}
        [0.23\linewidth]{\includegraphics[scale = 0.20]{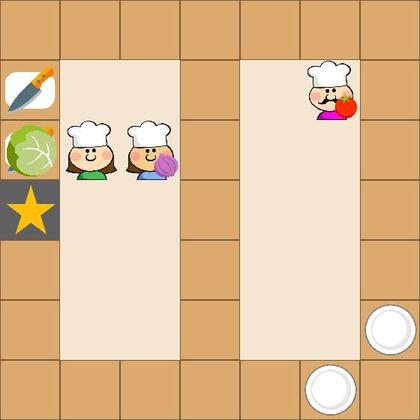}}
    ~    
    \subcaptionbox{(h) The blue agent executes \textbf{\emph{Chop}} to cut the onion to pieces.}
        [0.23\linewidth]{\includegraphics[scale = 0.20]{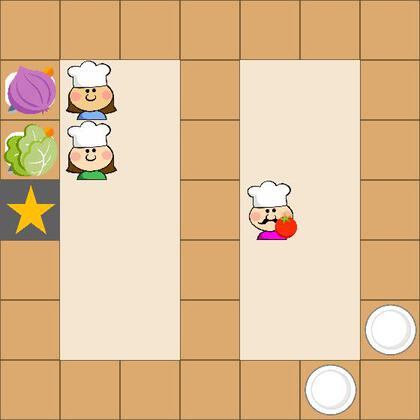}}
\end{figure*}

\clearpage

\begin{figure*}[h!]
    \centering
    \captionsetup[subfigure]{labelformat=empty}
    \centering
    ~    
    \subcaptionbox{(i) After putting the tomato on the counter, the pink agent executes \textbf{\emph{Get-Plate-2}}. The green agent executes \textbf{\emph{Get-Tomato}}. }
        [0.23\linewidth]{\includegraphics[scale = 0.20]{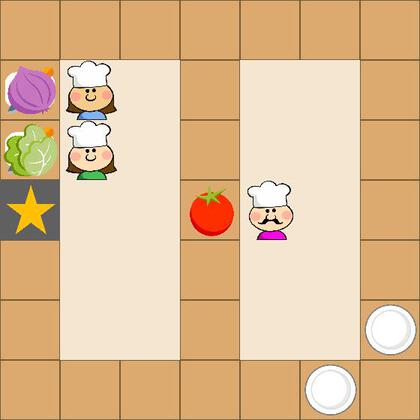}}
    ~    
    \subcaptionbox{(j) The blue agent finishes chopping the onion, and then picks it up by executing \textbf{\emph{Get-Onion}} again.}
        [0.23\linewidth]{\includegraphics[scale = 0.20]{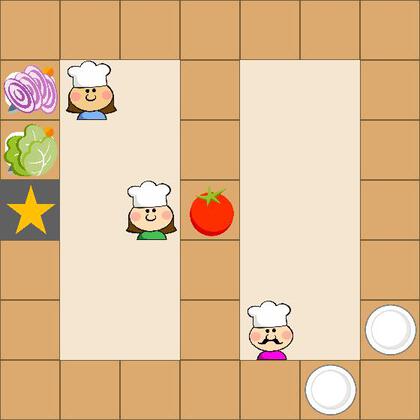}}
    ~    
    \subcaptionbox{(k) The blue agent keeps executing \textbf{\emph{Moving-Down}} to make room for the green agent to chop tomato later on. Meanwhile the green agent moves towards the upper cutting board by executing \textbf{\emph{Go-Cut-Board-1}}.}
        [0.23\linewidth]{\includegraphics[scale = 0.20]{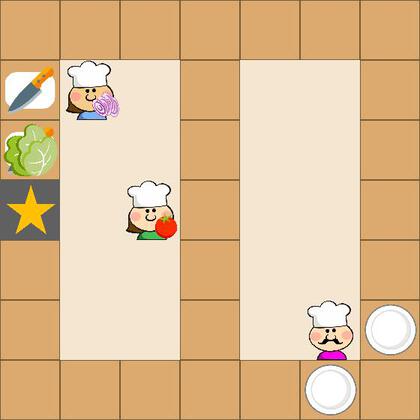}}
    ~    
    \subcaptionbox{(l) After getting the plate, the pink agent executes \textbf{\emph{Go-Counter}}.}
        [0.23\linewidth]{\includegraphics[scale = 0.20]{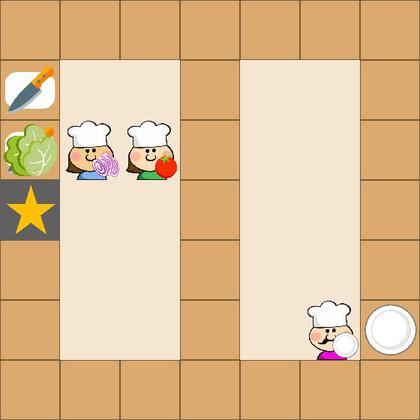}}
    ~    

    \subcaptionbox{(m) After putting the tomato on the cutting board, the green agent executes \textbf{\emph{Chop}}. }
        [0.23\linewidth]{\includegraphics[scale = 0.20]{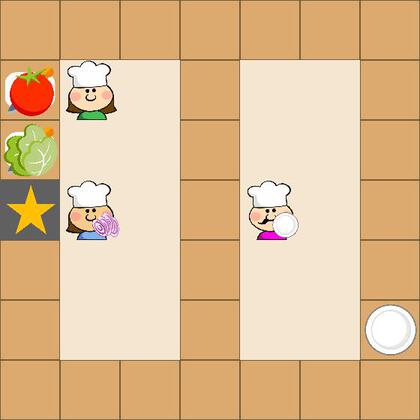}}
    ~    
    \subcaptionbox{(n) The pink agent puts the plate on the counter. The blue agent executes \textbf{\emph{Go-Counter}} to get the plate.}
        [0.23\linewidth]{\includegraphics[scale = 0.20]{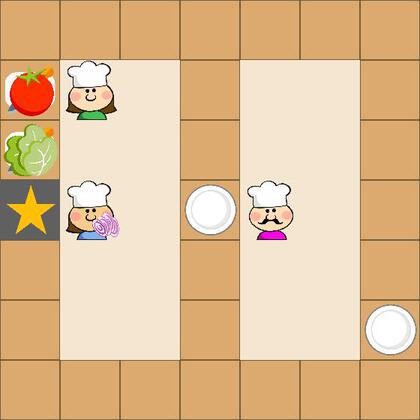}}
    ~    
    \subcaptionbox{(o) The green agent finishes chopping the tomato, while the blue agent puts chopped onion on the plate.}
        [0.23\linewidth]{\includegraphics[scale = 0.20]{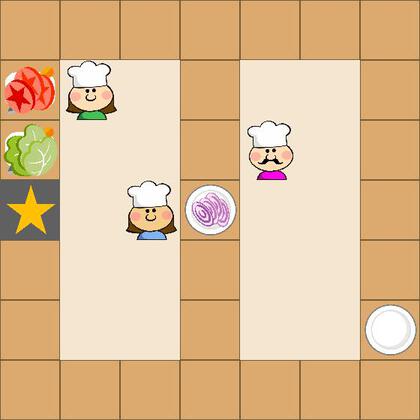}}
    ~    
    \subcaptionbox{(p) The green agent executes \emph{Go-Cut-Board-2} to make room for the blue agent to merge the tomato in to the plate.}
        [0.23\linewidth]{\includegraphics[scale = 0.20]{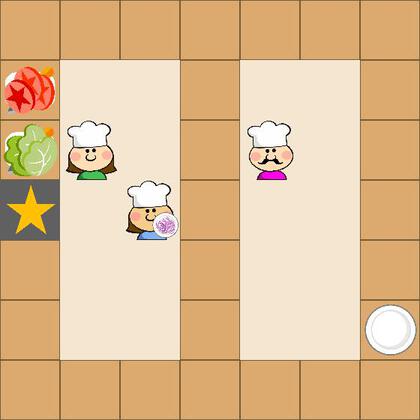}}

    ~    
    \subcaptionbox{(q) The blue agent executes \textbf{\emph{Get-Lettuce}}. The green agent executes \textbf{\emph{Go-Cut-Board-1}}.}
        [0.23\linewidth]{\includegraphics[scale = 0.20]{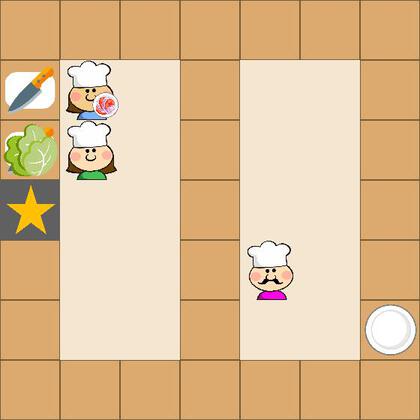}}
    ~    
    \subcaptionbox{(r) After putting the lettuce on the plate, the blue agent executes \textbf{\emph{Deliver}}.}
        [0.23\linewidth]{\includegraphics[scale = 0.20]{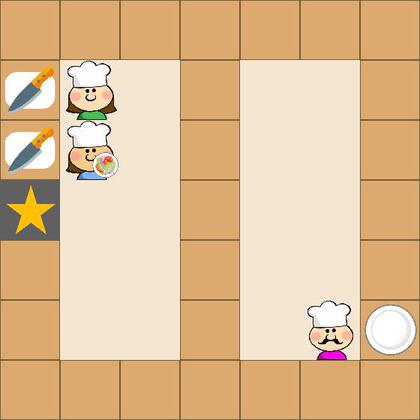}}
    ~
    \subcaptionbox{(s) The blue agent successfully delivers the tomato-lettuce-onion salad.}
        [0.23\linewidth]{\includegraphics[scale = 0.20]{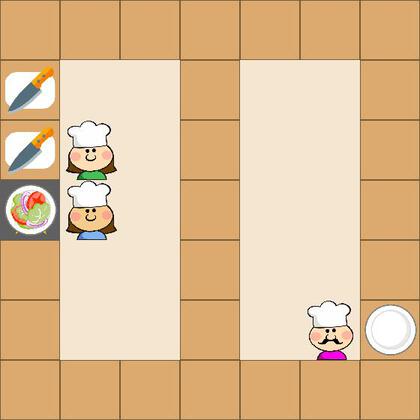}}
    ~
    \caption{Visualization of running decentralized policies learned by Mac-IAICC in Overcooked-B.}
    \label{fig:Overcooked_visual_mapC}
\end{figure*}

\clearpage

\textbf{\emph{Map C:}} In this map, the best strategy is that the pink agent should take the advantage of the middle counters to pass vegetables to the other agent. Our method learns a sub-optimal policy such that the blue agent still crosses the narrow passage to get the vegetable at the right side of the map.

\begin{figure*}[h!]
    \centering
    \captionsetup[subfigure]{labelformat=empty}
    \centering
    \subcaptionbox{(a) The blue agent executes \textbf{\emph{Get-Lettuce}}. The pink agent executes \textbf{\emph{Get-Onion}}.  The green agent executes \textbf{\emph{Get-Tomato}}.}
        [0.23\linewidth]{\includegraphics[scale = 0.15]{results/Overcooked/3_agent_E.png}}
    ~  
    \subcaptionbox{(b) After getting the onion, the pink agent executes \textbf{\emph{Go-Cut-Board-2}}.}
        [0.23\linewidth]{\includegraphics[scale = 0.20]{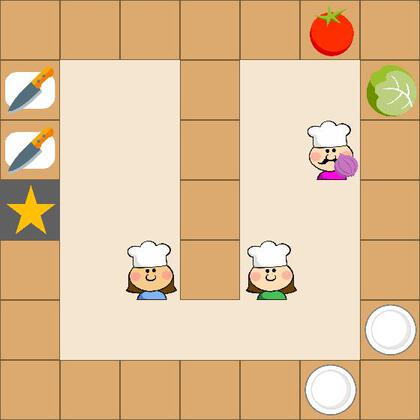}}
    ~    
    \subcaptionbox{(c) The green agent gets the tomato, and it executes \textbf{\emph{Go-Cut-Board-2}}.}
        [0.23\linewidth]{\includegraphics[scale = 0.20]{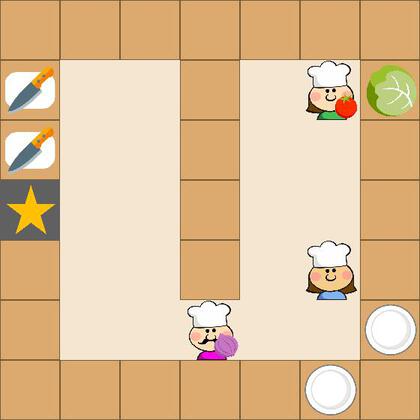}}
    ~    
    \subcaptionbox{(d) The blue agent gets the lettuce, and it executes \textbf{\emph{Go-Cut-Board-1}}.}
        [0.23\linewidth]{\includegraphics[scale = 0.20]{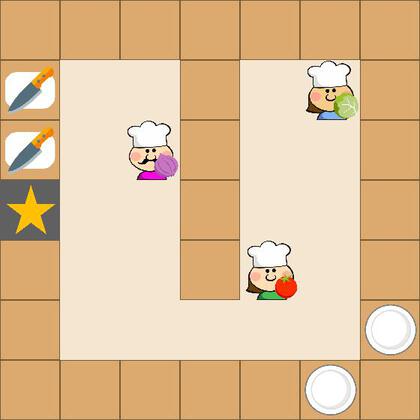}}
    ~    
    \subcaptionbox{(e) After putting the onion on the cutting board, the pink agent executes \textbf{\emph{Chop}}.}
        [0.23\linewidth]{\includegraphics[scale = 0.20]{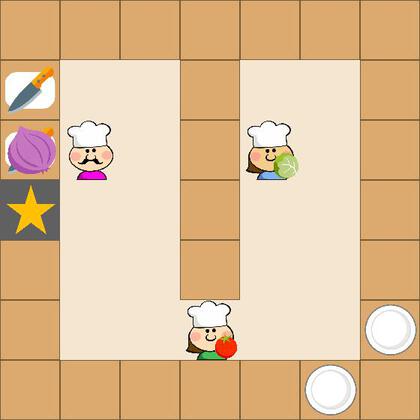}}
    ~    
    \subcaptionbox{(f) The pink agent finishes chopping the onion, and then it executes \textbf{\emph{Get-Onion}} to pick it up.}
        [0.23\linewidth]{\includegraphics[scale = 0.20]{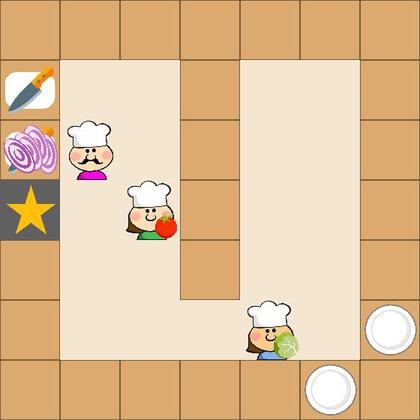}}
    ~    
    \subcaptionbox{(g) After picking up the onion, the pink agent executes \textbf{\emph{Get-Plate-1}}.}
        [0.23\linewidth]{\includegraphics[scale = 0.20]{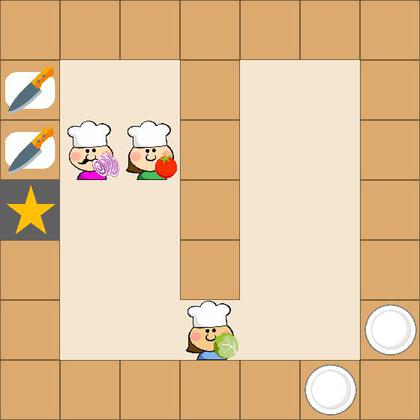}}
    ~    
    \subcaptionbox{(h) After putting the tomato on the cutting board, the green agent executes \textbf{\emph{Chop}}.}
        [0.23\linewidth]{\includegraphics[scale = 0.20]{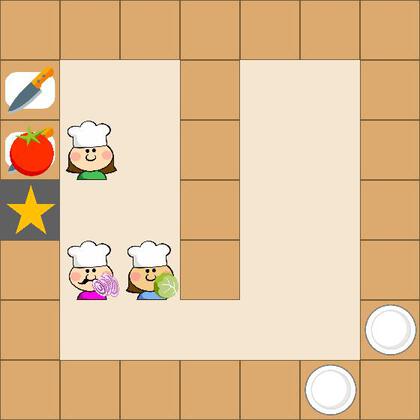}}
    ~    
    \subcaptionbox{(i) After putting the lettuce on the cutting board, the blue agent executes \textbf{\emph{Chop}}.}
        [0.23\linewidth]{\includegraphics[scale = 0.20]{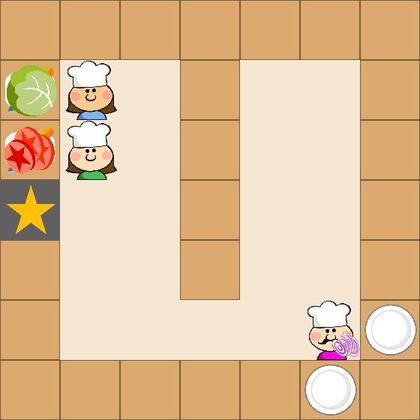}}
    ~    
    \subcaptionbox{(j) The pink agent puts the onion on the plate, and then executes \textbf{\emph{Go-Cut-Board-2}}. The blue agent executes \textbf{\emph{Go-Cut-Board-2}}. The green agent executes \textbf{\emph{Go-Cut-Board-1}}.}
        [0.23\linewidth]{\includegraphics[scale = 0.20]{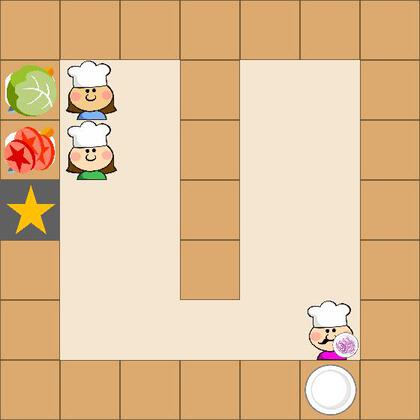}}
    ~    
    \subcaptionbox{(k) The green agent executes \textbf{\emph{Get-Lettuce}}.}
        [0.23\linewidth]{\includegraphics[scale = 0.20]{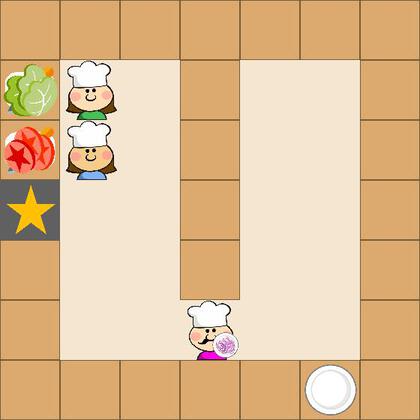}}
    ~    
    \subcaptionbox{(l) After picking up the lettuce, the green agent executes \textbf{\emph{Go-Counter}}.}
        [0.23\linewidth]{\includegraphics[scale = 0.20]{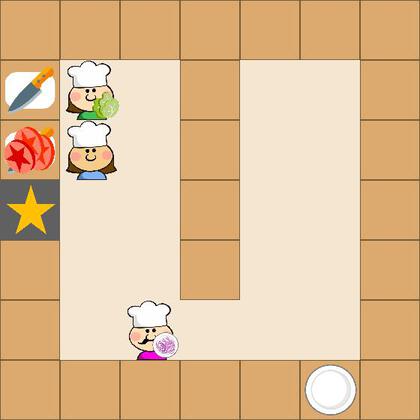}}
    ~    
    \label{fig:Overcooked_visual_mapB}
\end{figure*}

\clearpage

\begin{figure*}[h!]
    \centering
    \captionsetup[subfigure]{labelformat=empty}
    \centering

    \subcaptionbox{(m) The blue agent executes \textbf{\emph{Go-Cut-Board-1}} to make room for the pink agent.}
        [0.23\linewidth]{\includegraphics[scale = 0.20]{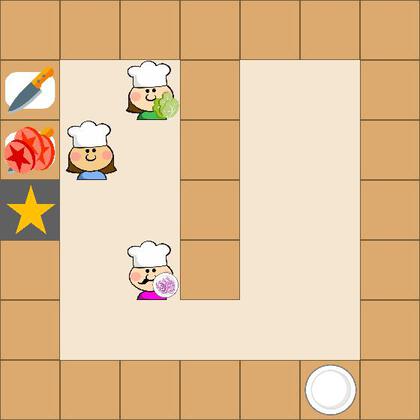}}
    ~    
    \subcaptionbox{(n) The green agent puts the lettuce on the counter}
        [0.23\linewidth]{\includegraphics[scale = 0.20]{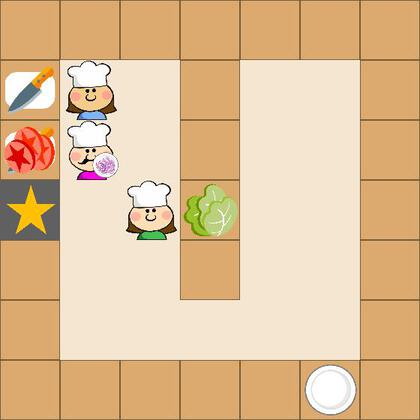}}
    ~
    \subcaptionbox{(o) After merging the tomato into the plate, the pink agent executes \textbf{\emph{Get-Lettuce}}. Meanwhile the green agent steps away to make room for the pink agent.}
        [0.23\linewidth]{\includegraphics[scale = 0.20]{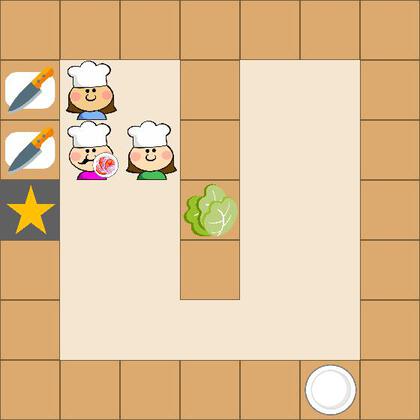}}
    ~    
    \subcaptionbox{(p) After getting the lettuce, the pink agent executes \textbf{\emph{Deliver}}.}
        [0.23\linewidth]{\includegraphics[scale = 0.20]{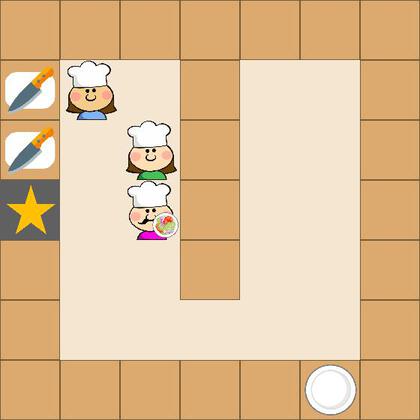}}
    ~

    \subcaptionbox{(q) The pink agent successfully delivers the tomato-lettuce-onion salad.}
        [0.23\linewidth]{\includegraphics[scale = 0.20]{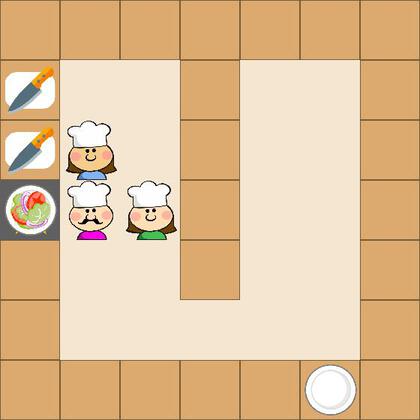}}
    ~
    \caption{Visualization of running decentralized policies learned by Mac-IAICC in Overcooked-C.}
    \label{fig:Overcooked_visual_mapB}
\end{figure*}

\clearpage
\subsection{Warehouse Tool Delivery}

\textbf{\emph{Warehouse A:}} 

\begin{figure*}[h!]
    \centering
    \captionsetup[subfigure]{labelformat=empty}
    \centering
    \subcaptionbox{(a) Initial State.\vspace{2mm}}
        [0.30\linewidth]{\includegraphics[scale=0.26]{results/WTD/wtd_a_small.png}}
    \quad
    \centering
    \subcaptionbox{(b) Mobile robots moves towards the table by running \emph{\textbf{Get-Tool}}, and arm robot runs \emph{\textbf{Search-Tool(0)}} to find Tool-0.\vspace{2mm}}
        [0.30\linewidth]{\includegraphics[scale=0.18]{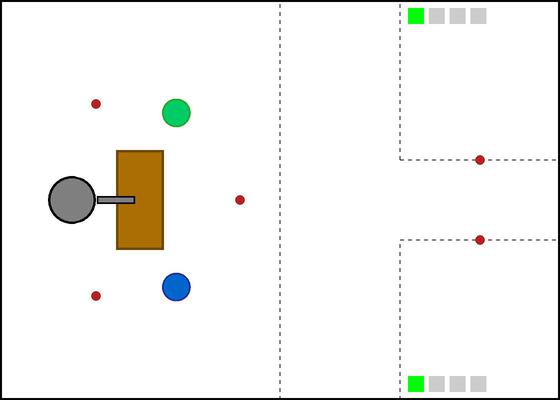}}
    \quad
    \centering
    \subcaptionbox{(c) Mobile robots wait there and arm robot keeps looking for Tool-0.\vspace{2mm}}
        [0.30\linewidth]{\includegraphics[scale=0.18]{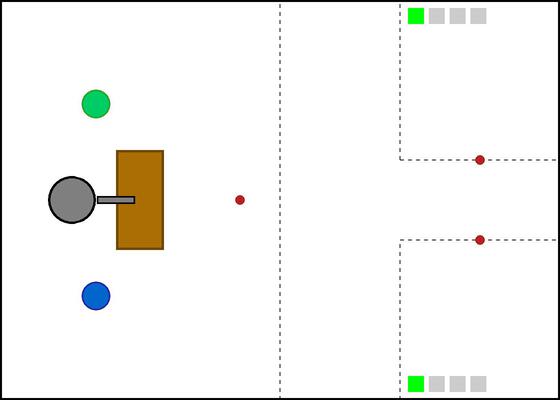}}
    \quad
    \centering
    \subcaptionbox{(d) Arm robot executes \emph{\textbf{Pass-to-M(1)}} to pass Tool-0 to the blue robot.\vspace{2mm}}
        [0.30\linewidth]{\includegraphics[scale=0.18]{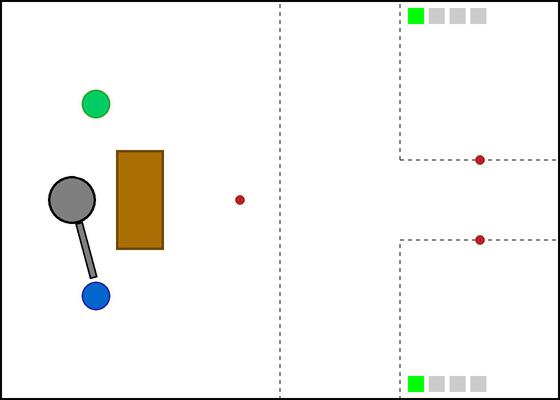}}
    \quad
    \centering
    \subcaptionbox{(e) Arm robot executes \emph{\textbf{Search-Tool(0)}} to find Tool-0, and blue robot moves to workshop-1 by executing \emph{\textbf{Go-W(1)}}.\vspace{2mm}}
        [0.30\linewidth]{\includegraphics[scale=0.18]{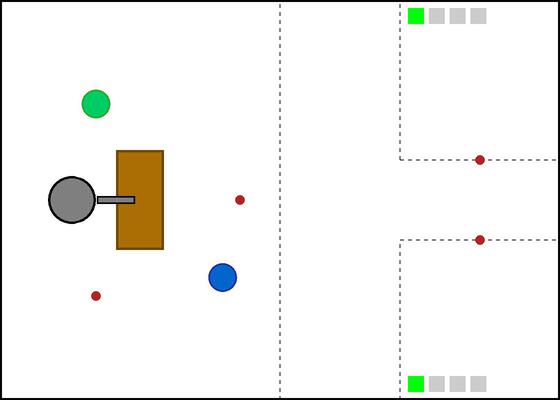}}
    \quad
    \centering
    \subcaptionbox{(f) Blue robot successfully delivers Tool-0 to workshop-1.\vspace{2mm}}
        [0.30\linewidth]{\includegraphics[scale=0.18]{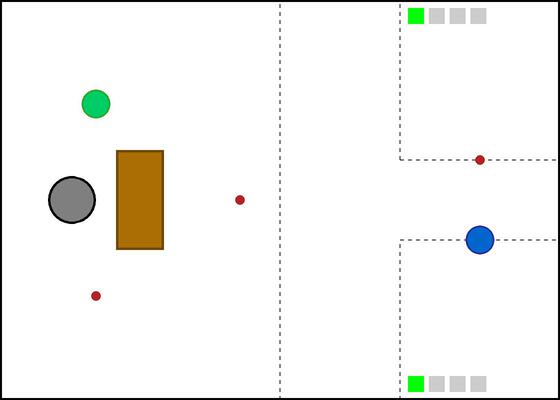}}
    \quad
    \centering
    \subcaptionbox{(g) Blue robot runs \emph{\textbf{Get-Tool}} to go back table, and arm robot executes \emph{\textbf{Pass-to-M(0)}} to pass Tool-0 to green robot.\vspace{2mm}}
        [0.30\linewidth]{\includegraphics[scale=0.18]{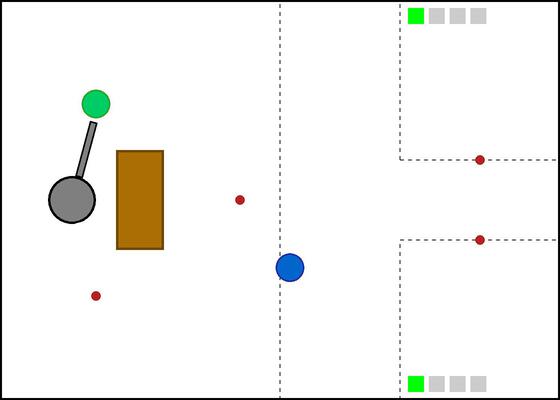}}
    \quad
    \centering
    \subcaptionbox{(h) Green robot executes \emph{\textbf{Go-W(0)}} and arm robot runs \emph{\textbf{Search-Tool(1)}}. \vspace{2mm}}
        [0.30\linewidth]{\includegraphics[scale=0.18]{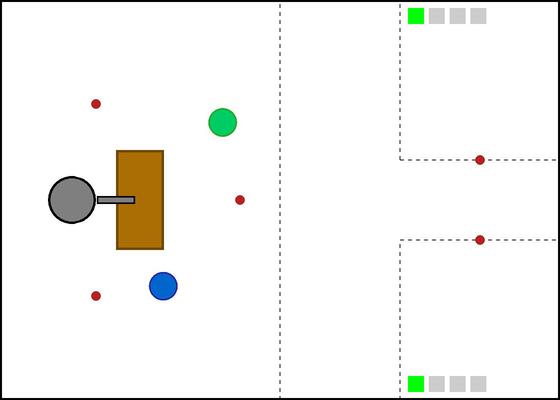}}
    \quad
    \centering
    \subcaptionbox{(i) Green robot successfully delivers Tool-0 to workshop-0. Human-0 and human-1 finish subtask-0 and start to do subtask-1 with delivered Tool-0.\vspace{2mm}}
        [0.30\linewidth]{\includegraphics[scale=0.18]{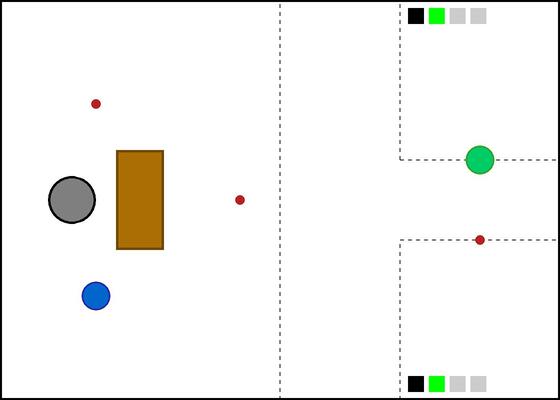}}
    \quad
    \centering
    \subcaptionbox{(j) Green robot runs \emph{\textbf{Get-Tool}} to go back table, and arm robot executes \emph{\textbf{Pass-to-M(1)}} to pass a Tool-1 to blue robot.\vspace{2mm}}
        [0.30\linewidth]{\includegraphics[scale=0.18]{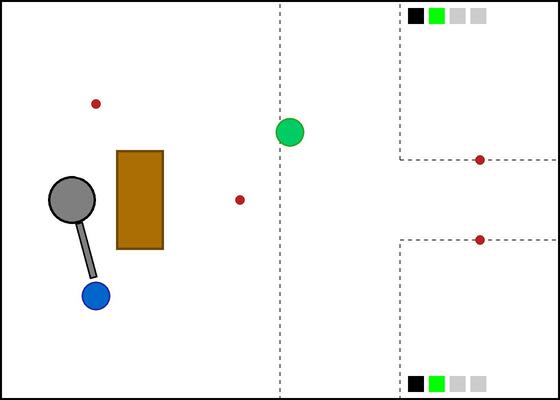}}
    \quad
    \centering
    \subcaptionbox{(k) Blue robot executes \emph{\textbf{Go-W(1)}} and arm robot runs \emph{\textbf{Search-Tool(1)}}. \vspace{2mm}}
        [0.30\linewidth]{\includegraphics[scale=0.18]{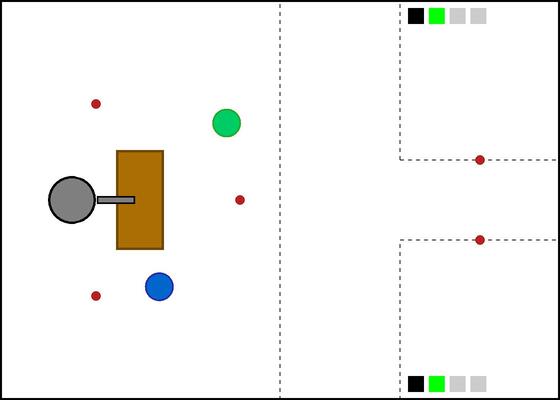}}
    \quad
    \centering
    \subcaptionbox{(l)  Blue robot successfully delivers a Tool-1 to workshop-1.\vspace{2mm}}
        [0.30\linewidth]{\includegraphics[scale=0.18]{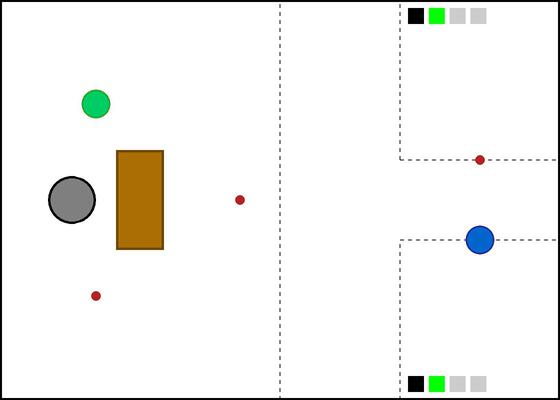}}
\end{figure*}
\begin{figure*}[h!]
    \centering
    \captionsetup[subfigure]{labelformat=empty}
    \centering
    \subcaptionbox{(m) Arm robot executes \emph{\textbf{Pass-to-M(0)}} to pass Tool-1 to green robot. Blue robot runs \emph{\textbf{Get-Tool}} to go back table. \vspace{2mm}}
        [0.30\linewidth]{\includegraphics[scale=0.18]{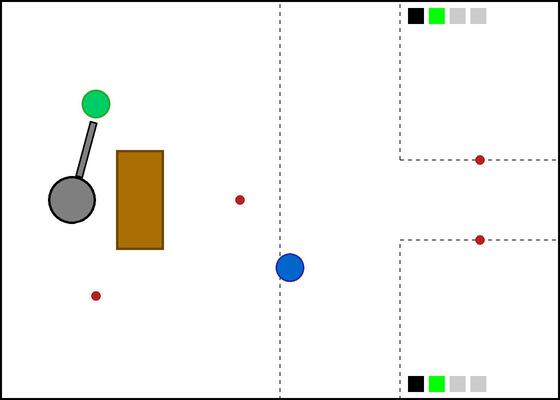}}
    \quad
    \centering
    \subcaptionbox{(n) Green robot successfully delivers Tool-1 to workshop-0. Human-0 and human-1 finish subtask-1 and start to do subtask-2 with delivered Tool-1.\vspace{2mm}}
        [0.30\linewidth]{\includegraphics[scale=0.18]{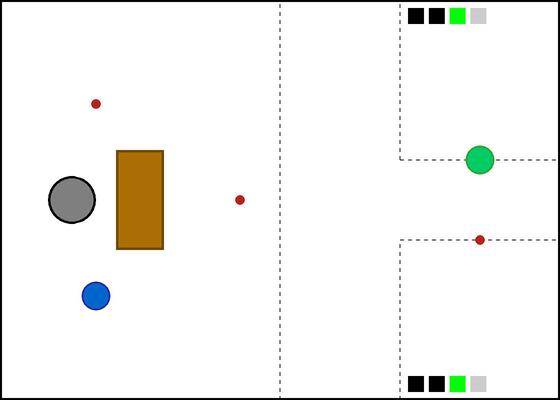}}
    \quad
    \centering
    \subcaptionbox{(o) Arm robot executes \emph{\textbf{Pass-to-M(1)}} to pass Tool-2 to blue robot. Green robot runs \emph{\textbf{Get-Tool}} to go back table.\vspace{2mm}}
        [0.30\linewidth]{\includegraphics[scale=0.18]{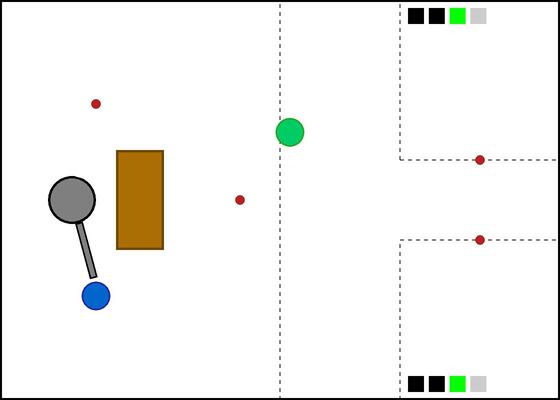}}
    \quad
    \centering
    \subcaptionbox{(p) Blue robot executes \emph{\textbf{Go-W(1)}}. Arm robot runs \emph{\textbf{Search-Tool(2)}}. \vspace{2mm}}
        [0.30\linewidth]{\includegraphics[scale=0.18]{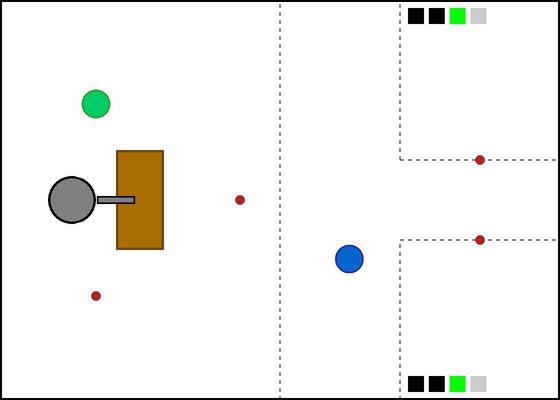}}
    \quad
    \centering
    \subcaptionbox{(q) Blue robot successfully delivers Tool-2 to human-0. \vspace{2mm}}
        [0.30\linewidth]{\includegraphics[scale=0.18]{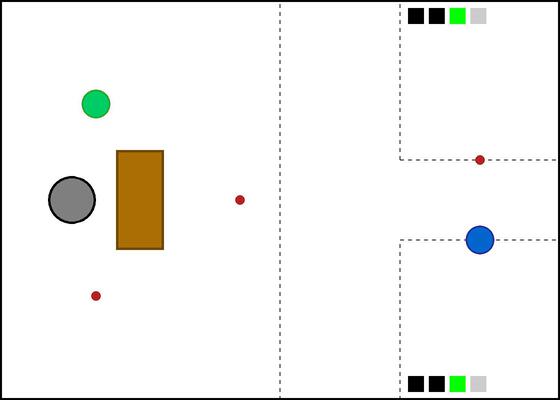}}
    \quad
    \centering
    \subcaptionbox{(r) Arm robot executes \emph{\textbf{Pass-to-M(0)}} to pass Tool-2 to green robot. Blue robot runs \emph{\textbf{Get-Tool}} to go back table.\vspace{2mm}}
        [0.30\linewidth]{\includegraphics[scale=0.18]{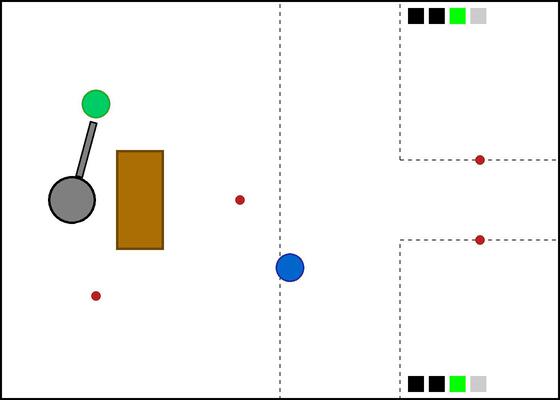}}

    \quad
    \centering
    \subcaptionbox{(s) Green robot directly goes to workshop-0 by running \emph{\textbf{Go-W(0)}} and finishes the last tool delivery for human-0. The entire task is done.}
        [0.30\linewidth]{\includegraphics[scale=0.18]{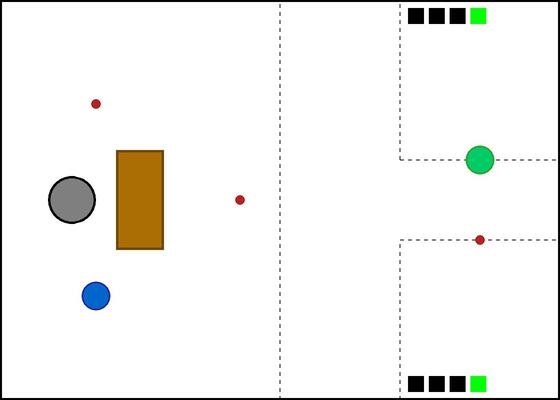}}
    \label{wtd_a_behavior}
\end{figure*}

\clearpage
\textbf{\emph{Warehouse-B:}}.  

\begin{figure*}[h!]
    \centering
    \captionsetup[subfigure]{labelformat=empty}
    \centering
    \subcaptionbox{(a) Initial State.\vspace{2mm}}
        [0.30\linewidth]{\includegraphics[scale=0.24]{results/WTD/wtd_c_small.png}}
    ~
    \centering
    \subcaptionbox{(b) Green robot moves towards the table by running \emph{\textbf{Get-Tool}}. Blue robot moves to workshop-0 by executing \emph{\textbf{Go-W(0)}}. Arm robot runs \emph{\textbf{Search-Tool(0)}} to find Tool-0.\vspace{2mm}}
        [0.30\linewidth]{\includegraphics[scale=0.16]{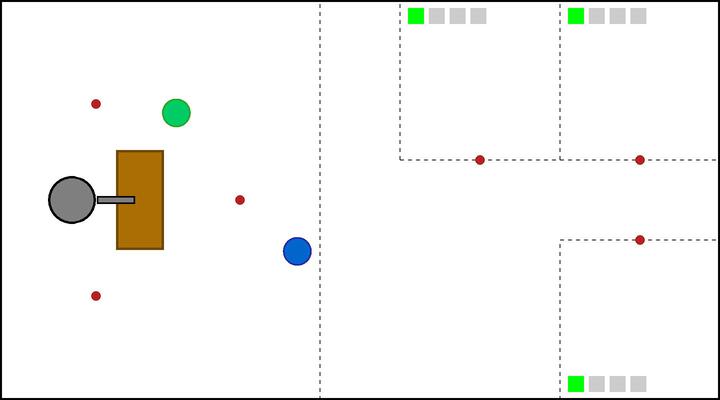}}
    ~
    \centering
    \subcaptionbox{(c) Green robot waits there and arm robot keeps looking for Tool-0.\vspace{2mm}}
        [0.30\linewidth]{\includegraphics[scale=0.16]{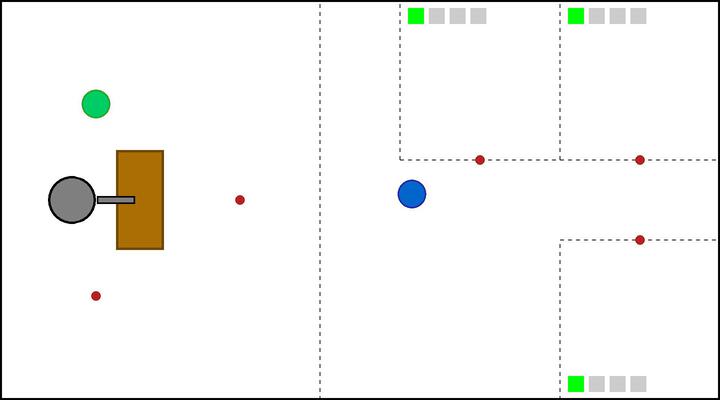}}
    ~
    \centering
    \subcaptionbox{(d) Blue robot reaches workshop-0.\vspace{2mm}}
        [0.30\linewidth]{\includegraphics[scale=0.16]{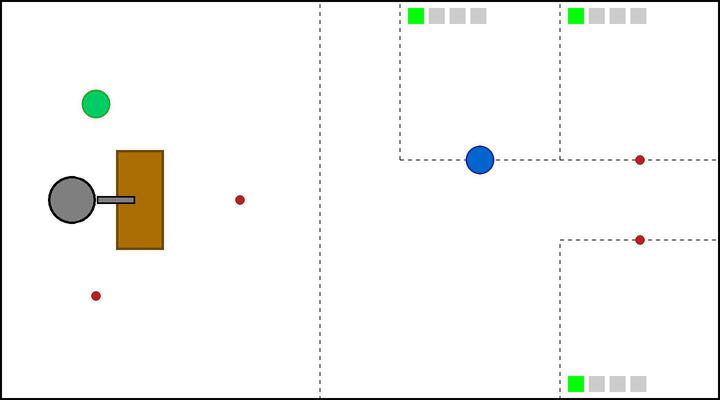}}
    ~
    \centering
    \subcaptionbox{(e) Blue robot runs \emph{\textbf{Get-Tool}} to go back table. \vspace{2mm}}
        [0.30\linewidth]{\includegraphics[scale=0.16]{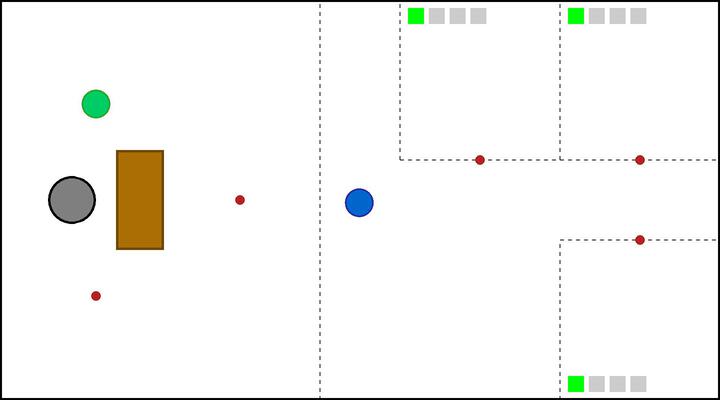}}  
    ~
    \centering
    \subcaptionbox{(f) Arm robot executes \emph{\textbf{Pass-to-M(0)}} to pass Tool-0 to green robot. \vspace{2mm}}
        [0.30\linewidth]{\includegraphics[scale=0.16]{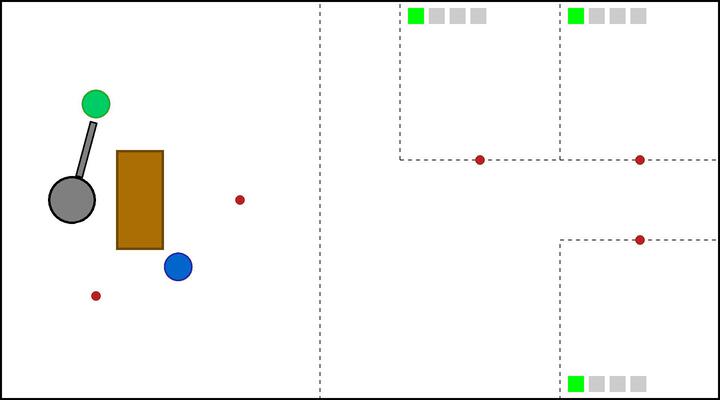}}
    ~
    \centering
    \subcaptionbox{(g)  Arm robot runs \emph{\textbf{Search-Tool(0)}} to find the 2nd Tool-0.\vspace{2mm}}
        [0.30\linewidth]{\includegraphics[scale=0.16]{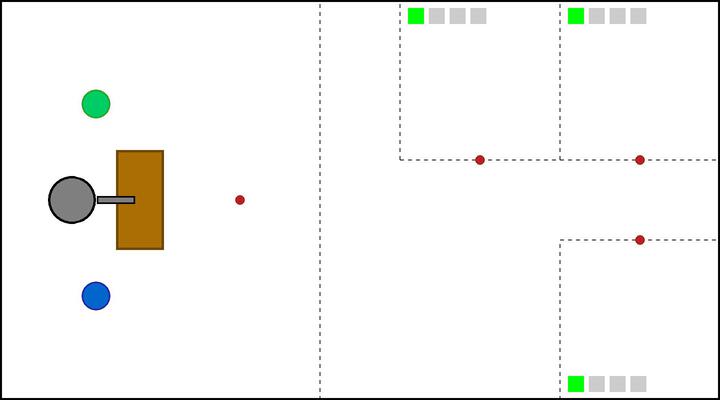}}
    ~
    \centering
    \subcaptionbox{(h) Arm robot executes \emph{\textbf{Pass-to-M(0)}} to pass the 2nd Tool-0 to green robot.  \vspace{2mm}}
        [0.30\linewidth]{\includegraphics[scale=0.16]{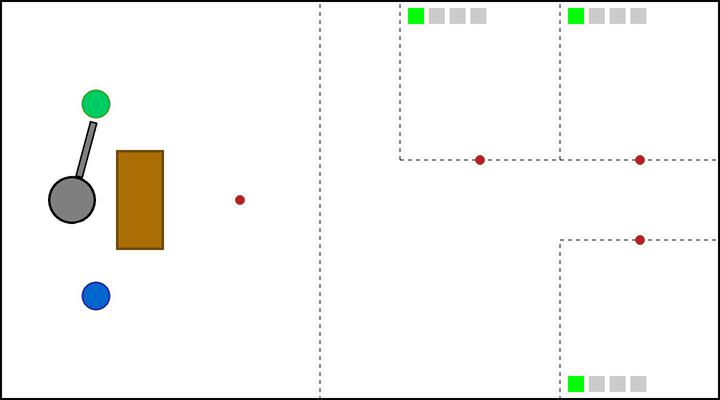}}
    ~
    \centering
    \subcaptionbox{(i) Arm robot runs \emph{\textbf{Search-Tool(0)}} to find the the 3rd Tool-0. Green robot moves to workshop-1 by executing \emph{\textbf{Go-W(1)}}.
    \vspace{2mm}}
        [0.30\linewidth]{\includegraphics[scale=0.16]{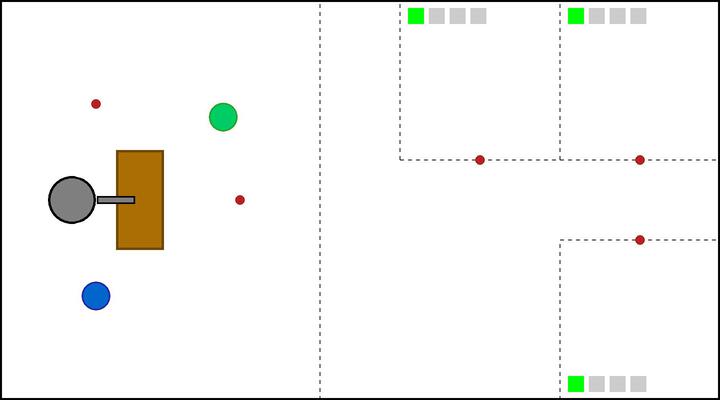}}
    ~
    \centering
    \subcaptionbox{(j) Green robot successfully delivers Tool-0 to workshop-1. \vspace{2mm}}
        [0.30\linewidth]{\includegraphics[scale=0.16]{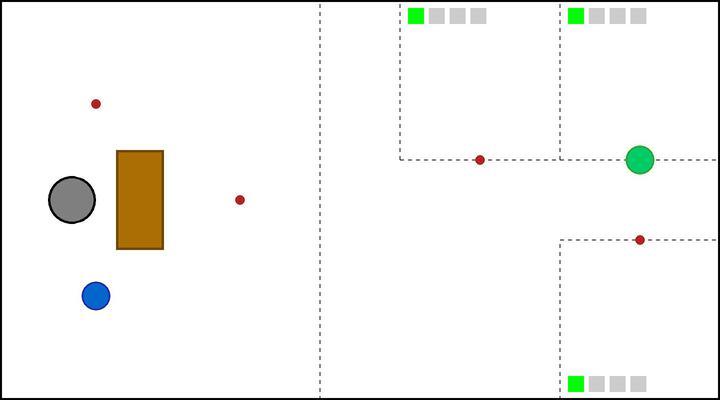}}
    ~
    \centering
    \subcaptionbox{(k) Arm robot executes \emph{\textbf{Pass-to-M(1)}} to pass the 3rd Tool-0 to blue robot. Green robot moves to workshop-0 by executing \emph{\textbf{Go-W(0)}}. \vspace{2mm}}
        [0.30\linewidth]{\includegraphics[scale=0.16]{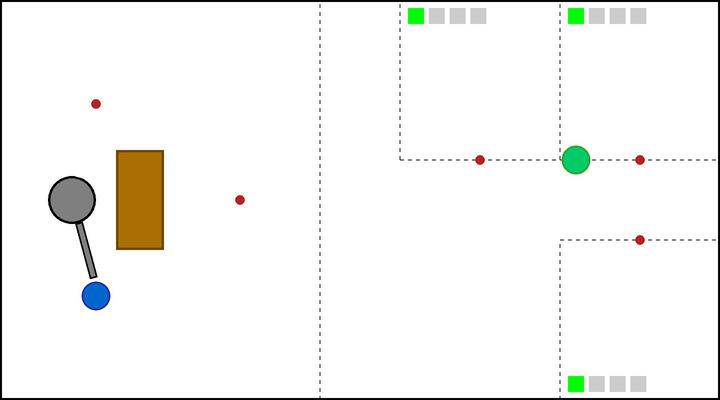}}  
    ~
    \centering
    \subcaptionbox{(l) Arm robot runs \emph{\textbf{Search-Tool(1)}} to find Tool-1. Blue robot moves to workshop-2 by executing \emph{\textbf{Go-W(2)}}. \vspace{2mm}}
        [0.30\linewidth]{\includegraphics[scale=0.16]{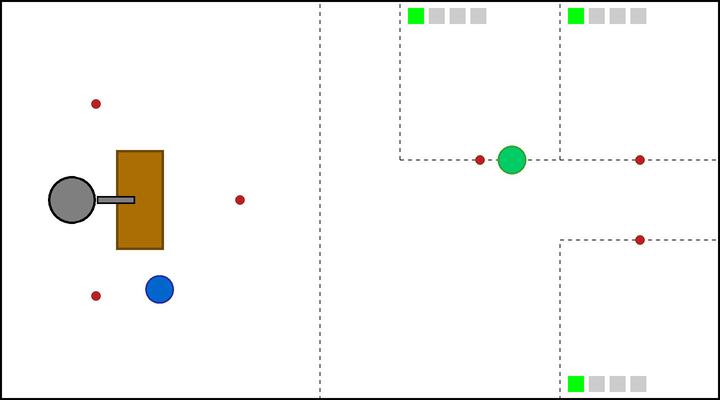}}
\end{figure*}
\begin{figure*}[h!]
    \centering
    \captionsetup[subfigure]{labelformat=empty}          
    ~
    \centering        
    ~
    \centering
    \subcaptionbox{(m) Green robot successfully delivers a Tool-0 to workshop-0. \vspace{2mm}}
        [0.30\linewidth]{\includegraphics[scale=0.16]{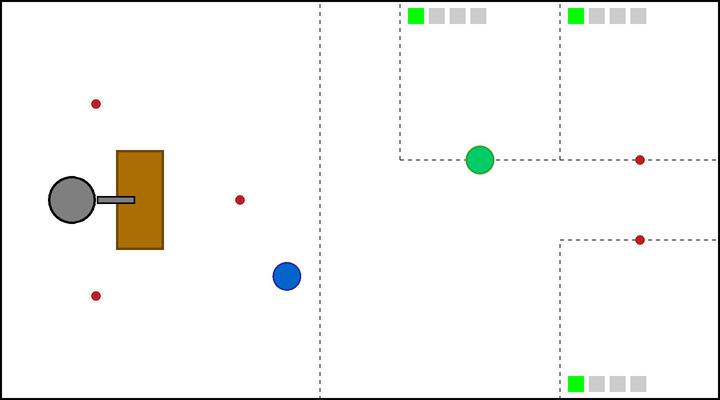}}
    ~
    \centering
    \subcaptionbox{(n) Blue robot successfully delivers a Tool-0 to workshop-2. Arm robot runs \emph{\textbf{Search-Tool(1)}} to find another Tool-1.\vspace{2mm}}
        [0.30\linewidth]{\includegraphics[scale=0.16]{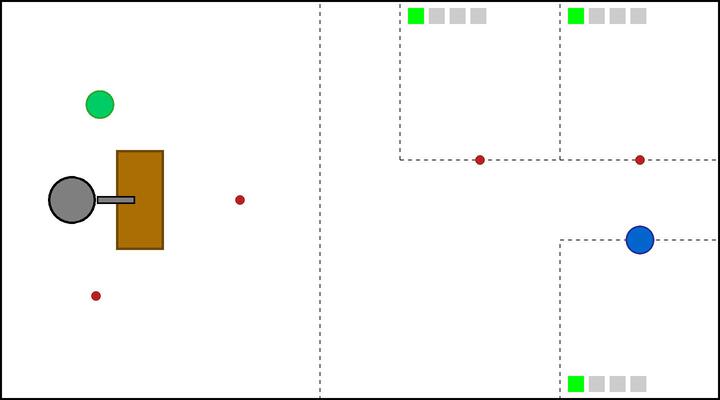}}
    ~
    \centering
    \subcaptionbox{(o) Blue robot runs \emph{\textbf{Get-Tool}} to go back table. All humans finish subtask-0 and start to do subtask-1.\vspace{2mm}}
        [0.30\linewidth]{\includegraphics[scale=0.16]{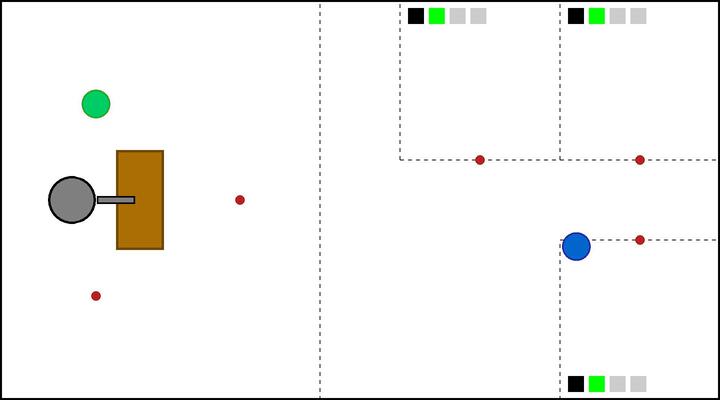}}
    ~
    \subcaptionbox{(p)Arm robot executes \emph{\textbf{Pass-to-M(0)}} to pass a Tool-1 to green robot.  \vspace{2mm}}
        [0.30\linewidth]{\includegraphics[scale=0.16]{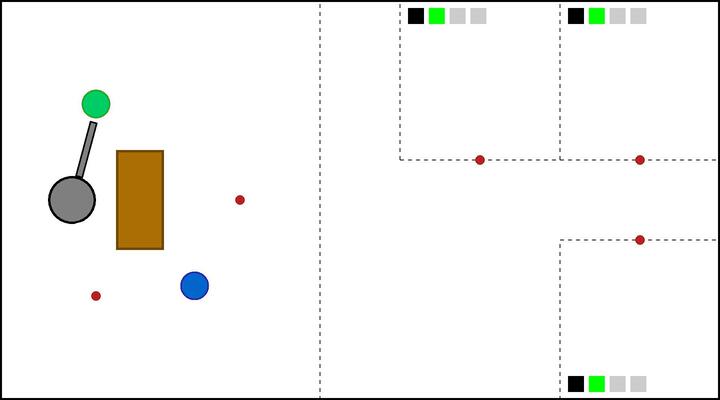}}
    ~
    \centering
    \subcaptionbox{(q) Arm robot executes \emph{\textbf{Pass-to-M(0)}} to pass another Tool-1 to green robot.\vspace{2mm}}
        [0.30\linewidth]{\includegraphics[scale=0.16]{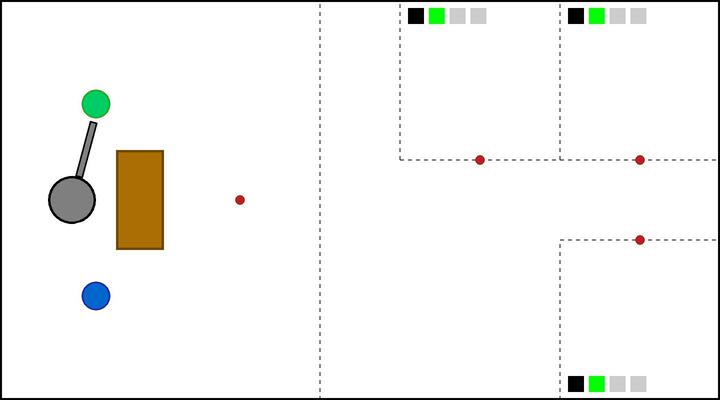}}
    ~
    \centering
    \subcaptionbox{(r) Green robot moves to workshop-1 by executing \emph{\textbf{Go-W(1)}}. Arm robot runs \emph{\textbf{Search-Tool(1)}} to find the 3rd Tool-1.\vspace{2mm}}
        [0.30\linewidth]{\includegraphics[scale=0.16]{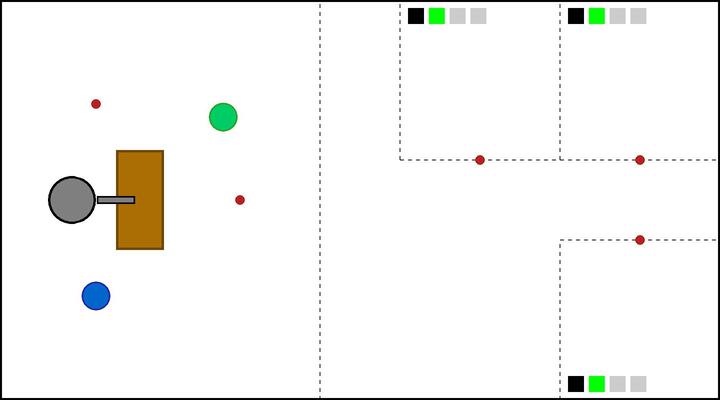}}
    ~
    \centering
    \subcaptionbox{(s) Green robot successfully delivers a Tool-0 to workshop-0. \vspace{2mm}}
        [0.30\linewidth]{\includegraphics[scale=0.16]{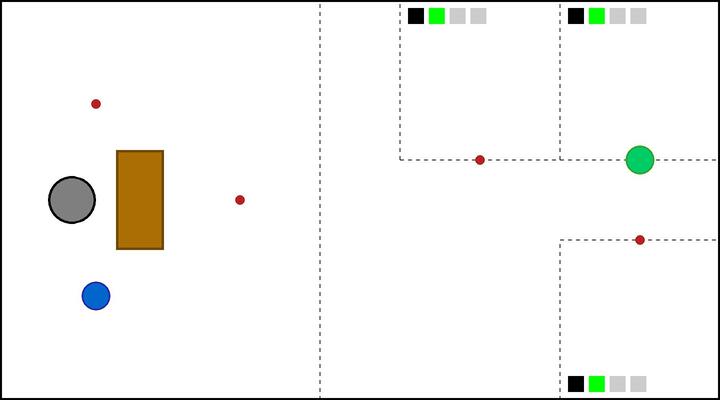}}
    ~
    \centering
    \subcaptionbox{(t) Green robot moves to workshop-0 by executing \emph{\textbf{Go-W(0)}}. Arm robot executes \emph{\textbf{Pass-to-M(1)}} to pass the 3rd Tool-1 to blue robot. \vspace{2mm}}
        [0.30\linewidth]{\includegraphics[scale=0.16]{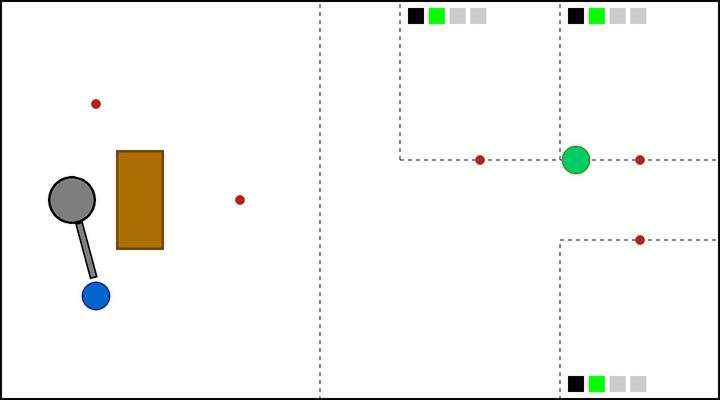}}
    ~
    \centering
    \subcaptionbox{(u) Arm robot runs \emph{\textbf{Search-Tool(2)}} to find Tool-2. Green robot successfully delivers a Tool-1 to workshop-0. Blue robot moves to workshop-2 by executing \emph{\textbf{Go-W(2)}}.\vspace{2mm}}
        [0.30\linewidth]{\includegraphics[scale=0.16]{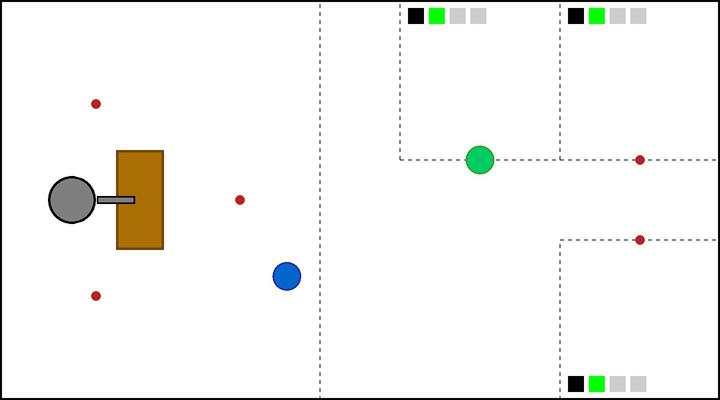}}
    ~
    \centering
    \subcaptionbox{(v) Green robot runs \emph{\textbf{Get-Tool}} to go back table.\vspace{2mm}}
        [0.30\linewidth]{\includegraphics[scale=0.16]{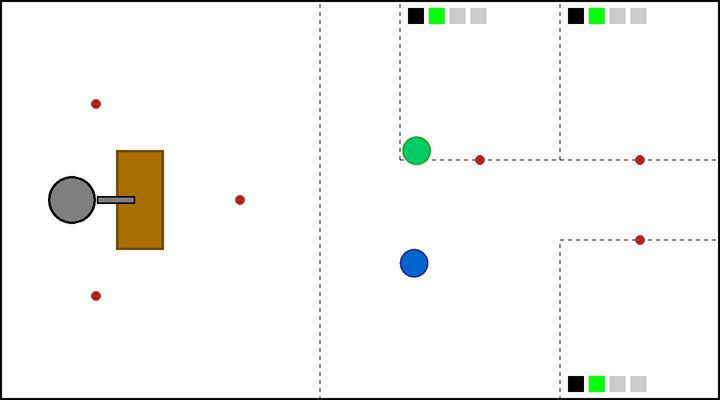}}
    ~
    \centering
    \subcaptionbox{(w) Arm robot runs \emph{\textbf{Search-Tool(2)}} to find another Tool-2. Blue robot successfully delivers a Tool-1 to workshop-2. \vspace{2mm}}
        [0.30\linewidth]{\includegraphics[scale=0.16]{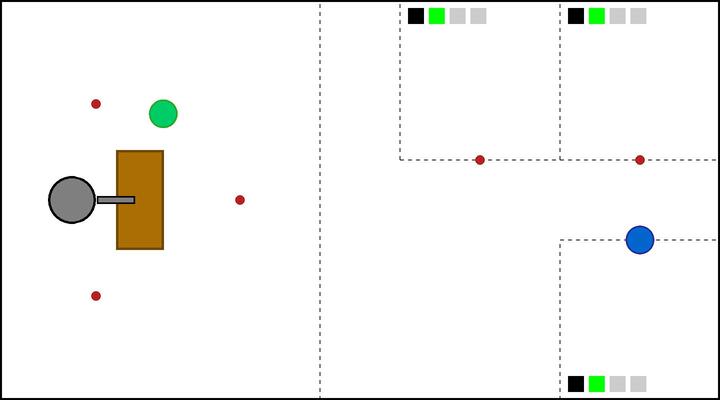}}
    ~
    \centering
    \subcaptionbox{(x) Blue robot runs \emph{\textbf{Get-Tool}} to go back table. \vspace{2mm}}
        [0.30\linewidth]{\includegraphics[scale=0.16]{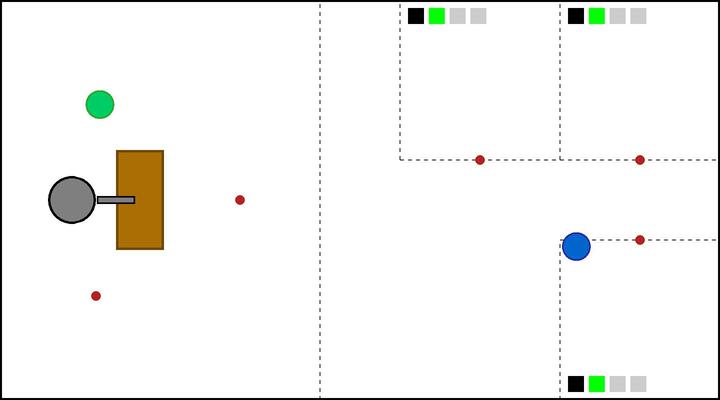}}
    ~
    \centering
    \subcaptionbox{(y) Arm robot executes \emph{\textbf{Pass-to-M(0)}} to pass a Tool-2 to green robot.\vspace{2mm}}
        [0.30\linewidth]{\includegraphics[scale=0.16]{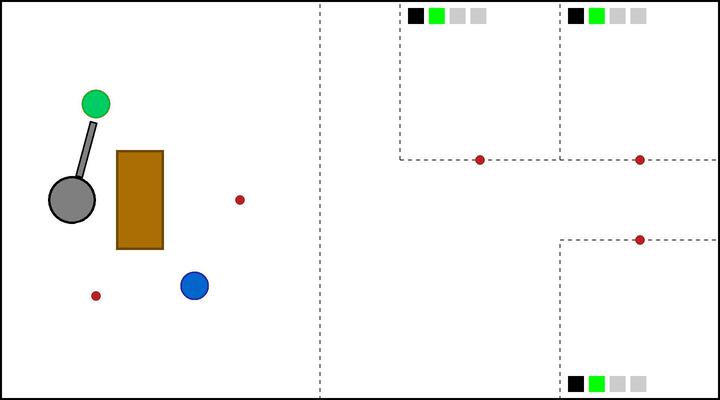}}  
    ~
    \centering
    \subcaptionbox{(z) Arm robot executes \emph{\textbf{Pass-to-M(0)}} to pass another Tool-2 to green robot. All humans finish subtask-1 and start to do subtask-2.\vspace{2mm}}
        [0.30\linewidth]{\includegraphics[scale=0.16]{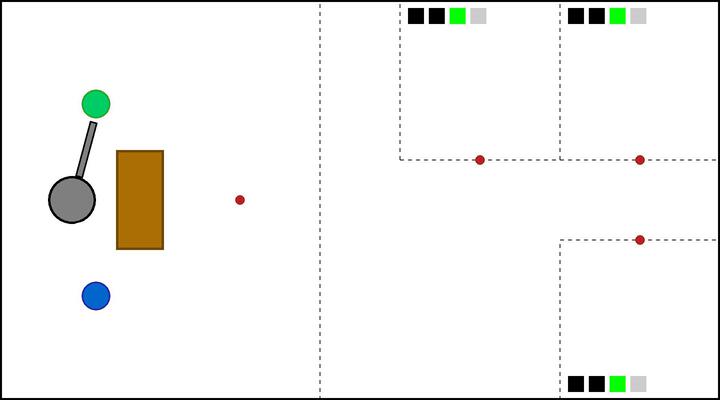}}
    ~
    \centering
    \subcaptionbox{(A) Arm robot runs \emph{\textbf{Search-Tool(2)}} to find the 3rd Tool-2. Green robot moves to workshop-1 by executing \emph{\textbf{Go-W(1)}}.\vspace{2mm}}
        [0.30\linewidth]{\includegraphics[scale=0.16]{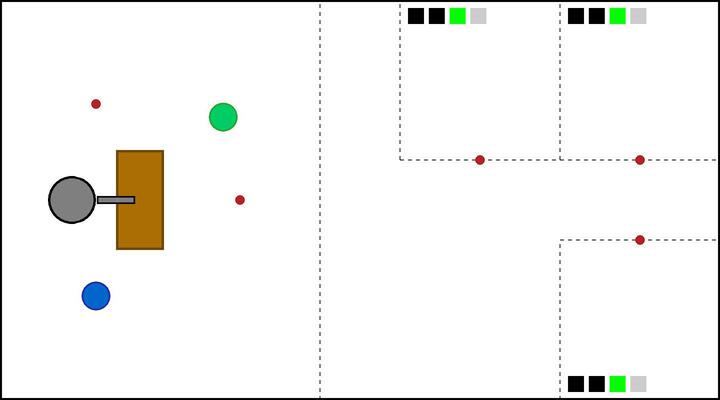}}  
        
\end{figure*}
\begin{figure*}[h!]
    \centering
    \captionsetup[subfigure]{labelformat=empty}          
    ~
    \centering
    ~
    \centering
    \subcaptionbox{(B) Green robot successfully delivers a Tool-2 to workshop-1. \vspace{2mm}}
        [0.30\linewidth]{\includegraphics[scale=0.16]{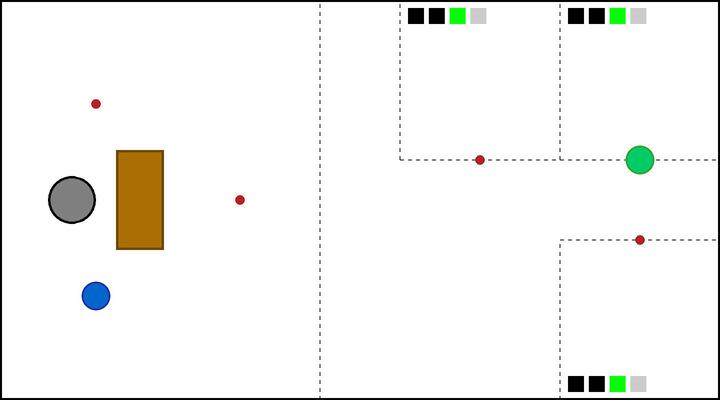}}
    ~
    \centering
    \subcaptionbox{(C) Arm robot executes \emph{\textbf{Pass-to-M(1)}} to pass the 3rd Tool-2 to blue robot.\vspace{2mm}}
        [0.30\linewidth]{\includegraphics[scale=0.16]{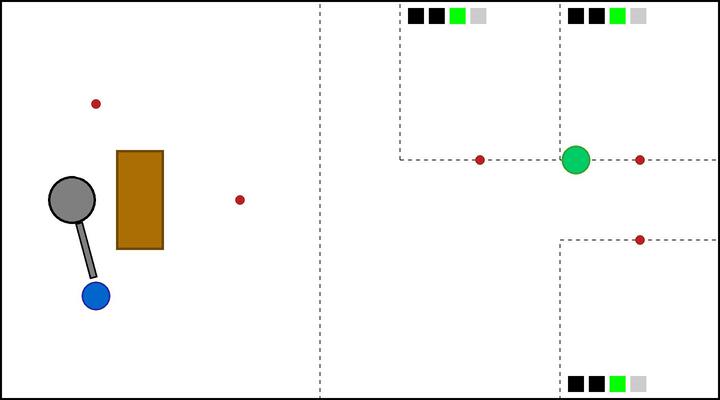}}  
    ~
    \centering
    \subcaptionbox{(D) Green robot successfully delivers a Tool-2 to workshop-0. Blue robot moves to workshop-2 by executing \emph{\textbf{Go-W(2)}}.\vspace{2mm}}
        [0.30\linewidth]{\includegraphics[scale=0.16]{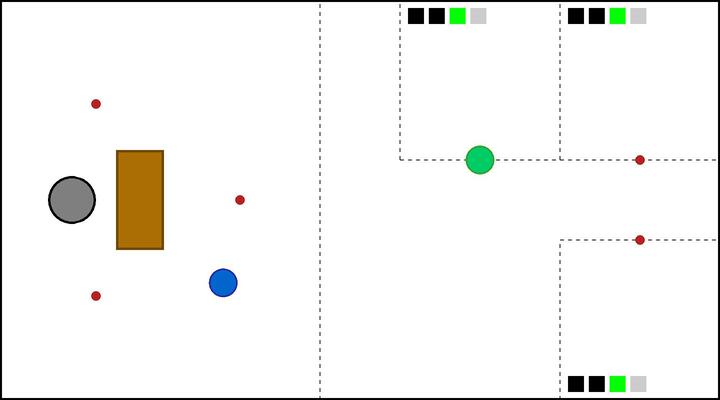}}
    ~
    \centering
    \subcaptionbox{(E) Blue robot successfully delivers a Tool-2 to workshop-2. Humans have received all tools, and for robots, the task is done.\vspace{2mm}}
        [0.30\linewidth]{\includegraphics[scale=0.16]{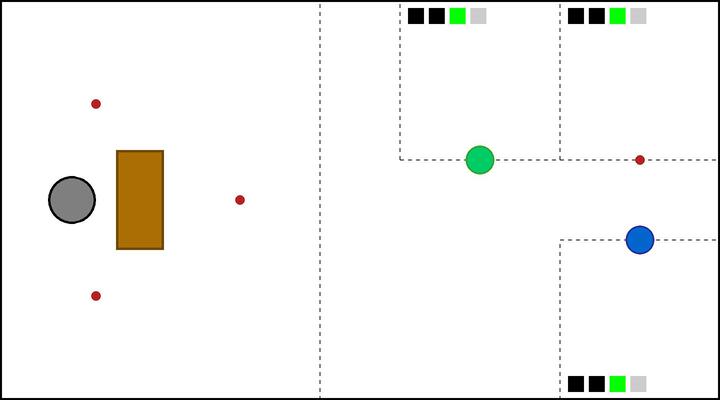}}
    \label{wtd_c_behavior}
\end{figure*}

\clearpage

\textbf{\emph{Warehouse-C:}}  

\begin{figure*}[h!]
    \centering
    \captionsetup[subfigure]{labelformat=empty}
    \centering
    \subcaptionbox{(a) Initial State.\vspace{2mm}}
        [0.30\linewidth]{\includegraphics[scale=0.24]{results/WTD/wtd_e_small.png}}
    ~
    \centering
    \subcaptionbox{(b) Mobile robots move towards the table by running \emph{\textbf{Get-Tool}}. Arm robot runs \emph{\textbf{Search-Tool(0)}} to find the 1st Tool-0.\vspace{2mm}}
        [0.30\linewidth]{\includegraphics[scale=0.16]{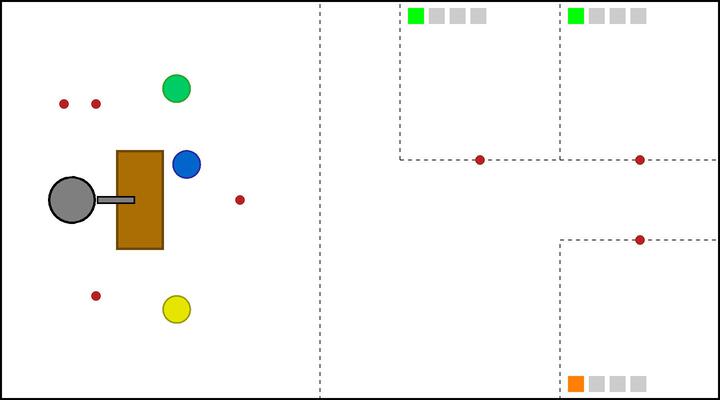}}
    ~
    \centering
    \subcaptionbox{(c) Mobile robots wait there and arm robot keeps looking for the 1st Tool-0.\vspace{2mm}}
        [0.30\linewidth]{\includegraphics[scale=0.16]{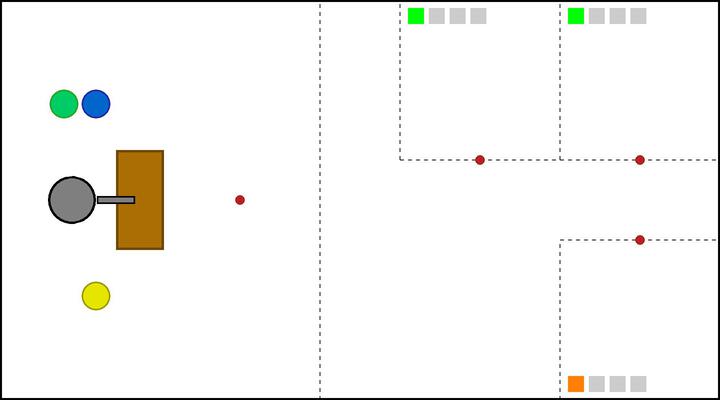}}
    ~
    \centering
    \subcaptionbox{(d) Arm robot executes \emph{\textbf{Pass-to-M(0)}} to pass a Tool-0 to green robot.\vspace{2mm}}
        [0.30\linewidth]{\includegraphics[scale=0.16]{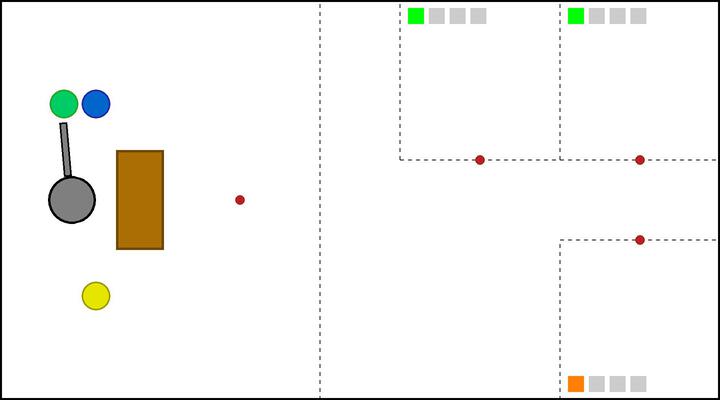}}
    ~
    \centering
    \subcaptionbox{(e) Arm robot runs \emph{\textbf{Search-Tool(0)}} to find the 2nd Tool-0. Green robot moves to workshop-0 by executing \emph{\textbf{Go-W(0)}}.\vspace{2mm}}
        [0.30\linewidth]{\includegraphics[scale=0.16]{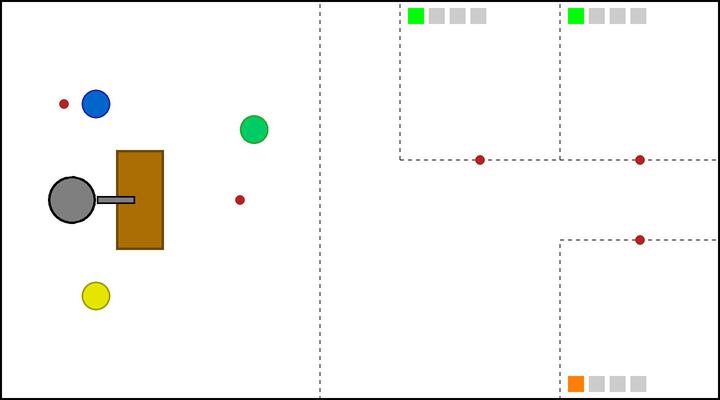}}  
    ~
    \centering
    \subcaptionbox{(f) Green robot successfully delivers the a Tool-0 to workshop-0.  \vspace{2mm}}
        [0.30\linewidth]{\includegraphics[scale=0.16]{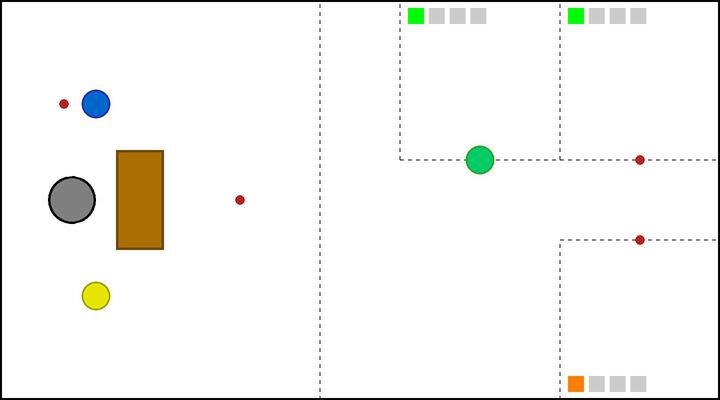}}
    ~
    \centering
    \subcaptionbox{(g) Green robot runs \emph{\textbf{Get-Tool}} to go back table. Arm robot executes \emph{\textbf{Pass-to-M(1)}} to pass a Tool-0 to blue robot.
\vspace{2mm}}
        [0.30\linewidth]{\includegraphics[scale=0.16]{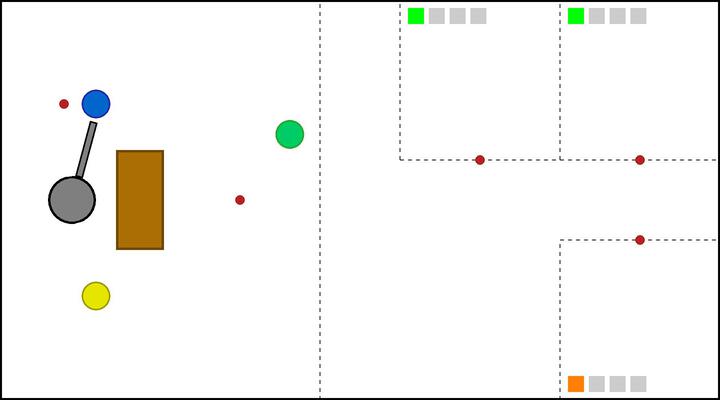}}
    ~
    \centering
    \subcaptionbox{(h) Arm robot runs \emph{\textbf{Search-Tool(0)}} to find the 3rd Tool-0. Blue robot moves to workshop-1 by executing \emph{\textbf{Go-W(1)}}. Yellow robot moves to workshop-0 by executing \emph{\textbf{Go-W(0)}}. \vspace{2mm}}
        [0.30\linewidth]{\includegraphics[scale=0.16]{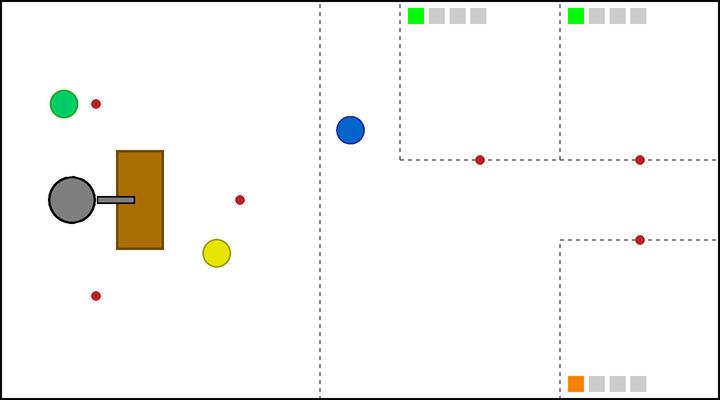}}
    ~
    \centering
    \subcaptionbox{(i) Blue robot successfully delivers the a Tool-0 to workshop-1. Yellow robot reaches workshop-0 and observes that human-0 has got Tool-0. Human-2 finishes subtask-0 and waits for Tool-0.
    \vspace{2mm}}
        [0.30\linewidth]{\includegraphics[scale=0.16]{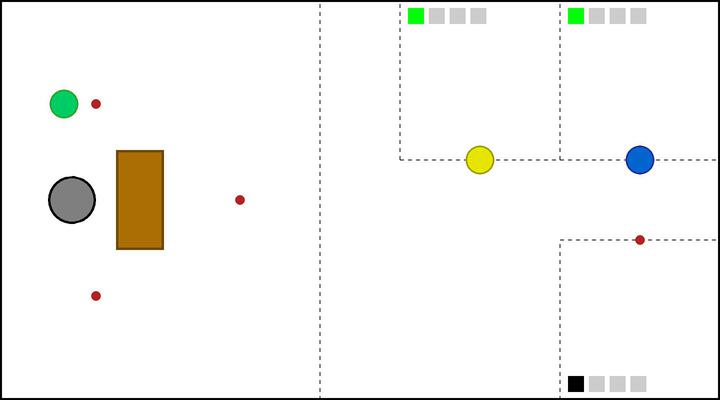}}
    ~
    \centering
    \subcaptionbox{(j) Arm robot executes \emph{\textbf{Pass-to-M(0)}} to pass a Tool-0 to green robot. Yellow and blue robots run \emph{\textbf{Get-Tool}} to go back table.\vspace{2mm}}
        [0.30\linewidth]{\includegraphics[scale=0.16]{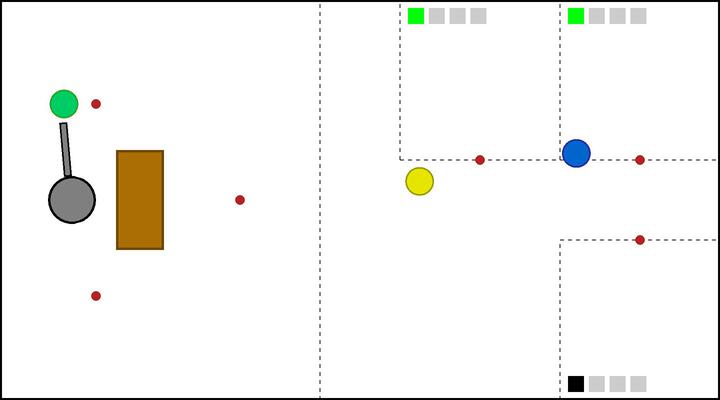}}
    ~
    \centering
    \subcaptionbox{(k) Arm robot runs \emph{\textbf{Search-Tool(1)}} to find the 1st Tool-1. Green robot moves to workshop-0 by executing \emph{\textbf{Go-W(0)}}. \vspace{2mm}}
        [0.30\linewidth]{\includegraphics[scale=0.16]{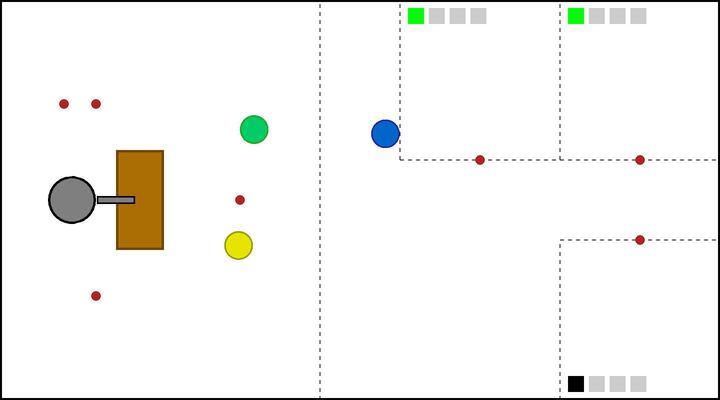}}  
    ~
    \centering
    \subcaptionbox{(l) Green robot reaches workshop-0 and observes that human-0 does not need Tool-0.  \vspace{2mm}}
        [0.30\linewidth]{\includegraphics[scale=0.16]{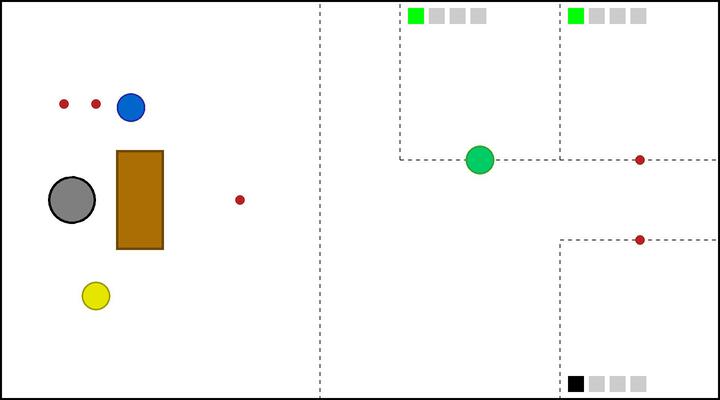}}
    ~
\end{figure*}
\begin{figure*}[h!]
    \centering
    \captionsetup[subfigure]{labelformat=empty}          
    ~
    \centering
    \centering
    \subcaptionbox{(m) Green robot runs \emph{\textbf{Get-Tool}} to go back table. \vspace{2mm}}
        [0.30\linewidth]{\includegraphics[scale=0.16]{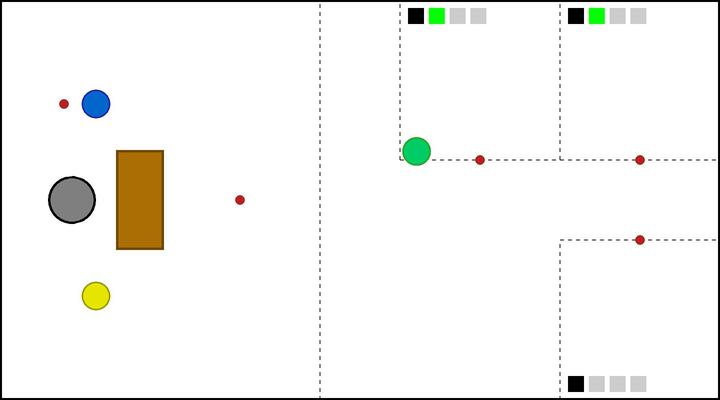}}
    ~
    \centering
    \subcaptionbox{(n) Arm robot executes \emph{\textbf{Pass-to-M(2)}} to pass a Tool-1 to yellow robot.\vspace{2mm}}
        [0.30\linewidth]{\includegraphics[scale=0.16]{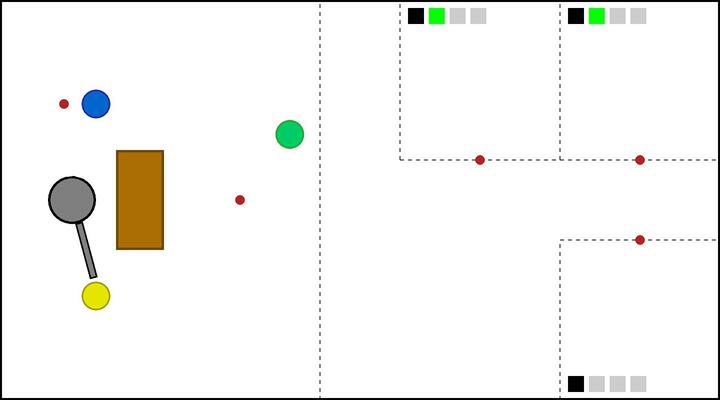}}
    ~
    \centering
    \subcaptionbox{(o) Arm robot runs \emph{\textbf{Search-Tool(1)}} to find the 2nd Tool-1. Yellow robot moves to workshop-0 by executing \emph{\textbf{Go-W(0)}}. \vspace{2mm}}
        [0.30\linewidth]{\includegraphics[scale=0.16]{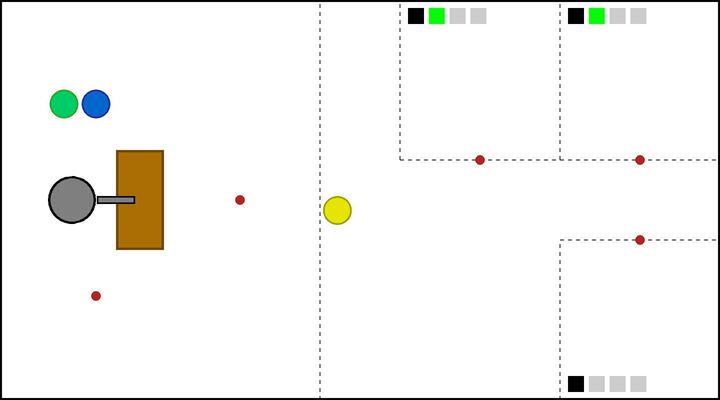}}
    ~
    \subcaptionbox{(p) Yellow robot successfully delivers the a Tool-1 to workshop-0.   \vspace{2mm}}
        [0.30\linewidth]{\includegraphics[scale=0.16]{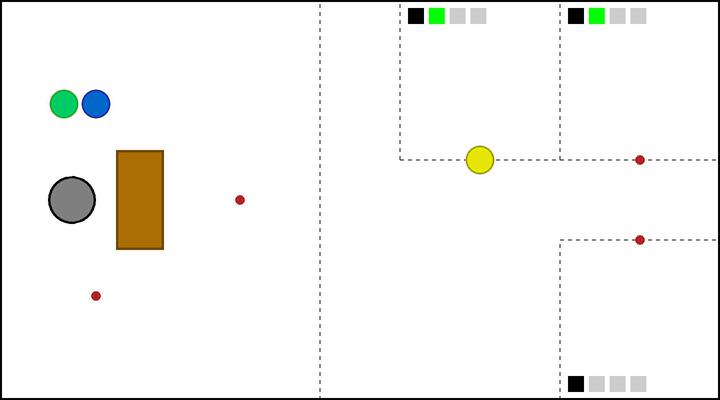}}
    ~
    \centering
    \subcaptionbox{(q) Yellow robot moves to workshop-1 by executing \emph{\textbf{Go-W(1)}}. \vspace{2mm}}
        [0.30\linewidth]{\includegraphics[scale=0.16]{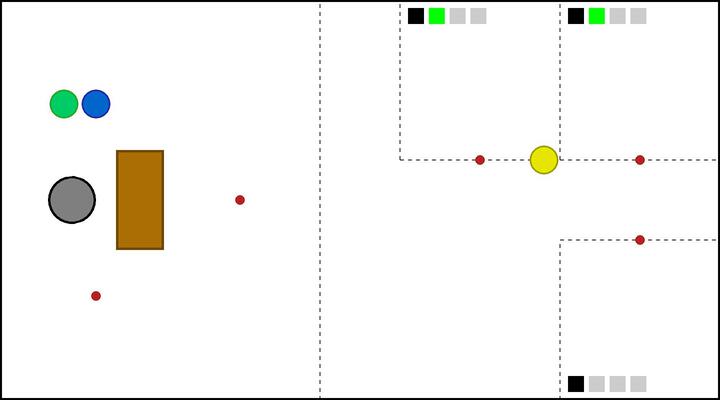}}
    ~
    \centering
    \subcaptionbox{(r) Arm robot executes \emph{\textbf{Pass-to-M(0)}} to pass a Tool-1 to green robot. \vspace{2mm}}
        [0.30\linewidth]{\includegraphics[scale=0.16]{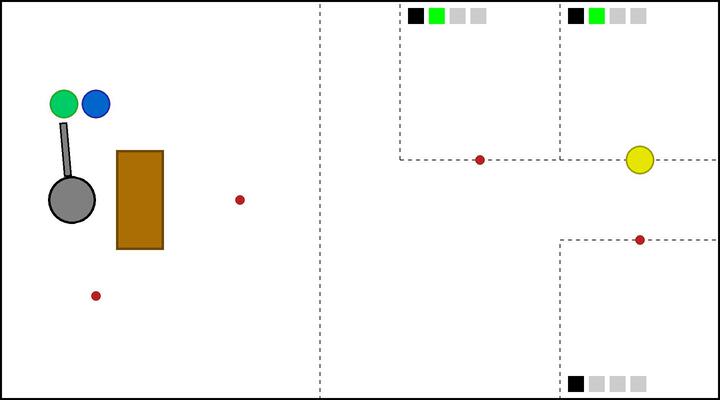}}
    ~
    \centering
    \subcaptionbox{(s) Arm robot runs \emph{\textbf{Search-Tool(1)}} to find 3st Tool-1. Yellow robot runs \emph{\textbf{Get-Tool}} to go back table. Green robot moves to workshop-0 by executing \emph{\textbf{Go-W(0)}}. \vspace{2mm}}
        [0.30\linewidth]{\includegraphics[scale=0.16]{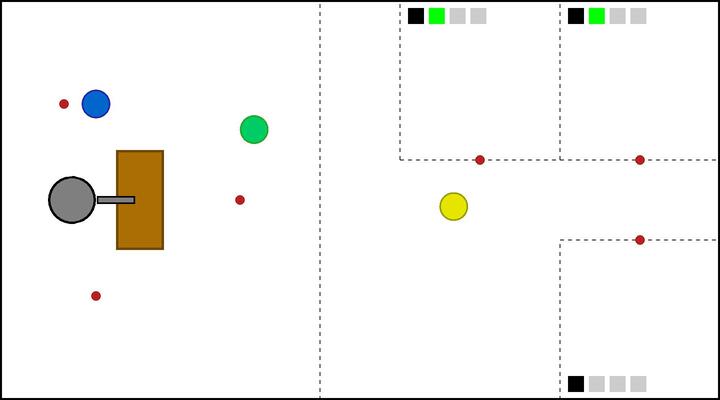}}
    ~
    \centering
    \subcaptionbox{(t) Green robot successfully delivers a Tool-1 to workshop-0.  \vspace{2mm}}
        [0.30\linewidth]{\includegraphics[scale=0.16]{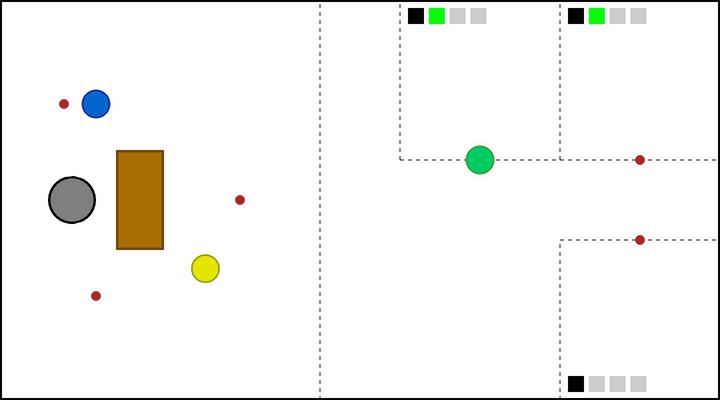}}
    ~
    \centering
    \subcaptionbox{(u) Green robot moves to workshop-2 by executing \emph{\textbf{Go-W(2)}}.\vspace{2mm}}
        [0.30\linewidth]{\includegraphics[scale=0.16]{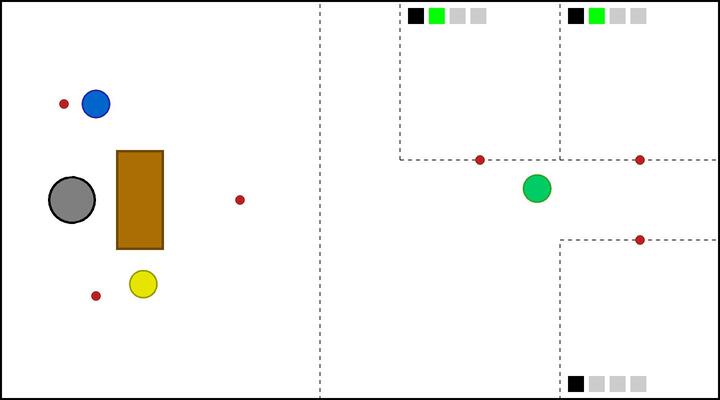}}
    ~
    \centering
    \subcaptionbox{(v) Green robot successfully delivers the a Tool-0 to workshop-2. Human-2 finishes subtask-0 and starts to do subtask-1. Arm robot executes \emph{\textbf{Pass-to-M(1)}} to pass a Tool-1 to blue robot.\vspace{2mm}}
        [0.30\linewidth]{\includegraphics[scale=0.16]{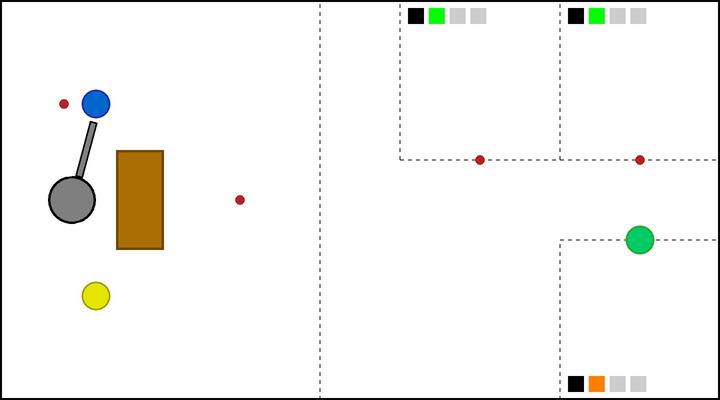}}
    ~
    \centering
    \subcaptionbox{(w) Green robot runs \emph{\textbf{Get-Tool}} to go back table.  Arm robot runs \emph{\textbf{Search-Tool(2)}} to find the 1st Tool-2. Blue robot moves to workshop-1 by executing \emph{\textbf{Go-W(1)}}. \vspace{2mm}}
        [0.30\linewidth]{\includegraphics[scale=0.16]{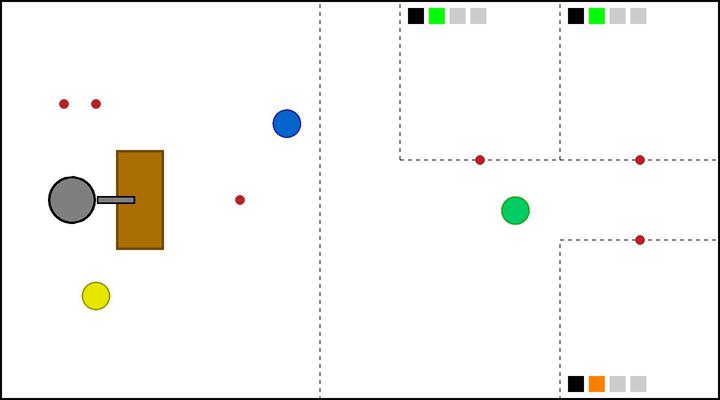}}
    ~
    \centering
    \subcaptionbox{(x) Blue robot successfully delivers a Tool-1 to workshop-1.  \vspace{2mm}}
        [0.30\linewidth]{\includegraphics[scale=0.16]{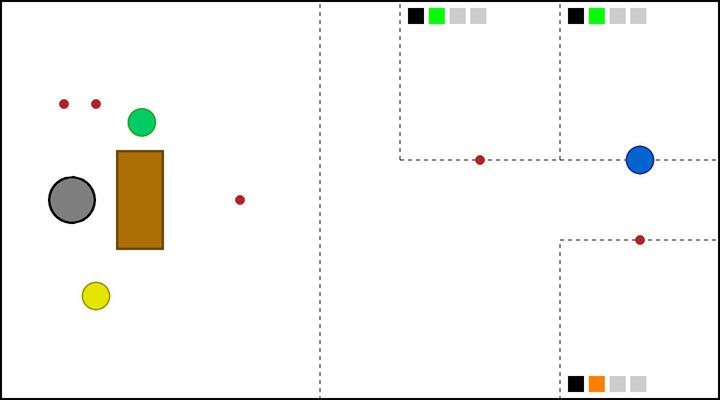}}
    ~
    \centering
    \subcaptionbox{(y) Blue robot runs \emph{\textbf{Get-Tool}} to go back table. Arm robot executes \emph{\textbf{Pass-to-M(2)}} to pass a Tool-2 to yellow robot.\vspace{2mm}}
        [0.30\linewidth]{\includegraphics[scale=0.16]{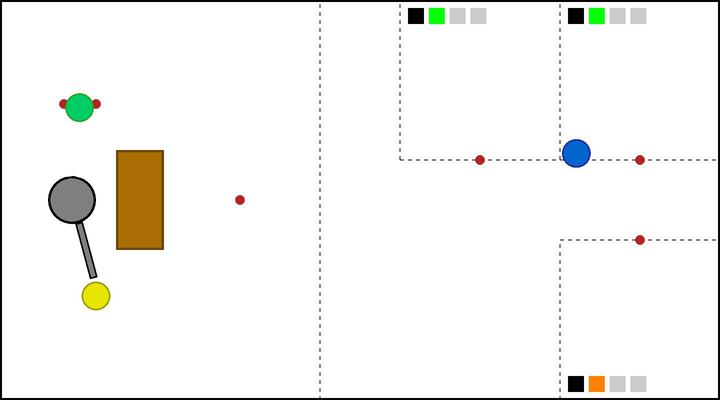}}  
    ~
    \centering
    \subcaptionbox{(z) Arm robot runs \emph{\textbf{Search-Tool(2)}} to find the 2nd Tool-2. Yellow robot moves to workshop-1 by executing \emph{\textbf{Go-W(1)}}.\vspace{2mm}}
        [0.30\linewidth]{\includegraphics[scale=0.16]{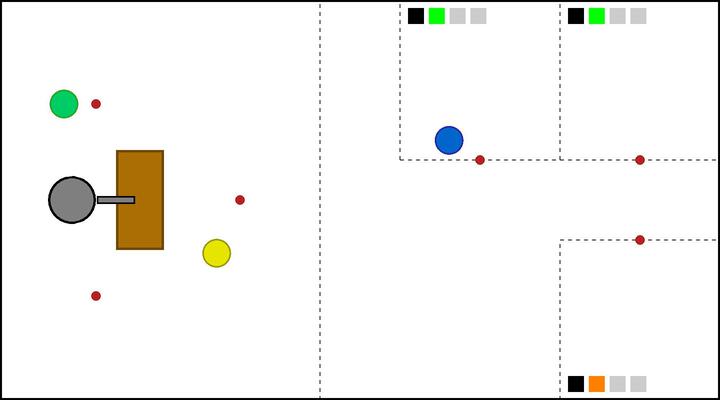}}
    ~
    \centering
    \subcaptionbox{(A) Yellow robot successfully delivers a Tool-2 to workshop-0. Human-0 and human-1 finish subtask-1 and start to do subtask-2. \vspace{2mm}}
        [0.30\linewidth]{\includegraphics[scale=0.16]{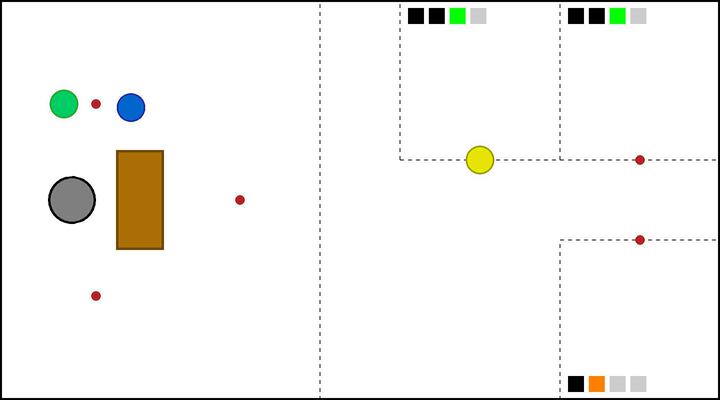}}  
\end{figure*}
\begin{figure*}[h!]
    \centering
    \captionsetup[subfigure]{labelformat=empty}          
    ~
    \centering
    \subcaptionbox{(B) Yellow robot moves to workshop-1 by executing \emph{\textbf{Go-W(1)}}. \vspace{2mm}}
        [0.30\linewidth]{\includegraphics[scale=0.16]{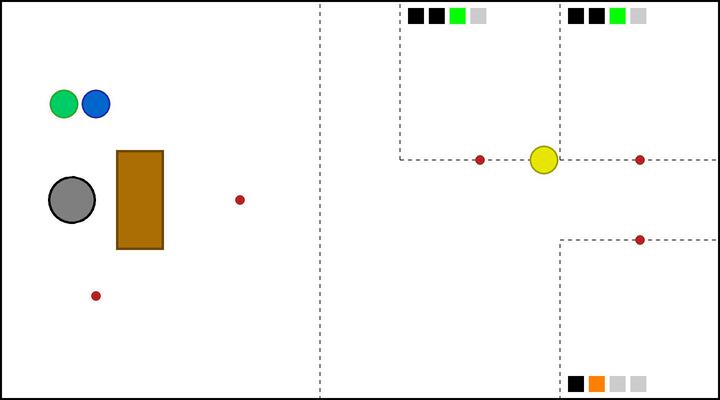}}
    ~
    \centering
    \subcaptionbox{(C) Arm robot executes \emph{\textbf{Pass-to-M(0)}} to pass a Tool-2 to green robot. Yellow robot reaches workshop-1 but it does not have any tool.\vspace{2mm}}
        [0.30\linewidth]{\includegraphics[scale=0.16]{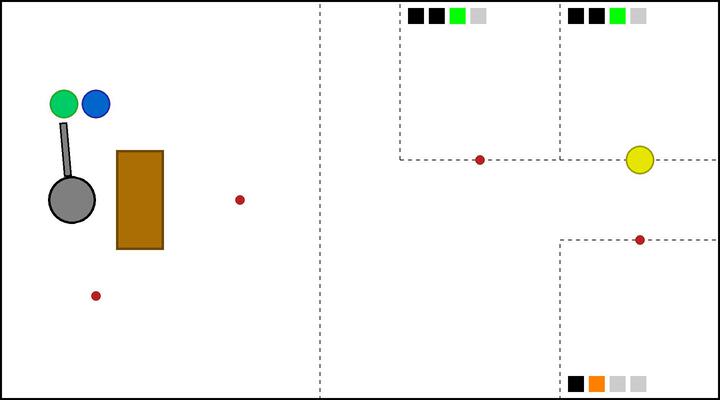}}  
    ~
    \centering
    \subcaptionbox{(D) Arm robot runs \emph{\textbf{Search-Tool(2)}} to find the 3rd Tool-2. Green robot moves to workshop-0 by executing \emph{\textbf{Go-W(0)}}. Yellow robot runs \emph{\textbf{Get-Tool}} to go back table.\vspace{2mm}}
        [0.30\linewidth]{\includegraphics[scale=0.16]{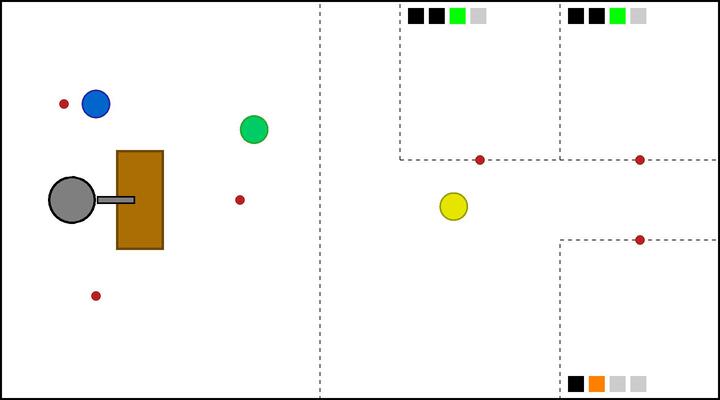}}
    ~
    \centering
    \subcaptionbox{(E) Green robot reaches workshop-0 and observes that human-0 does not need Tool-2.  Human-2 finishes subtask-1 and starts to do subtask-2.\vspace{2mm}}
        [0.30\linewidth]{\includegraphics[scale=0.16]{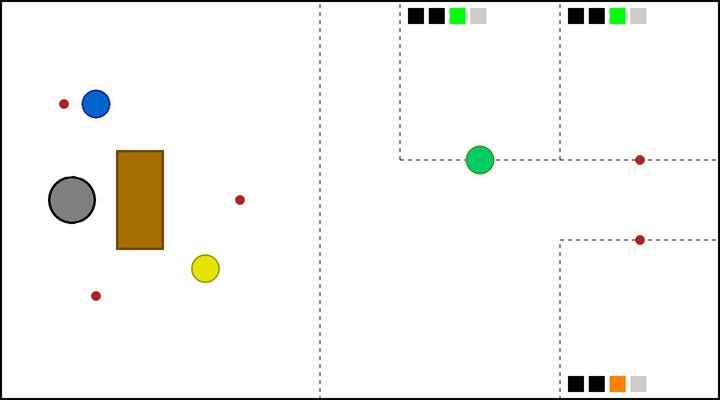}}
    ~
    \centering
    \subcaptionbox{(F) Green robot moves to workshop-2 by executing \emph{\textbf{Go-W(2)}}.\vspace{2mm}}
        [0.30\linewidth]{\includegraphics[scale=0.16]{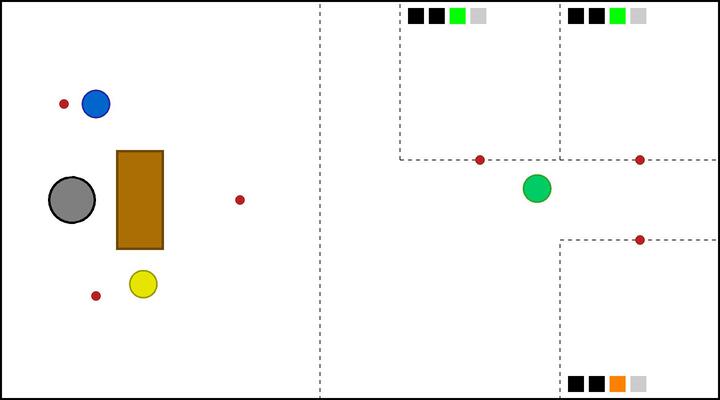}}  
    ~
    \centering
    \subcaptionbox{(G) Green robot successfully delivers a Tool-2 to workshop-2. Arm robot executes \emph{\textbf{Pass-to-M(1)}} to pass a Tool-2 to blue robot. \vspace{2mm}}
        [0.30\linewidth]{\includegraphics[scale=0.16]{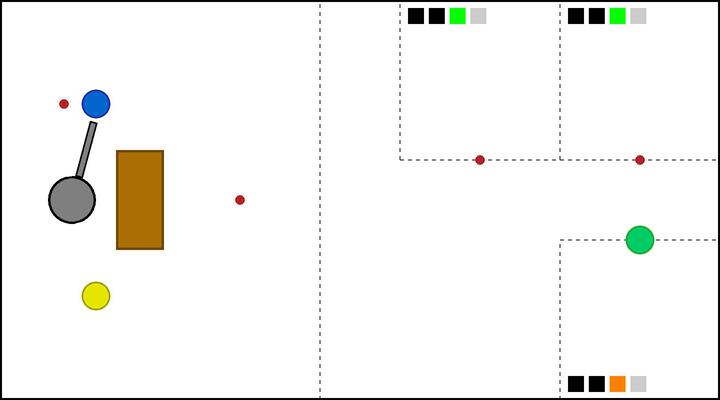}}

    ~
    \centering
    \subcaptionbox{(H) Blue robot moves to workshop-1 by executing \emph{\textbf{Go-W(1)}}.\vspace{2mm}}
        [0.30\linewidth]{\includegraphics[scale=0.16]{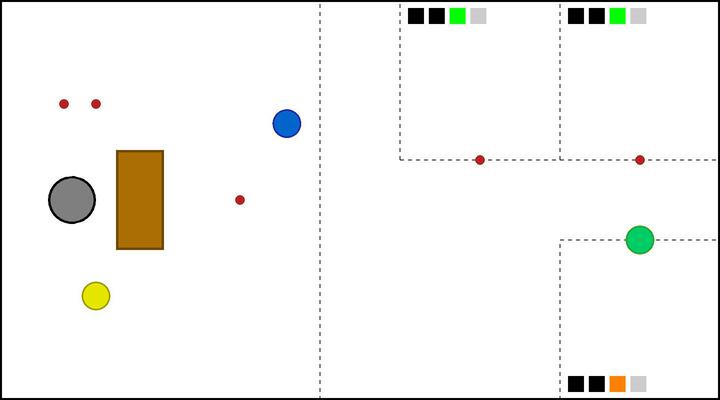}}
    ~
    \centering
    \subcaptionbox{(I) Blue robot successfully delivers a Tool-2 to workshop-1. Humans have received all tools, and for robots, the task is done.\vspace{2mm}}
        [0.30\linewidth]{\includegraphics[scale=0.16]{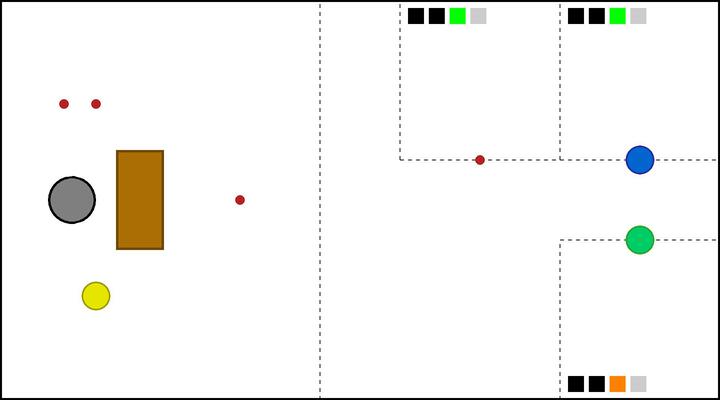}}
    \label{wtd_e_behavior}
\end{figure*}

\clearpage
\textbf{\emph{Warehouse-D:}}  

\begin{figure*}[h!]
    \centering
    \captionsetup[subfigure]{labelformat=empty}
    \centering
    \subcaptionbox{(a) Initial State.\vspace{2mm}}
        [0.30\linewidth]{\includegraphics[scale=0.24]{results/WTD/wtd_d_small.png}}
    ~
    \centering
    \subcaptionbox{(b) Green and yellow robots move towards the table by running \emph{\textbf{Get-Tool}}. Blue robot moves to workshop-3 by executing \emph{\textbf{Go-W(3)}}. Arm robot runs \emph{\textbf{Search-Tool(0)}} to find the 1st Tool-0.\vspace{2mm}}
        [0.30\linewidth]{\includegraphics[scale=0.16]{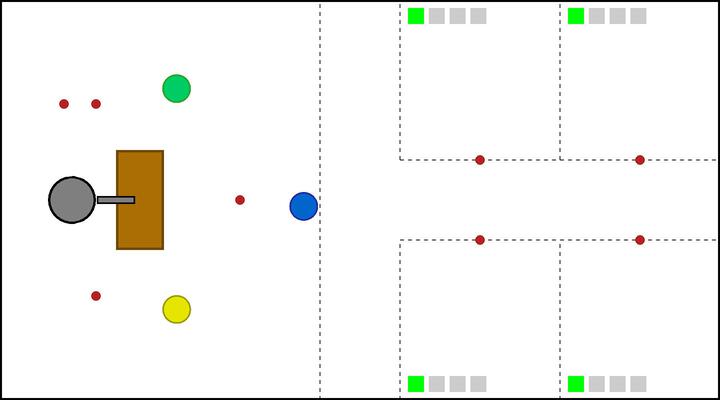}}
    ~
    \centering
    \subcaptionbox{(c) Blue robot reaches workshop-3.\vspace{2mm}}
        [0.30\linewidth]{\includegraphics[scale=0.16]{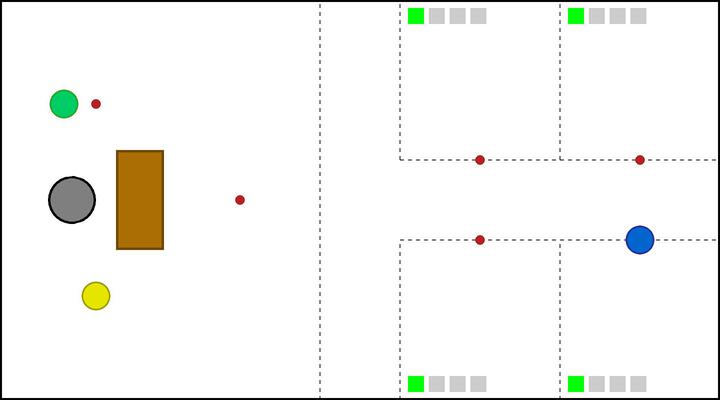}}
    ~
    \centering
    \subcaptionbox{(d) Blue robot moves to workshop-1 by executing \emph{\textbf{Go-W(1)}}.\vspace{2mm}}
        [0.30\linewidth]{\includegraphics[scale=0.16]{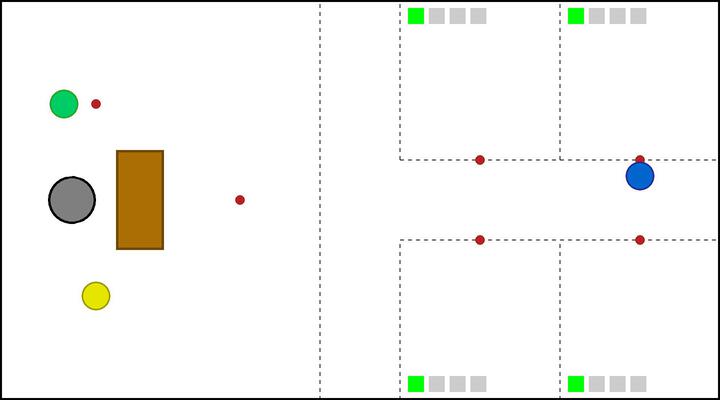}}
    ~
    \centering
    \subcaptionbox{(e) Blue robot reaches workshop-1. Arm robot executes \emph{\textbf{Pass-to-M(2)}} to pass a Tool-0 to yellow robot. \vspace{2mm}}
        [0.30\linewidth]{\includegraphics[scale=0.16]{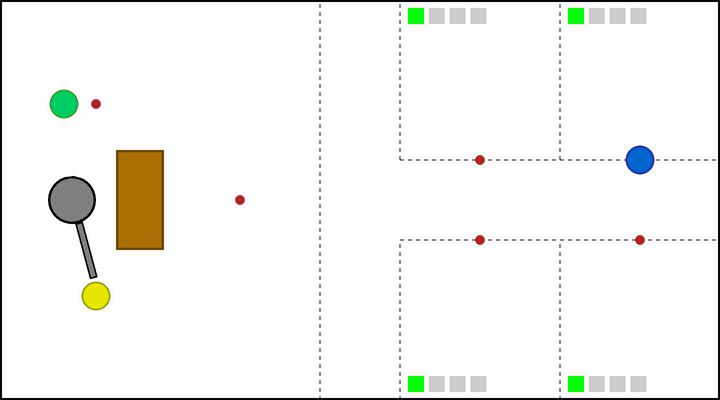}}  
    ~
    \centering
    \subcaptionbox{(f) Arm robot runs \emph{\textbf{Search-Tool(0)}} to find the 2nd Tool-0. Blue robot runs \emph{\textbf{Get-Tool}} to go back table. Yellow robot moves to workshop-1 by executing \emph{\textbf{Go-W(1)}}. \vspace{2mm}}
        [0.30\linewidth]{\includegraphics[scale=0.16]{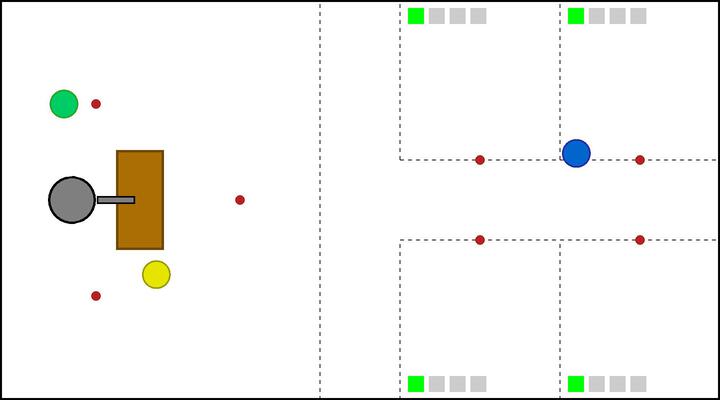}}
    ~
    \centering
    \subcaptionbox{(g)  Yellow robot successfully delivers a Tool-0 to workshop-0. \vspace{2mm}}
        [0.30\linewidth]{\includegraphics[scale=0.16]{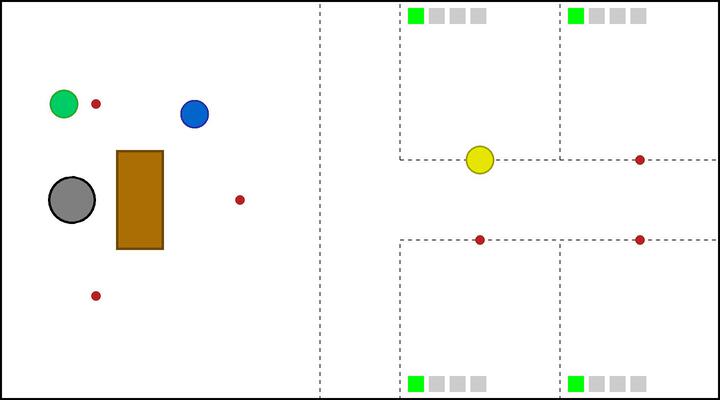}}
    ~
    \centering
    \subcaptionbox{(h) Arm robot executes \emph{\textbf{Pass-to-M(0)}} to pass a Tool-0 to green robot. Yellow robot runs \emph{\textbf{Get-Tool}} to go back table.  \vspace{2mm}}
        [0.30\linewidth]{\includegraphics[scale=0.16]{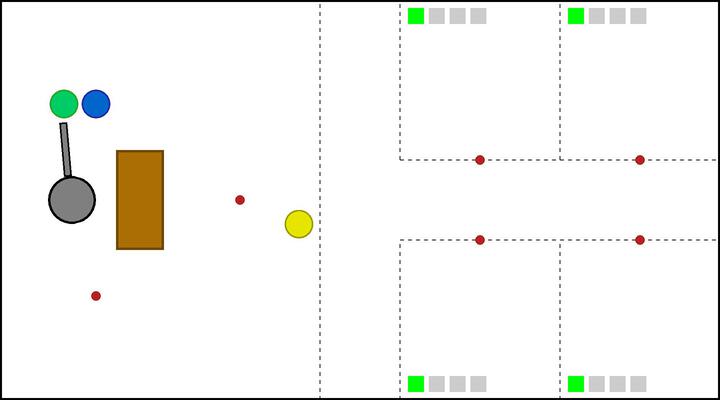}}
    ~
    \centering
    \subcaptionbox{(i) Arm robot runs \emph{\textbf{Search-Tool(0)}} to find the 3rd Tool-0. Green robot moves to workshop-1 by executing \emph{\textbf{Go-W(1)}}.
    \vspace{2mm}}
        [0.30\linewidth]{\includegraphics[scale=0.16]{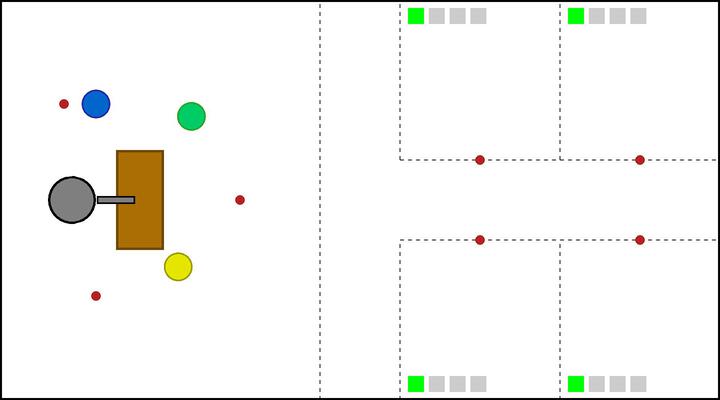}}
    ~
    \centering
    \subcaptionbox{(j) Green robot successfully delivers the a Tool-0 to workshop-1. Arm robot executes \emph{\textbf{Pass-to-M(2)}} to pass a Tool-0 to yellow robot.\vspace{2mm}}
        [0.30\linewidth]{\includegraphics[scale=0.16]{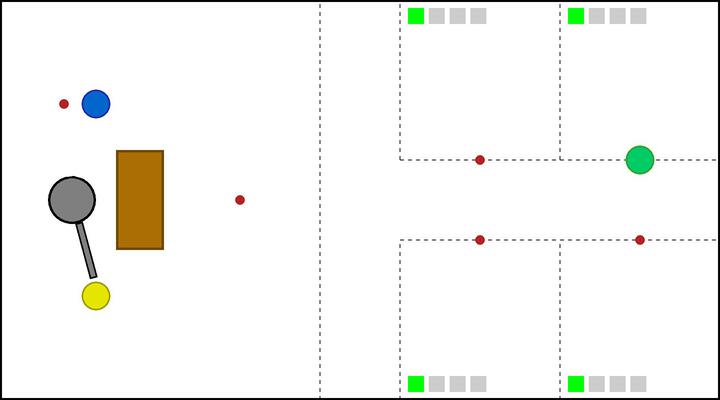}}
    ~
    \centering
    \subcaptionbox{(k) Arm robot runs \emph{\textbf{Search-Tool(1)}} to find the 1st Tool-1. Yellow robot moves to workshop-2 by executing \emph{\textbf{Go-W(2)}}. Green robot runs \emph{\textbf{Get-Tool}} to go back table. \vspace{2mm}}
        [0.30\linewidth]{\includegraphics[scale=0.16]{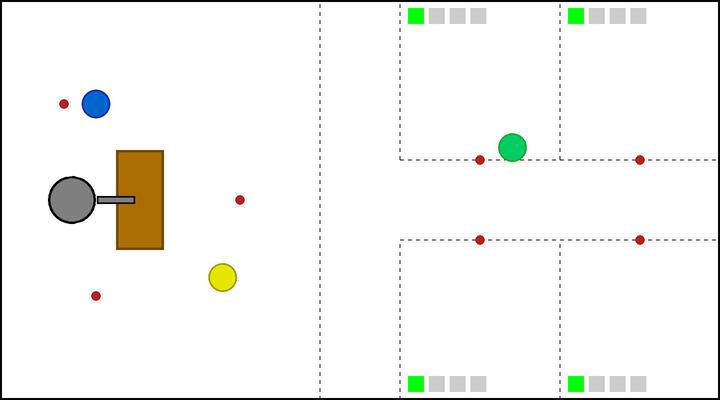}}  
    ~
    \centering
    \subcaptionbox{(l) Yellow robot successfully delivers a Tool-0 to workshop-2.  \vspace{2mm}}
        [0.30\linewidth]{\includegraphics[scale=0.16]{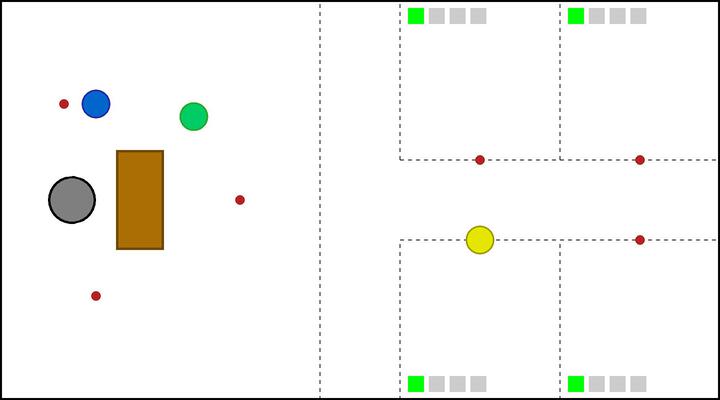}}
\end{figure*}
\begin{figure*}[h!]
    \centering
    \captionsetup[subfigure]{labelformat=empty}          
    ~
    \centering
    \subcaptionbox{(m) Yellow robot runs \emph{\textbf{Get-Tool}} to go back table. Arm robot executes \emph{\textbf{Pass-to-M(1)}} to pass the a Tool-1 to blue robot. Human-0, human-1 and human-2 finish subtask-0 and start to do subtask-1.\vspace{2mm}}
        [0.30\linewidth]{\includegraphics[scale=0.16]{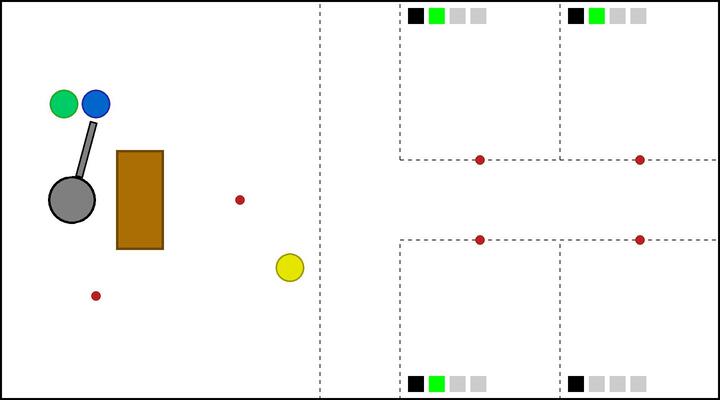}}
    ~
    \centering
    \subcaptionbox{(n) Arm robot runs \emph{\textbf{Search-Tool(1)}} to find the 2nd Tool-1. Blue robot moves to workshop-1 by executing \emph{\textbf{Go-W(1)}}.\vspace{2mm}}
        [0.30\linewidth]{\includegraphics[scale=0.16]{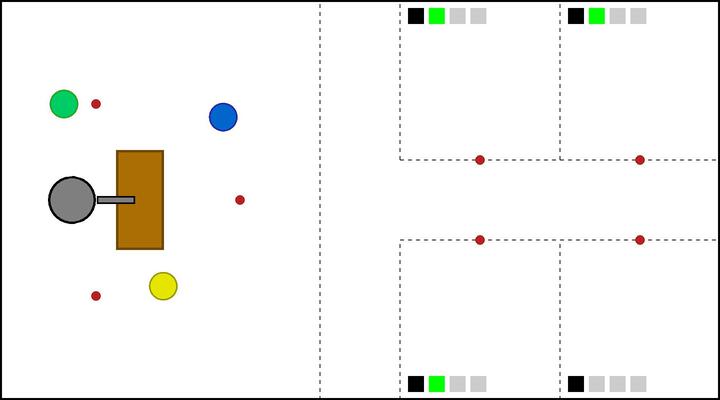}}
    ~
    \centering
    \subcaptionbox{(o) Blue robot successfully delivers a Tool-1 to workshop-1.  \vspace{2mm}}
        [0.30\linewidth]{\includegraphics[scale=0.16]{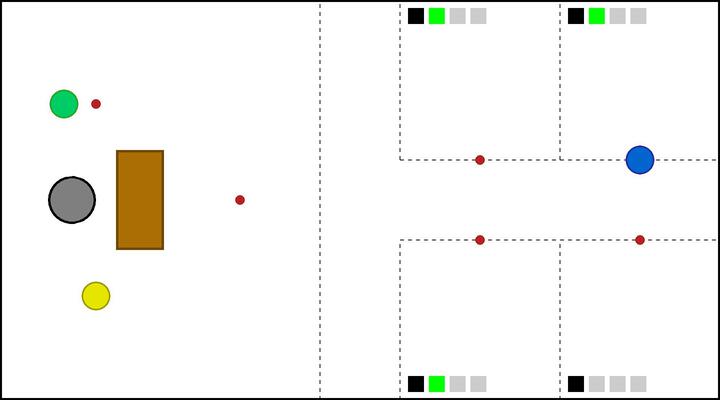}}
    ~
    \subcaptionbox{(p)Arm robot executes \emph{\textbf{Pass-to-M(2)}} to pass a Tool-1 to yellow robot. Blue robot runs \emph{\textbf{Get-Tool}} to go back table. \vspace{2mm}}
        [0.30\linewidth]{\includegraphics[scale=0.16]{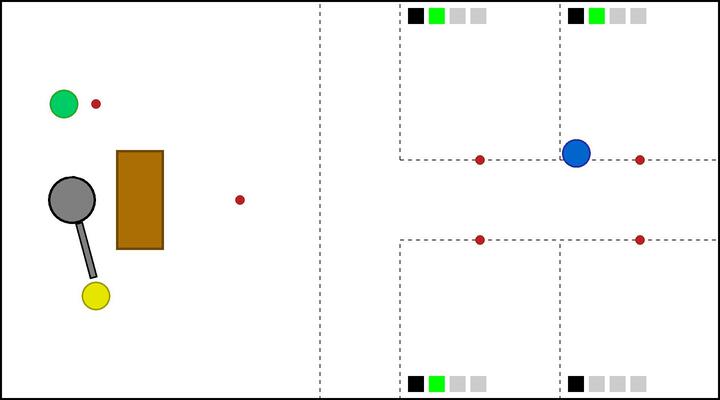}}
    ~
    \centering
    \subcaptionbox{(q) Arm robot runs \emph{\textbf{Search-Tool(0)}} to find 4th Tool-0. Yellow robot moves to workshop-0 by executing \emph{\textbf{Go-W(0)}}.\vspace{2mm}}
        [0.30\linewidth]{\includegraphics[scale=0.16]{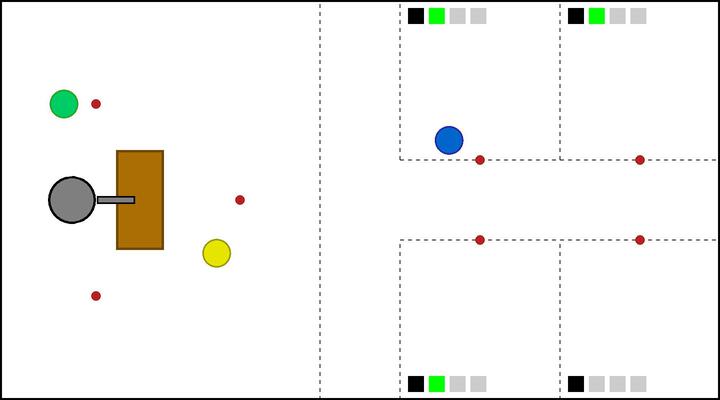}}
    ~
    \centering
    \subcaptionbox{(r) Yellow robot successfully delivers a Tool-1 to workshop-0. \vspace{2mm}}
        [0.30\linewidth]{\includegraphics[scale=0.16]{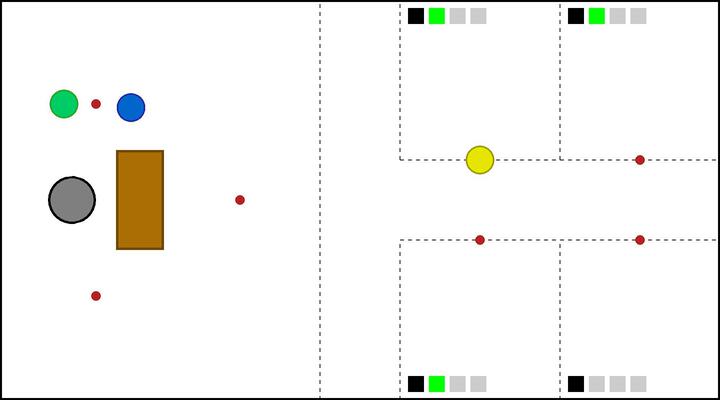}}
    ~
    \centering
    \subcaptionbox{(s) Yellow robot runs \emph{\textbf{Get-Tool}} to go back table. Arm robot executes \emph{\textbf{Pass-to-M(0)}} to pass a Tool-0 to green robot. \vspace{2mm}}
        [0.30\linewidth]{\includegraphics[scale=0.16]{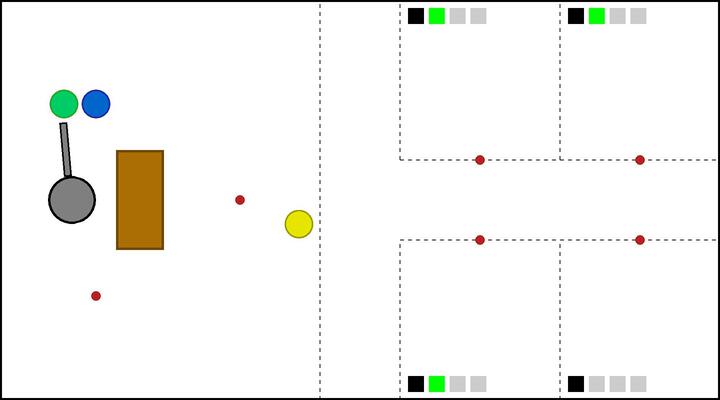}}
    ~
    \centering
    \subcaptionbox{(t) Arm robot runs \emph{\textbf{Search-Tool(1)}} to find the 3rd Tool-1. Green robot moves to workshop-1 by executing \emph{\textbf{Go-W(1)}}. \vspace{2mm}}
        [0.30\linewidth]{\includegraphics[scale=0.16]{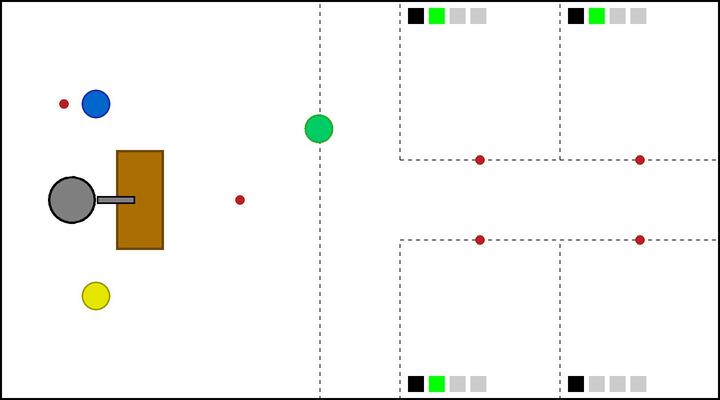}}
    ~
    \centering
    \subcaptionbox{(u) Green robot reaches workshop-1 and observes that human-1 does not need Tool-0  and it moves to workshop-3 by executing \emph{\textbf{Go-W(3)}}. Arm robot executes \emph{\textbf{Pass-to-M(2)}} to pass a Tool-1 to yellow robot. \vspace{2mm}}
        [0.30\linewidth]{\includegraphics[scale=0.16]{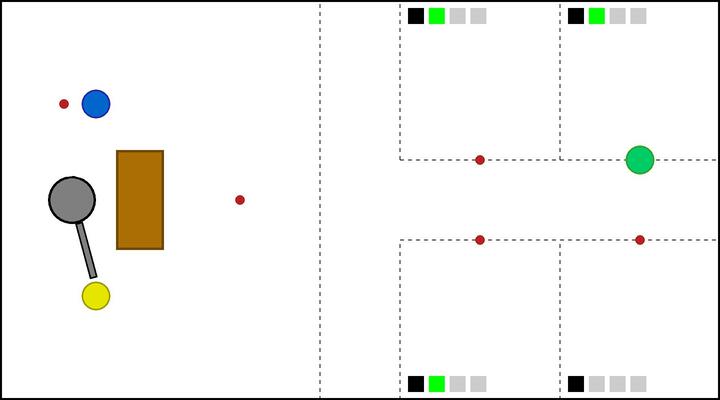}}
    ~
    \centering
    \subcaptionbox{(v) Arm robot runs \emph{\textbf{Search-Tool(1)}} to find the 4th Tool-1. Green robot successfully delivers a Tool-0 to workshop-3. Human-3 finishes subtask-0 and starts to do subtask-1.  Yellow robot moves to workshop-2 by executing \emph{\textbf{Go-W(2)}}.\vspace{2mm}}
        [0.30\linewidth]{\includegraphics[scale=0.16]{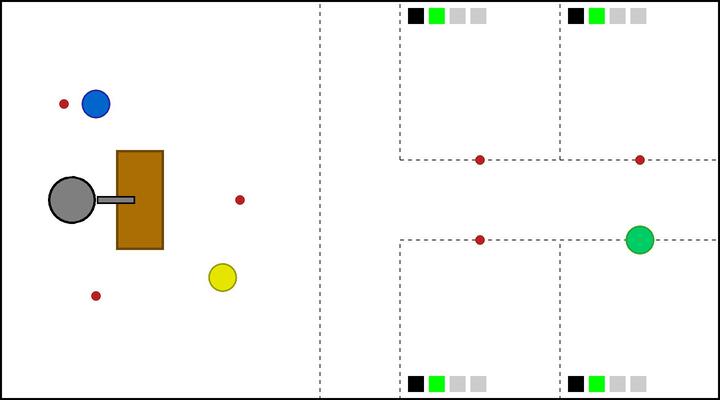}}
    ~
    \centering
    \subcaptionbox{(w) Yellow robot successfully delivers a Tool-1 to workshop-2. Green robot runs \emph{\textbf{Get-Tool}} to go back table. \vspace{2mm}}
        [0.30\linewidth]{\includegraphics[scale=0.16]{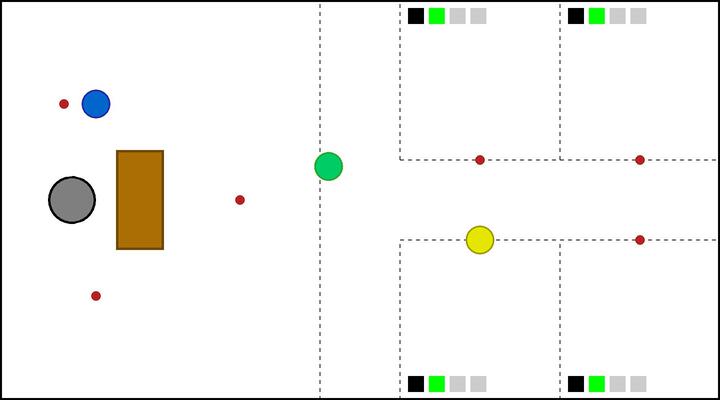}}
    ~
    \centering
    \subcaptionbox{(x) Arm robot executes \emph{\textbf{Pass-to-M(1)}} to pass a Tool-1 to blue robot. Yellow robot runs \emph{\textbf{Get-Tool}} to go back table.\vspace{2mm}}
        [0.30\linewidth]{\includegraphics[scale=0.16]{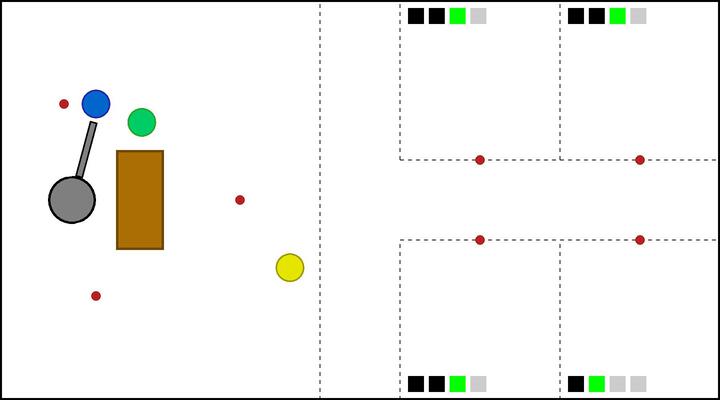}}
\end{figure*}
\begin{figure*}[h!]
    \centering
    \captionsetup[subfigure]{labelformat=empty}          
    ~
    \centering
    \subcaptionbox{(y) Arm robot runs \emph{\textbf{Search-Tool(2)}} to find the 1st Tool-2. Blue robot moves to workshop-1 by executing \emph{\textbf{Go-W(1)}}. \vspace{2mm}}
        [0.30\linewidth]{\includegraphics[scale=0.16]{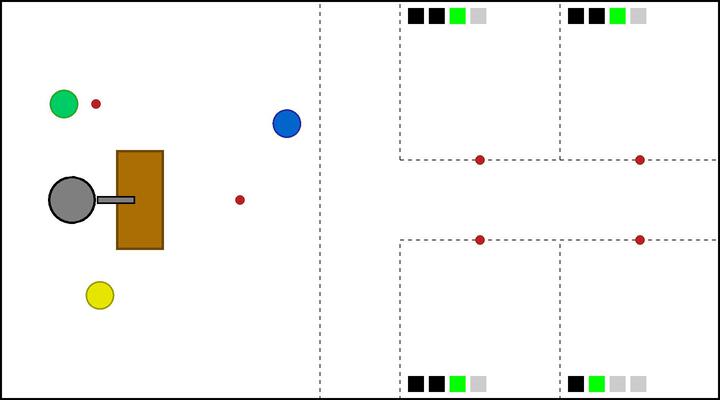}}  
    ~
    \centering
    \subcaptionbox{(z) Blue robot reaches workshop-1 and observes that human-1 does not need Tool-1.\vspace{2mm}}
        [0.30\linewidth]{\includegraphics[scale=0.16]{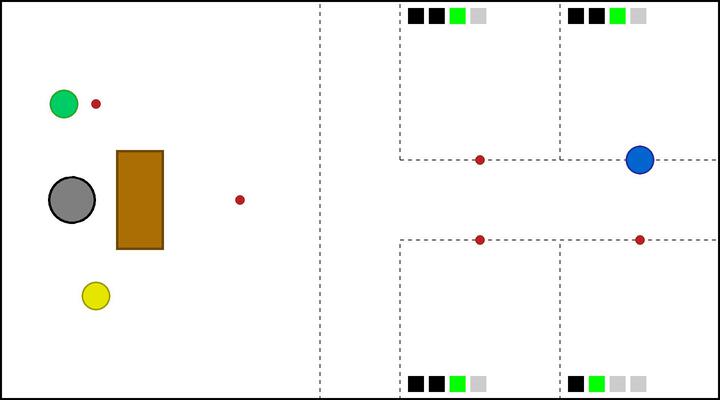}}
    ~
    \centering
    \subcaptionbox{(A) Arm robot executes \emph{\textbf{Pass-to-M(2)}} to pass a Tool-2 to yellow robot. Blue robot moves to workshop-3 by executing \emph{\textbf{Go-W(3)}}.\vspace{2mm}}
        [0.30\linewidth]{\includegraphics[scale=0.16]{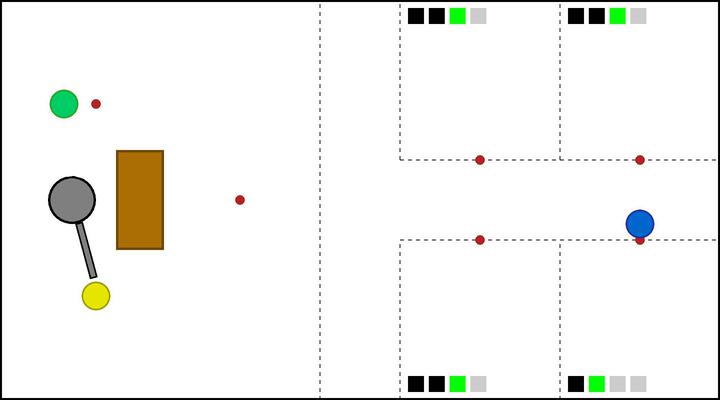}}  
    ~
    \centering
    \subcaptionbox{(B) Arm robot runs \emph{\textbf{Search-Tool(2)}} to find the 2nd Tool-2. Blue robot successfully delivers a Tool-1 to workshop-3. \vspace{2mm}}
        [0.30\linewidth]{\includegraphics[scale=0.16]{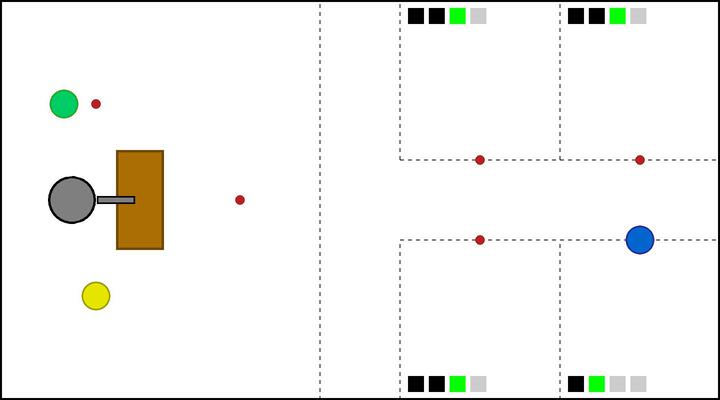}}
    ~
    \centering
    \subcaptionbox{(C) Blue robot moves to workshop-1 by executing \emph{\textbf{Go-W(1)}}.\vspace{2mm}}
        [0.30\linewidth]{\includegraphics[scale=0.16]{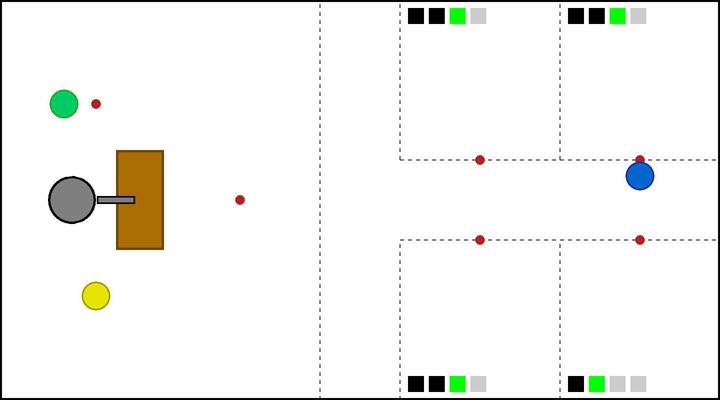}}  
    ~
    \centering
    \subcaptionbox{(D) Blue robot reaches workshop-1. \vspace{2mm}}
        [0.30\linewidth]{\includegraphics[scale=0.16]{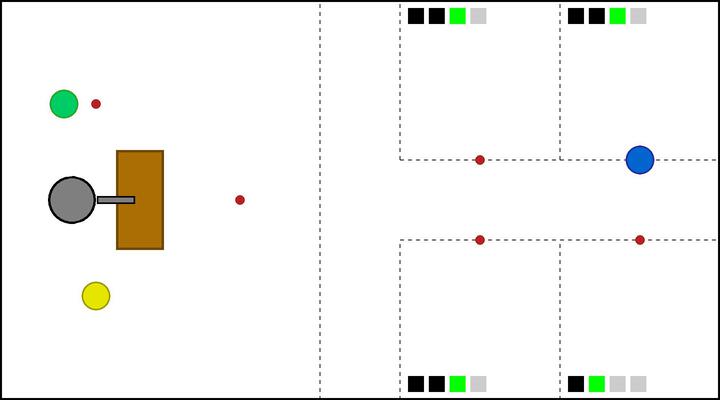}}
    ~
    \centering
    \subcaptionbox{(E) Blue robot runs \emph{\textbf{Get-Tool}} to go back table.\vspace{2mm}}
        [0.30\linewidth]{\includegraphics[scale=0.16]{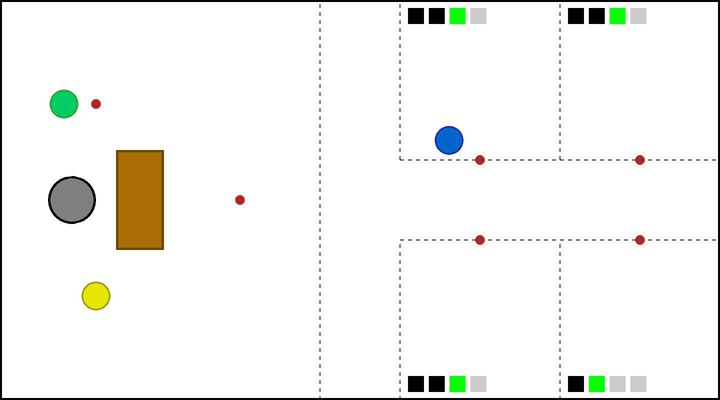}}
    ~
    \centering
    \subcaptionbox{(F) Arm robot executes \emph{\textbf{Pass-to-M(2)}} to pass the a Tool-2 to yellow robot.\vspace{2mm}}
        [0.30\linewidth]{\includegraphics[scale=0.16]{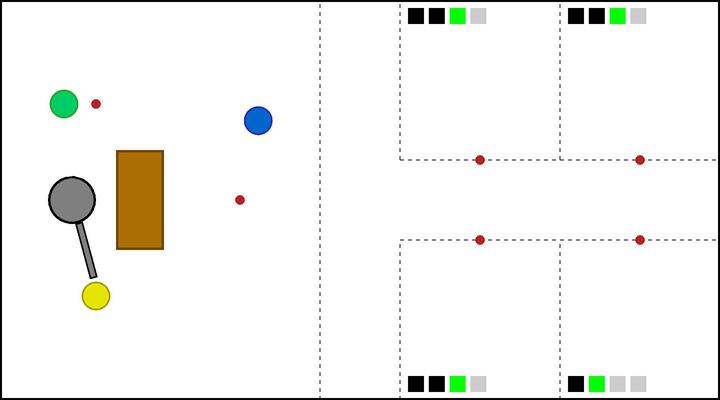}}  
    ~
    \centering
    \subcaptionbox{(G) Arm robot runs \emph{\textbf{Search-Tool(2)}} to find the 3rd Tool-2. Yellow robot moves to workshop-1 by executing \emph{\textbf{Go-W(1)}}. \vspace{2mm}}
        [0.30\linewidth]{\includegraphics[scale=0.16]{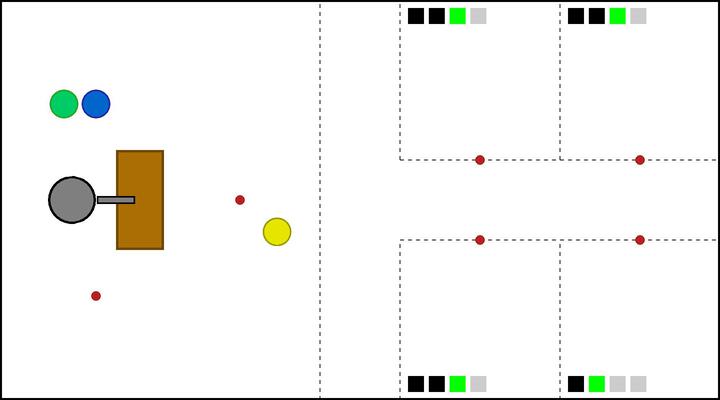}}
    ~
    \centering
    \subcaptionbox{(H) Yellow robot reaches workshop-0 and observes that human-0 has got a Tool-2.\vspace{2mm}}
        [0.30\linewidth]{\includegraphics[scale=0.16]{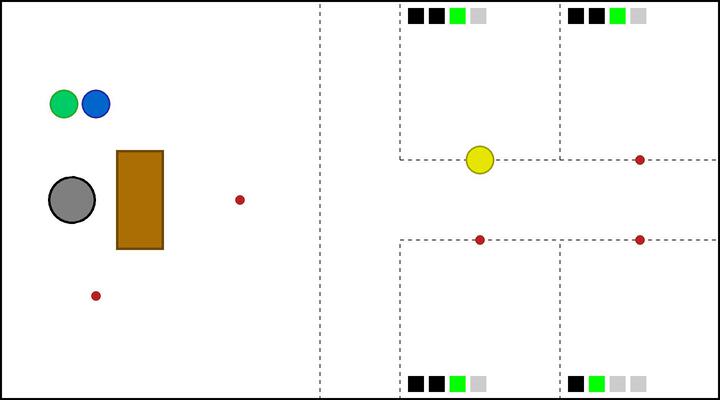}}
    ~
    \centering
    \subcaptionbox{(I) Yellow robot moves to workshop-2 by executing \emph{\textbf{Go-W(2)}}.\vspace{2mm}}
        [0.30\linewidth]{\includegraphics[scale=0.16]{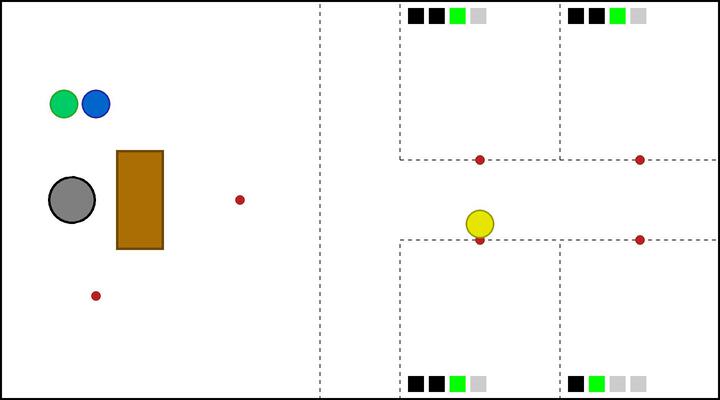}}
    ~
    \centering
    \subcaptionbox{(J) Yellow robot successfully delivers a Tool-2 to workshop-2. \vspace{2mm}}
        [0.30\linewidth]{\includegraphics[scale=0.16]{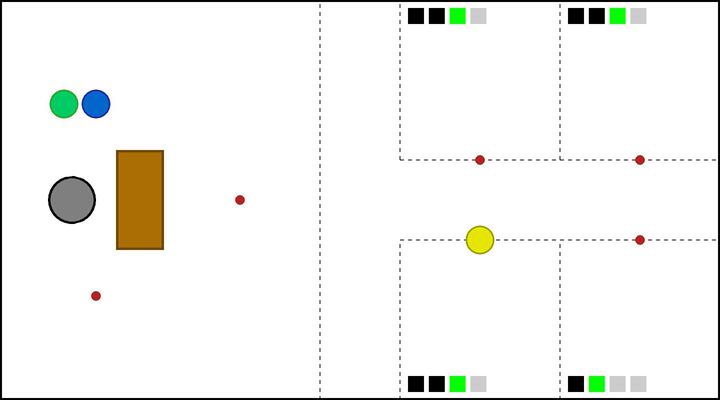}}
    ~
    \centering
    \subcaptionbox{(K) Arm robot executes \emph{\textbf{Pass-to-M(0)}} to pass a Tool-2 to green robot. Yellow robot runs \emph{\textbf{Get-Tool}} to go back table.\vspace{2mm}}
        [0.30\linewidth]{\includegraphics[scale=0.16]{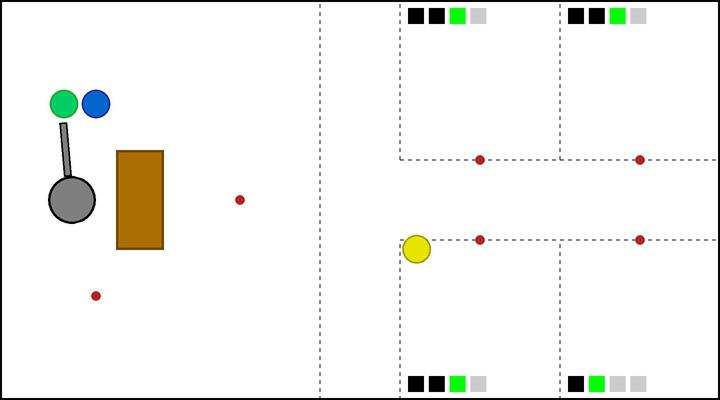}}
    ~
    \centering
    \subcaptionbox{(L) Arm robot runs \emph{\textbf{Search-Tool(2)}} to find the 4th Tool-2. Green robot moves to workshop-1 by executing \emph{\textbf{Go-W(1)}}.\vspace{2mm}}
        [0.30\linewidth]{\includegraphics[scale=0.16]{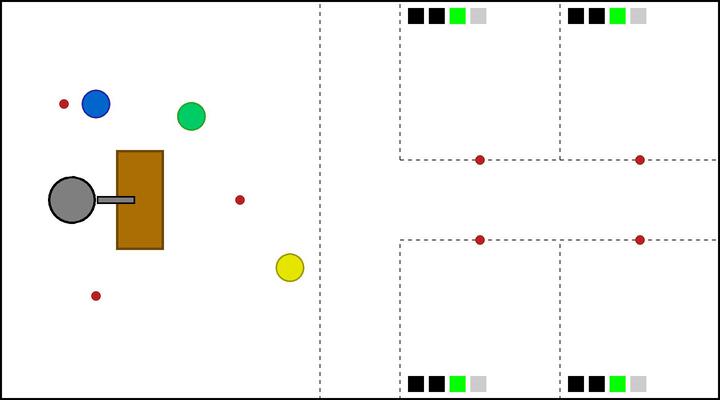}}
    ~
    \centering
    \subcaptionbox{(M) Arm robot executes \emph{\textbf{Pass-to-M(2)}} to pass a Tool-2 to yellow robot. Green robot successfully delivers a Tool-2 to workshop-1. Human-0, human-1 and human-2 finish subtask-2 and starts to do subtask-3.\vspace{2mm}}
        [0.30\linewidth]{\includegraphics[scale=0.16]{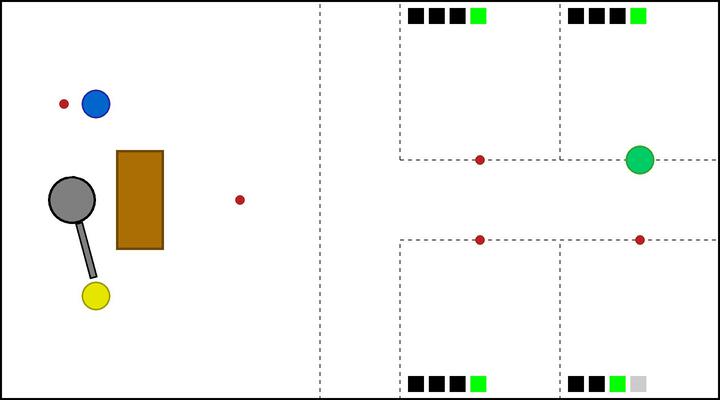}}
\end{figure*}
\begin{figure*}[h!]
    \centering
    \captionsetup[subfigure]{labelformat=empty}          
    ~
    \centering
    \subcaptionbox{(N) Yellow robot moves to workshop-0 by executing \emph{\textbf{Go-W(0)}}. Green robot moves to workshop-3 by executing \emph{\textbf{Go-W(3)}}. \vspace{2mm}}
        [0.30\linewidth]{\includegraphics[scale=0.16]{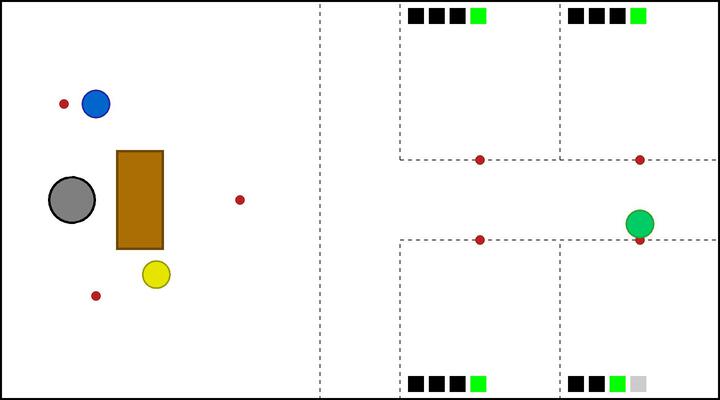}}
    ~
    \centering
    \subcaptionbox{(O) Yellow robot reaches workshop-0 and observes that human-0 does not need Tool-2.\vspace{2mm}}
        [0.30\linewidth]{\includegraphics[scale=0.16]{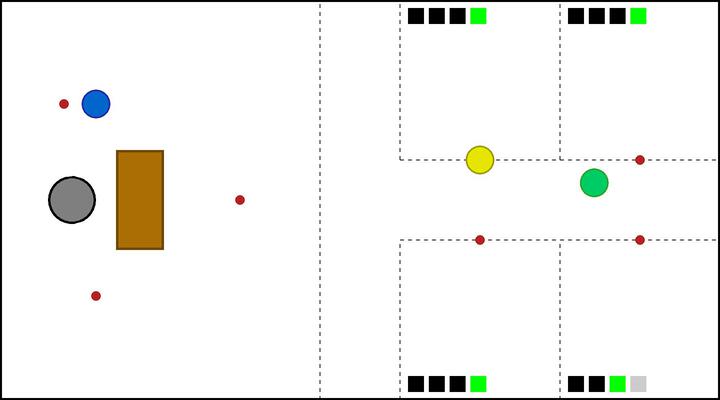}}
    ~
    \centering
    \subcaptionbox{(P) Yellow and green robot move to workshop-3 by executing \emph{\textbf{Go-W(3)}}. \vspace{2mm}}
        [0.30\linewidth]{\includegraphics[scale=0.16]{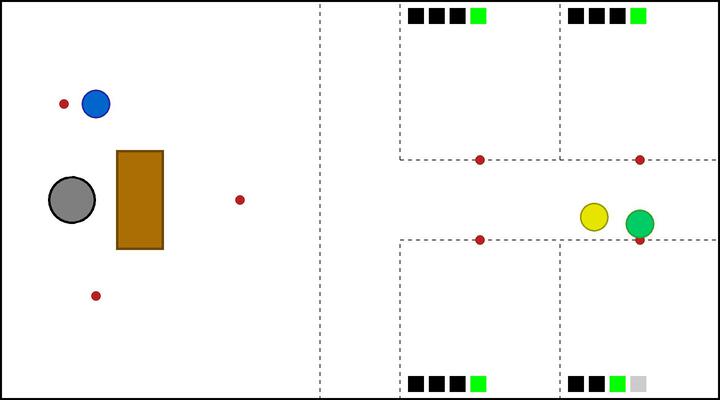}}
    ~
    \centering
    \subcaptionbox{(Q) Yellow robot successfully delivers a Tool-2 to workshop-3.  Humans have received all tools, and for robots, the task is done.\vspace{2mm}}
        [0.30\linewidth]{\includegraphics[scale=0.16]{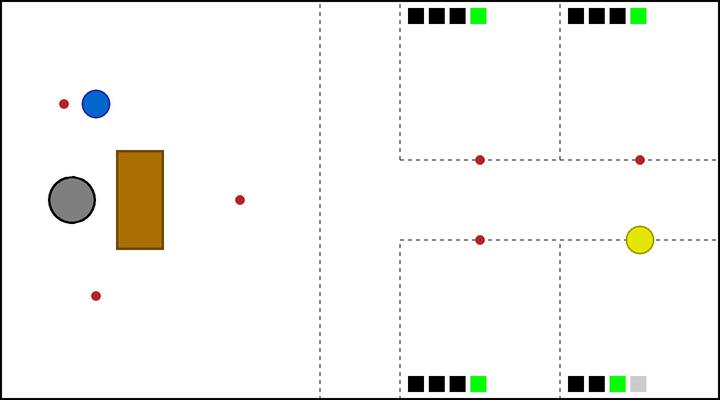}}
    \label{wtd_e_behavior}
\end{figure*}

\clearpage
\textbf{\emph{Warehouse-E:}}  

\begin{figure*}[h!]
    \centering
    \captionsetup[subfigure]{labelformat=empty}
    \centering
    \subcaptionbox{(a) Initial State.\vspace{2mm}}
        [0.30\linewidth]{\includegraphics[scale=0.26]{results/WTD/wtd_b_small.png}}
    ~
    \centering
    \subcaptionbox{(b) Mobile robots moves towards the table by running \emph{\textbf{Get-Tool}}, and arm robot runs \emph{\textbf{Search-Tool(0)}} to find Tool-0.\vspace{2mm}}
        [0.30\linewidth]{\includegraphics[scale=0.18]{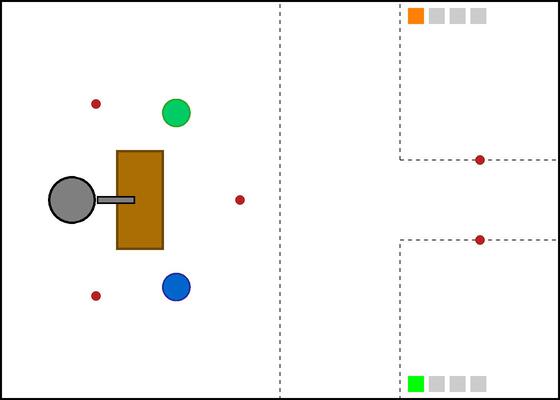}}
    ~
    \centering
    \subcaptionbox{(c) Mobile robots wait there and arm robot keeps looking for Tool-0.\vspace{2mm}}
        [0.30\linewidth]{\includegraphics[scale=0.18]{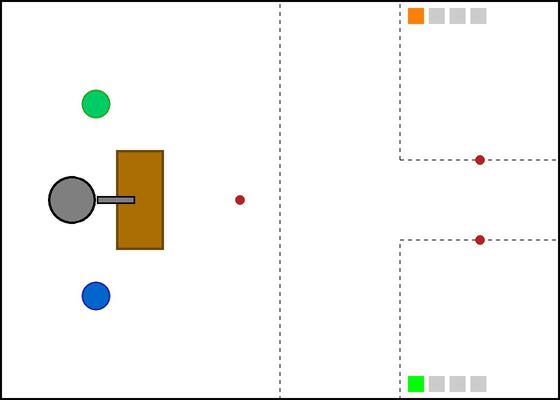}}
    ~
    \centering
    \subcaptionbox{(d) Arm robot executes \emph{\textbf{Pass-to-M(1)}} to pass Tool-0 to the blue robot.\vspace{2mm}}
        [0.30\linewidth]{\includegraphics[scale=0.18]{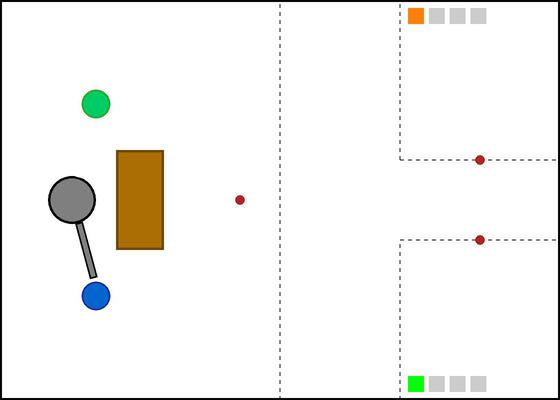}}
    ~
    \centering
    \subcaptionbox{(e) Arm robot runs \emph{\textbf{Search-Tool(1)}} to find Tool-1. Blue robot executes \emph{\textbf{Go-W(0)}} to go to workshop-0. \vspace{2mm}}
        [0.30\linewidth]{\includegraphics[scale=0.18]{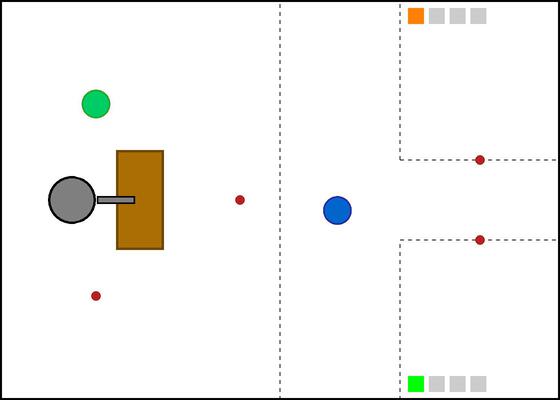}}  
    ~
    \centering
    \subcaptionbox{(f) Blue robot successfully delivers Tool-0 to workshop-0. \vspace{2mm}}
        [0.30\linewidth]{\includegraphics[scale=0.18]{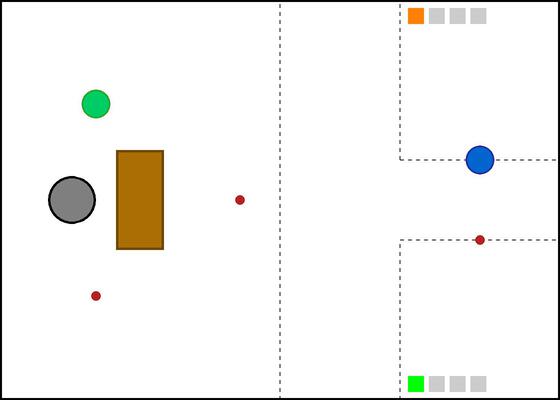}}
    ~
    \centering
    \subcaptionbox{(g)  Blue robot runs \emph{\textbf{Get-Tool}} to go back table. Human-0 finishes subtask-0 and starts to do subtask-1.\vspace{2mm}}
        [0.30\linewidth]{\includegraphics[scale=0.18]{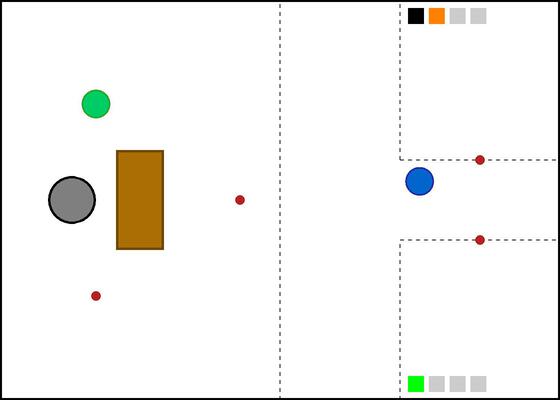}}
    ~
    \centering
    \subcaptionbox{(h) Arm robot executes \emph{\textbf{Pass-to-M(0)}} to pass Tool-1 to green robot.  \vspace{2mm}}
        [0.30\linewidth]{\includegraphics[scale=0.18]{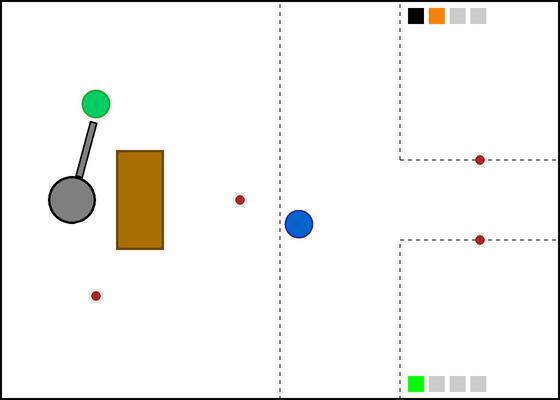}}
    ~
    \centering
    \subcaptionbox{(i) Arm robot runs \emph{\textbf{Search-Tool(0)}} to find Tool-0. Green robot moves to workshop-0 by executing \emph{\textbf{Go-W(0)}}.
    \vspace{2mm}}
        [0.30\linewidth]{\includegraphics[scale=0.18]{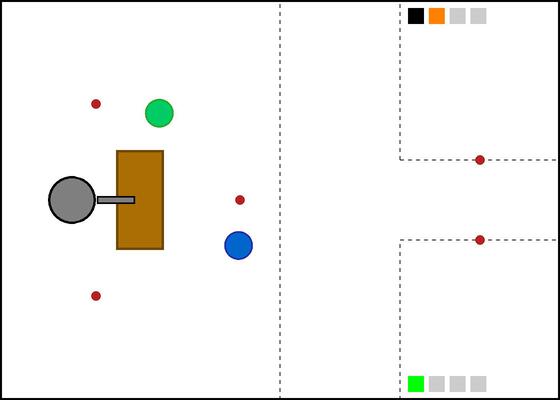}}
    ~
    \centering
    \subcaptionbox{(j) Green robot successfully delivers Tool-1 to workshop-0. \vspace{2mm}}
        [0.30\linewidth]{\includegraphics[scale=0.18]{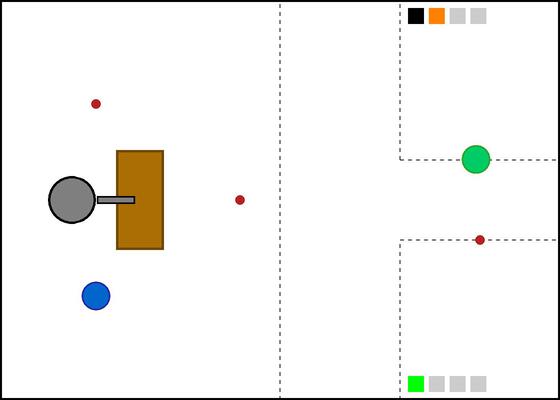}}
    ~
    \centering
    \subcaptionbox{(k) Arm robot executes \emph{\textbf{Pass-to-M(1)}} to pass Tool-0 to blue robot. Green robot runs \emph{\textbf{Get-Tool}} to go back table. \vspace{2mm}}
        [0.30\linewidth]{\includegraphics[scale=0.18]{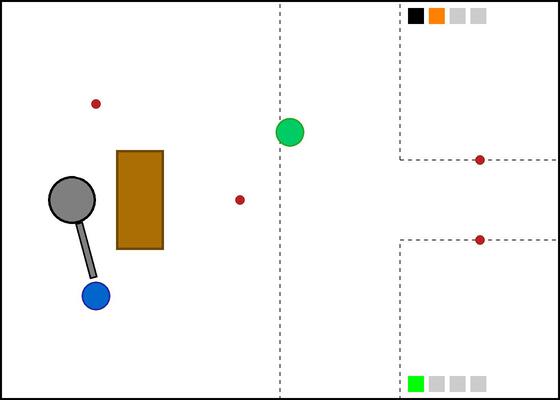}}  
    ~
    \centering
    \subcaptionbox{(l) Arm robot runs \emph{\textbf{Search-Tool(2)}} to find Tool-2. Blue robot moves to workshop-1 by executing \emph{\textbf{Go-W(1)}}. Human-0 finishes subtask-1 and starts to do subtask-2.\vspace{2mm}}
        [0.30\linewidth]{\includegraphics[scale=0.18]{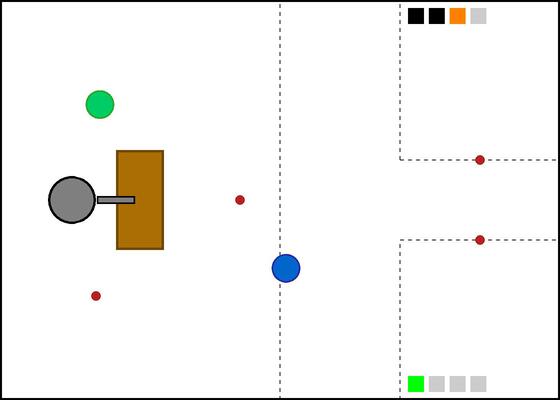}}
\end{figure*}
\begin{figure*}[h!]
    \centering
    \captionsetup[subfigure]{labelformat=empty}          
    ~
    \centering        
    ~
    \centering
    \subcaptionbox{(m) Blue robot successfully delivers Tool-0 to workshop-1. \vspace{2mm}}
        [0.30\linewidth]{\includegraphics[scale=0.18]{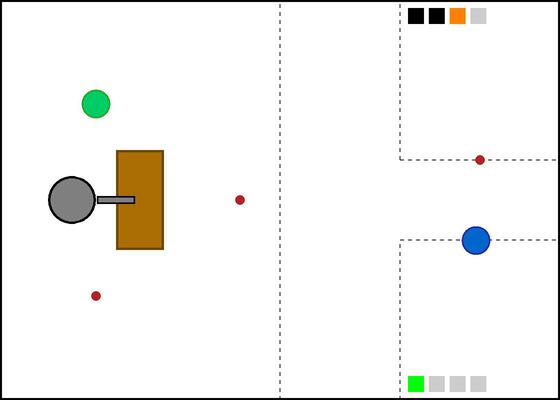}}
    ~
    \centering
    \subcaptionbox{(n) Arm robot executes \emph{\textbf{Pass-to-M(0)}} to pass Tool-2 to green robot. Blue robot runs \emph{\textbf{Get-Tool}} to go back table.\vspace{2mm}}
        [0.30\linewidth]{\includegraphics[scale=0.18]{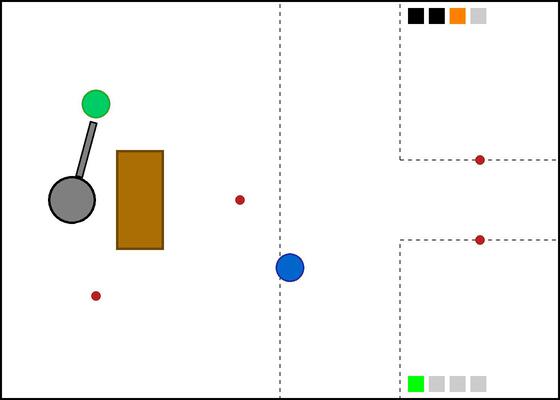}}
    ~
    \centering
    \subcaptionbox{(o) Arm robot runs \emph{\textbf{Search-Tool(1)}} to find Tool-1. Green robot moves to workshop-0 by executing \emph{\textbf{Go-W(0)}}.\vspace{2mm}}
        [0.30\linewidth]{\includegraphics[scale=0.18]{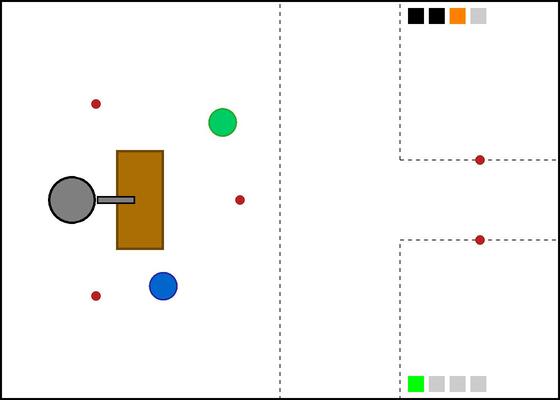}}
    ~
    \subcaptionbox{(p) Green robot successfully delivers Tool-2 to workshop-0.  \vspace{2mm}}
        [0.30\linewidth]{\includegraphics[scale=0.18]{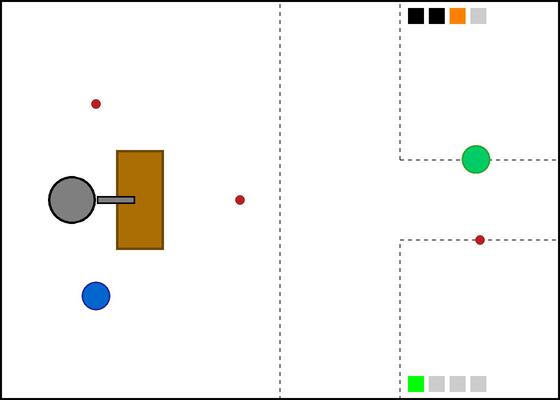}}
    ~
    \centering
    \subcaptionbox{(q) Green robot moves to workshop-1 by executing \emph{\textbf{Go-W(1)}} to observe human-1's status. Human-0 finishes subtask-2 and starts to do subtask-3.\vspace{2mm}}
        [0.30\linewidth]{\includegraphics[scale=0.18]{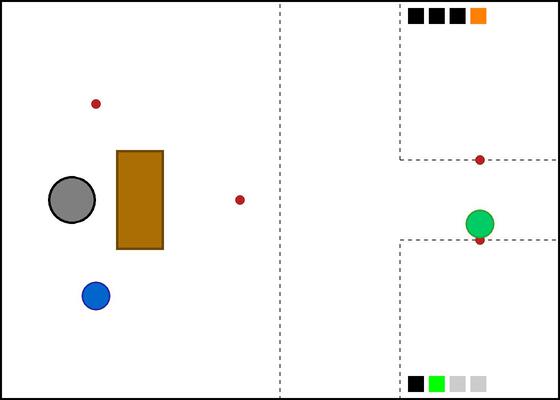}}
    ~
    \centering
    \subcaptionbox{(r) Green robot reaches workshop-1.\vspace{2mm}}
        [0.30\linewidth]{\includegraphics[scale=0.18]{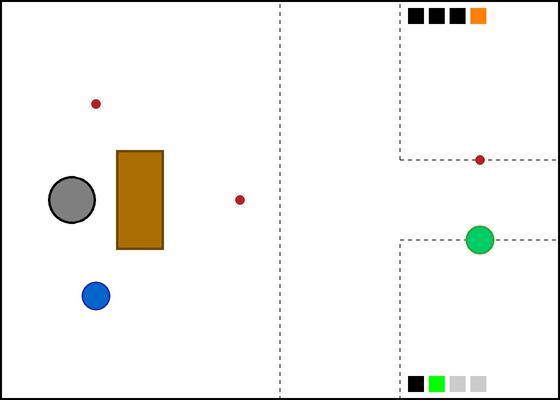}}
    ~
    \centering
    \subcaptionbox{(s) Arm robot executes \emph{\textbf{Pass-to-M(1)}} to pass Tool-1 to blue robot. Green robot runs \emph{\textbf{Get-Tool}} to go back table.\vspace{2mm}}
        [0.30\linewidth]{\includegraphics[scale=0.18]{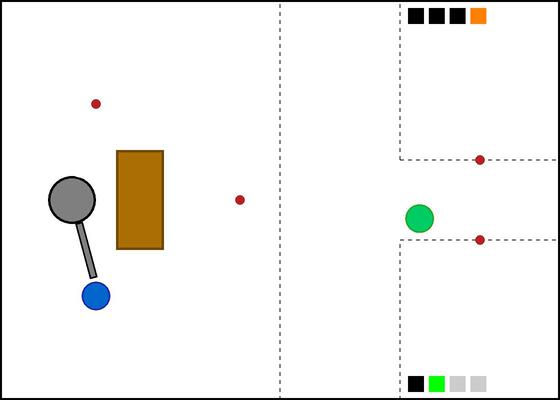}}
    ~
    \centering
    \subcaptionbox{(t) Arm robot runs \emph{\textbf{Search-Tool(2)}} to find Tool-2. Blue robot moves to workshop-1 by executing \emph{\textbf{Go-W(1)}}. \vspace{2mm}}
        [0.30\linewidth]{\includegraphics[scale=0.18]{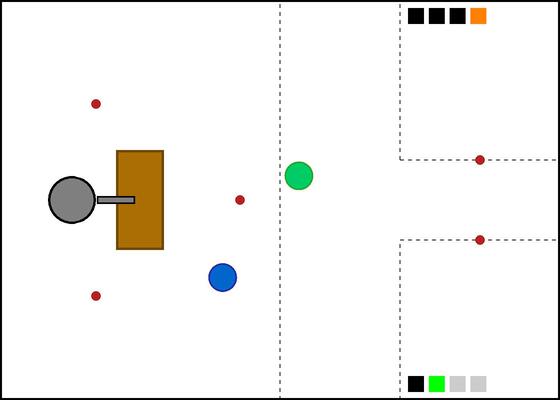}}
    ~
    \centering
    \subcaptionbox{(u) Blue robot successfully delivers Tool-1 to workshop-1. \vspace{2mm}}
        [0.30\linewidth]{\includegraphics[scale=0.18]{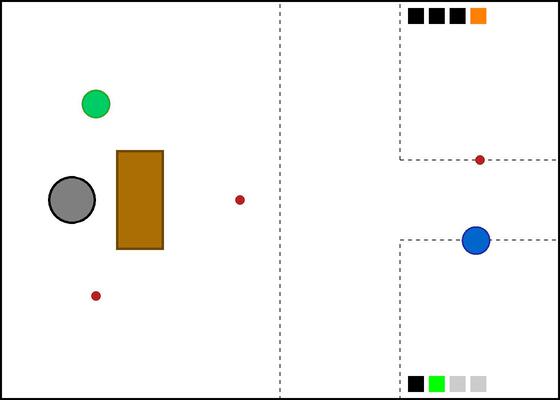}}
    ~
    \centering
    \subcaptionbox{(v) Blue robot runs \emph{\textbf{Get-Tool}} to go back table. Arm robot executes \emph{\textbf{Pass-to-M(0)}} to pass Tool-2 to green robot. Blue robot runs \emph{\textbf{Get-Tool}} to go back table.\vspace{2mm}}
        [0.30\linewidth]{\includegraphics[scale=0.18]{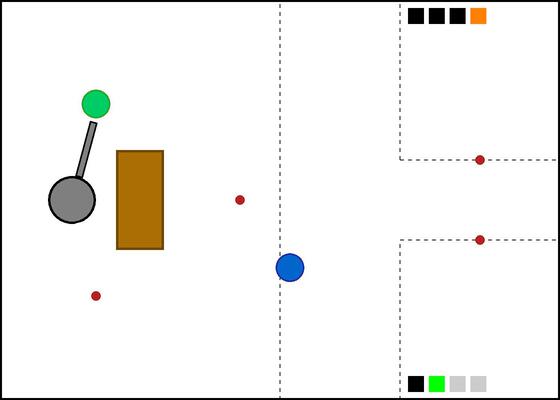}}
    ~
    \centering
    \subcaptionbox{(w) Green robot moves to workshop-1 by executing \emph{\textbf{Go-W(1)}}.\vspace{2mm}}
        [0.30\linewidth]{\includegraphics[scale=0.18]{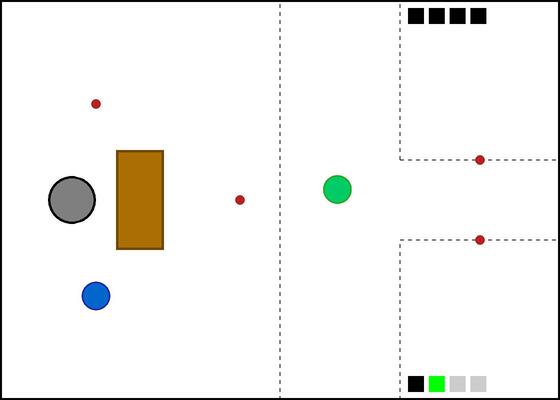}}
    ~
    \centering
    \subcaptionbox{(x) Human-1 finishes subtask-1 and start to do subtask-2. \vspace{2mm}}
        [0.30\linewidth]{\includegraphics[scale=0.18]{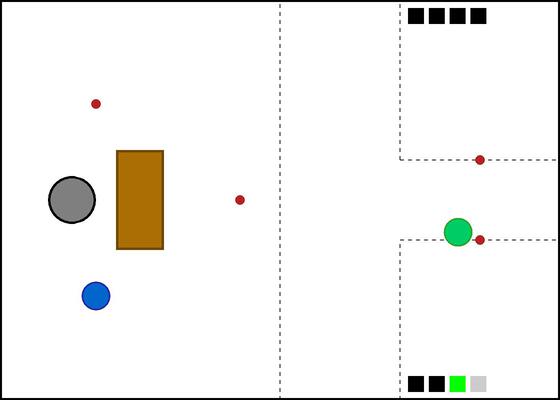}}
    ~
    \centering
    \subcaptionbox{(y) Green robot successfully delivers Tool-2 to workshop-1. Humans have received all tools, and for robots, the task is done.\vspace{2mm}}
        [0.30\linewidth]{\includegraphics[scale=0.18]{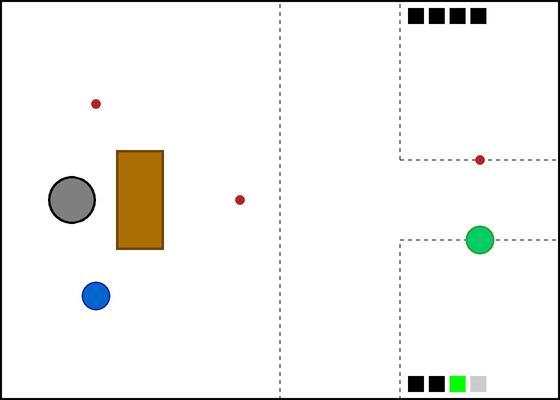}}  
    \label{wtd_b_behavior}
\end{figure*}





\end{document}